\documentclass[11pt]{article}

\usepackage[preprint]{acl}

\usepackage{times}
\usepackage{latexsym}
\usepackage{amssymb}
\usepackage{amsmath}

\usepackage[T1]{fontenc}

\usepackage[utf8]{inputenc}

\usepackage{microtype}
\usepackage{url}
\usepackage{inconsolata}

\usepackage{graphicx}
\usepackage{xspace}
\usepackage{booktabs}
\usepackage{multirow}
\usepackage{nccmath}
\usepackage[most]{tcolorbox}
\usepackage{listings}
\usepackage{authblk}

\def\secref#1{\S\ref{sec:#1}}
\def\seclabel#1{\label{sec:#1}}

\def\copsd{\textsc{COPSD}\xspace}

\title{Crosslingual On-Policy Self-Distillation for Multilingual Reasoning}

\author[]{\bf{Yihong Liu}$^{\text *}$}
\author[]{\bf{Raoyuan Zhao}$^{\text *}$}
\author[]{{\bf Michael A. Hedderich}}
\author[]{{\bf Hinrich Sch\"utze}}

\affil[]{Center for Information and Language Processing, LMU Munich \\Munich Center for Machine Learning (MCML)
 \protect\\ \texttt{\{yihong, rzhao, hedderich\}@cis.lmu.de}} 

\begin{document}
\maketitle

\begin{abstract}
Large language models (LLMs) have achieved remarkable progress in mathematical reasoning, but this ability is not equally accessible across languages.
Especially low-resource languages exhibit much lower reasoning performance.
To address this, we propose \emph{\textbf{C}rosslingual \textbf{O}n-\textbf{P}olicy \textbf{S}elf-\textbf{D}istillation} (\textbf{\copsd}), which transfers a model's own high-resource reasoning behavior to low-resource languages.
\copsd uses the same model as student and teacher: the student sees only the low-resource problem, while the teacher receives privileged crosslingual context, including the problem translation and reference solution in English.
Training minimizes full-distribution token-level divergence on the student's own rollouts, providing dense supervision while avoiding the sparsity and instability of outcome-only reinforcement learning (RL).
Experiments on 17 low-resource African languages show that \copsd consistently improves low-resource mathematical reasoning across model sizes and substantially outperforms Group Relative Policy Optimization (GRPO).
Further analyses show that \copsd improves answer-format adherence, strengthens test-time scaling, and generalizes to harder multilingual reasoning benchmarks, with especially large gains for lower-resource languages.
We make our code and data available at \url{https://github.com/cisnlp/COPSD}.
\end{abstract}

\def\thefootnote{*}\footnotetext{Equal contribution.}\def\thefootnote{\arabic{footnote}}

\section{Introduction}

Large language models (LLMs) have achieved remarkable progress in mathematical reasoning \citep{ahn-etal-2024-large,yang2025qwen3technicalreport,Guo2025deepseek}.
A key driver of this progress is their ability to generate step-by-step reasoning traces, which can elicit strong problem-solving behavior \citep{wei2022cot}.
However, this capability remains far from multilingual.
Models often struggle when reasoning in underrepresented languages \citep{hwang2025learngloballyspeaklocally,yong2025crosslingualreasoningtesttimescaling,ghosh-etal-2025-survey}, which receive limited exposure during pretraining and are rarely represented in high-quality reasoning supervision during post-training \citep{qin2024multilinguallargelanguagemodel,yang2025languageimbalance}.
As a result, a model may possess the latent ability to solve a problem, yet fail to access that ability when the problem and reasoning traces are expressed in a low-resource language.

\begin{figure}
    \centering
    \includegraphics[width=0.95\columnwidth]{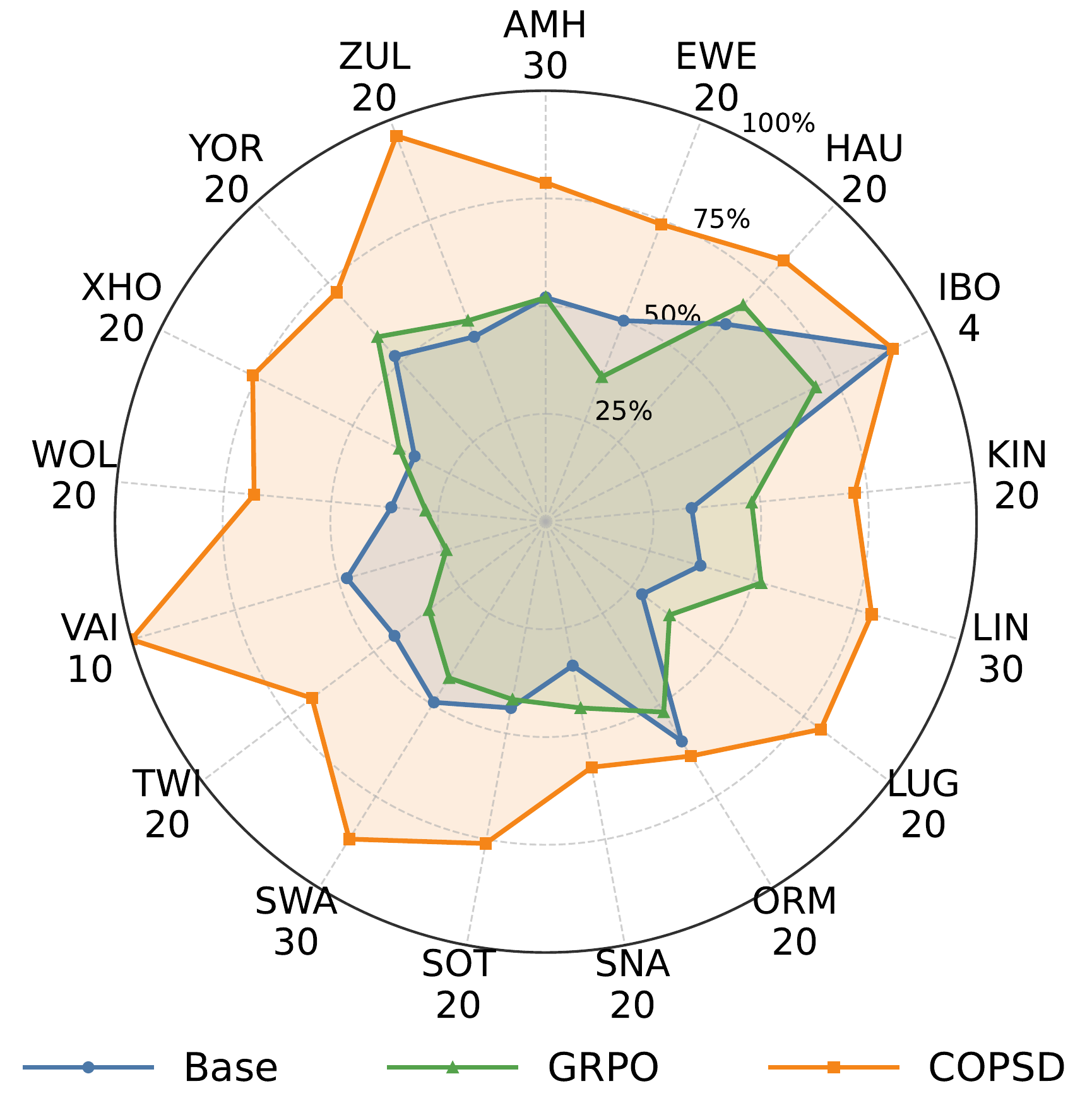}
    \caption{
    Radar comparison of \texttt{Qwen3-1.7B} performance on AfriMGSM under a 4096-token generation budget.
    Each axis corresponds to one of the 17 low-resource African languages, with axis-specific scaling based on the maximum observed performance for that language.
    COPSD consistently outperforms both the base and GRPO-trained models across languages.
    }
    \label{fig:first_page}
\end{figure}

A natural approach to this issue is to construct reasoning supervision directly in low-resource languages, e.g., by translating English reasoning traces into target languages and then performing supervised fine-tuning (SFT) \citep{wu-etal-2025-english,barua2026longchainofthoughtreasoninglanguages}.
Yet this approach faces several limitations.
Machine translation can introduce noise and is prone to inconsistencies or errors in mathematical expressions, quantities, and logical dependencies \citep{petersen-etal-2023-neural,zhang-etal-2024-enhancing-multilingual}.
Moreover, translated reasoning traces may not match the model's own reasoning behavior and therefore can suffer from train-inference distribution mismatch \citep{Agarwal2024OPD,gu2024minillm}.
Another possibility is to use reinforcement learning (RL) with outcome-based rewards, where the model is rewarded when its final answer matches the ground truth \citep{schulman2017proximalpolicyoptimizationalgorithms,shao2024deepseekmathpushinglimitsmathematical}.
However, such rewards can become extremely sparse in low-resource settings: if the model rarely produces correct answers, then binary outcome feedback provides little information about how intermediate reasoning should be improved, making RL sample-inefficient and potentially unstable \citep{lightman2024verify}.
These limitations suggest the need for a training signal that is both dense and scalable, while remaining aligned with the reasoning trajectories the model actually produces in low-resource languages.

To this end, we build on \emph{on-policy self-distillation}, where a single model acts as both student and teacher under different contexts and learns from dense feedback on its own generated trajectories \citep{zhao2026selfdistilledreasoneronpolicyselfdistillation,zhang2026opsdlonpolicyselfdistillationlongcontext,sang2026crispcompressedreasoningiterative}.
We extend this idea to multilingual reasoning and propose \emph{\textbf{C}rosslingual \textbf{O}n-\textbf{P}olicy \textbf{S}elf-\textbf{D}istillation} (\textbf{\copsd}), which transfers reasoning behavior from high-resource languages such as English to low-resource languages.
Specifically, in \copsd, the student observes only the low-resource problem, while the teacher is additionally conditioned on privileged crosslingual information, including the English translation of the problem and the English reference solution.
The student first generates its own reasoning trajectory, and \copsd then minimizes a full-distribution token-level divergence between the student and teacher policies along this trajectory.
This provides dense supervision at every decoding step while keeping training aligned with the reasoning paths the student policy actually explores.
Intuitively, \copsd enables the model to use its own English-accessible reasoning behavior to correct and improve its reasoning in low-resource languages.

We train Qwen3 models at three scales (1.7B, 4B, and 8B) with \copsd on 17 low-resource African languages and evaluate them on AfriMGSM \citep{adelani-etal-2025-irokobench}.
Our results show that \copsd consistently improves over the base models and substantially outperforms
GRPO (cf. Figure~\ref{fig:first_page}).
Further analyses show that \copsd converges rapidly, improves answer-format adherence, and enables models to better leverage larger test-time generation budgets.
We also evaluate \copsd on 8 languages from the more challenging PolyMath benchmark \citep{wang2025polymathevaluatingmathematicalreasoning}, finding that its gains generalize beyond AfriMGSM and are especially pronounced for lower-resource languages.


Our contributions are summarized as follows:
(\textbf{i}) We propose \copsd, a crosslingual on-policy self-distillation framework that uses high-resource language context as privileged information to improve low-resource reasoning.
(\textbf{ii}) We demonstrate consistent improvements over base models and substantial gains over GRPO across 17 low-resource African languages and multiple model sizes.
(\textbf{iii}) We analyze training dynamics, answer-format adherence, and test-time scaling, showing that \copsd improves both accuracy and the effectiveness of low-resource reasoning trajectories.
(\textbf{iv}) We show that \copsd generalizes to harder multilingual reasoning settings, with especially strong gains for lower-resource languages.
(\textbf{v}) We release our code and data to support future research on multilingual reasoning in low-resource languages.


\section{Related Work}

\paragraph{On-Policy Distillation.}
On-policy distillation (OPD) \citep{gu2024minillm,Agarwal2024OPD,lu2025onpolicydistillation,yang2026learningteachergeneralizedonpolicy} has emerged as an effective alternative to both SFT \citep{yang-etal-2024-self,Chung2024sft,ye-etal-2025-analyzing} and outcome-based RL for improving LLM reasoning \citep{shao2024deepseekmathpushinglimitsmathematical,liu2025understandingr1zeroliketrainingcritical,wen2025reinforcementlearningverifiablerewards}.
Compared to SFT and RL, OPD combines on-policy supervision from student-generated trajectories with dense token-level teacher feedback, thereby reducing train-inference distribution mismatch while avoiding sparse sequence-level rewards \citep{Agarwal2024OPD,gu2024minillm,zhao2026selfdistilledreasoneronpolicyselfdistillation}.
Recent work shows that effective OPD requires compatible teacher-student thinking patterns, as mismatches can hinder reasoning capability transfer \citep{li2026rethinkingonpolicydistillationlarge}.
This motivates \emph{on-policy self-distillation}, where a single model serves as both student and teacher under different contexts to improve reasoning behavior \citep{zhao2026selfdistilledreasoneronpolicyselfdistillation,zhang2026opsdlonpolicyselfdistillationlongcontext,kim2026doesselfdistillationsometimesdegrade,sang2026crispcompressedreasoningiterative}.
Our work extends OPSD to the multilingual setting, enabling the model to transfer its English-accessible reasoning behavior to low-resource languages and offering an effective, novel approach to improving low-resource reasoning.


\paragraph{Multilingual Reasoning.}
Multilingual reasoning concerns the ability of language models to solve reasoning problems consistently across languages, rather than relying primarily on English or other high-resource languages \citep{ghosh-etal-2025-survey}.
Prior work shows that LLMs exhibit substantial crosslingual performance gaps \citep{tam2025languagemattersmultilingualinput,zhao-etal-2026-comprehensive,liu2026largereasoningmodelsnot,ki2026makesgoodmultilingualreasoning}, especially in low-resource languages, and may generate inconsistent or language-mixed reasoning traces \citep{qi-etal-2025-models,wang-etal-2025-language-mixing}.
To address these issues, existing methods often use \emph{translate-and-test} pipelines \citep{qin-etal-2023-cross,huang-etal-2023-languages,zhu-etal-2024-question,kang2026multilingualreasoninggapsemerge}, \emph{supervised fine-tuning} \citep{zhao2024llamaenglishempiricalstudy,zhang-etal-2024-enhancing-multilingual,ustun-etal-2024-aya,lai-nissim-2024-mcot}, \emph{self-training} \citep{ranaldi-pucci-2025-multilingual,sutawika2026gainedtranslationprivilegedpairwise}, and \emph{reinforcement learning} \citep{she-etal-2024-mapo,ranaldi-pucci-2025-multilingual,wang-etal-2025-demystifying-multilingual,huang2025englishcentrictrainingreinforcementlearning,faisal2025aligningmultilingualreasoningverifiable,zhang2026thinknativelyunlockingmultilingual}.
However, these approaches typically require translated reasoning rationales or sparse outcome rewards.
In contrast, \copsd improves low-resource reasoning by using high-resource language context as privileged information and distilling dense token-level supervision from the same model on its own low-resource reasoning.

\section{Preliminary: On-Policy Self-Distillation}

\subsection{Teacher and Student Policies}

On-Policy Self-Distillation (OPSD) is a framework for improving reasoning without requiring a separate teacher model \citep{zhao2026selfdistilledreasoneronpolicyselfdistillation,zhang2026opsdlonpolicyselfdistillationlongcontext}.
Instead of distilling knowledge from an external model \citep{Agarwal2024OPD,lu2025onpolicydistillation}, OPSD instantiates the same model as both a student and a teacher under different conditioning contexts.
Given a reasoning dataset $\mathcal{D}=\{(x,y^*)\}$, where $x$ is a problem and $y^*$ is privileged information such as a reference solution, OPSD defines two policies from the same model $p_\theta$:
\begin{align*}
p_S(\cdot \mid x)
&\triangleq p_\theta(\cdot \mid x), \\
p_T(\cdot \mid x,y^*)
&\triangleq p_\theta(\cdot \mid x,y^*).
\end{align*}
The student policy $p_S$ observes only the problem, matching the inference-time setting, while the teacher policy $p_T$ additionally conditions on privileged information.
Although both policies share the same parameters, the teacher distribution is expected to provide a stronger learning signal because it can rationalize the problem with access to the reference solution.

\subsection{On-Policy Trajectory Sampling}

OPSD preserves the on-policy training paradigm by sampling trajectories from the \textbf{student} rather than from the teacher.
For a problem $x$, the student generates a response
\begin{equation*}
\hat{y}
=
(\hat{y}_1,\ldots,\hat{y}_{|\hat{y}|})
\sim p_S(\cdot \mid x).
\end{equation*}
Both the student and teacher then evaluate this same student-generated trajectory.
At each decoding step $n$, they produce next-token distributions conditioned on the same prefix $\hat{y}_{<n}$:
\begin{align*}
p_S^n
&\triangleq
p_S(\cdot \mid x,\hat{y}_{<n}), \\
p_T^n
&\triangleq
p_T(\cdot \mid x,y^*,\hat{y}_{<n}).
\end{align*}

\begin{figure*}[t]
    \centering
    \setlength{\belowcaptionskip}{-0.4cm}
    \includegraphics[width=\textwidth]{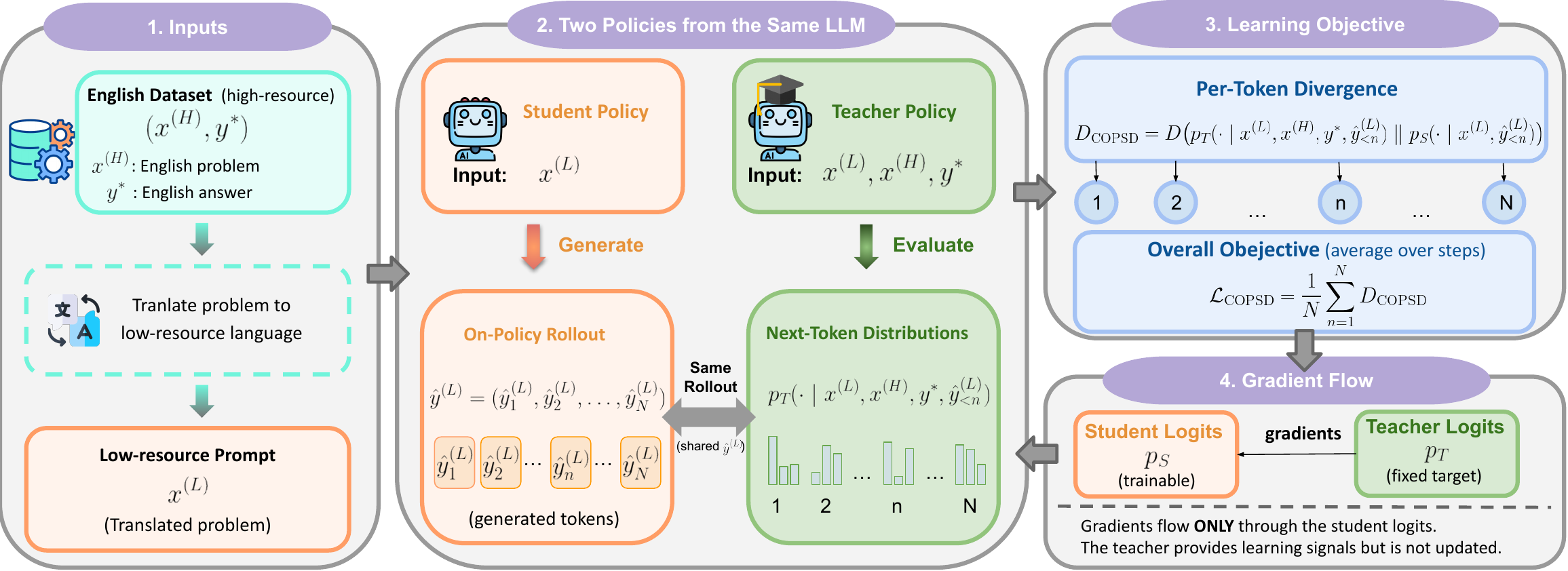}
    \caption{
Overview of \copsd.
Each problem is translated into a low-resource language as the student's input.
The same LLM acts as both student and teacher: the student generates an on-policy rollout, while the teacher evaluates it with privileged English context and the reference solution.
By minimizing per-token divergence along the rollout, \copsd transfers English-accessible reasoning behavior to improve reasoning in low-resource languages.
    }
    \label{fig:copsd}
\end{figure*}

\subsection{Distillation Objective}

The training objective minimizes the trajectory-averaged token-level divergence between the teacher and student distributions:
\begin{equation*}
D(p_T \parallel p_S)(\hat{y}\mid x)
=
\frac{1}{|\hat{y}|}
\sum_{n=1}^{|\hat{y}|}
D\!\left(
p_T^n \parallel p_S^n
\right),
\end{equation*}
where $D$ can be instantiated as a distributional divergence such as KL divergence \citep{kullback1951information}.
The overall OPSD objective is
\begin{equation*}
\begin{aligned}
\mathcal{L}_{\mathrm{OPSD}}(\theta)
= {}&
\mathbb{E}_{(x,y^*)\sim \mathcal{D}}
\,
\mathbb{E}_{\hat{y}\sim p_S(\cdot\mid x)}
\\
&\left[
D(p_T \parallel p_S)(\hat{y}\mid x)
\right].
\end{aligned}
\end{equation*}
Gradients flow only through the student policy, while the teacher serves as a fixed distributional target conditioned on privileged information.

\subsection{Discussion}

OPSD is attractive because: (i) It learns from on-policy student-generated trajectories, exploits privileged information, and avoids the need for an external teacher.
(ii) Compared with SFT/off-policy distillation, it reduces train-test mismatch by training on the student's own generations.
(iii) Compared with outcome-based RL, it provides dense teacher feedback over intermediate reasoning steps rather than relying only on sparse final-answer rewards.



\section{Methodology}\seclabel{method}

We introduce \emph{\textbf{C}rosslingual \textbf{O}n-\textbf{P}olicy \textbf{S}elf-\textbf{D}istillation} (\textbf{\copsd}), which extends OPSD to multilingual reasoning.
The key idea is to leverage high-resource language information as \emph{privileged context}.
During training, the student must reason from the low-resource problem alone, while the teacher is given additional high-resource, English context that helps elicit a stronger reasoning distribution from the same model, as shown in Figure~\ref{fig:copsd}.
This allows the model to transfer its own English-accessible reasoning behavior to low-resource languages without relying on an external teacher or target-language rationales.

\subsection{Crosslingual Learning Setup}

We consider a multilingual reasoning dataset
\[
\mathcal{D}
=
\{(x^{(L)}, x^{(H)}, y^*)\},
\]
where $x^{(L)}$ denotes a problem in a low-resource language, $x^{(H)}$ denotes its high-resource language counterpart, and $y^*$ is the reference solution in high-resource language.
In this work, we use English as the high-resource language, reflecting the English-centric nature of common LLM post-training \citep{shaham-etal-2024-multilingual,dang-etal-2024-rlhf}.

Following OPSD, \copsd instantiates two policies from the same language model $p_\theta$.
The student policy observes only the low-resource problem:
\begin{equation*}
p_S(\cdot \mid x^{(L)})
\triangleq
p_\theta(\cdot \mid x^{(L)}).
\end{equation*}
The teacher policy receives privileged crosslingual information:
\begin{equation*}
p_T(\cdot \mid x^{(L)},x^{(H)},y^*)
\triangleq
p_\theta(\cdot \mid x^{(L)},x^{(H)},y^*).
\end{equation*}
Thus, the student matches the inference-time condition, while the teacher has access to information that can induce more reliable reasoning behavior.\footnote{During training, we control the explicit reasoning language of both the student and teacher policies to match the low-resource language of the student input; see \secref{control}.}

\subsection{On-Policy Crosslingual Distillation}

Given a low-resource problem $x^{(L)}$, the student generates an on-policy reasoning trajectory:
\begin{equation*}
\hat{y}^{(L)}
=
(\hat{y}_1^{(L)},\ldots,\hat{y}^{(L)}_{|\hat{y}^{(L)}|})
\sim
p_S(\cdot \mid x^{(L)}).
\end{equation*}
Both policies then evaluate the same student-generated prefix.
At each step $n$, we have
\begin{align*}
p_S^n
&\triangleq
p_S(\cdot \mid x^{(L)}, \hat{y}^{(L)}_{<n}), \\
p_T^n
&\triangleq
p_T(\cdot \mid x^{(L)}, x^{(H)}, y^*, \hat{y}^{(L)}_{<n}).
\end{align*}

\copsd then minimizes the token-level divergence between the teacher and student distributions along the student's own rollout:
\begin{equation*}
D_{\copsd}(\hat{y}^{(L)}\mid x^{(L)})
=
\frac{1}{|\hat{y}^{(L)}|}
\sum_{n=1}^{|\hat{y}^{(L)}|}
D\!\left(
p_T^n \parallel p_S^n
\right),
\end{equation*}
where $D$ is a distributional divergence, such as KL divergence.
The training objective is formulated as
\begin{equation*}
\begin{aligned}
\mathcal{L}_{\copsd}&(\theta) = \mathbb{E}_{(x^{(L)},x^{(H)},y^*)\sim \mathcal{D}}
\\
&\mathbb{E}_{\hat{y}^{(L)}\sim p_S(\cdot\mid x^{(L)})}
\left[
D_{\copsd}(\hat{y}^{(L)}\mid x^{(L)})
\right].
\end{aligned}
\end{equation*}
Gradients are backpropagated only through the student policy, enabling the student to improve its reasoning in the low-resource language $L$.

\begin{table*}[t]
\centering
\setlength{\belowcaptionskip}{-0.5cm}
\small
\resizebox{\textwidth}{!}{%
\begin{tabular}{llrrrrrrrrrrrrrrrrrr}
\toprule
Model & Method & AMH & EWE & HAU & IBO & KIN & LIN & LUG & ORM & SNA & SOT & SWA & TWI & VAI & WOL & XHO & YOR & ZUL & Avg. \\
\midrule
\multirow{3}{*}{Qwen3-1.7B} & Base & 15.60 & 10.00 & 12.40 & \textbf{3.60} & 6.80 & 11.20 & 5.60 & 12.00 & 6.80 & 8.80 & 14.80 & 8.80 & 4.80 & 7.20 & 6.80 & 10.40 & 9.20 & 9.11 \\
 & GRPO & 15.60 & 7.20 & 13.60 & 2.80 & 9.60 & 15.60 & 7.20 & 10.40 & 8.80 & 8.40 & 12.80 & 6.80 & 2.40 & 5.60 & 7.60 & 11.60 & 10.00 & 9.18 \\
 & COPSD & \textbf{23.60} & \textbf{14.80} & \textbf{16.40} & \textbf{3.60} & \textbf{14.40} & \textbf{23.60} & \textbf{16.00} & \textbf{12.80} & \textbf{11.60} & \textbf{15.20} & \textbf{26.00} & \textbf{13.60} & \textbf{10.00} & \textbf{13.60} & \textbf{15.20} & \textbf{14.40} & \textbf{19.20} & \textbf{15.53} \\
\midrule
\multirow{3}{*}{Qwen3-4B} & Base & 29.20 & 16.40 & 20.80 & 4.40 & \textbf{17.60} & 21.60 & 14.80 & 24.40 & 14.00 & \textbf{16.80} & 47.60 & 12.80 & \textbf{18.00} & 17.60 & \textbf{18.40} & 15.20 & 16.80 & 19.20 \\
 & GRPO & 29.60 & \textbf{18.80} & \textbf{22.40} & 5.60 & \textbf{17.60} & 24.00 & 17.20 & 22.40 & \textbf{17.20} & \textbf{16.80} & 46.00 & 10.00 & 15.60 & 15.20 & 16.00 & \textbf{17.20} & 17.60 & 19.36 \\
 & COPSD & \textbf{38.00} & 15.60 & 20.80 & \textbf{9.20} & 16.00 & \textbf{26.00} & \textbf{18.80} & \textbf{27.20} & \textbf{17.20} & 16.40 & \textbf{48.40} & \textbf{14.80} & 12.80 & \textbf{20.00} & 14.00 & \textbf{17.20} & \textbf{18.00} & \textbf{20.61} \\
\midrule
\multirow{3}{*}{Qwen3-8B} & Base & 43.20 & 9.60 & 16.00 & 3.60 & \textbf{20.80} & 17.20 & 15.20 & 23.20 & 12.80 & 18.40 & 69.20 & 9.20 & 7.20 & 16.00 & 15.60 & 16.80 & 16.00 & 19.41 \\
 & GRPO & 42.00 & 8.00 & 14.40 & \textbf{4.00} & 17.60 & 20.40 & 8.40 & 26.00 & 15.20 & \textbf{21.20} & \textbf{70.80} & 8.00 & \textbf{8.40} & 15.20 & 15.20 & 16.00 & 16.00 & 19.22 \\
 & COPSD & \textbf{46.80} & \textbf{22.40} & \textbf{21.60} & \textbf{4.00} & 18.80 & \textbf{22.00} & \textbf{20.80} & \textbf{29.20} & \textbf{20.40} & 20.00 & 66.80 & \textbf{15.20} & 8.00 & \textbf{18.00} & \textbf{19.20} & \textbf{19.60} & \textbf{27.60} & \textbf{23.55} \\
\bottomrule
\end{tabular}%
}
\caption{
Pass@12 performance on 17 low-resource AfriMGSM languages under a 4,096-token generation budget.
\textbf{Bold} values indicate the best result among Base, GRPO, and \copsd for each model size and language.
\copsd outperforms both the base model and GRPO on most languages, with large gains for \texttt{Qwen3-1.7B} and \texttt{Qwen3-4B}.
}
\label{tab:afrimgsm_passk_4096}
\end{table*}

\section{Experiments}\seclabel{experiments}

\subsection{Models}
We conduct experiments with the Qwen3 model family \citep{yang2025qwen3technicalreport} of three sizes: \texttt{Qwen3-1.7B}, \texttt{Qwen3-4B}, and \texttt{Qwen3-8B}.
Qwen3 models are pretrained on multilingual corpora and further post-trained with SFT and RL on data dominated by high-resource languages such as English.

\subsection{Controlling Reasoning Language}\seclabel{control}
LLMs may switch to English in their reasoning traces, even when prompted in a different target language \citep{yong2025crosslingualreasoningtesttimescaling,wang-etal-2025-language-mixing}.
Since our goal is to improve reasoning in specific low-resource languages, we control the reasoning language with a prompt-hacking strategy \citep{qi-etal-2025-models,zhao-etal-2026-comprehensive}.
Specifically, we insert a language-specific prefix immediately after the \texttt{<think>} token, encouraging the model to reason in the target language during both \textbf{training} and \textbf{inference}.
Further details are provided in \secref{language_control}.

\subsection{Training}

\paragraph{Data}
We use OpenThoughts \citep{guha2025openthoughtsdatarecipesreasoning} as our training source, which provides math reasoning problems paired with English step-by-step reference solutions.
We sample 0.5K examples and translate the questions into the 17 low-resource African languages which are covered by AfriMGSM benchmark \citep{adelani-etal-2025-irokobench}.\footnote{Translations are produced with \texttt{Gemini-3-Flash}. The translation prompt template is provided in \secref{prompt}.}
The English questions and solutions are used as privileged information for the teacher policy, while the translated questions are used for the student policy.

\paragraph{Implementation}
Following \citet{zhao2026selfdistilledreasoneronpolicyselfdistillation}, we fix the teacher policy during training and use full-vocabulary logit distillation.
We instantiate the distributional divergence with reverse KL.
For all models, we set the maximum generation length for the student policy to 2048 tokens and train with Low-Rank Adaptation (LoRA) \citep{hu2022lora}.
All experiments are conducted on NVIDIA A100 or H200 GPUs.
Details are provided in \secref{environment}.

\subsection{Evaluation}

\paragraph{Benchmarks}
We primarily evaluate on \textbf{AfriMGSM} \citep{adelani-etal-2025-irokobench}, a human-translated version of MGSM \citep{shi2023mgsm} covering 17 African languages.
Each language contains 250 math reasoning problems.
In \secref{harder_benchmark}, we further evaluate on \textbf{PolyMath} \citep{wang2025polymathevaluatingmathematicalreasoning}, a more challenging multilingual reasoning benchmark with problems of varying difficulty.
For PolyMath, each language contains 125 questions.

\paragraph{Metrics}
We report pass@$k$ \citep{Kulal2019passk,chen2021evaluatinglargelanguagemodels} with $k=12$ throughout the paper.
For each problem, we sample 12 responses and compute whether at least one response yields the correct final answer.
We instruct models to enclose their final answers in \texttt{\textbackslash boxed\{\}}, extract the boxed content, and then compare it with the gold answer using \texttt{Math-Verify}.\footnote{\url{https://github.com/huggingface/Math-Verify}}

\paragraph{Baselines}
We compare \copsd against two baselines.
First, we evaluate the \textbf{original Qwen3 models}, which already exhibit strong reasoning capability in high-resource languages.
Second, we train Qwen3 models with \textbf{GRPO} \citep{shao2024deepseekmathpushinglimitsmathematical} using binary outcome rewards verified against gold numerical answers, where we set the maximum generation length to 16K tokens during training.

\subsection{Results and Discussion}

\paragraph{\copsd consistently improves low-resource mathematical reasoning across model scales.}
As shown in Table~\ref{tab:afrimgsm_passk_4096}, \copsd achieves the best average Pass@12 performance for all evaluated model sizes, improving \texttt{Qwen3-1.7B} from 9.11 to 15.53, \texttt{Qwen3-4B} from 19.20 to 20.61, and \texttt{Qwen3-8B} from 19.41 to 23.55.
The gains are especially pronounced for the smaller model, where \copsd improves performance on nearly every language and yields a relative improvement of over 70\% in average Pass@12 over the base model.
This suggests that low-resource reasoning performance can be substantially improved even without target-language reasoning rationales, as long as the model is provided with dense crosslingual supervision during training.
Notably, \copsd also improves performance across typologically and orthographically diverse languages, indicating that the benefit is not limited to language family or script.

\paragraph{Outcome-based RL provides limited gains in low-resource languages, while \copsd offers a denser and more reliable learning signal.}
For \texttt{Qwen3-1.7B}, GRPO only marginally improves the score from 9.11 to 9.18, and for \texttt{Qwen3-4B}, the improvement is similarly modest.
In several languages, GRPO even underperforms the base model, suggesting that binary rewards provide weak supervision when correct low-resource reasoning trajectories are rarely sampled.
This indicates that sparse rewards become a severe bottleneck in low-resource settings: If most sampled responses are incorrect, the reward signal gives little guidance about which intermediate reasoning steps should change.
In contrast, \copsd provides token-level distributional feedback along the student's own rollouts.
By conditioning the teacher on privileged English information and a reference solution, the same model can serve as an effective crosslingual teacher, guiding the student toward better reasoning behavior in the target low-resource language.

\begin{table}[t]
\centering
\setlength{\abovecaptionskip}{-0.01cm}
\setlength{\belowcaptionskip}{-0.5cm}
\footnotesize
\resizebox{\columnwidth}{!}{%
\begin{tabular}{lcccc}
\toprule
Model & $\rho_P^{\mathrm{mean}}$ & $\rho_S^{\mathrm{mean}}$ & $\rho_P^{\mathrm{pool}}$ & $\rho_S^{\mathrm{pool}}$ \\
\midrule
\texttt{Qwen3-1.7B} & 0.628 & 0.607 & 0.352 & 0.421 \\
\texttt{Qwen3-4B}   & 0.838 & 0.805 & 0.453 & 0.563 \\
\texttt{Qwen3-8B}   & 0.728 & 0.693 & 0.232 & 0.479 \\
\bottomrule
\end{tabular}}
\caption{Correlation between format rate and Pass@12 for \copsd during training.
$\rho_P$ and $\rho_S$ denote Pearson and Spearman correlations, respectively. 
The mean correlation averages coefficients computed independently for each language trajectory, while the pooled correlation is computed over all language-step pairs. 
The consistently positive correlations indicate that better format adherence is strongly associated with higher Pass@12.}
\label{tab:training_format_pass_corr_1024}
\end{table}

\begin{figure}[h]
    \centering
    \setlength{\abovecaptionskip}{-0.01cm}
    \setlength{\belowcaptionskip}{-0.5cm}
    \setlength{\belowcaptionskip}{-0.3cm}
    \includegraphics[width=0.80\linewidth]{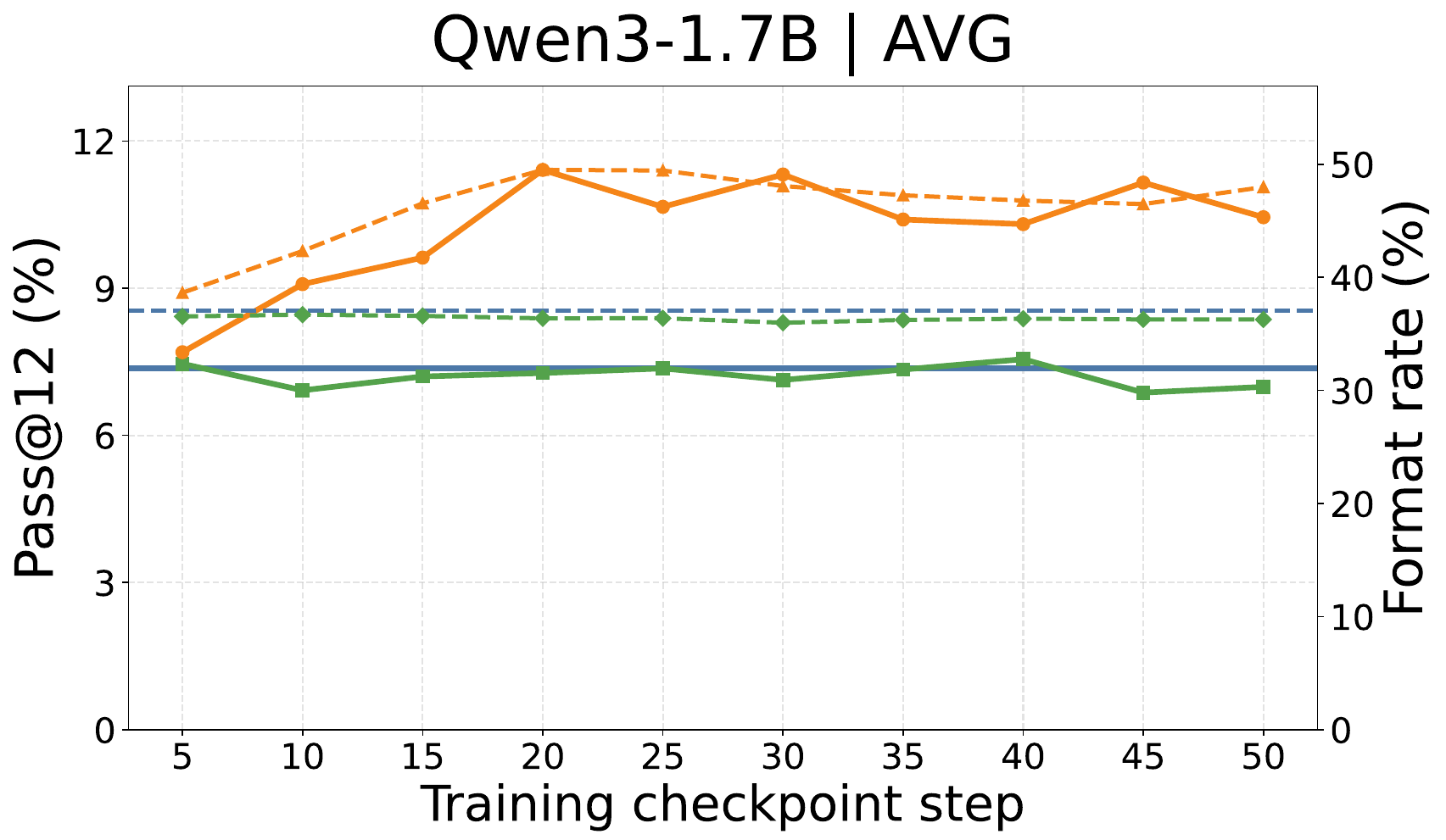}
    \includegraphics[width=0.80\linewidth]{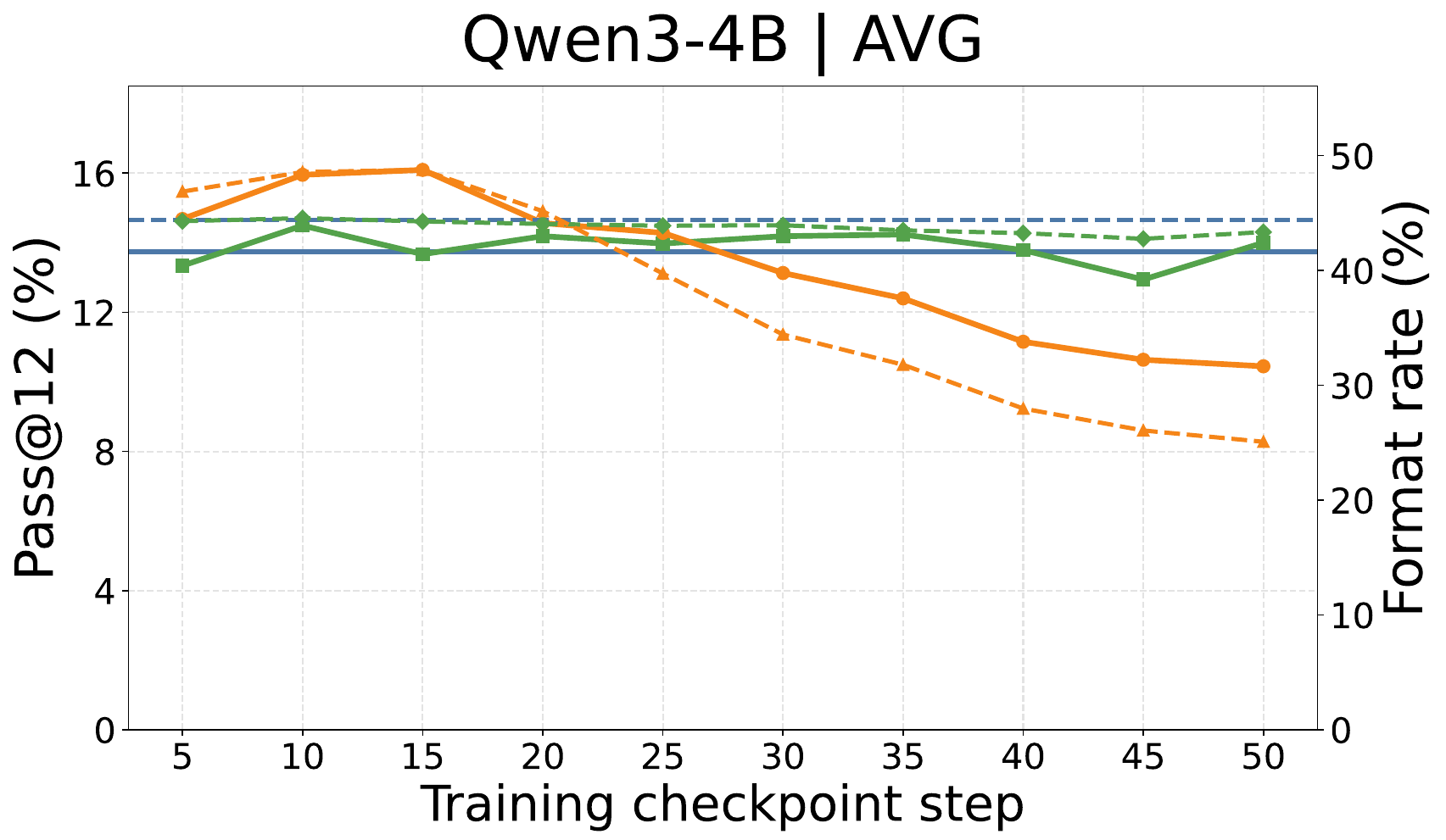}
    \includegraphics[width=0.80\linewidth]{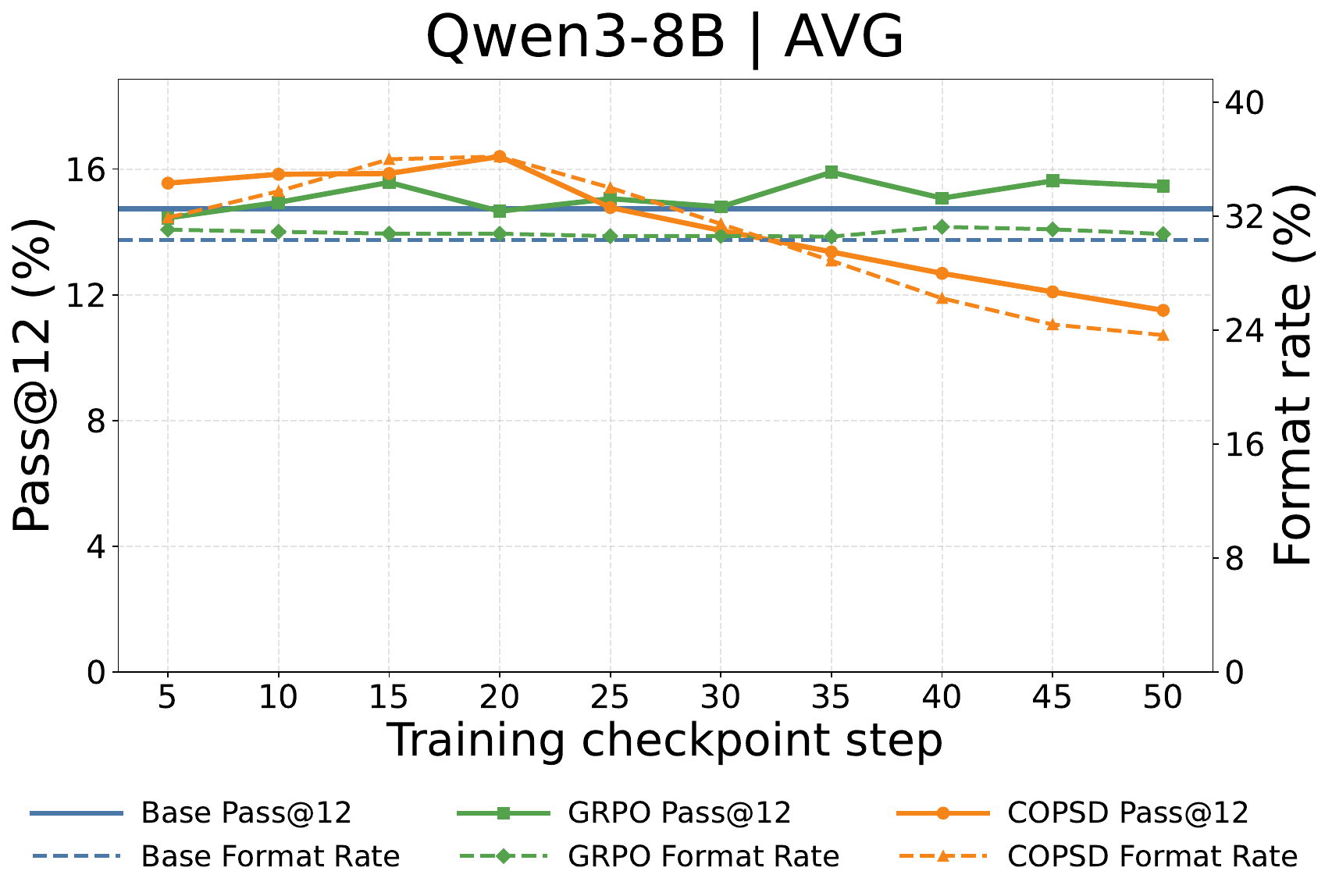}
    \caption{Average training dynamics across languages for \texttt{Qwen3-1.7B}, \texttt{Qwen3-4B}, and \texttt{Qwen3-8B}. Solid lines show Pass@12 and dashed lines show format rate. 
    Overall, \copsd converges quickly and often reaches its best performance within only a few training steps, while GRPO shows no clear improvement trend over training.
    }
    \label{fig:training_dynamics_average}
\end{figure}

\begin{figure*}
    \centering
    \setlength{\abovecaptionskip}{-0.03cm}
    \setlength{\belowcaptionskip}{-0.5cm}
    \includegraphics[width=0.32\linewidth]{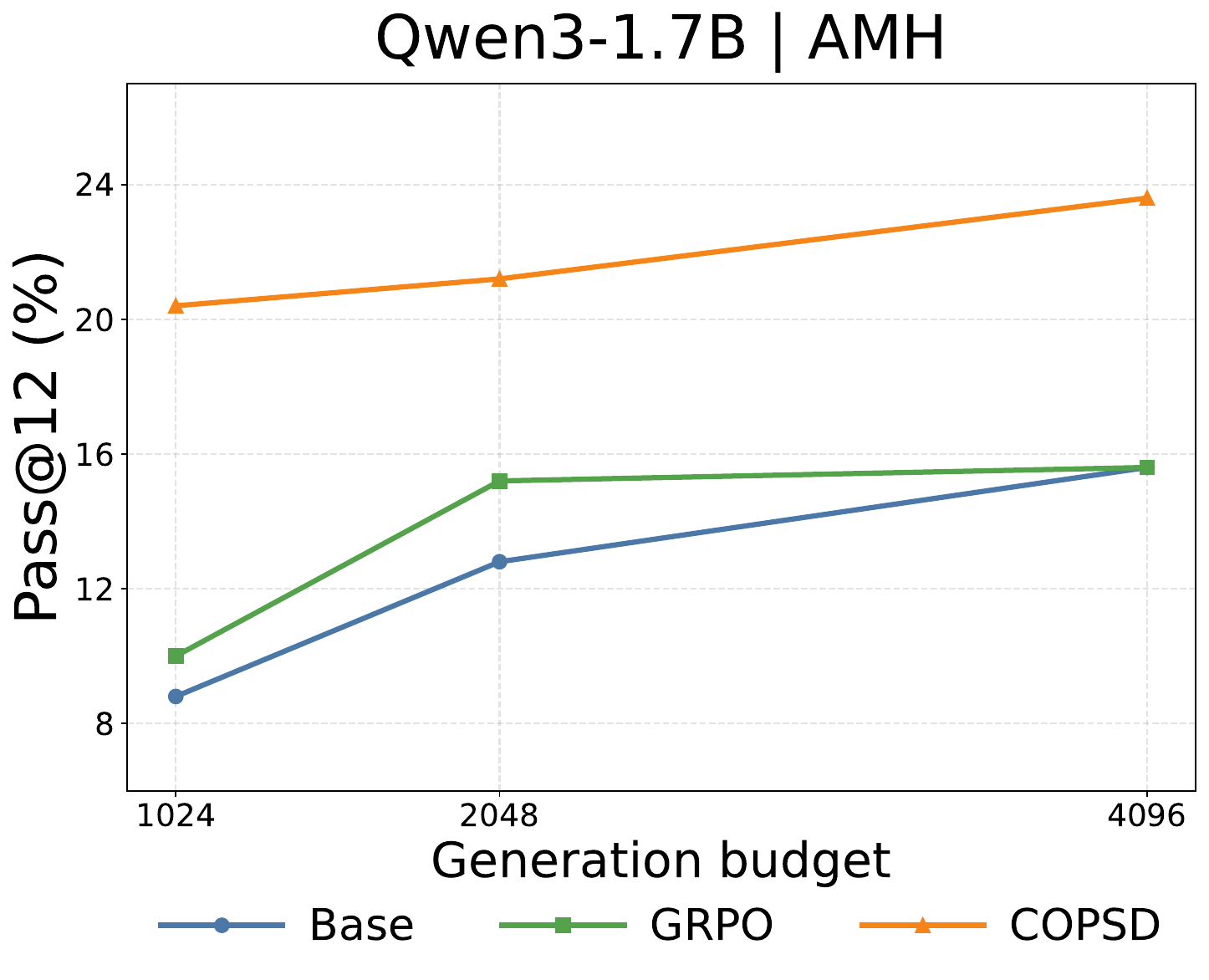}
    \includegraphics[width=0.32\linewidth]{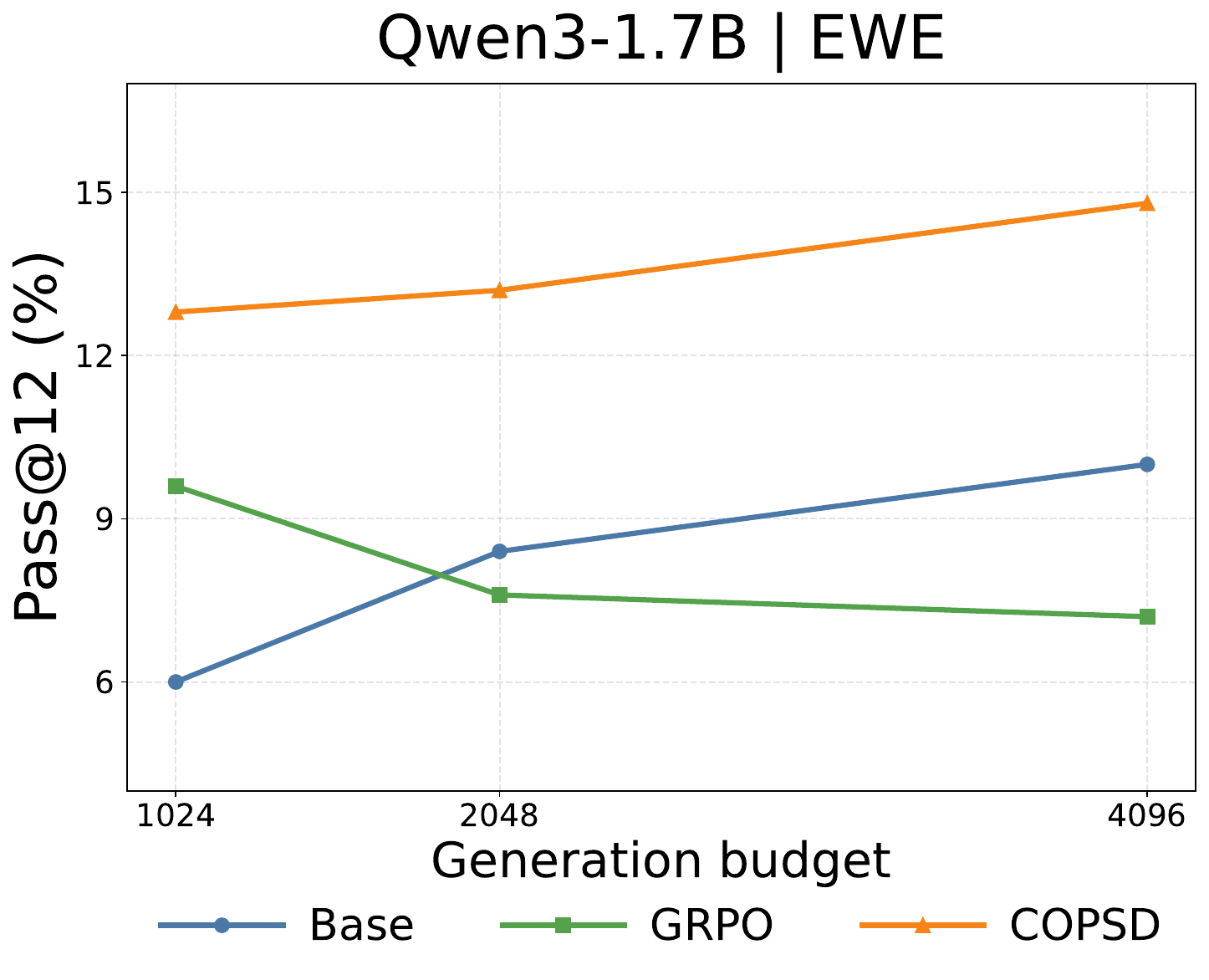}
    \includegraphics[width=0.32\linewidth]{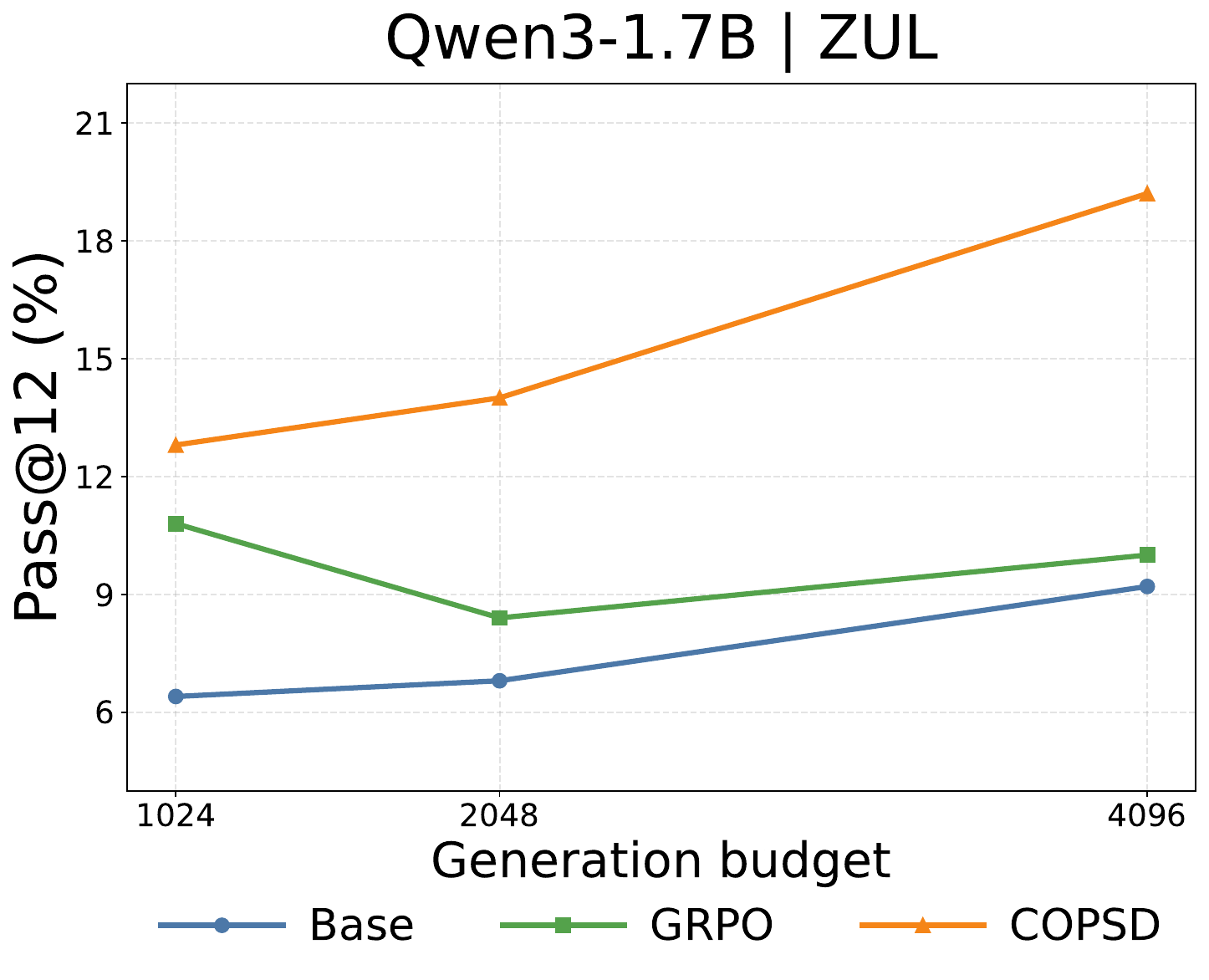}
    \includegraphics[width=0.32\linewidth]{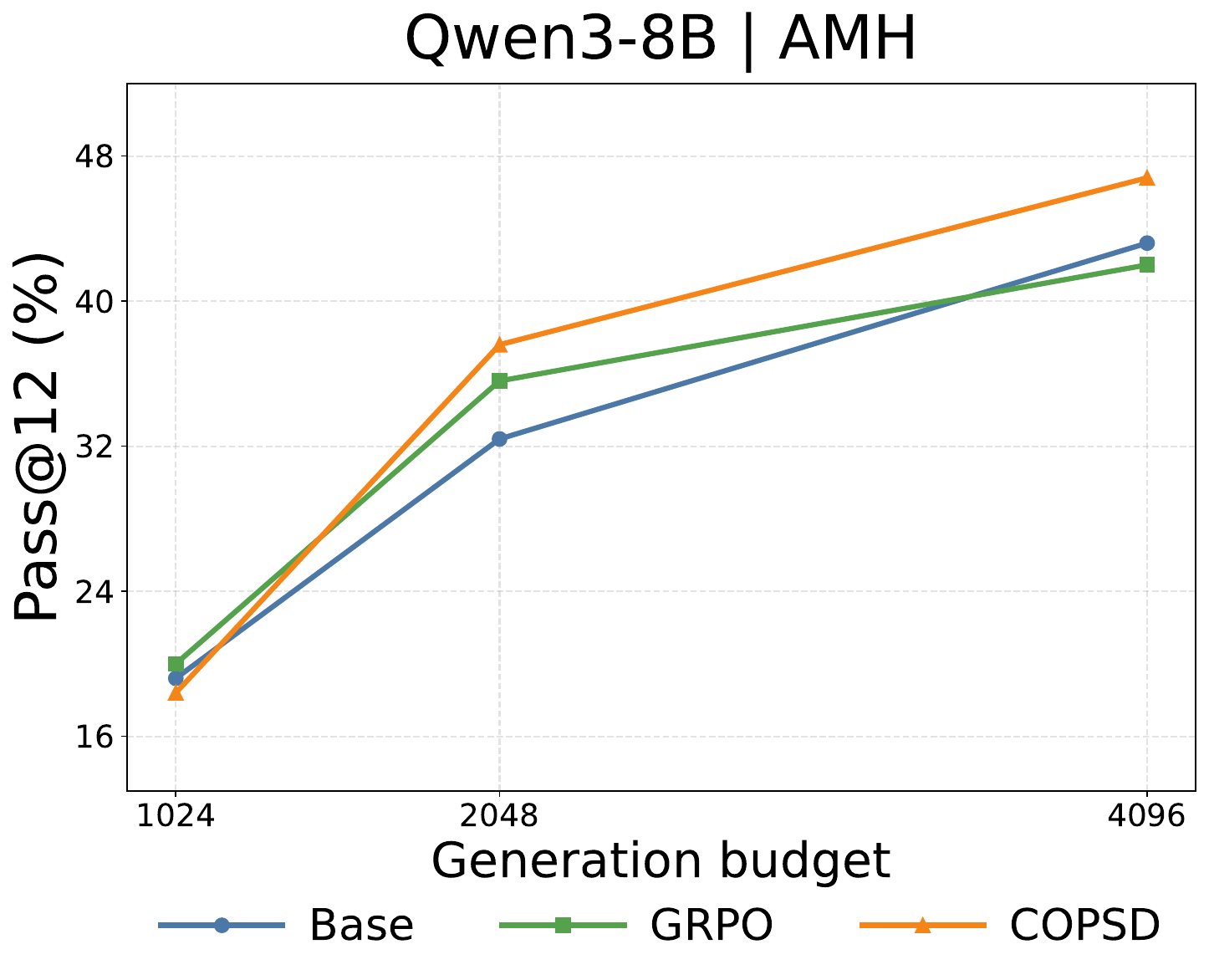}
    \includegraphics[width=0.32\linewidth]{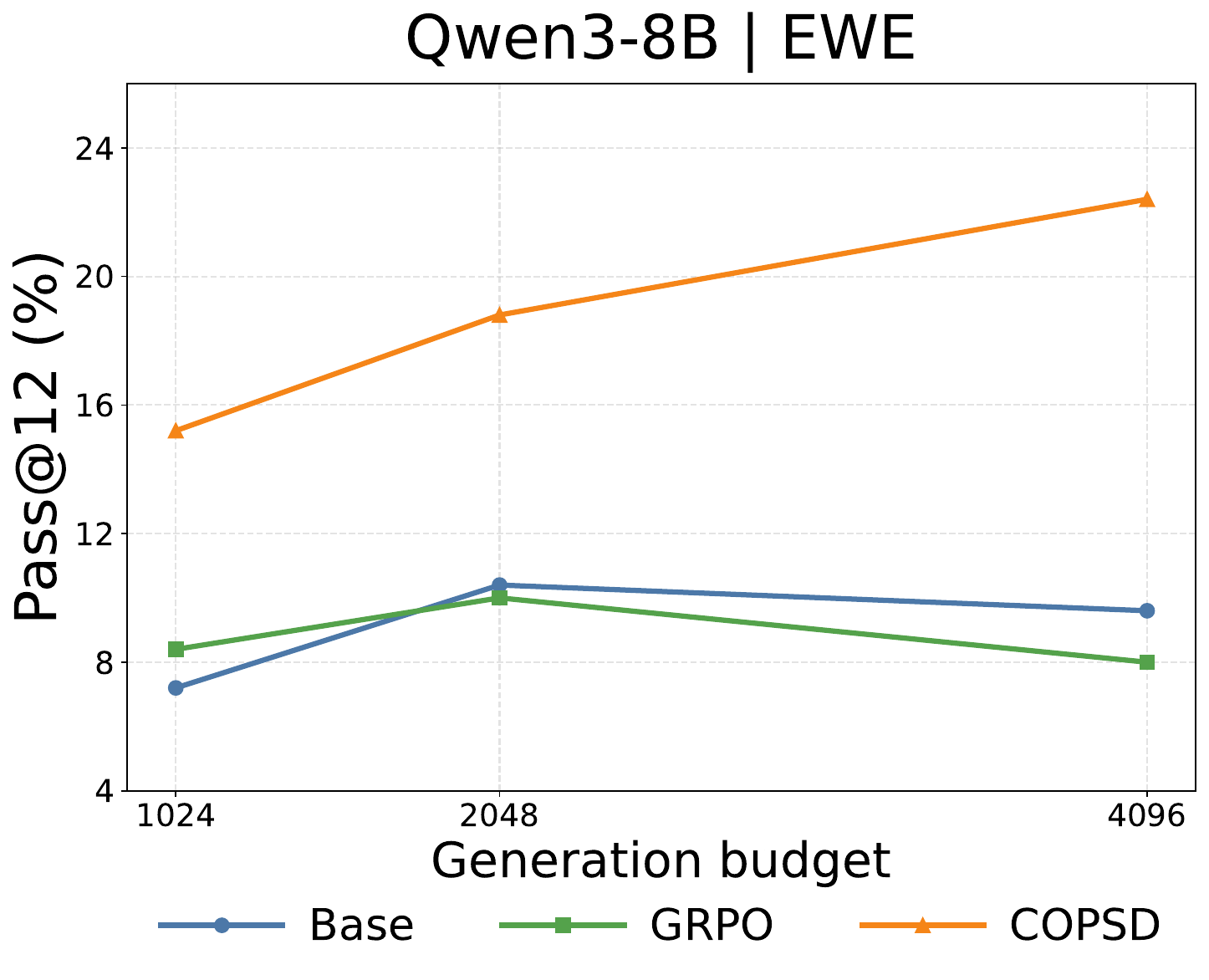}
    \includegraphics[width=0.32\linewidth]{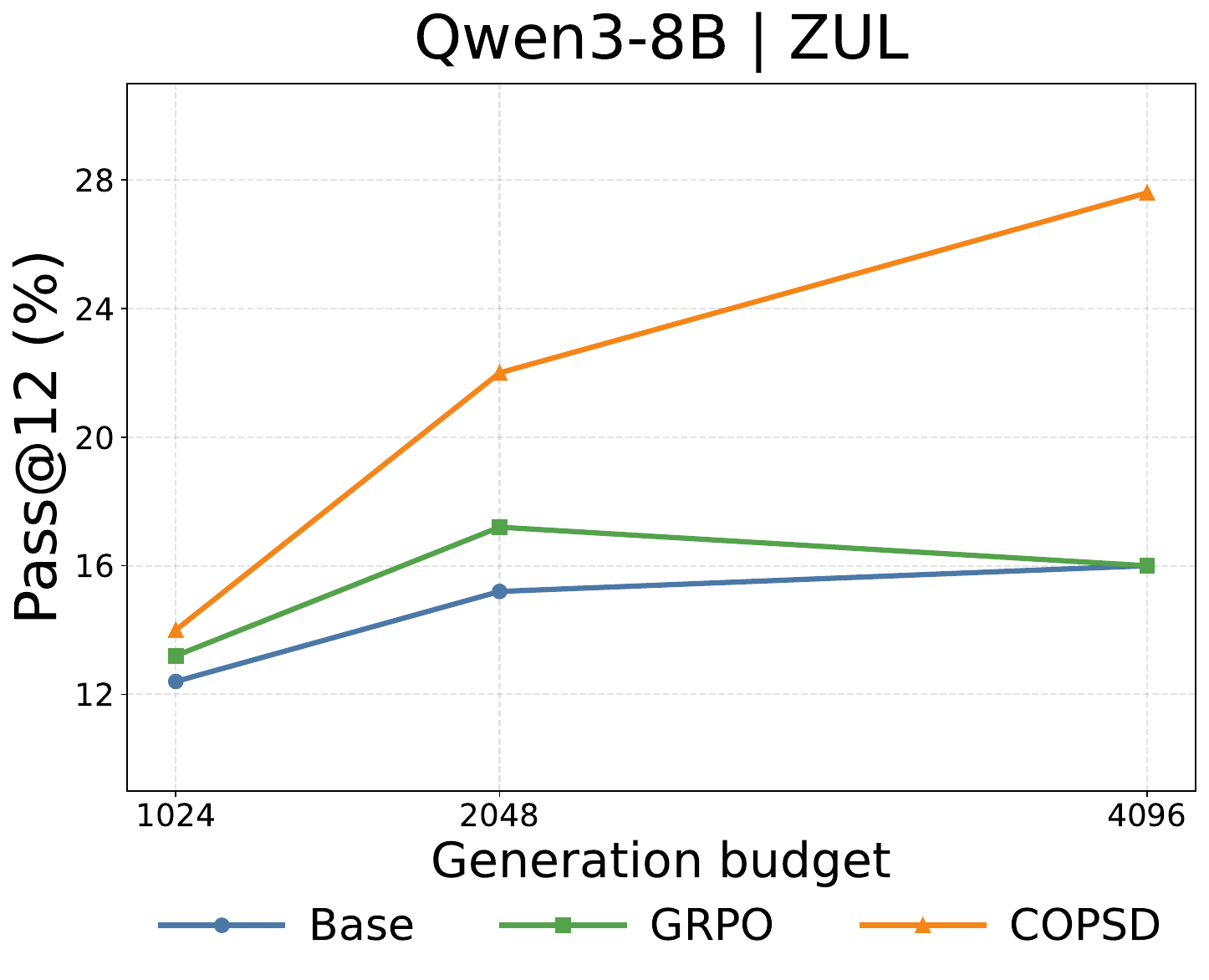}
    \caption{Test-time scaling results on Pass@12 for three representative languages: Amharic (AMH), Ewe (EWE), and Zulu (ZUL).
    The \texttt{Qwen3-8B} model exhibits a clearer and more consistent benefit from increased test-time computation.
    Across all languages and budgets, \copsd consistently outperforms both the Base model and GRPO.}
    \label{fig:test_time_scaling_selected}
\end{figure*}

\section{Complementary Analysis}

\subsection{Training Dynamics}

\paragraph{\copsd improves performance rapidly in early steps, while GRPO shows no clear upward trend.}
Figure~\ref{fig:training_dynamics_average} shows the average training dynamics across the 17 languages under the 1,024-token evaluation budget.\footnote{We provide complete dynamics for all languages in \secref{complete}.}
Across all model sizes, \copsd improves both Pass@12 and format rate in the early training steps.
While \texttt{Qwen3-1.7B} eventually plateaus, \texttt{Qwen3-4B} and \texttt{Qwen3-8B} reach their best performance within only a few gradient updates and then gradually decline.
This suggests that models can quickly absorb the dense distillation signal from the privileged teacher policy, but that the useful signal may be limited, possibly due to weak generation capability in the target low-resource languages.
As a result, continued updates may begin to overfit to imperfect teacher signals or otherwise hurt performance.
This behavior echoes prior observations that OPSD often converges rapidly \citep{zhao2026selfdistilledreasoneronpolicyselfdistillation}.
In contrast, GRPO shows no clear upward trend in either Pass@12 or format rate, consistent with its limited gains in Table~\ref{tab:afrimgsm_passk_4096}.
This further supports our hypothesis that binary outcome rewards are too sparse to provide reliable learning signals in low-resource reasoning settings.

\paragraph{Performance gains are closely tied to answer-format adherence.}
Figure~\ref{fig:training_dynamics_average} suggests a strong association between Pass@12 and format rate.
To further quantify this relationship, we report their correlations in Table~\ref{tab:training_format_pass_corr_1024}.
The mean per-language correlations are consistently high across model sizes, with Pearson correlations of 0.628, 0.838, and 0.728 for \texttt{Qwen3-1.7B}, \texttt{Qwen3-4B}, and \texttt{Qwen3-8B}, respectively.
Although the pooled correlations are lower, they remain positive, indicating that the relationship holds both within individual language learning trajectories and across all language--checkpoint pairs.
This suggests that low-resource reasoning failures can be partly caused by the model's inability to produce answers in the required format within a limited token budget.
The decline in format rate for larger models (4B and 8B) after early \copsd checkpoints therefore helps explain the corresponding drop in Pass@12 in Figure~\ref{fig:training_dynamics_average}.
These observations motivate our next analysis on test-time scaling (\secref{test_time}), where we examine whether larger generation budgets can recover or amplify the reasoning gains learned through \copsd.

\begin{table}[t]
\centering
\resizebox{\columnwidth}{!}{%
\begin{tabular}{llccc}
\toprule
Model & Method & 1,024 & 2,048 & 4,096 \\
\midrule
\multirow{3}{*}{\texttt{Qwen3-1.7B}} 
 & Base  & 7.36 & 8.33 (+13.1\%) & 9.11 (+23.6\%) \\
 & GRPO  & 9.13 & 7.84 (-14.2\%) & 9.18 (+0.5\%) \\
 & COPSD & 13.18 & 14.47 (+9.8\%) & 15.53 (+17.9\%) \\
\midrule
\multirow{3}{*}{\texttt{Qwen3-4B}} & Base & 13.74 & 17.62 (+28.3\%) & 19.20 (+39.7\%) \\
 & GRPO & 16.07 & 17.36 (+8.1\%) & 19.36 (+20.5\%) \\
 & COPSD & 18.16 & 19.51 (+7.4\%) & 20.61 (+13.5\%) \\
 \midrule
\multirow{3}{*}{\texttt{Qwen3-8B}} 
 & Base  & 14.73 & 18.42 (+25.1\%) & 19.41 (+31.8\%) \\
 & GRPO  & 16.89 & 18.19 (+7.7\%) & 19.22 (+13.8\%) \\
 & COPSD & 18.12 & 21.18 (+16.9\%) & 23.55 (+30.0\%) \\
\bottomrule
\end{tabular}
}
\caption{Average test-time scaling results on Pass@12 across languages under generation budgets of 1,024, 2,048, and 4,096 tokens. 
Values in parentheses indicate the relative change compared with the corresponding 1,024-token budget. 
\copsd consistently achieves the strongest performance across model sizes and generation budgets, showing larger gains from increased test-time computation than Base and GRPO.}
\label{tab:average-test-time-scaling}
\end{table}

\subsection{Test-Time Scaling}\seclabel{test_time}

\paragraph{Larger models benefit more consistently from increased test-time computation.}
Figure~\ref{fig:test_time_scaling_selected} shows test-time scaling trends for three representative languages (Amharic, Ewe, and Zulu),\footnote{We provide complete test-time scaling results for all languages and model sizes in \secref{complete}.} while Table~\ref{tab:average-test-time-scaling} reports average results across all 17 low-resource AfriMGSM languages.
Increasing the generation budget generally improves Pass@12, but the effect is clearer and more stable for larger models.
For example, the \texttt{Qwen3-8B} base model improves from 14.73 at 1,024 tokens to 19.41 at 4,096 tokens, while \copsd improves from 18.12 to 23.55.
By contrast, the gains for \texttt{Qwen3-1.7B} are relatively smaller, and GRPO shows unstable scaling behavior at the 2,048-token budget.
This suggests that effective crosslingual test-time scaling requires sufficient model capacity: larger models are better able to use additional generation budget to explore longer reasoning trajectories in low-resource languages, consistent with \citet{yong2025crosslingualreasoningtesttimescaling}.

\paragraph{\copsd strengthens the model's ability to use longer reasoning traces.}
Across all model sizes and generation budgets, \copsd achieves the highest average Pass@12, as shown in Table~\ref{tab:average-test-time-scaling}.
This indicates that the gains from \copsd persist as more test-time computation is allocated.
More importantly, \copsd often amplifies the benefit of longer generation budgets, especially for \texttt{Qwen3-8B}: its average performance increases by 30.0\% from 1,024 to 4,096 tokens, compared with 13.8\% for GRPO.
Figure~\ref{fig:test_time_scaling_selected} provides concrete examples.
For Amharic and Zulu with \texttt{Qwen3-8B}, \copsd starts close to the baselines at the 1,024-token budget, but separates more clearly as the budget increases.
This pattern is particularly strong for Zulu, where \copsd reaches roughly 28\% Pass@12 at 4,096 tokens, compared with about 16\% for the base and GRPO models.
These results suggest that \copsd improves not only low-resource reasoning accuracy, but also the model's ability to leverage longer target-language reasoning traces at inference time.

\subsection{Qualitative Analysis of Reasoning Trace}

Prior work has identified \emph{repetition} as a common failure mode in multilingual reasoning, particularly in low-resource languages~\citep{barua2026longchainofthoughtreasoninglanguages,tran2025reasoningtransferextremelylowresource}. 
Motivated by these findings, we examine whether model-generated reasoning traces exhibit repetitive degeneration and introduce a simple diagnostic metric, \emph{repeat rate}, to quantify this behavior.
Given a generated response, let $\mathcal{G}_n$ denote the multiset of all contiguous $n$-grams in the response, and let $\mathcal{G}_n^{\mathrm{unique}}$ denote the set of distinct $n$-grams. We define the $n$-gram repeat rate as
\begin{equation*}
\mathrm{RepeatRate}_n
= 1 - \frac{|\mathcal{G}_n^{\mathrm{unique}}|}{|\mathcal{G}_n|}.
\end{equation*}
A higher value indicates that a larger proportion of generated $n$-grams are repeated. 
We compute this metric for $n \in \{2,3,4,5,6\}$, which allows us to capture repetition at multiple granularities, ranging from short phrase-level duplication to longer repetitive reasoning fragments.

\begin{figure}[t]
    \centering
    \includegraphics[width=0.9\linewidth]{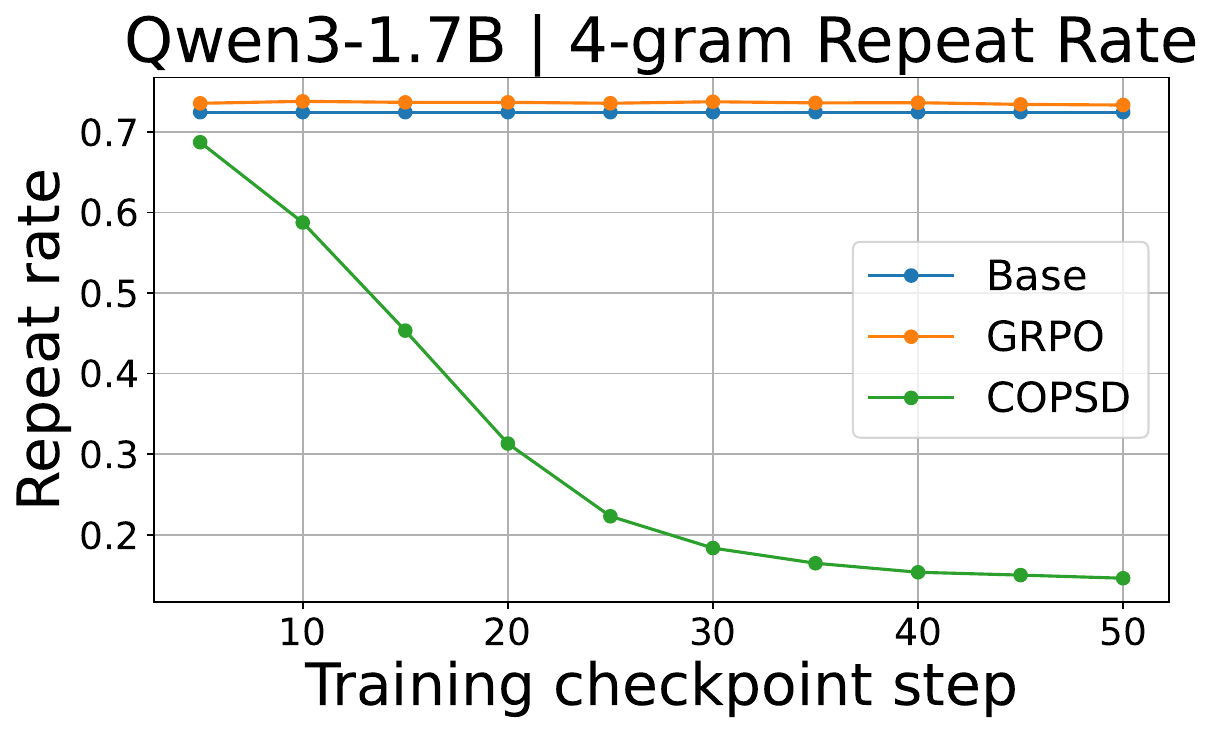}
    \caption{Average repeat rate comparison on \texttt{Qwen3-1.7B} with 4-grams.
    \copsd consistently reduces repetition compared to the base model and GRPO.}
    \label{fig:repeat_rate}
\end{figure}

\paragraph{\copsd effectively mitigates repetitive degeneration in reasoning traces.}
Figure~\ref{fig:repeat_rate} reports the average 4-gram repeat rate of \texttt{Qwen-1.7B} throughout training.\footnote{We report the full results for $n$-gram repeat rates in \secref{complete}.}
Compared with both the base model and GRPO, \copsd consistently maintains the lowest repeat rate across training steps.
Importantly, lower repeat rates should not be interpreted simply as greater lexical diversity; rather, for reasoning in low-resource languages, they typically indicate that the model is less likely to fall into repetitive loops or produce redundant reasoning fragments.
Together with the observed improvements in reasoning performance (cf. \secref{experiments}), this pattern suggests that \copsd encourages more coherent and structured reasoning, mitigating a failure mode in which multilingual reasoning traces collapse into meaningless or circular repetition.

\begin{figure*}
    \centering
    \setlength{\belowcaptionskip}{-0.4cm}
    \includegraphics[width=0.32\linewidth]{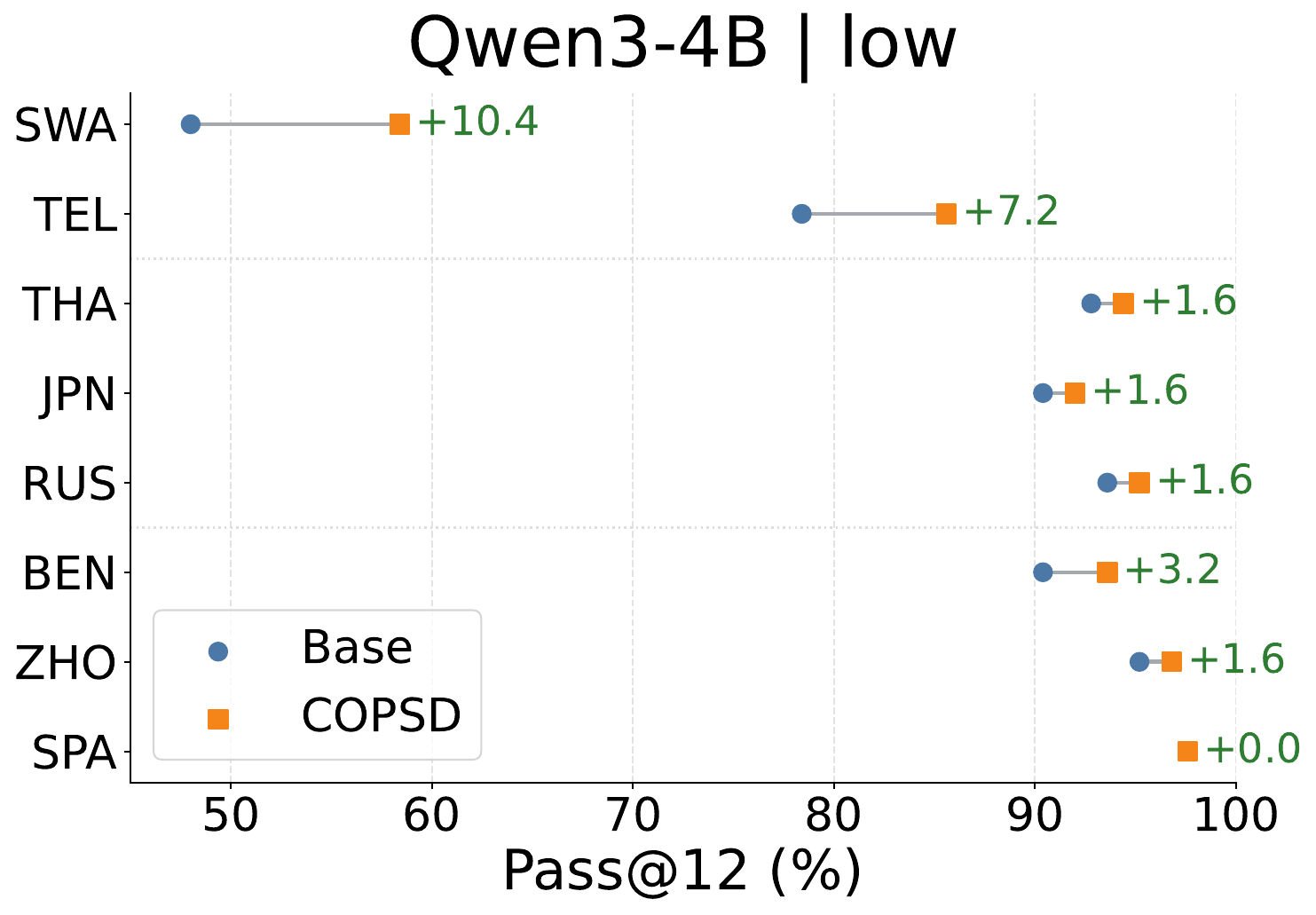}
    \includegraphics[width=0.32\linewidth]{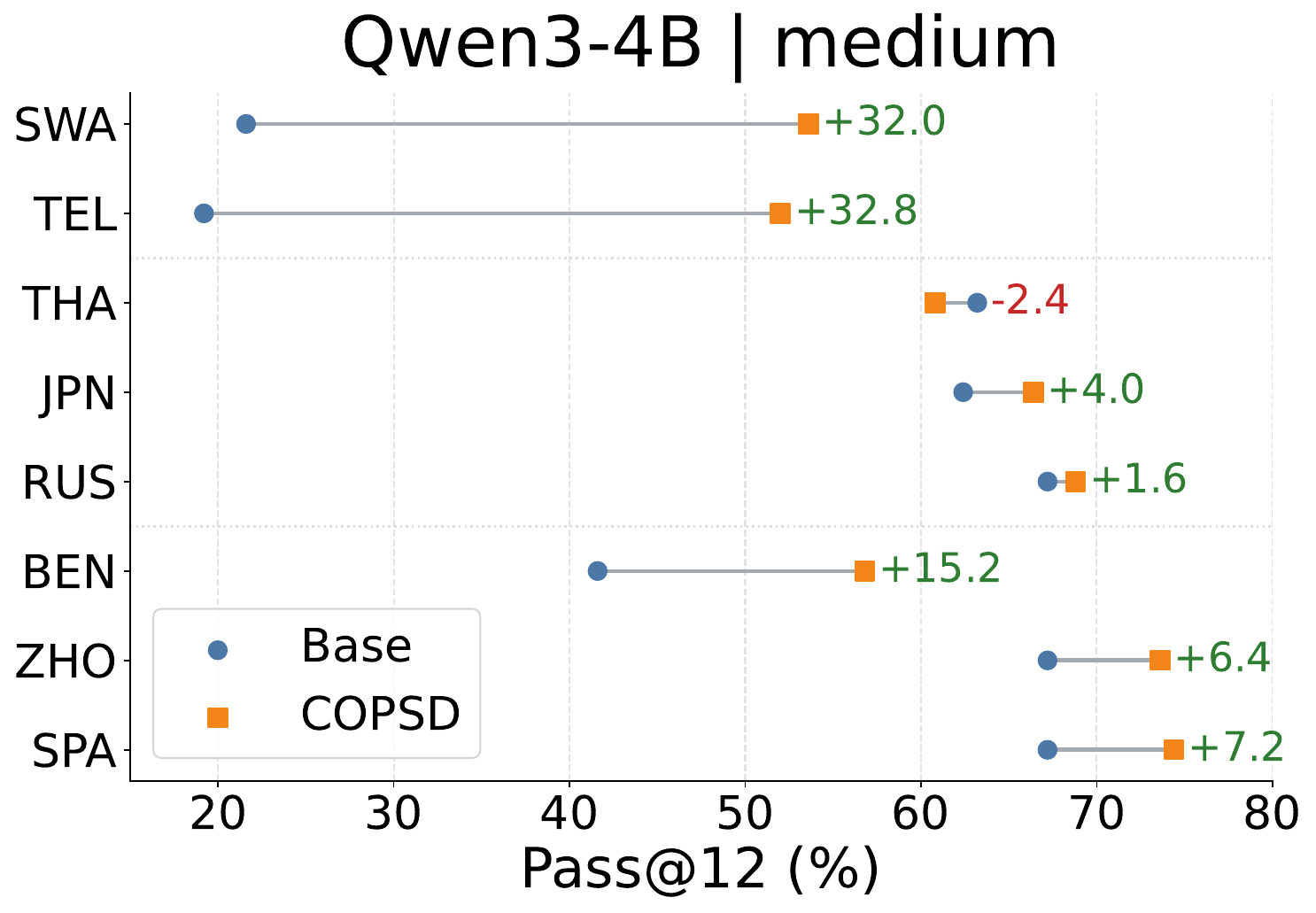}
    \includegraphics[width=0.32\linewidth]{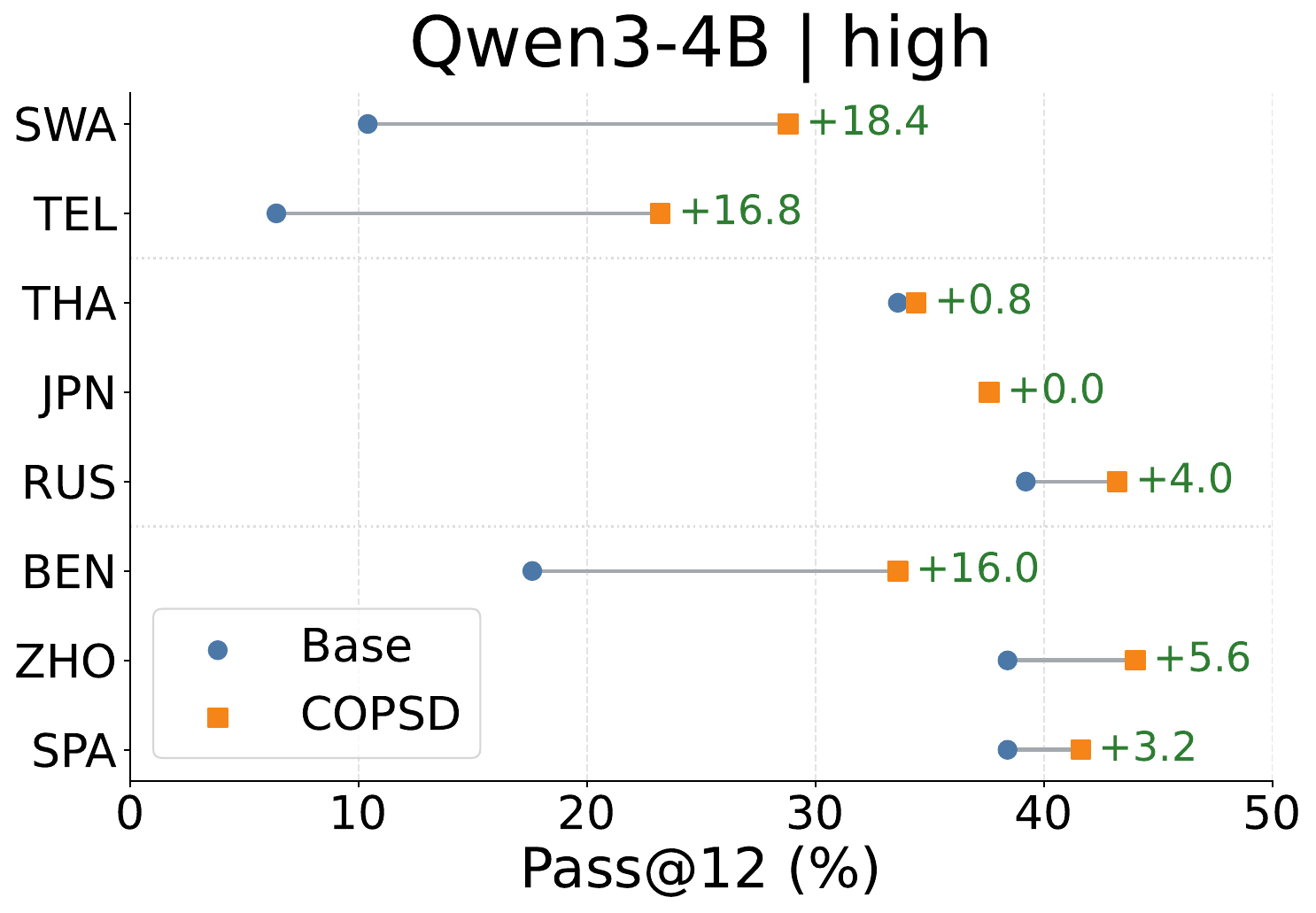}
    \caption{
    Pass@12 improvements of \copsd over the base model on PolyMath across low-, medium-, and high-difficulty settings under 8,192-token generation budget. 
    Each plot compares Base and \copsd for 8 languages spanning different resource levels. 
    \copsd yields consistent gains across resource levels and difficulty levels, with especially substantial improvements for lower-resource languages such as Swahili (SWA) and Telugu (TEL).
    }
    \label{fig:polymath_results}
\end{figure*}

\subsection{Generalization to Harder Benchmarks}\seclabel{harder_benchmark}

To examine whether the gains from \copsd transfer beyond AfriMGSM, we further evaluate on \textbf{PolyMath} \citep{wang2025polymathevaluatingmathematicalreasoning}, a more challenging multilingual mathematical reasoning benchmark with multiple difficulty levels.
We select 8 languages spanning different resource levels, including low-resource languages: Swahili (SWA) and Telugu (TEL), mid- or high-resource languages: Thai (THA), Russian (RUS), and Bengali (BEN), Japanese (JPN), Chinese (ZHO), and Spanish (SPA).
We train \texttt{Qwen3-4B} with \copsd on each language using the same training setup as in \secref{experiments}, and evaluate on the low-, medium-, and high-difficulty subsets of PolyMath.
For evaluation, we allow each model to generate up to 8,192 tokens.
The results are shown in Figure~\ref{fig:polymath_results}.

\paragraph{\copsd generalizes to harder reasoning settings, with the largest gains on lower-resource languages.}
Across difficulty levels, \copsd improves over the base model for almost all languages, indicating that the crosslingual reasoning behavior learned by \copsd is not limited to extremely low-resource languages in AfriMGSM.
Nevertheless, we observe that the gains are particularly large for lower-resource languages.
For example, on the medium-difficulty subset, \copsd improves Pass@12 by $+32.0$ points for Swahili and $+32.8$ points for Telugu, while also yielding a substantial gain of $+15.2$ points for Bengali.
On the high-difficulty subset, \copsd again produces large improvements for Swahili and Telugu, with gains of $+18.4$ and $+16.8$ points, respectively.
By contrast, improvements for higher-resource languages such as Japanese, Chinese, Russian, and Spanish are smaller, suggesting that these languages already benefit more from the base model's pretraining and post-training exposure, and therefore gain less from transferring English-accessible reasoning behavior.
Overall, these results suggest that \copsd is most effective when the model already possesses latent reasoning ability but struggles to express that ability through lower-resource language contexts.

\section{Conclusion}

We introduced \textbf{C}rosslingual \textbf{O}n-\textbf{P}olicy \textbf{S}elf-\textbf{D}istillation (\copsd), a framework for improving multilingual mathematical reasoning, with a particular focus on low-resource languages.
\copsd uses English question and reference solutions as privileged information: the student reasons from the low-resource problem alone, while the teacher, instantiated from the same model, provides dense token-level supervision on the student's own rollouts.
Across 17 African languages, \copsd consistently improves over base Qwen3 models and substantially outperforms GRPO-style outcome-based RL.
Further analyses show that \copsd improves format adherence, converges rapidly, strengthens test-time scaling, and generalizes to harder multilingual reasoning benchmarks.
These results suggest that low-resource reasoning failures are partly caused by difficulty accessing and expressing latent reasoning ability through underrepresented languages, and that \copsd offers an effective path toward more multilingual reasoning models.

\section*{Limitations}

While \copsd consistently improves over the baselines across languages, several limitations remain and point to directions for future work.

First, \copsd uses English as the high-resource privileged language and assumes access to English reference solutions. This may limit its applicability in settings where high-quality English supervision is unavailable or where another high-resource language would provide a better reasoning signal.

Second, our training questions are translated from English into the target low-resource languages. Although \copsd does not require translated reasoning traces, translation artifacts in the problem statements may still affect training quality and downstream performance.

Finally, \copsd relies on the same model as the privileged teacher. When the model has limited competence in a target language, the teacher distribution may still be imperfect, even with access to English context and reference solutions. This may cause the learning signal to saturate quickly or degrade with continued training, as observed for some languages and model sizes.

\section*{Ethical Considerations}

\paragraph{Use of AI Assistants.}
The authors used \texttt{ChatGPT} to assist with language polishing, including grammar, clarity, and coherence, as well as minor code implementation support.\footnote{\url{https://chatgpt.com/}}
All technical contributions, experimental design choices, and final decisions were made by the authors.

\section*{Acknowledgments}

This research was supported by the Munich Center for Machine Learning (MCML) and German Research Foundation (DFG, grant SCHU 2246/14-1).

\bibliography{custom}

\appendix

\section{Experimental Details}\seclabel{prompt_details}

\subsection{Language coverage}\seclabel{language_coverage}

\begin{table}[t]
\centering
\small
\begin{tabular}{llll}
\hline
\textbf{Code} & \textbf{Language} & \textbf{Family} & \textbf{Script} \\
\hline
AMH & Amharic & Semitic & Ethiopic \\
EWE & Ewe & Kwa & Latin \\
HAU & Hausa & Chadic & Latin \\
IBO & Igbo & Volta-Niger & Latin \\
KIN & Kinyarwanda & Bantu & Latin \\
LIN & Lingala & Bantu & Latin \\
LUG & Luganda & Bantu & Latin \\
ORM & Oromo & Cushitic & Latin \\
SNA & Shona & Bantu & Latin \\
SOT & Sesotho & Bantu & Latin \\
SWA & Swahili & Bantu & Latin \\
TWI & Twi & Kwa & Latin \\
VAI & Vai & Mande & Vai \\
WOL & Wolof & Senegambian & Latin \\
XHO & Xhosa & Bantu & Latin \\
YOR & Yoruba & Volta-Niger & Latin \\
ZUL & Zulu & Bantu & Latin \\
\hline
\end{tabular}
\caption{
Language coverage of our experiments.
We use the ISO 639-3 codes as language identifiers.
}
\label{tab:language_coverage}
\end{table}

\begin{figure*}
    \centering
    \setlength{\belowcaptionskip}{-0.4cm}
    \includegraphics[width=0.95\linewidth]{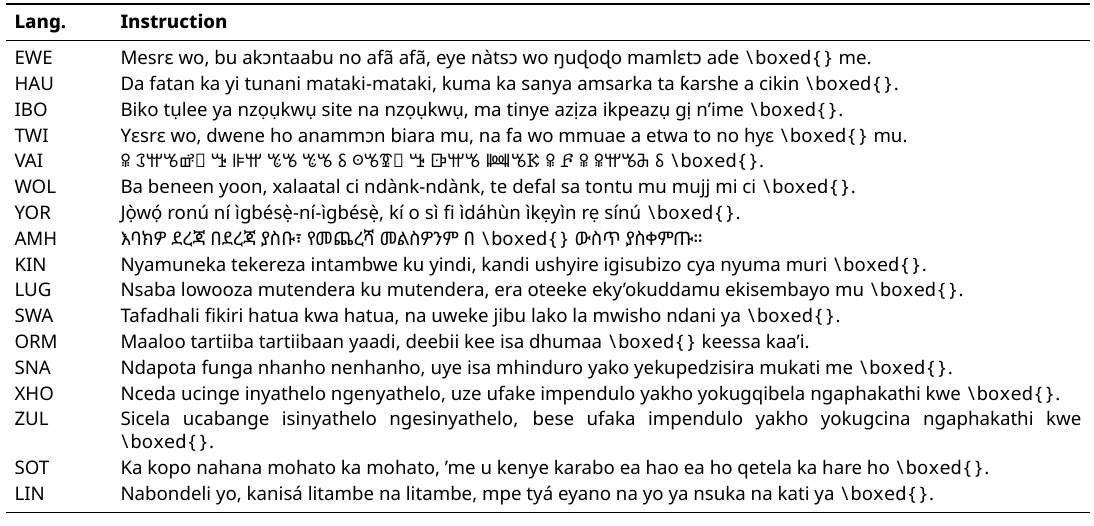}
    \caption{
    Instructions used for each target-language math problem to encourage step-by-step reasoning and require the final answer to be placed inside {\textbackslash boxed\{\}}.
    }
    \label{fig:language_instruction}
\end{figure*}

\begin{figure*}
    \centering
    \setlength{\belowcaptionskip}{-0.4cm}
    \includegraphics[width=0.95\linewidth]{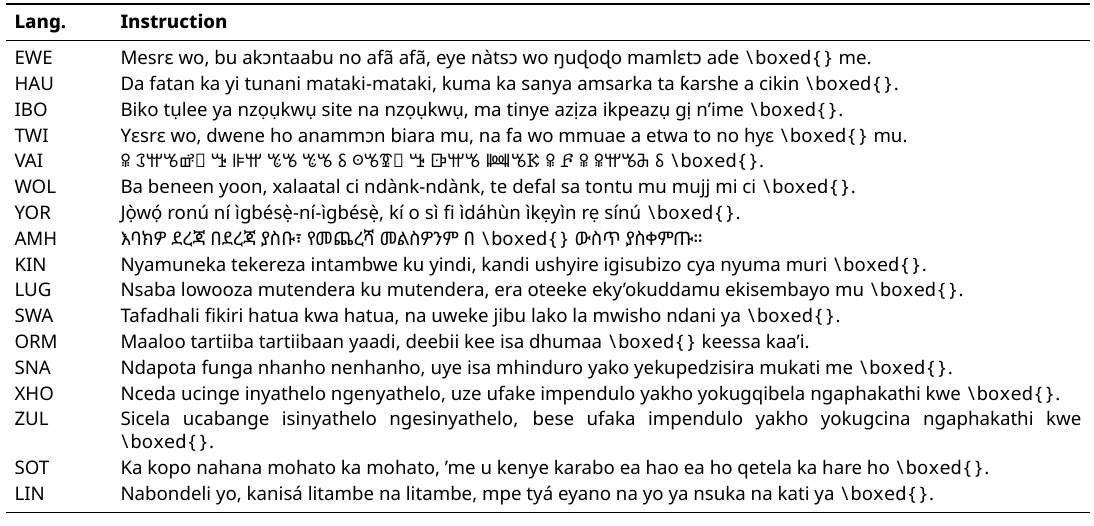}
    \caption{
   Prompt-hacking prefixes inserted immediately after \texttt{<think>} to steer the model's explicit reasoning trace in the target low-resource language.
    }
    \label{fig:language_hacking}
\end{figure*}

Our experiments cover all 17 African languages included in AfriMGSM \citep{adelani-etal-2025-irokobench}.
For each language, we use its ISO 639-3 code as the language identifier throughout training, evaluation, and result reporting.
The covered languages span multiple language families and writing systems.
This setting allows us to evaluate whether \copsd can improve reasoning not only across different languages, but also across substantially different orthographic and linguistic conditions.
Table~\ref{tab:language_coverage} lists the languages, ISO 639-3 codes, and target-language names used in our experiments.

\subsection{Language Control}\seclabel{language_control}

Following \citet{qi-etal-2025-models,zhao-etal-2026-comprehensive}, we use complementary prompting strategies to encourage the model to produce its explicit reasoning trace in the target low-resource language.

\paragraph{Language-Specific Instruction}
For each input, we prepend a language-specific instruction that specifies the desired reasoning language and asks the model to solve the problem step by step.
The language-specific instructions are shown in Figure~\ref{fig:language_instruction}.

\paragraph{Language-Specific Prompt Hacking}
Explicit language instructions alone do not always guarantee language-consistent reasoning: LLMs may still switch to English or mix languages in their reasoning traces, as observed in prior work on multilingual reasoning and language mixing~\citep{wang-etal-2025-language-mixing,qi-etal-2025-models,zhao-etal-2026-comprehensive}.
This behavior is undesirable in our setting because it makes it difficult to compare reasoning behavior across languages and may obscure whether improvements come from better low-resource reasoning or from implicit English reasoning.
To reduce such language drift, we adopt a \emph{prompt-hacking} strategy~\citep{schulhoff-etal-2023-ignore,benjamin2024systematicallyanalyzingpromptinjection}.
Specifically, following \citet{qi-etal-2025-models,zhao-etal-2026-comprehensive}, we insert a target-language prefix immediately after the opening \texttt{<think>} tag.
For example, for Swahili, we insert ``\textit{Kwa ombi, nitaanza kufikiria kwa Kiswahili.}'' immediately after \texttt{<think>}, which means ``As requested, I will begin thinking in Swahili.''
This prefix anchors the beginning of the reasoning trace in the target language and helps steer the model to continue reasoning in that language until the closing \texttt{</think>} tag.
The full set of language-specific prefixes used in our experiments is listed in Figure~\ref{fig:language_hacking}.
\section{Complete Results}\seclabel{complete}

\begin{figure*}[t]
    \centering
    \includegraphics[width=0.32\linewidth]{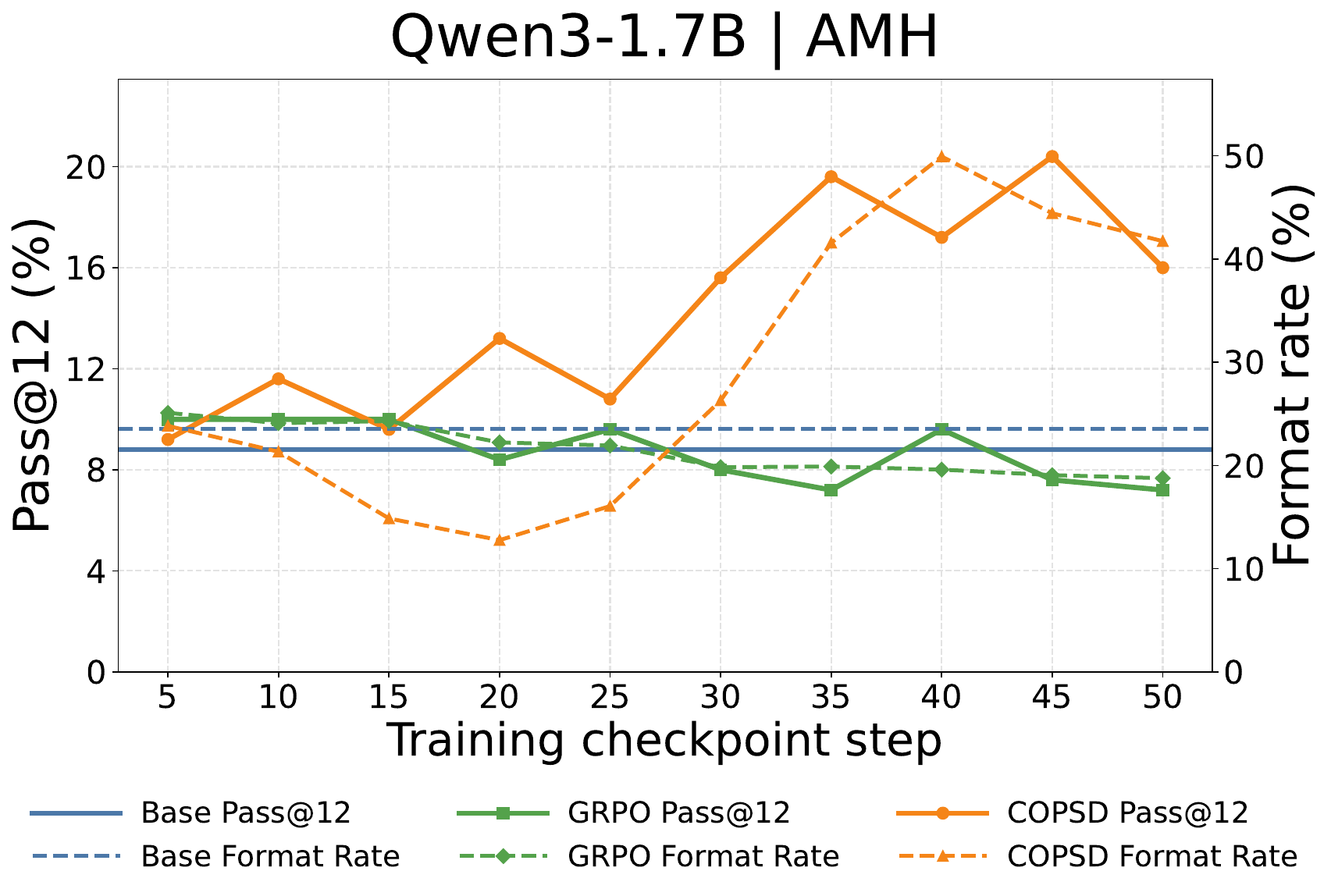}
    \includegraphics[width=0.32\linewidth]{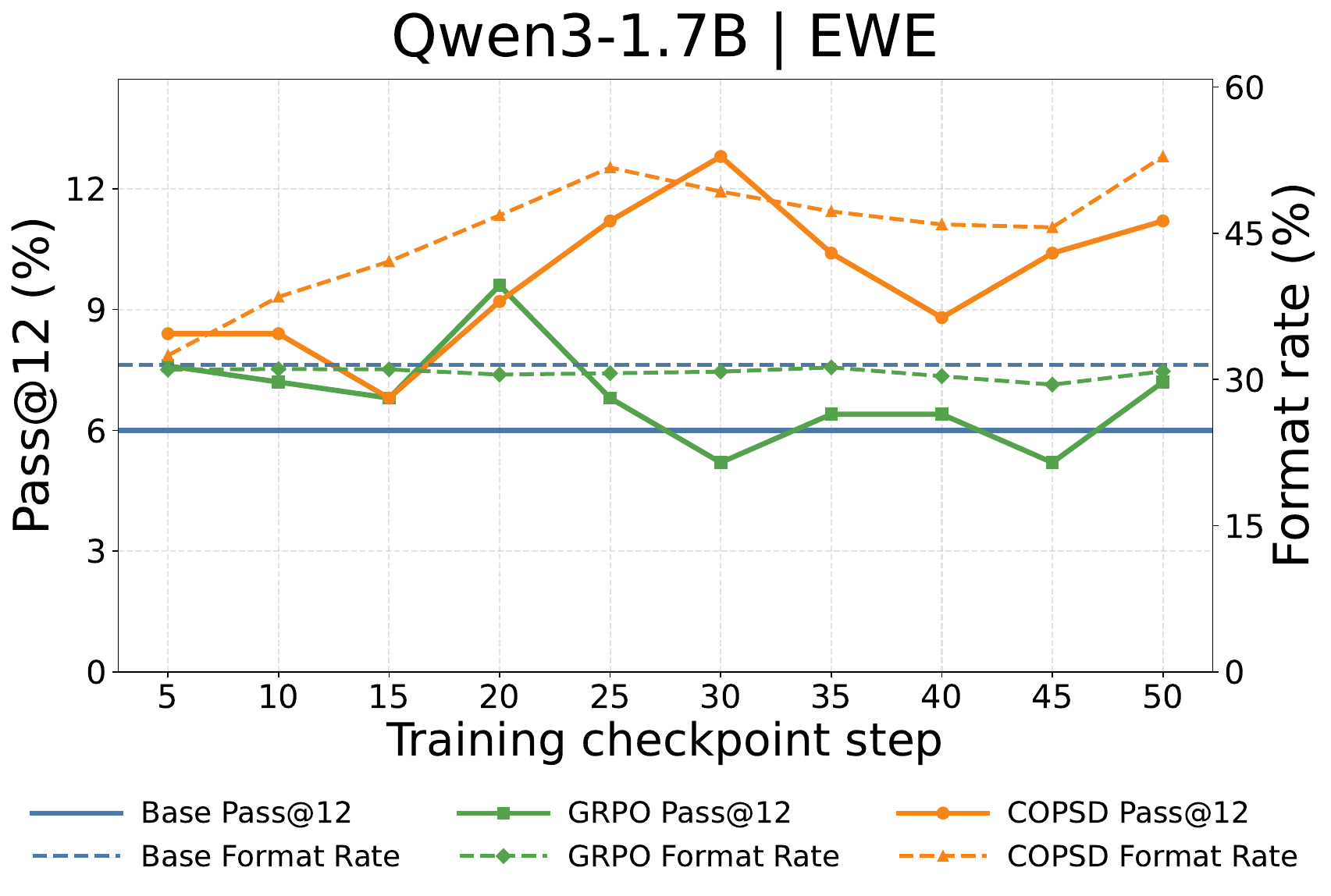}
    \includegraphics[width=0.32\linewidth]{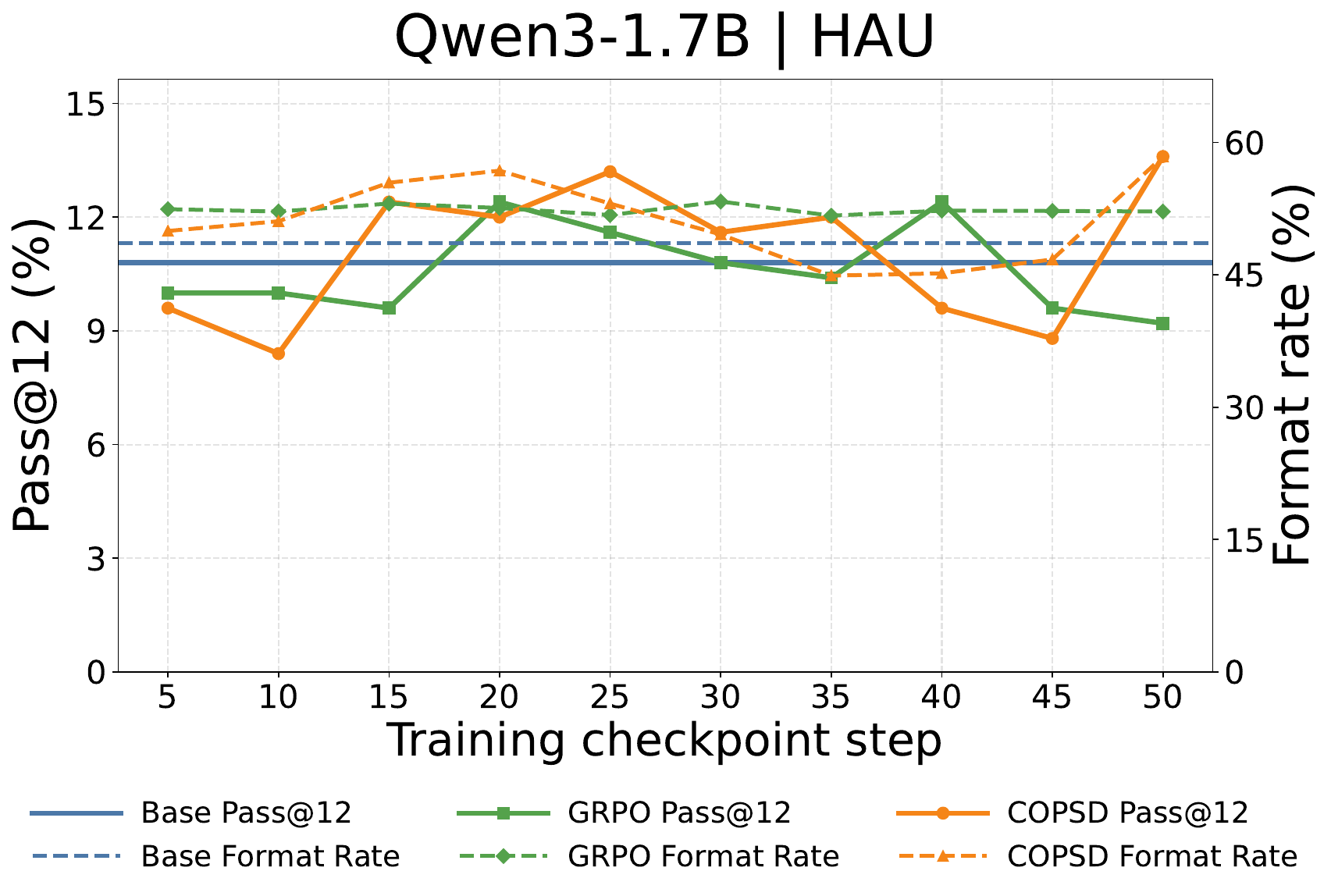}

    \includegraphics[width=0.32\linewidth]{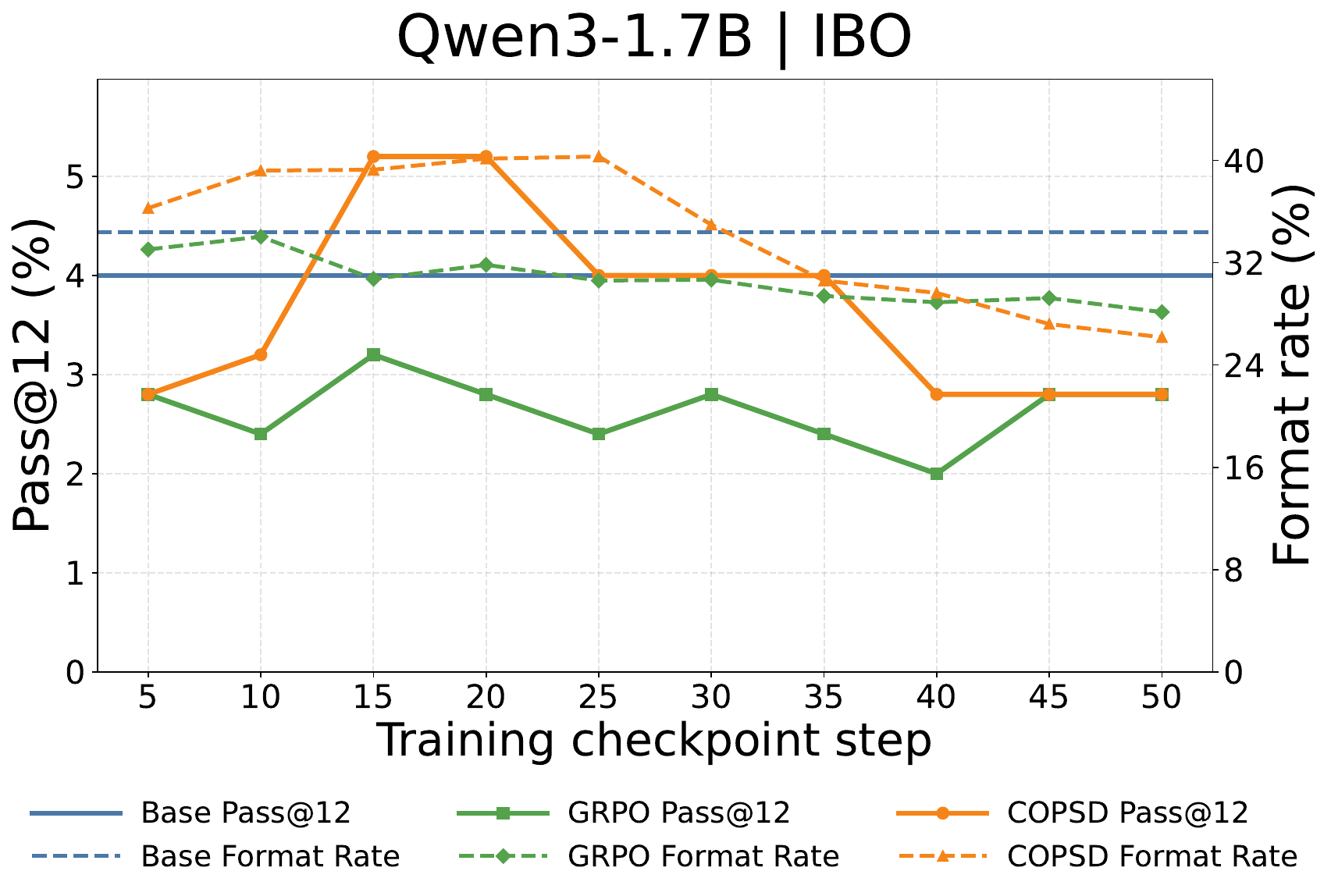}
    \includegraphics[width=0.32\linewidth]{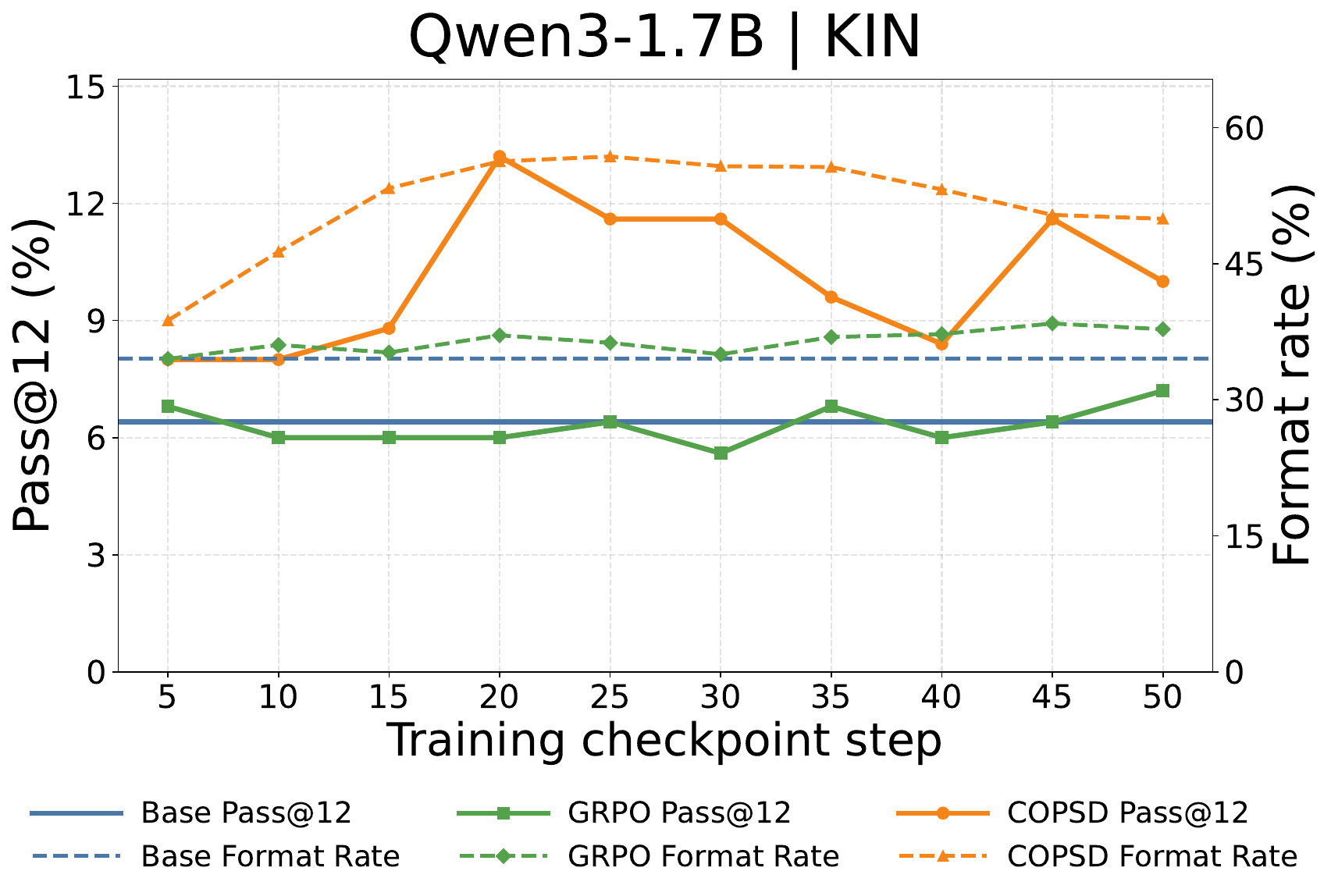}
    \includegraphics[width=0.32\linewidth]{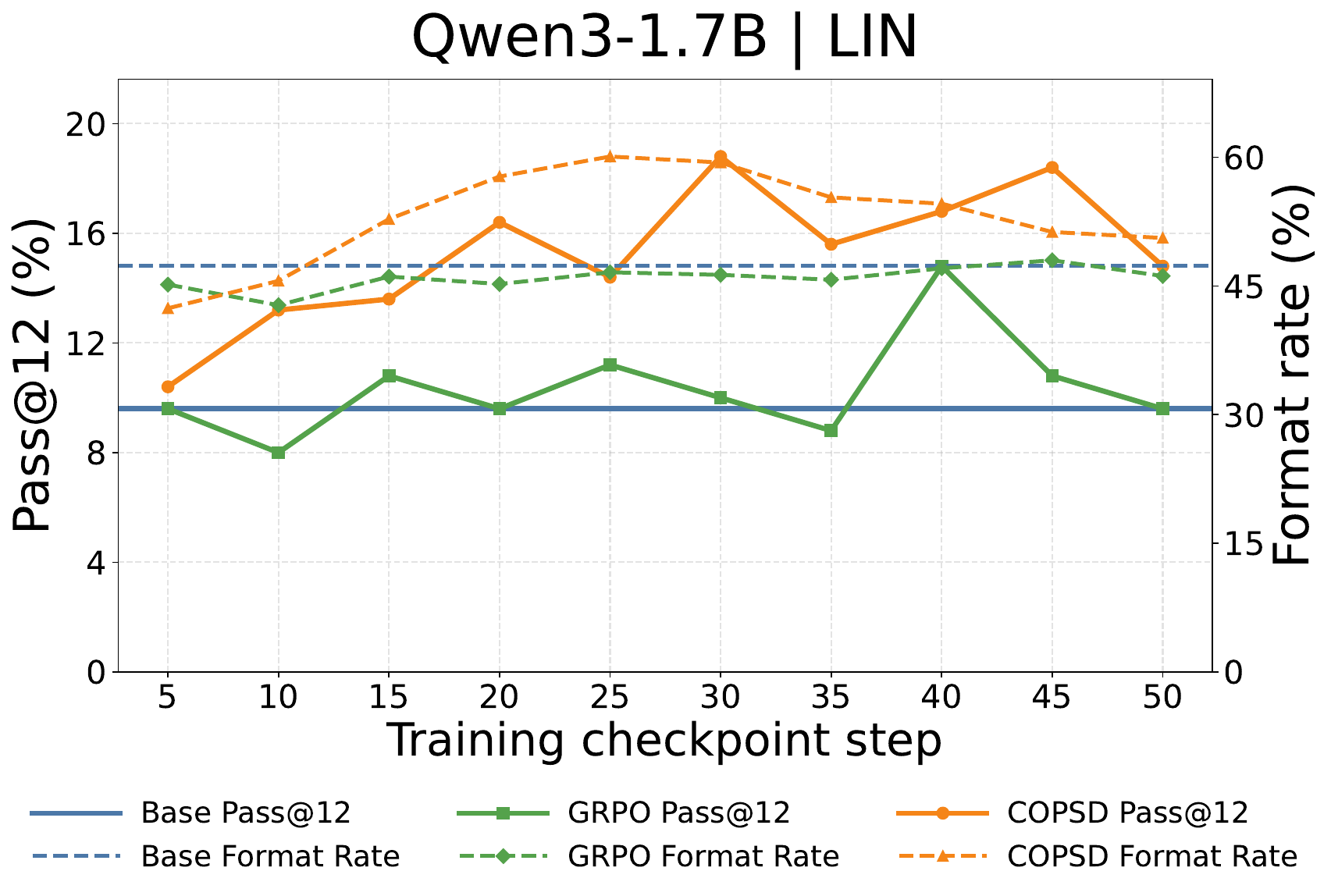}

    \includegraphics[width=0.32\linewidth]{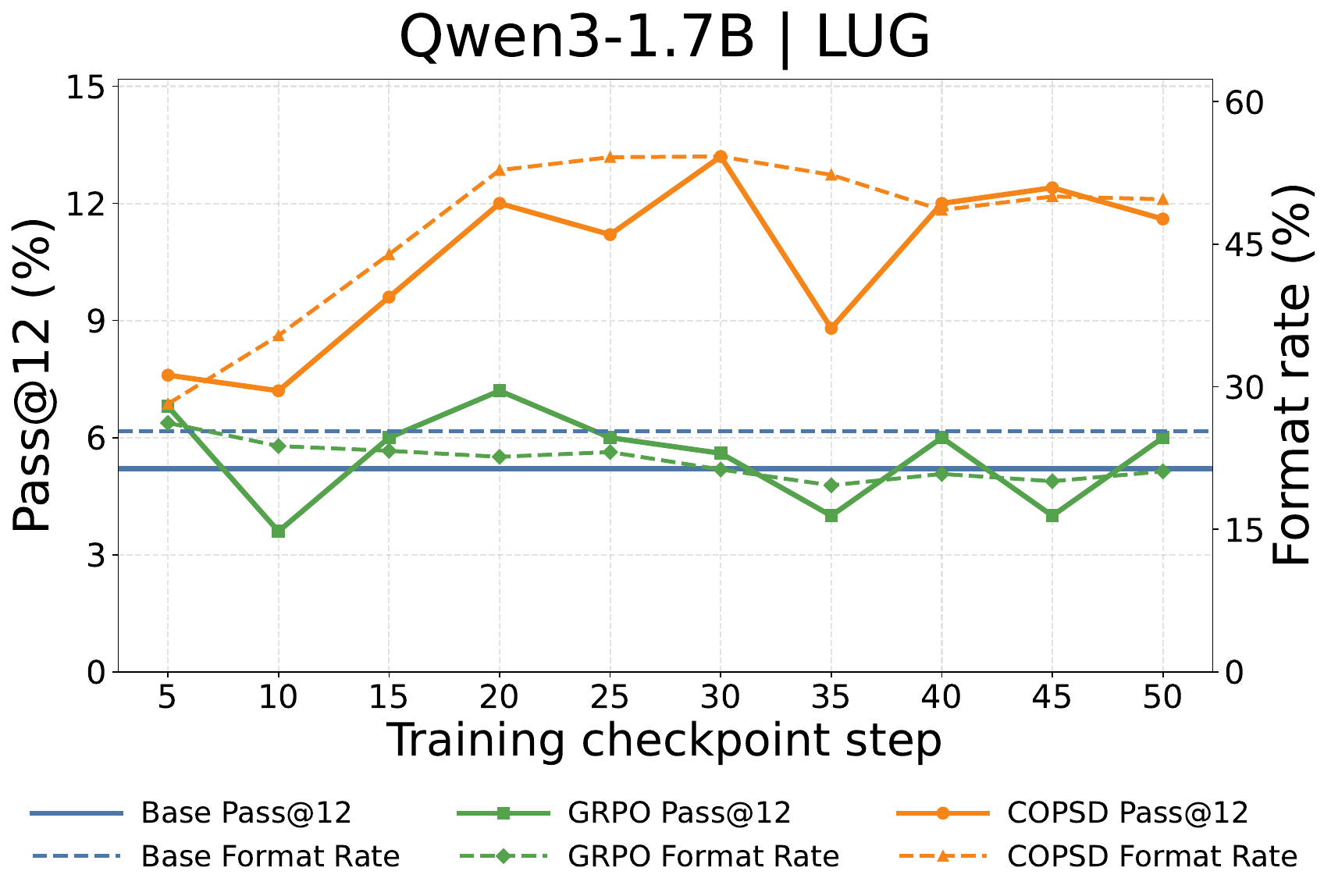}
    \includegraphics[width=0.32\linewidth]{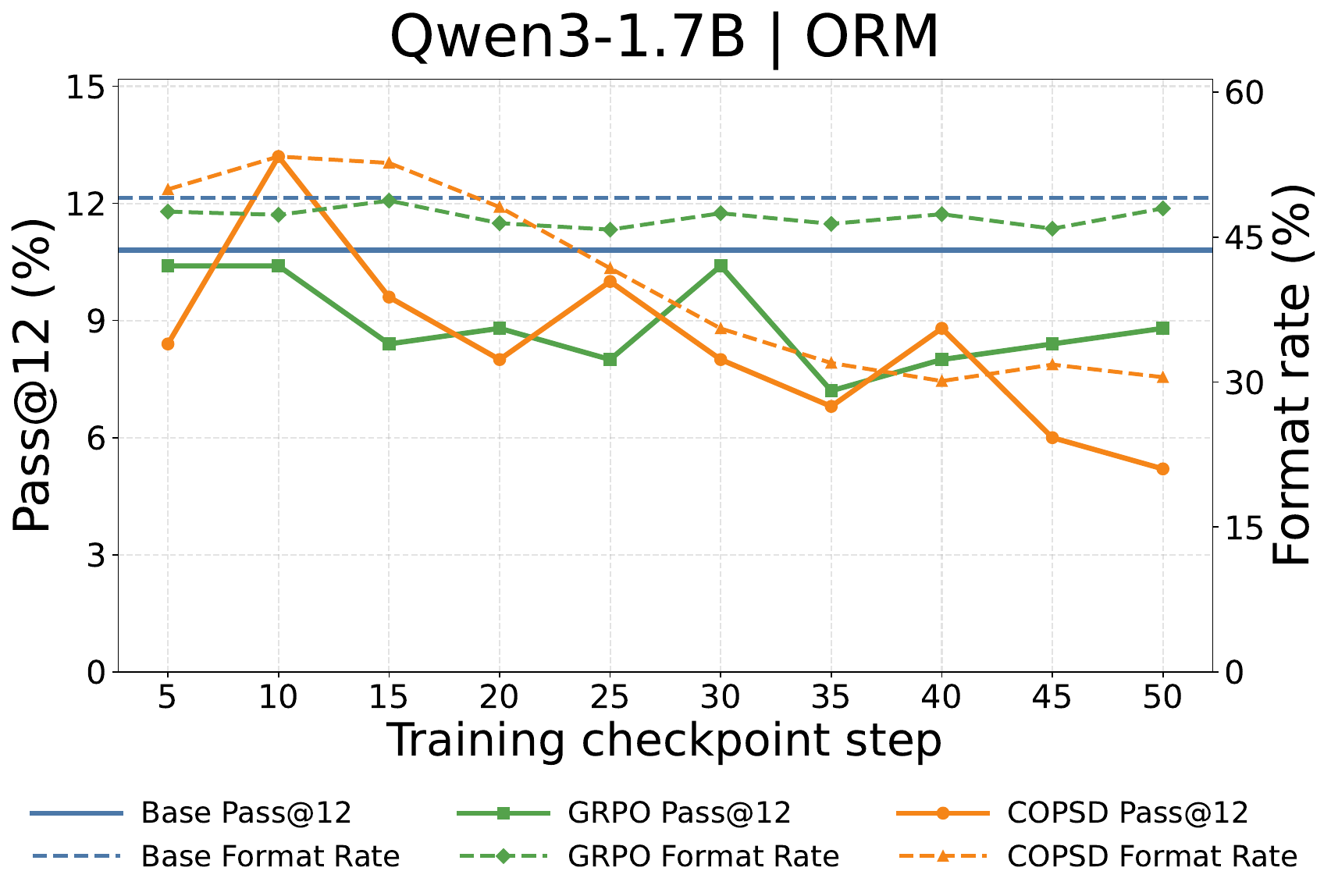}
    \includegraphics[width=0.32\linewidth]{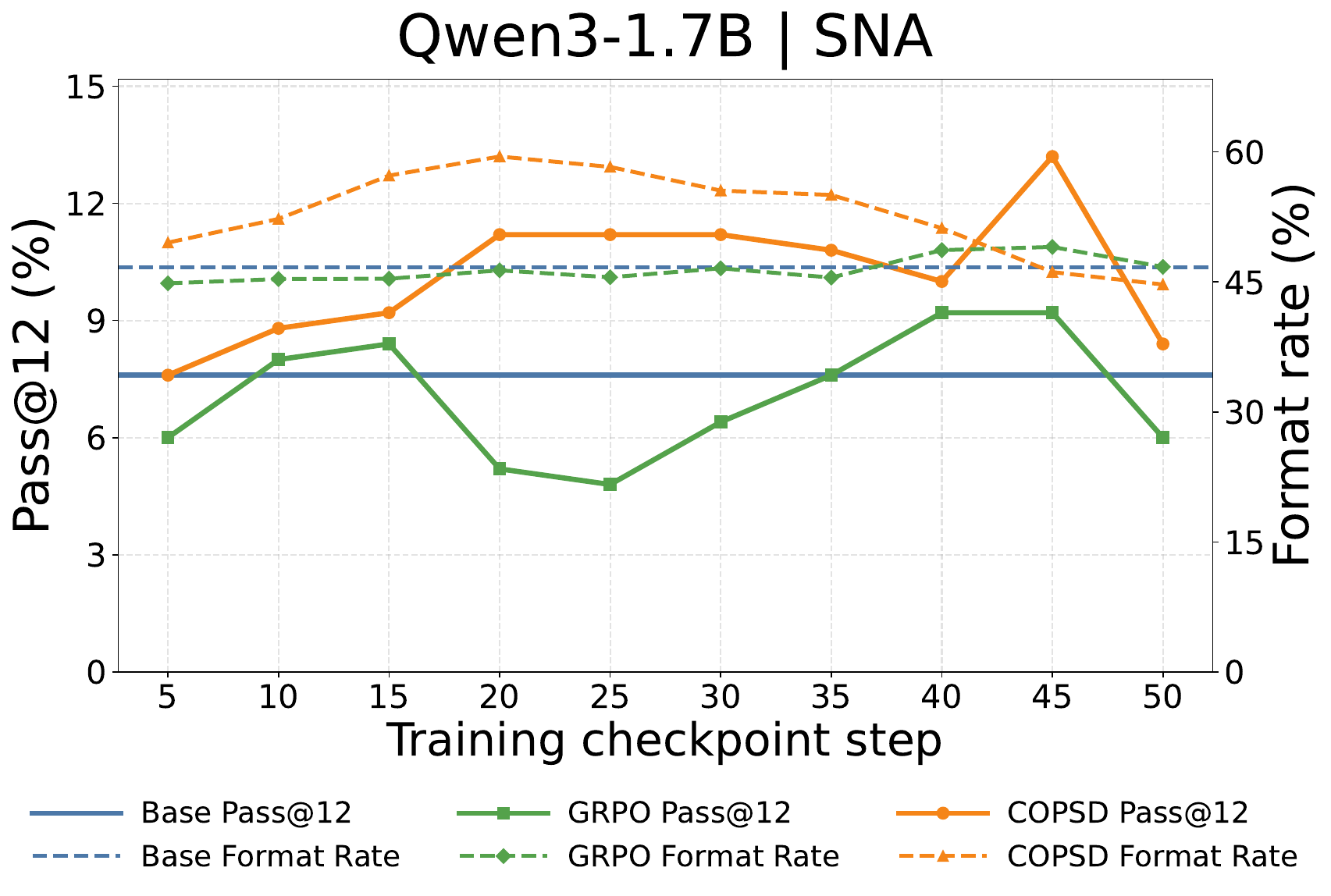}

    \includegraphics[width=0.32\linewidth]{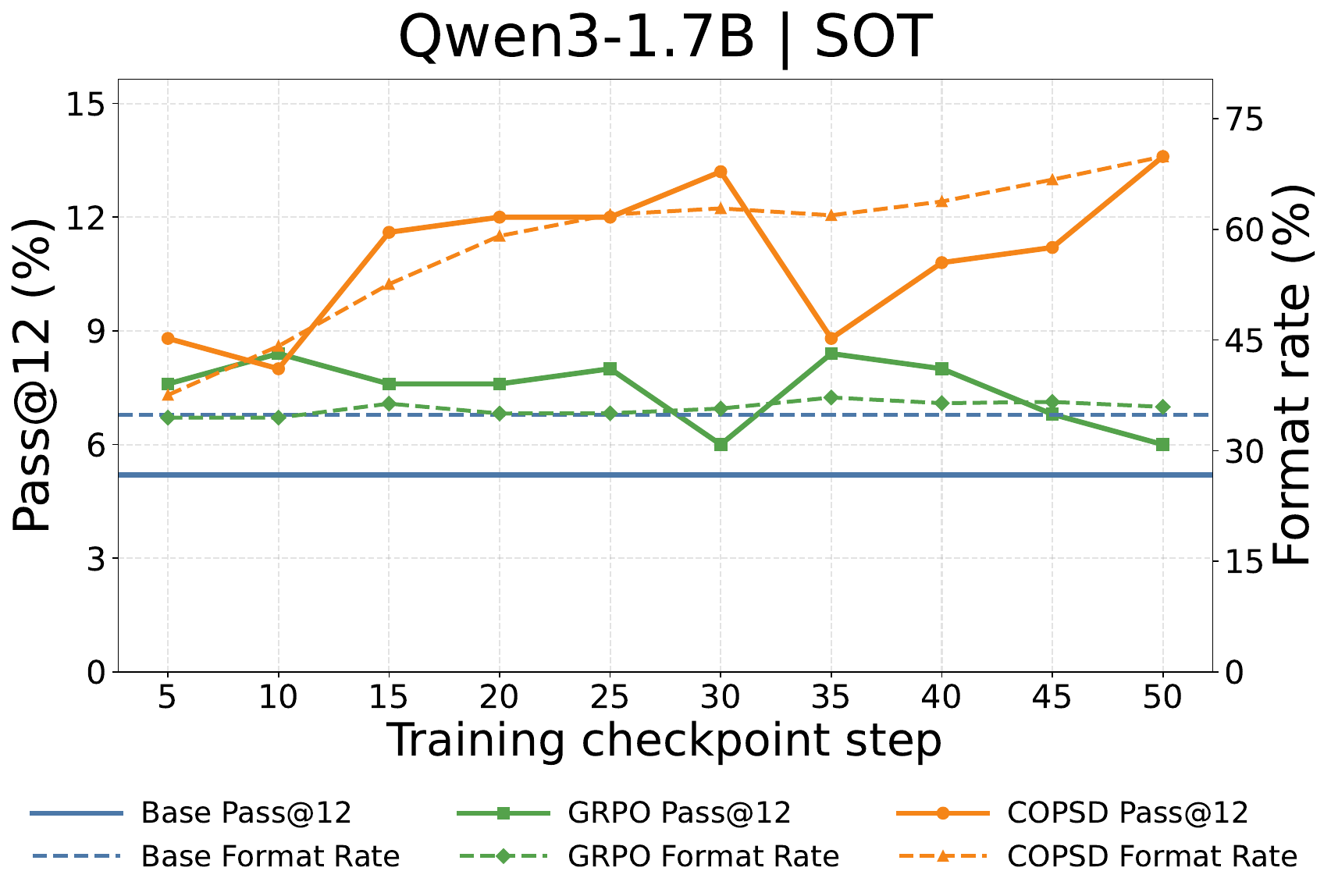}
    \includegraphics[width=0.32\linewidth]{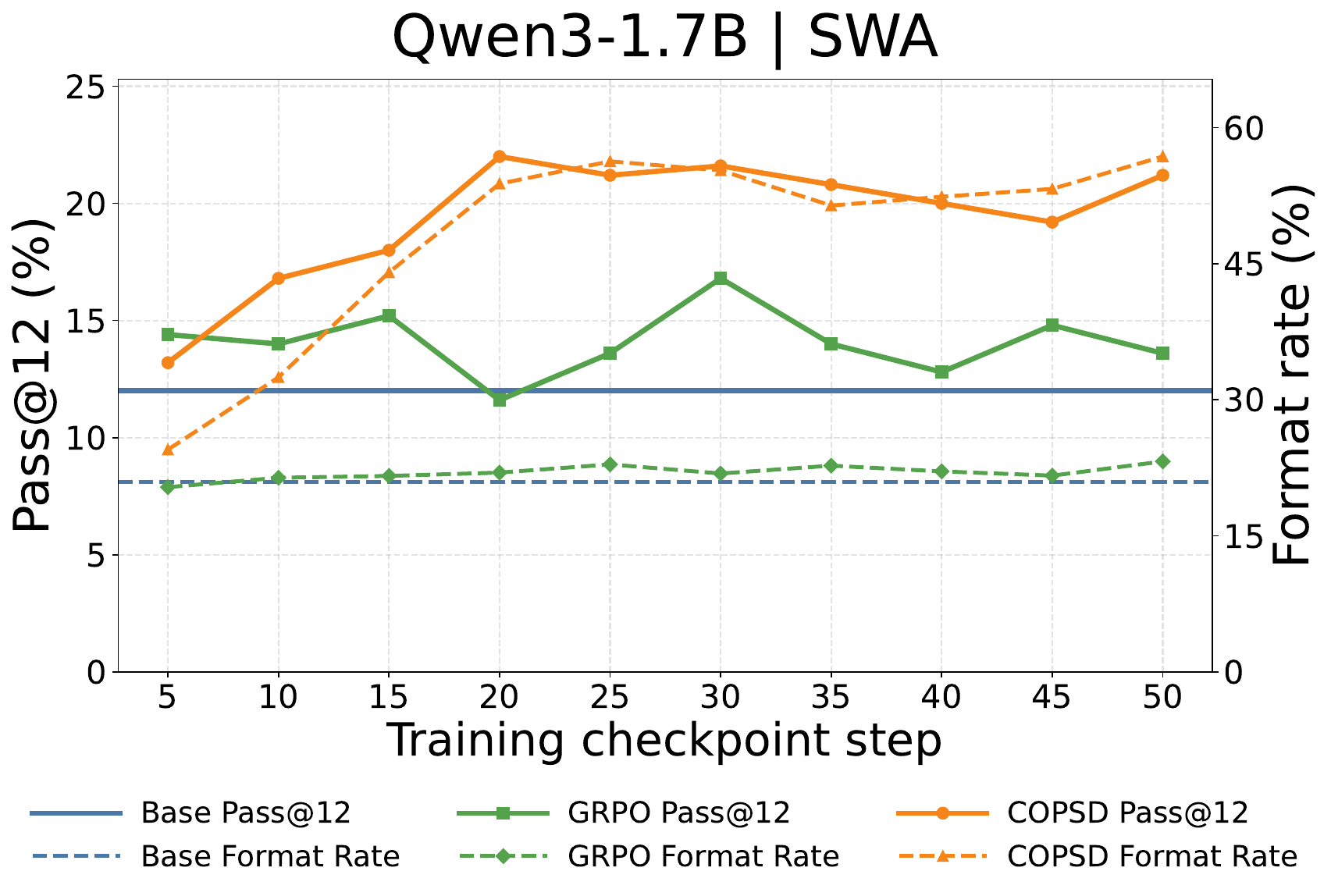}
    \includegraphics[width=0.32\linewidth]{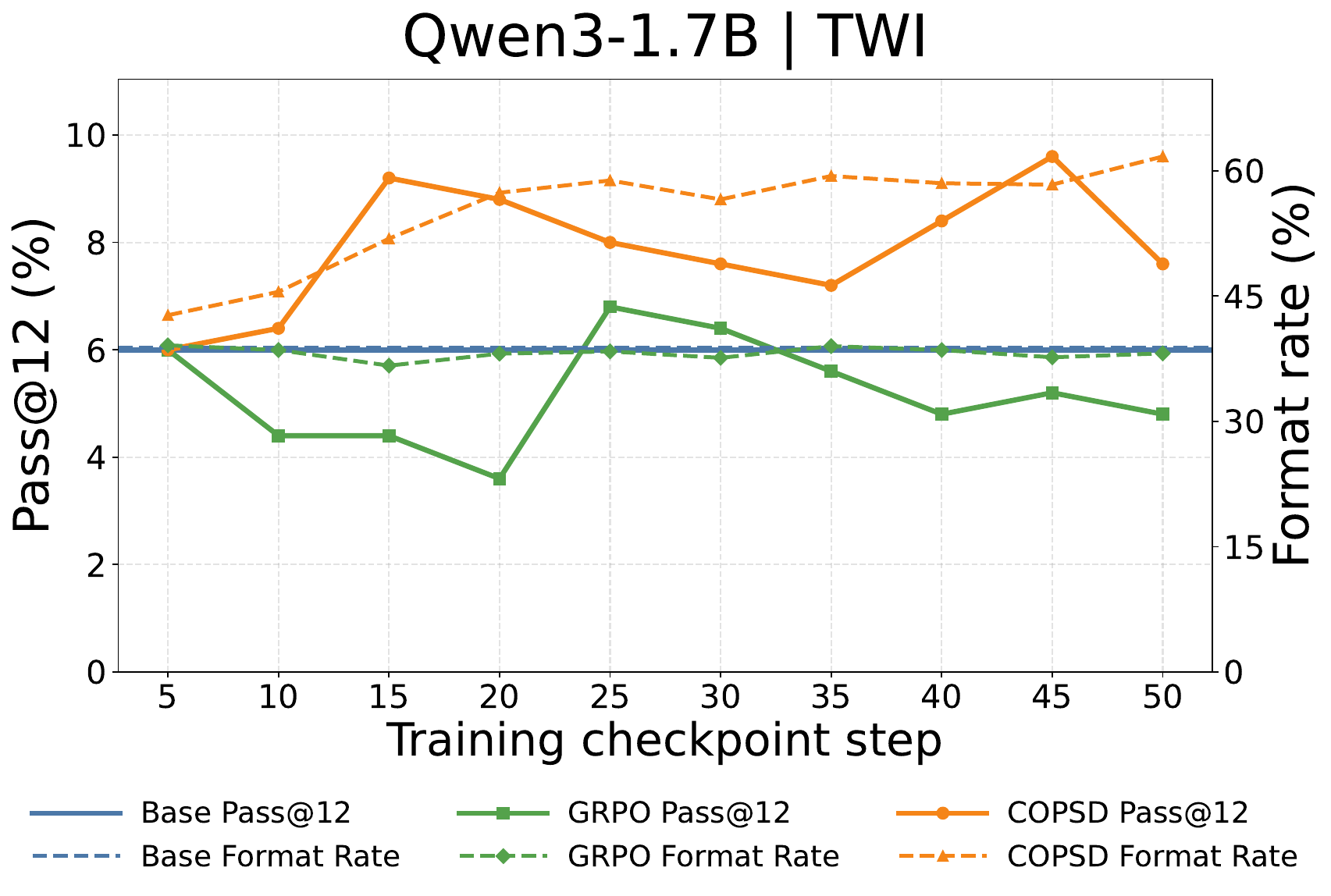}

    \includegraphics[width=0.32\linewidth]{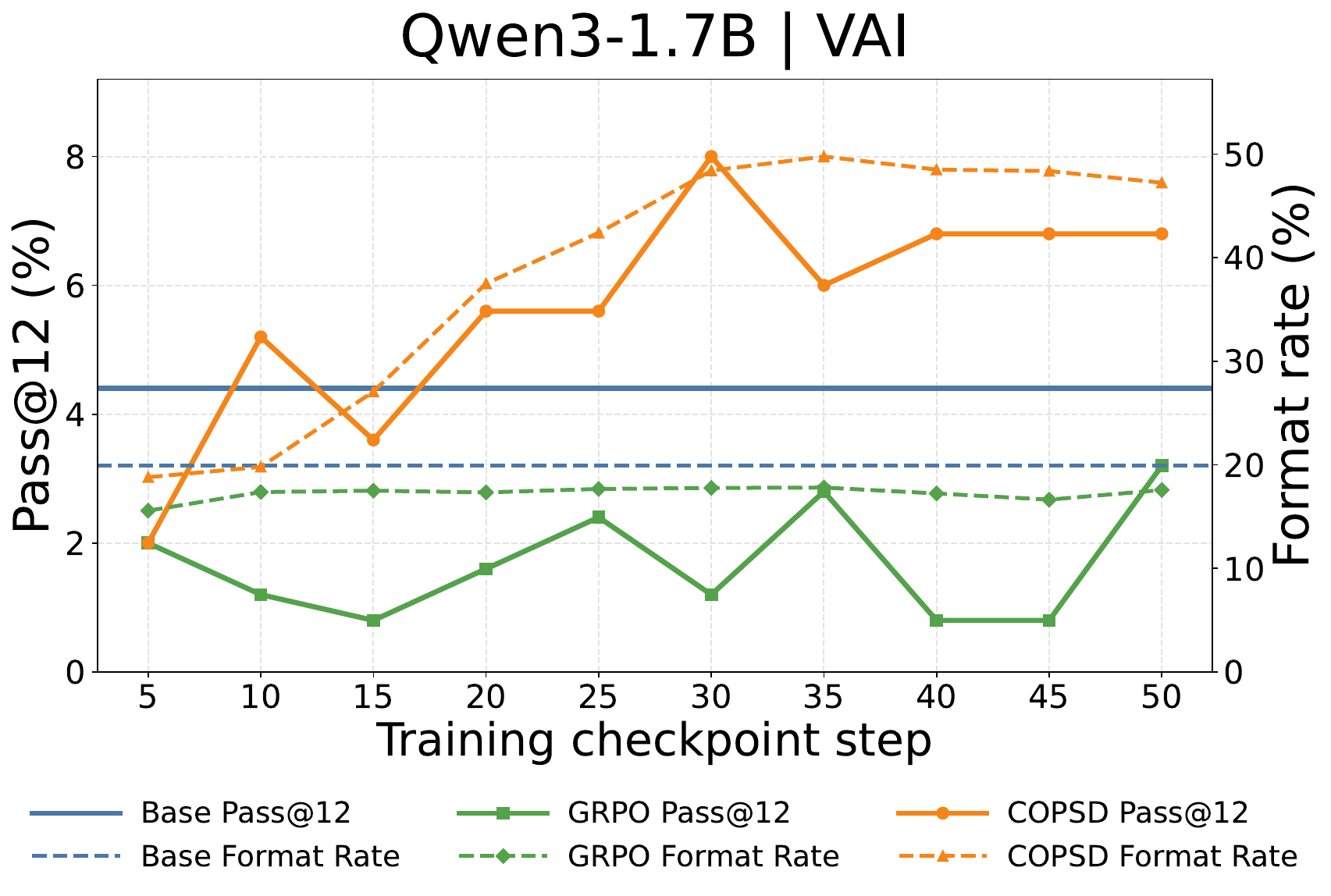}
    \includegraphics[width=0.32\linewidth]{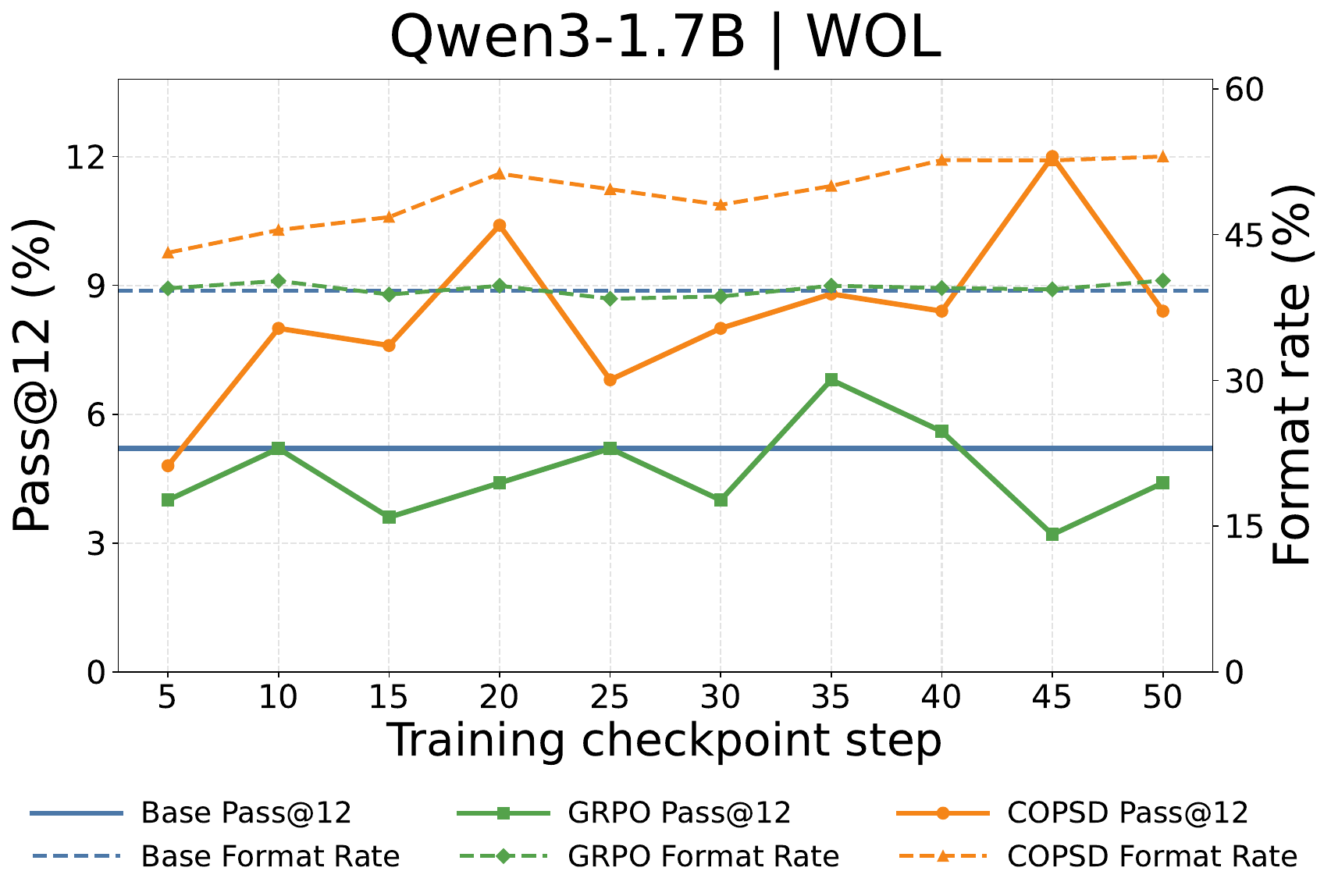}
    \includegraphics[width=0.32\linewidth]{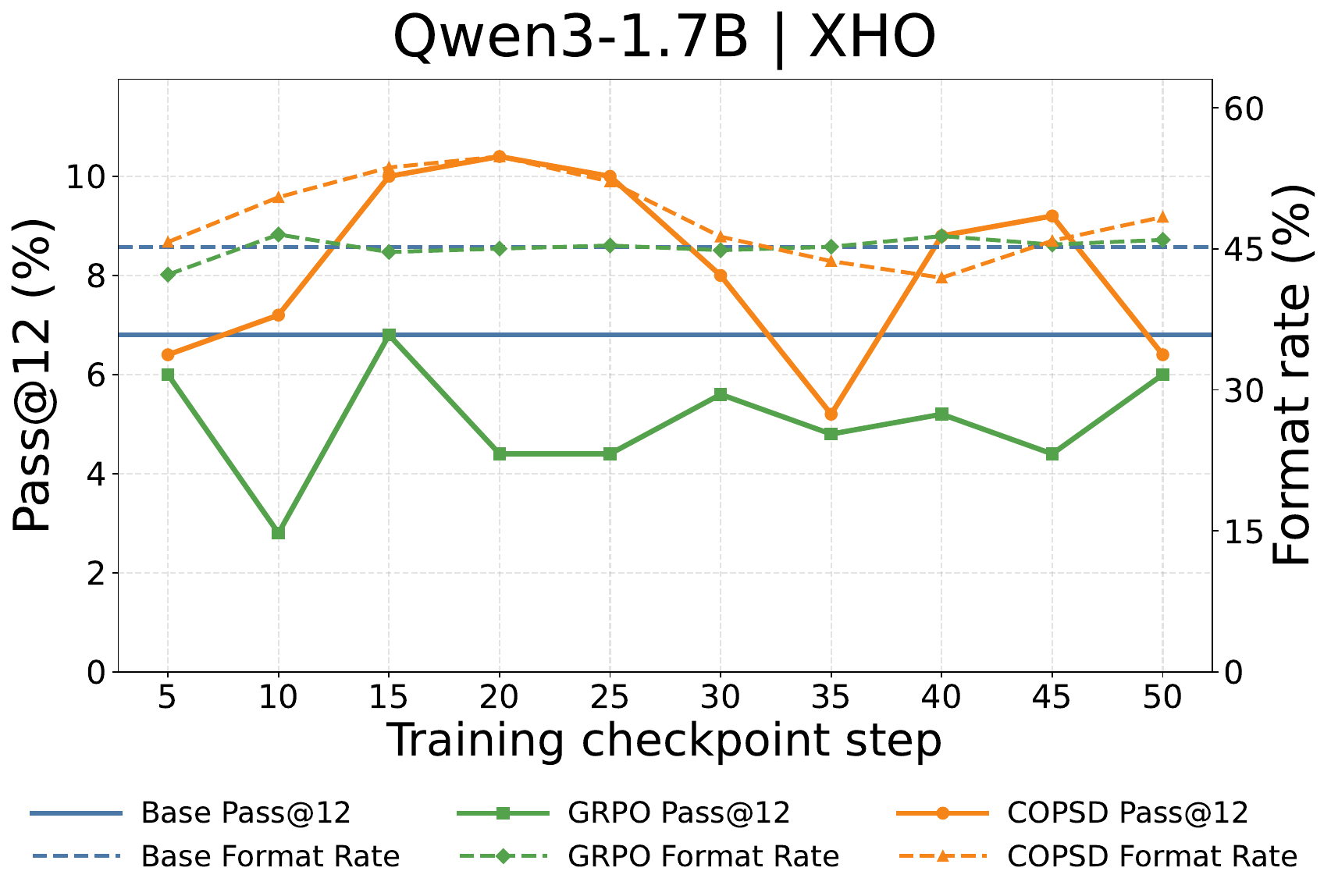}

    \includegraphics[width=0.32\linewidth]{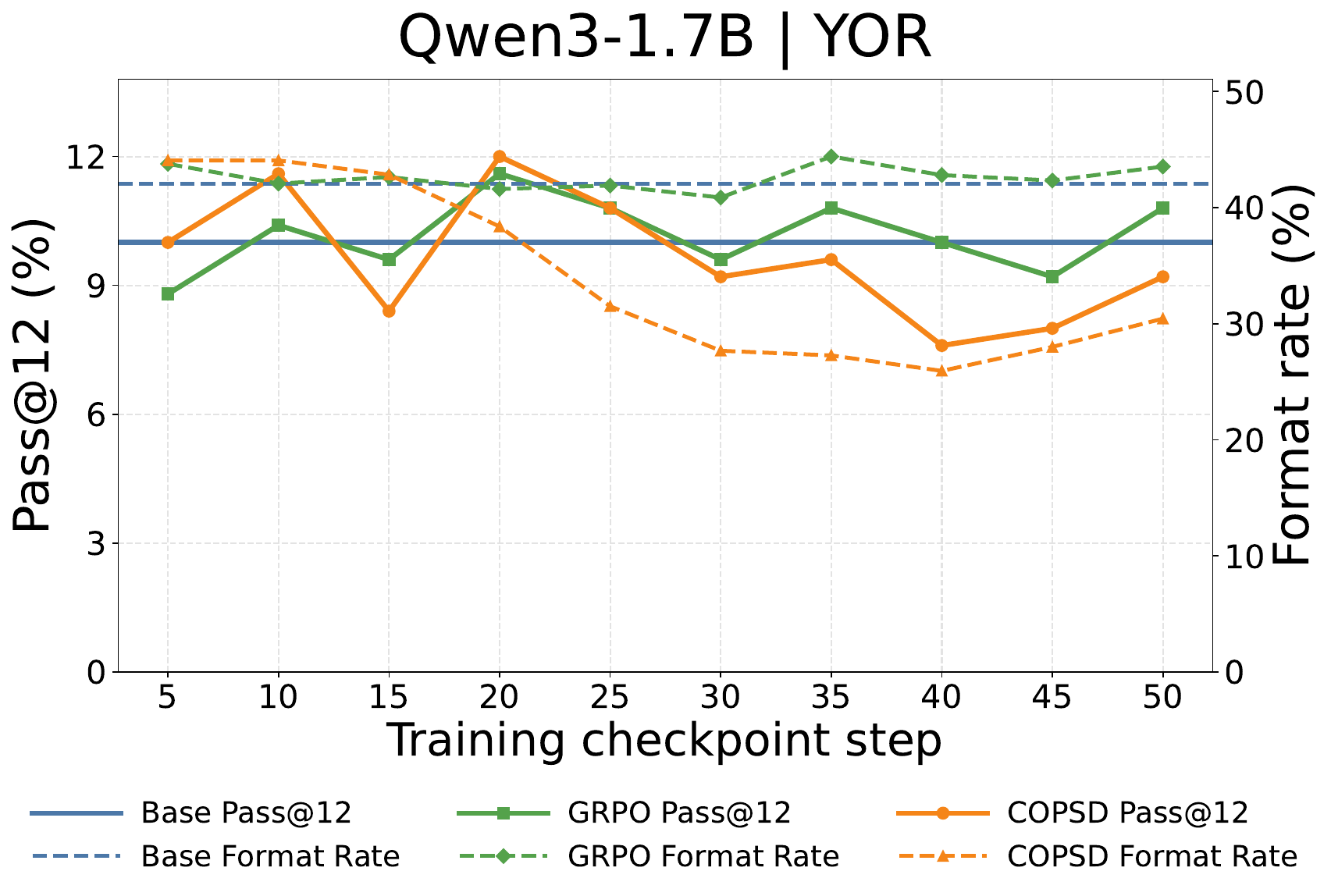}
    \includegraphics[width=0.32\linewidth]{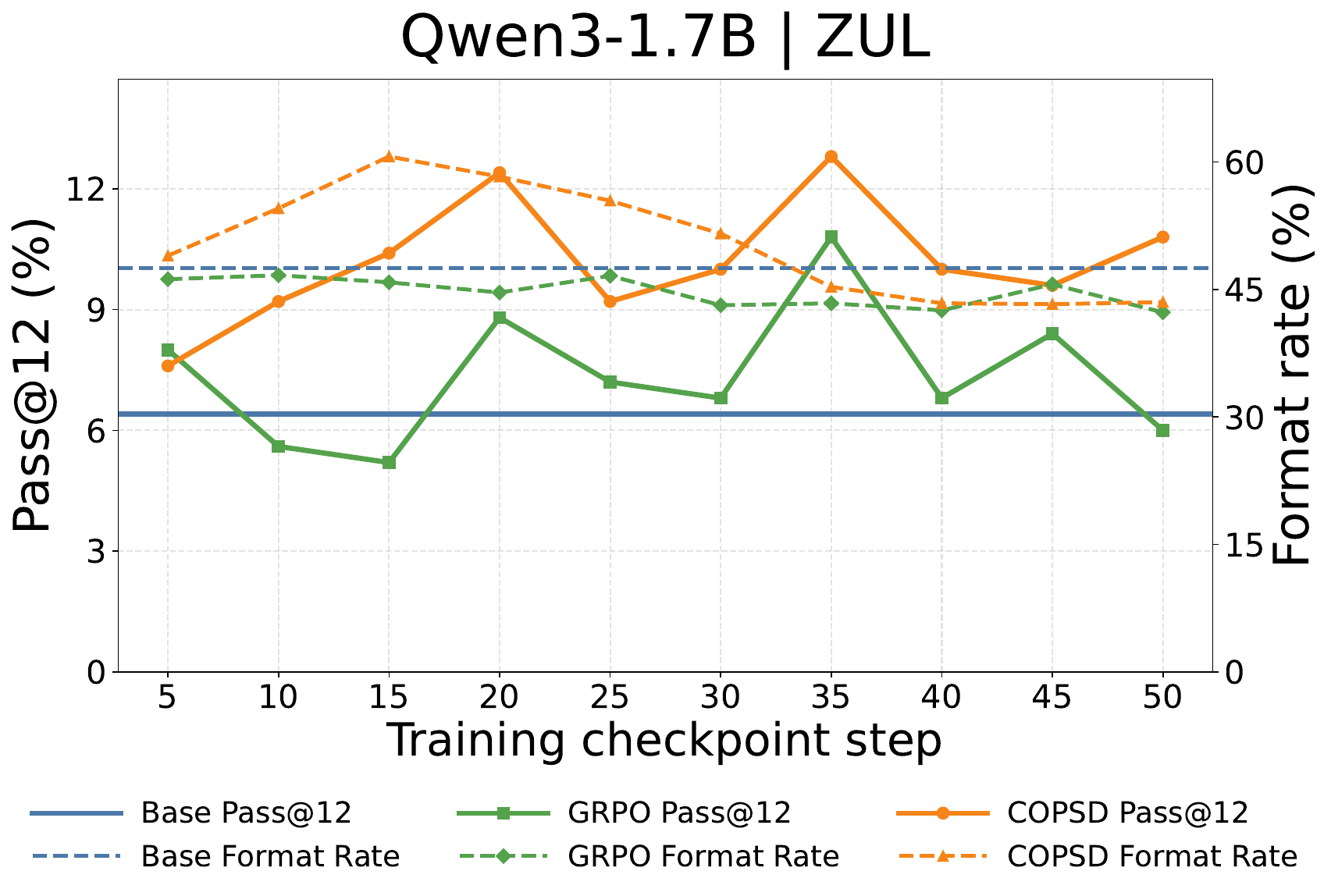}

    \caption{Per-language training dynamics for \texttt{Qwen3-1.7B} across all African languages under a 1024-token generation budget. Solid lines show Pass@12 and dashed lines show format rate.}
    \label{fig:training_dynamics_qwen3_1.7b_all_languages}
\end{figure*}

\begin{figure*}[t]
    \centering
    \includegraphics[width=0.32\linewidth]{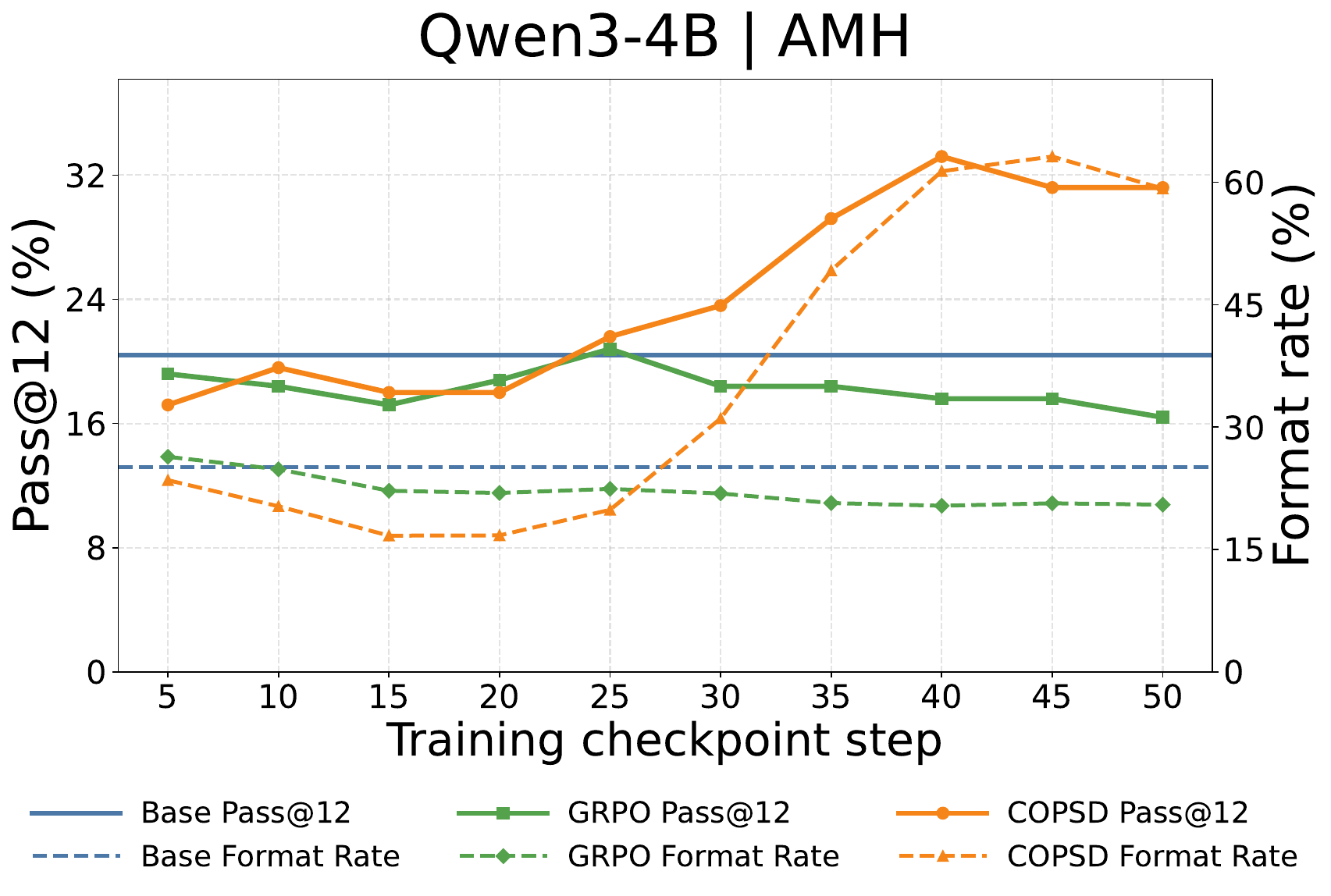}
    \includegraphics[width=0.32\linewidth]{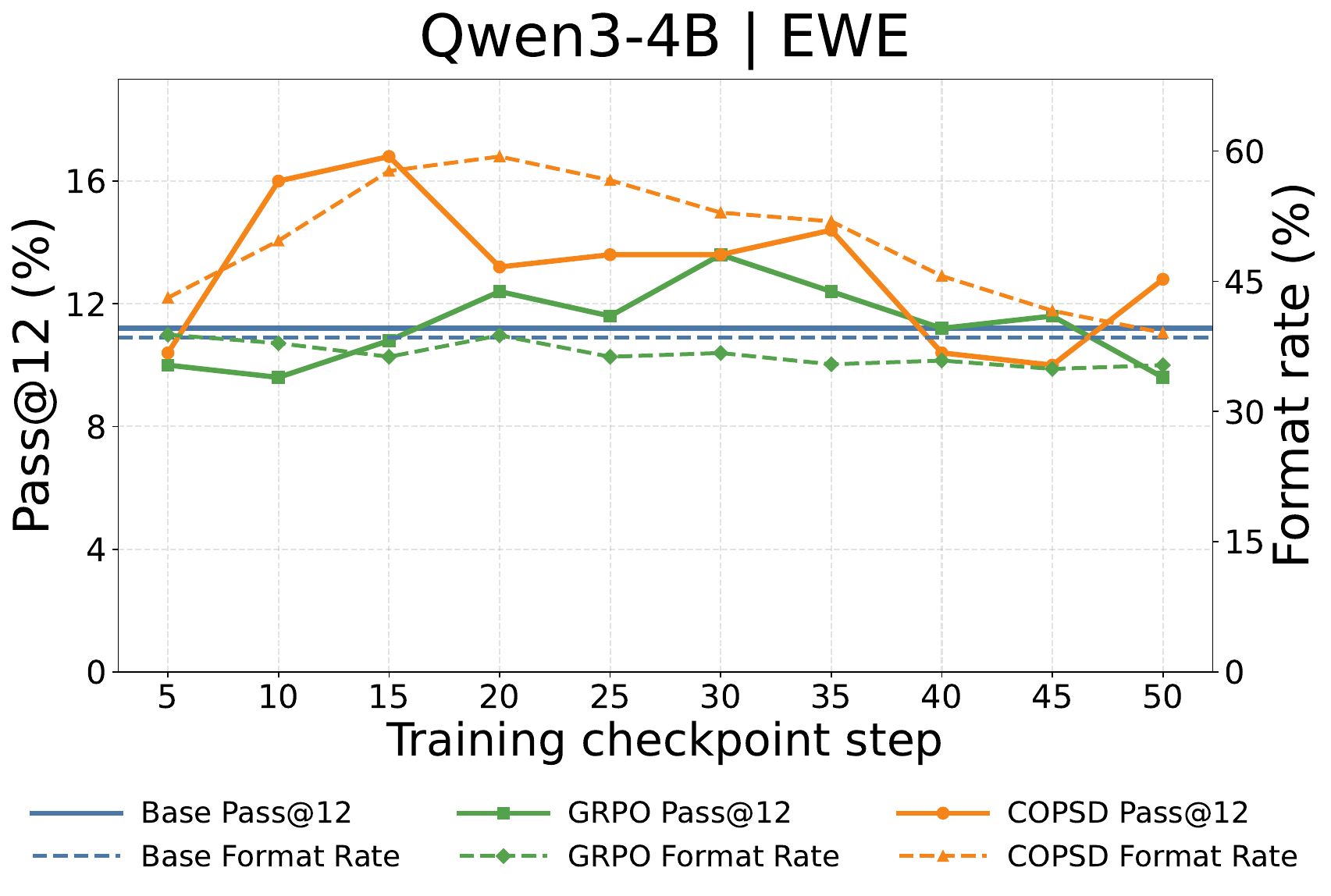}
    \includegraphics[width=0.32\linewidth]{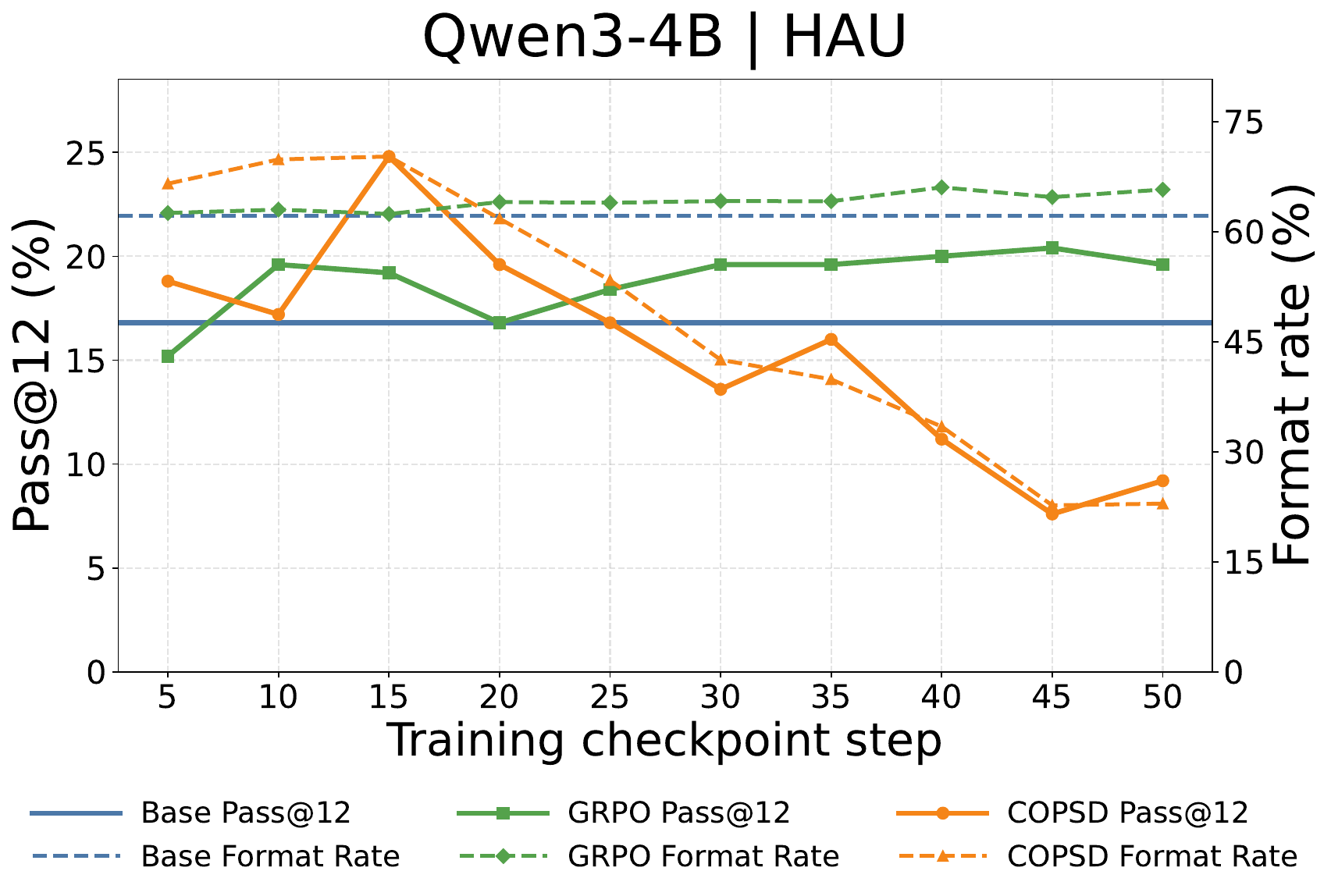}

    \includegraphics[width=0.32\linewidth]{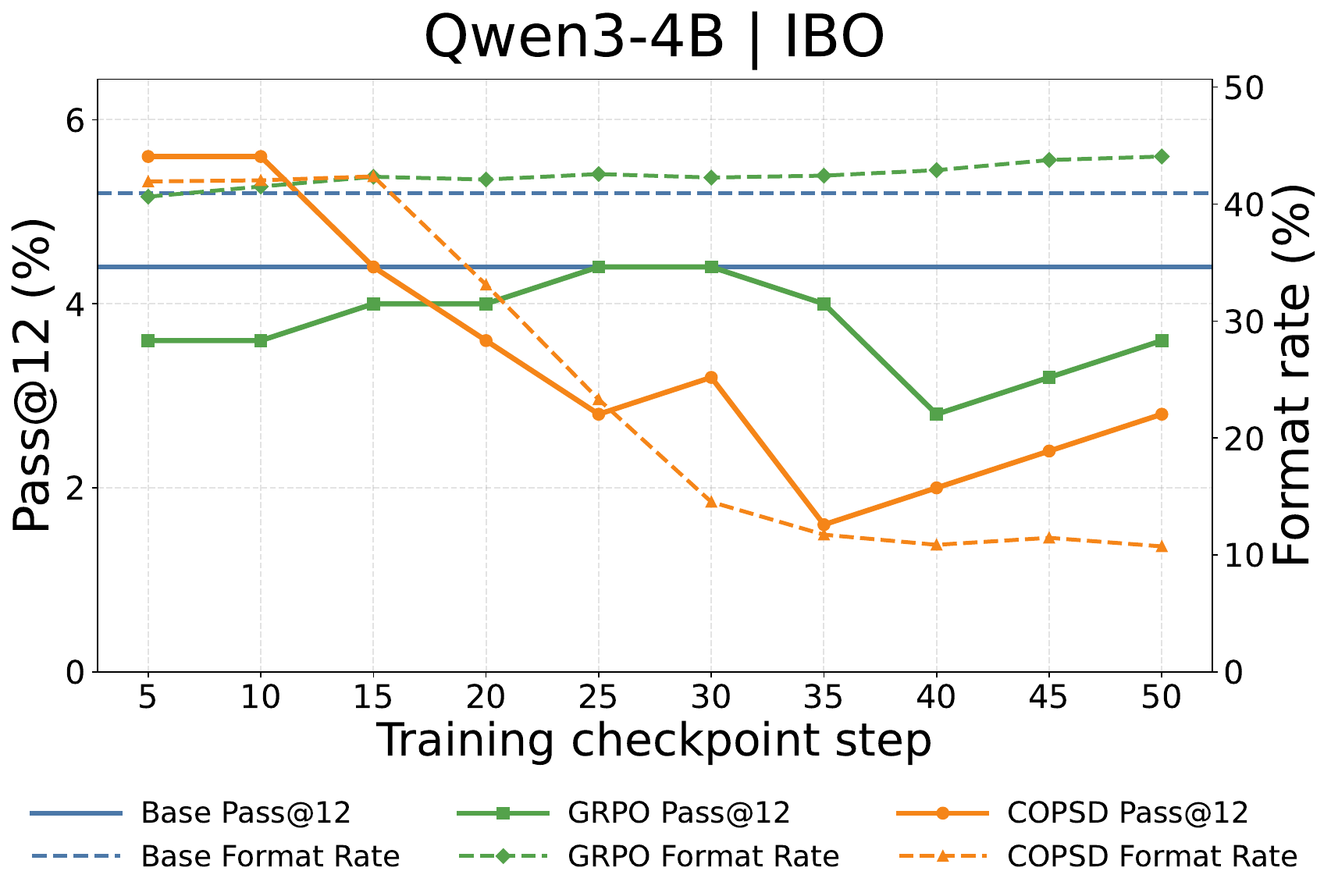}
    \includegraphics[width=0.32\linewidth]{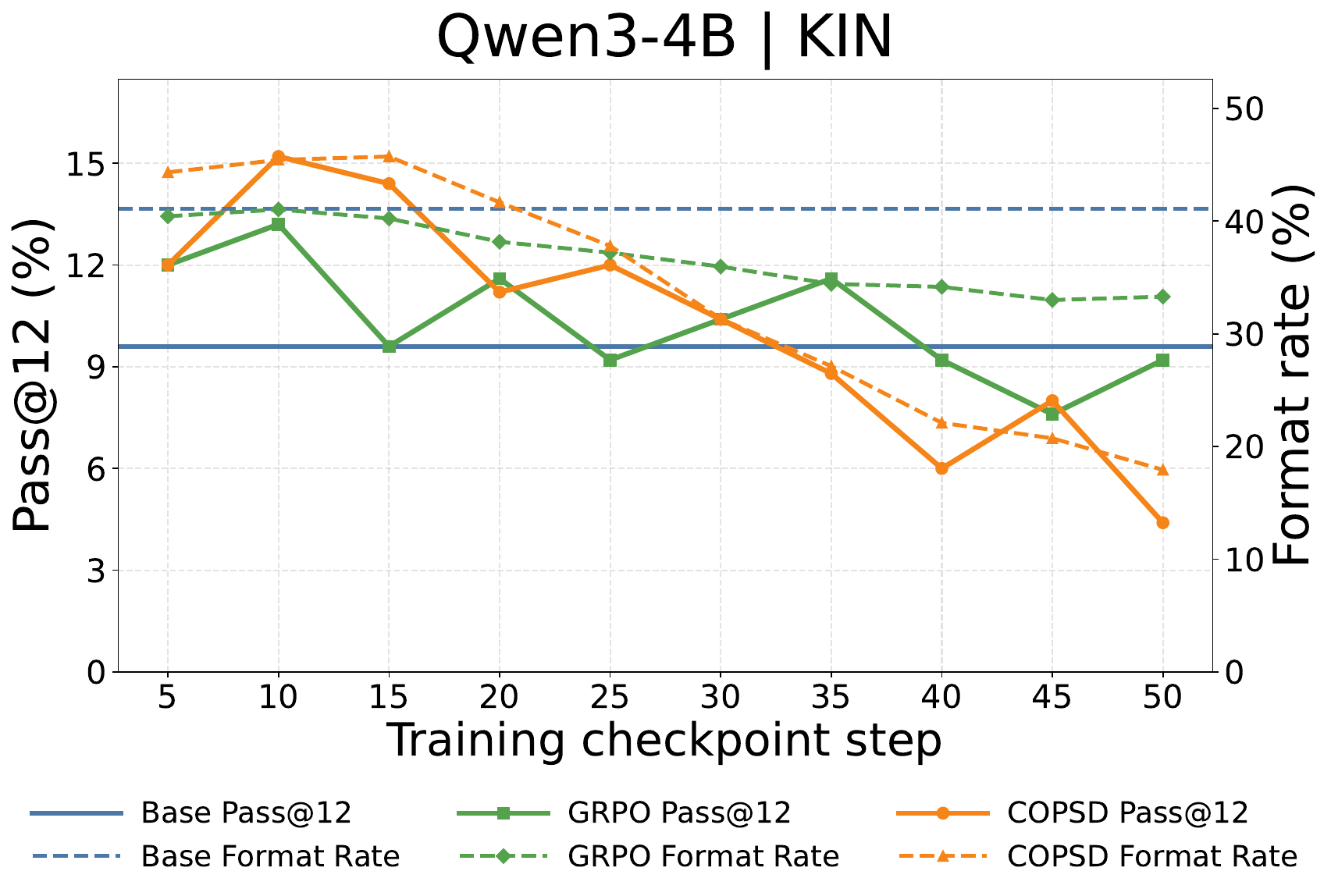}
    \includegraphics[width=0.32\linewidth]{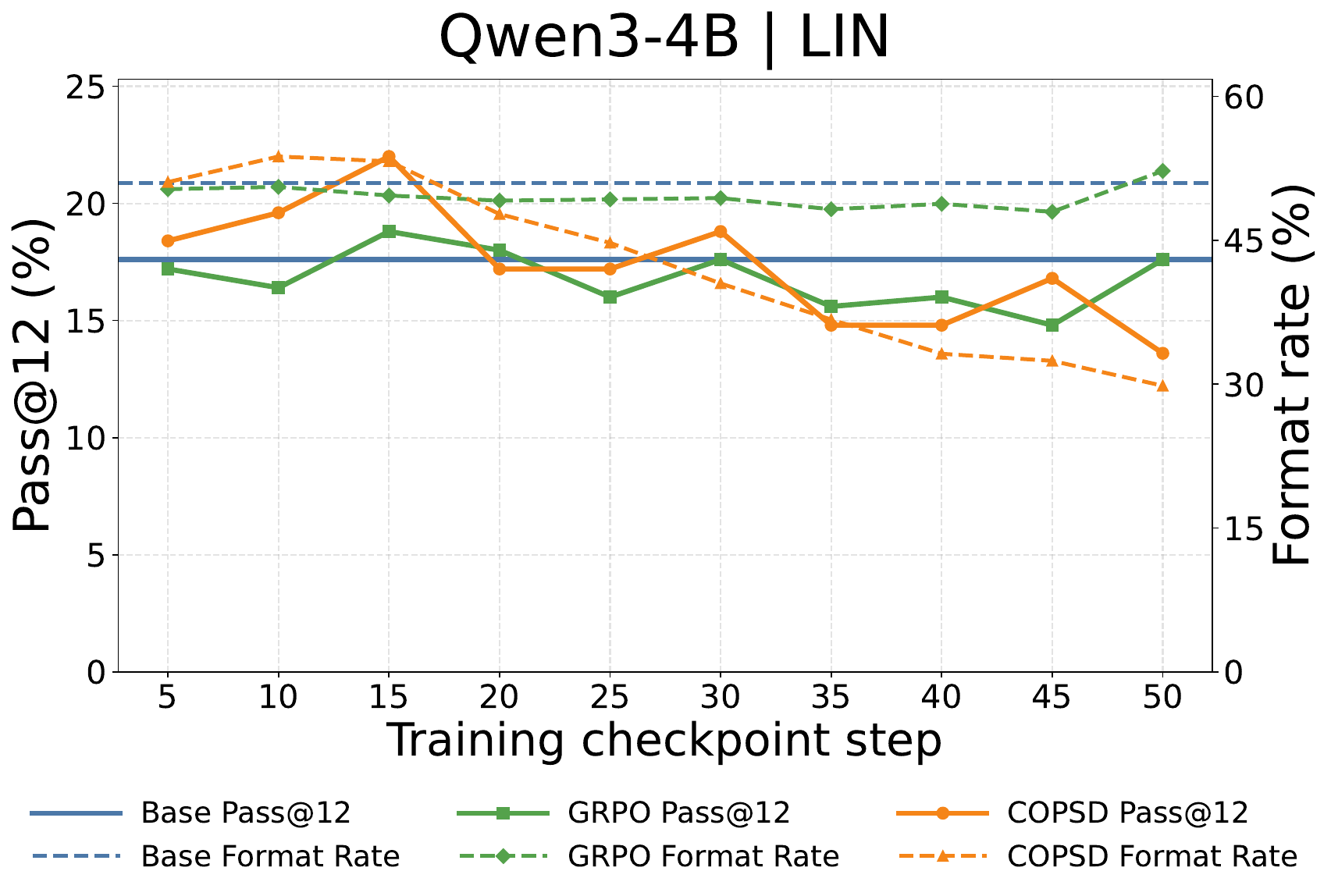}

    \includegraphics[width=0.32\linewidth]{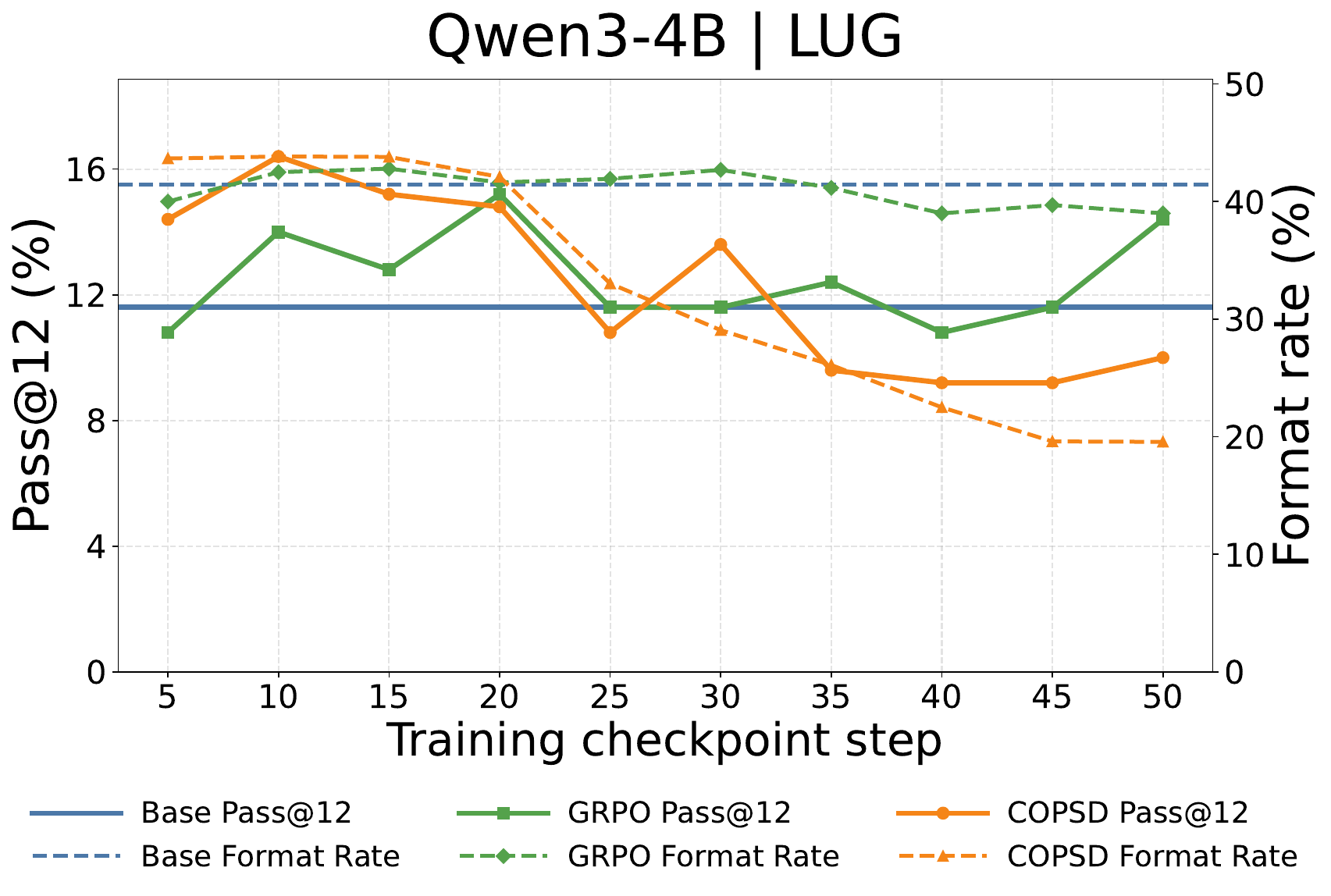}
    \includegraphics[width=0.32\linewidth]{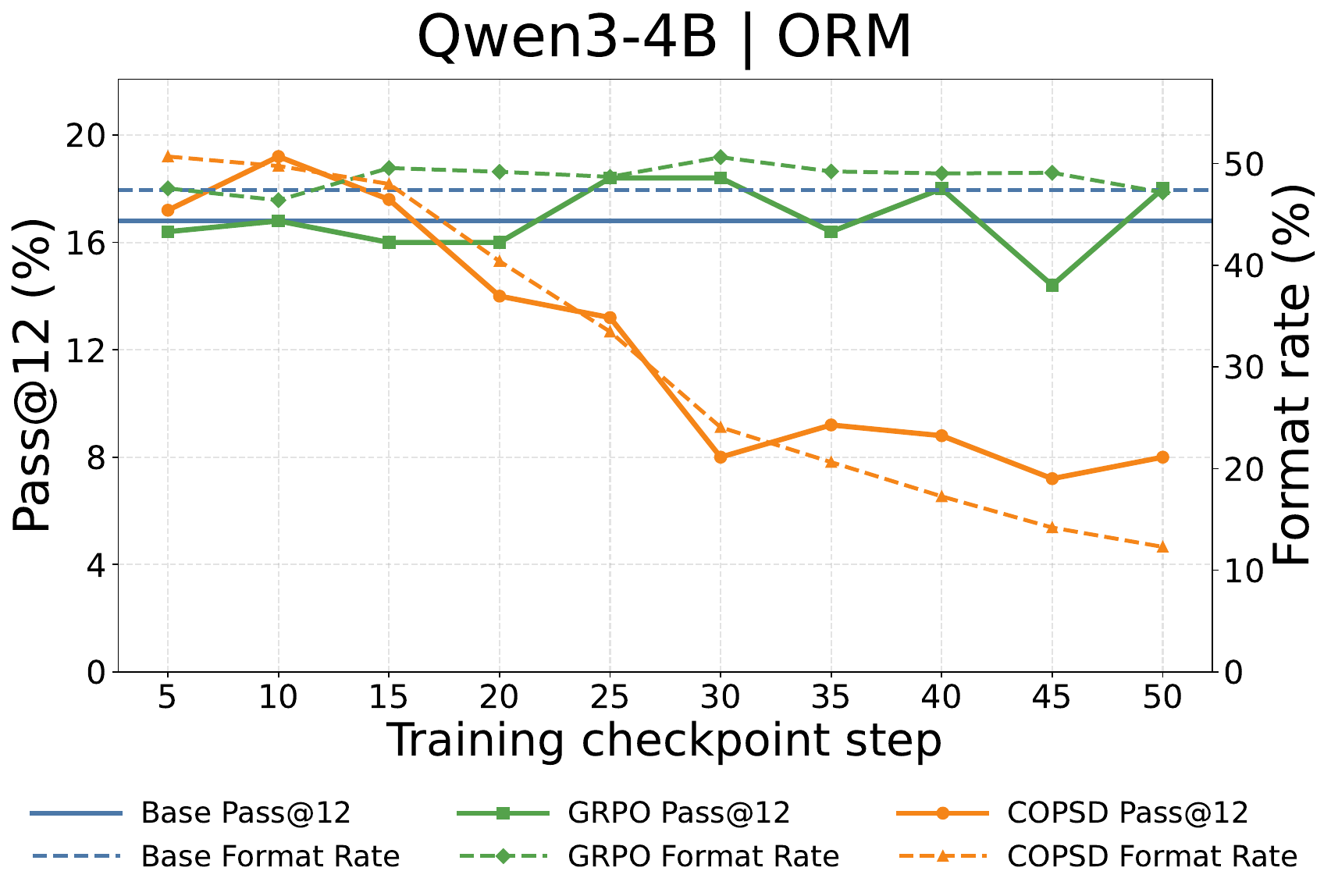}
    \includegraphics[width=0.32\linewidth]{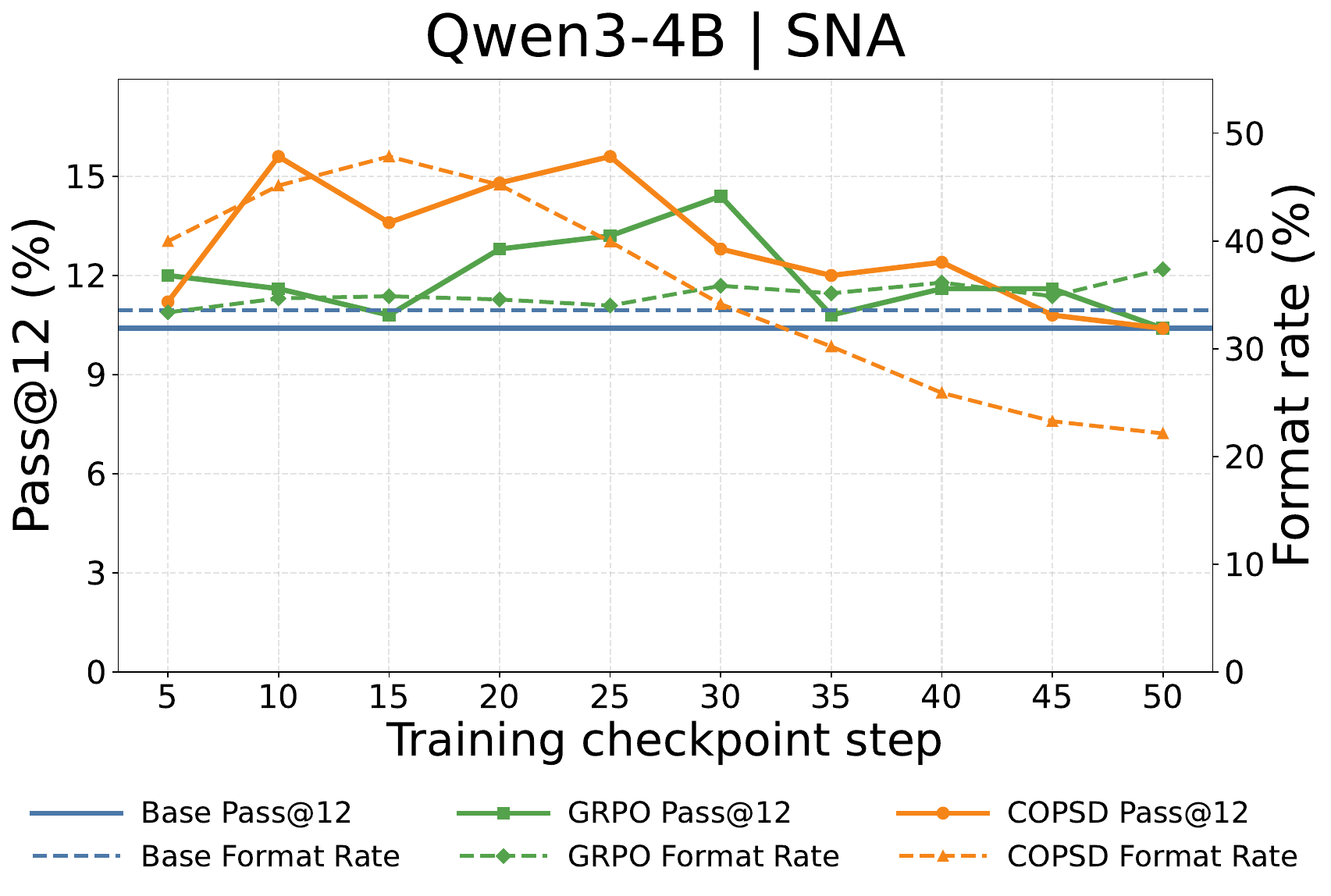}

    \includegraphics[width=0.32\linewidth]{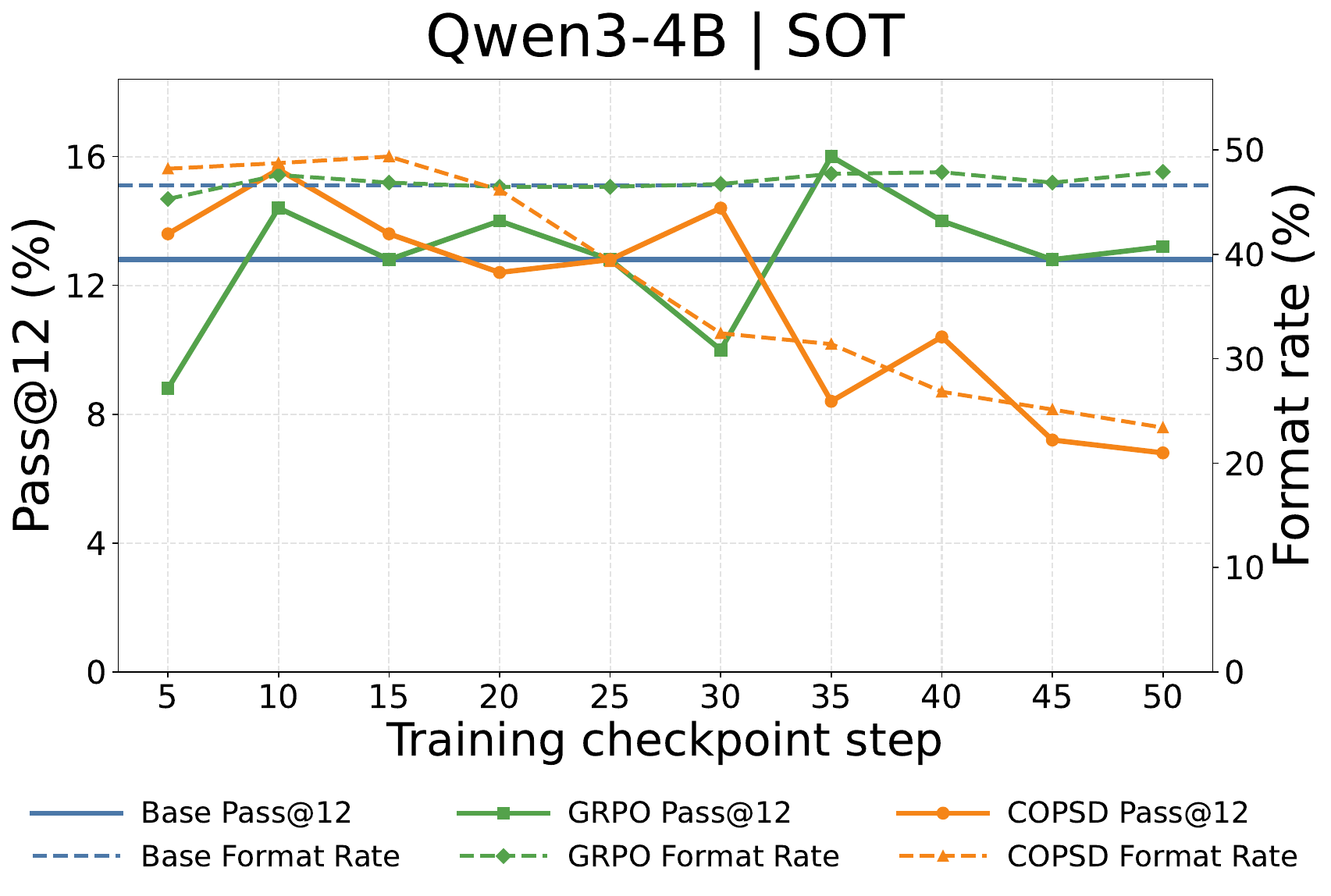}
    \includegraphics[width=0.32\linewidth]{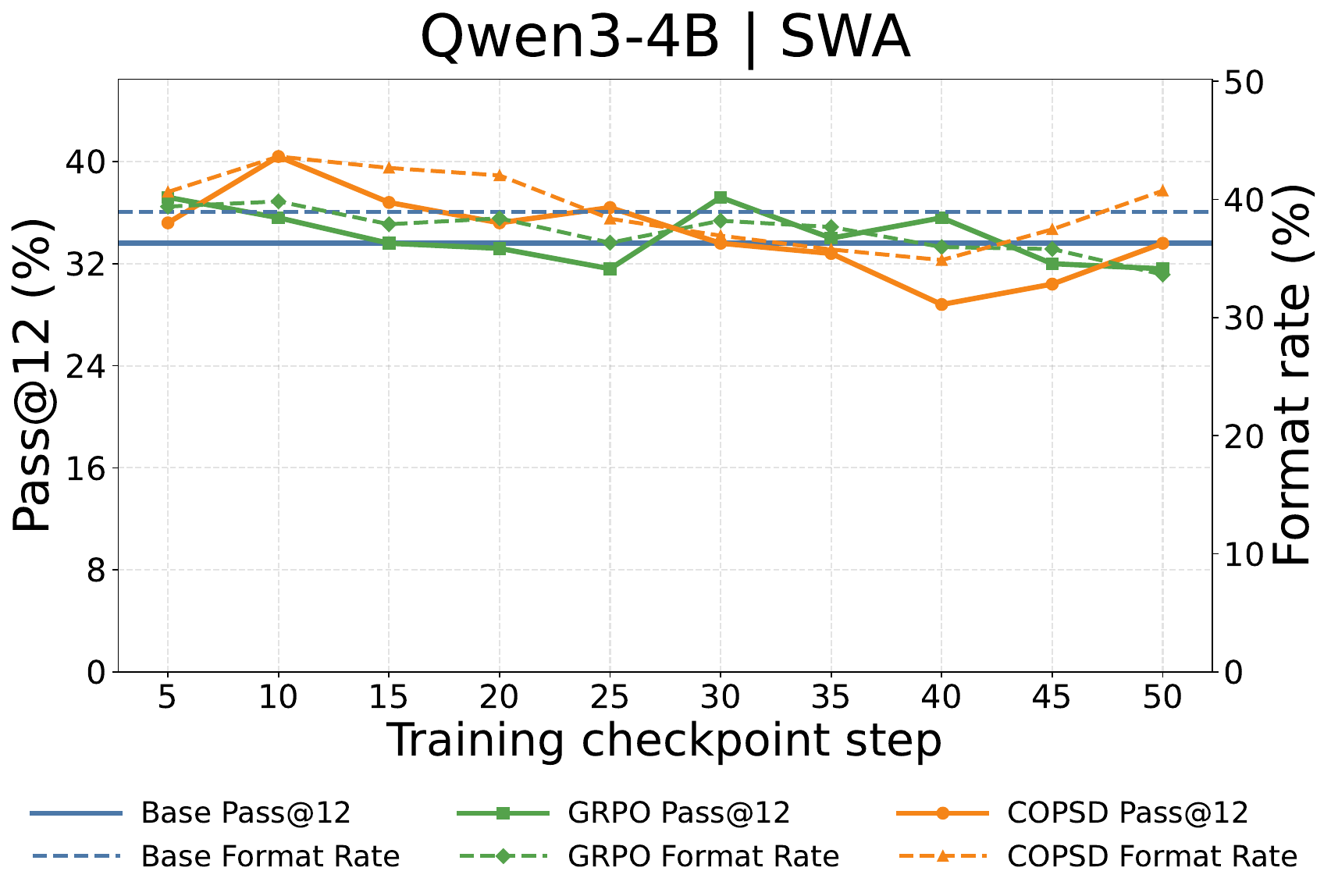}
    \includegraphics[width=0.32\linewidth]{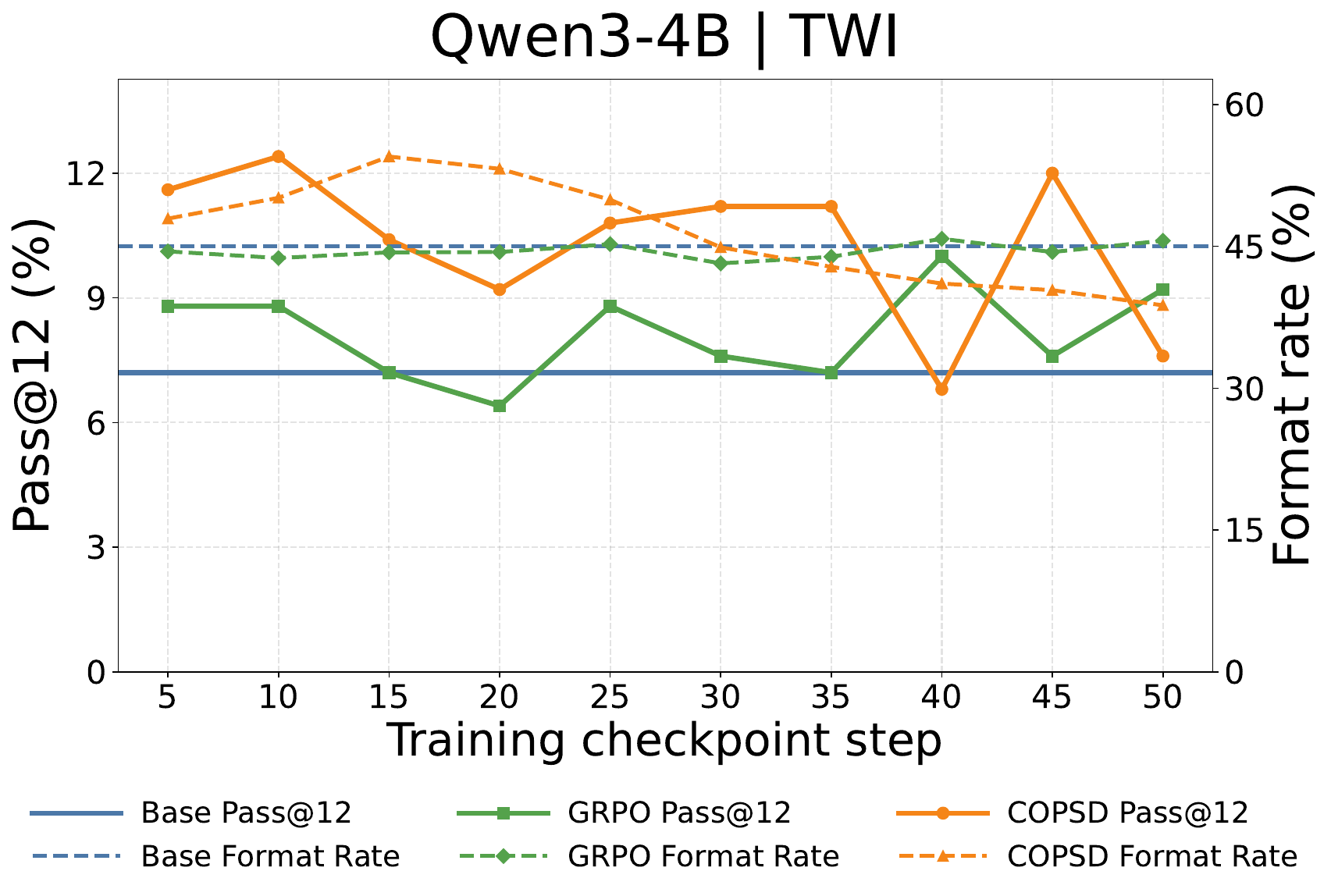}

    \includegraphics[width=0.32\linewidth]{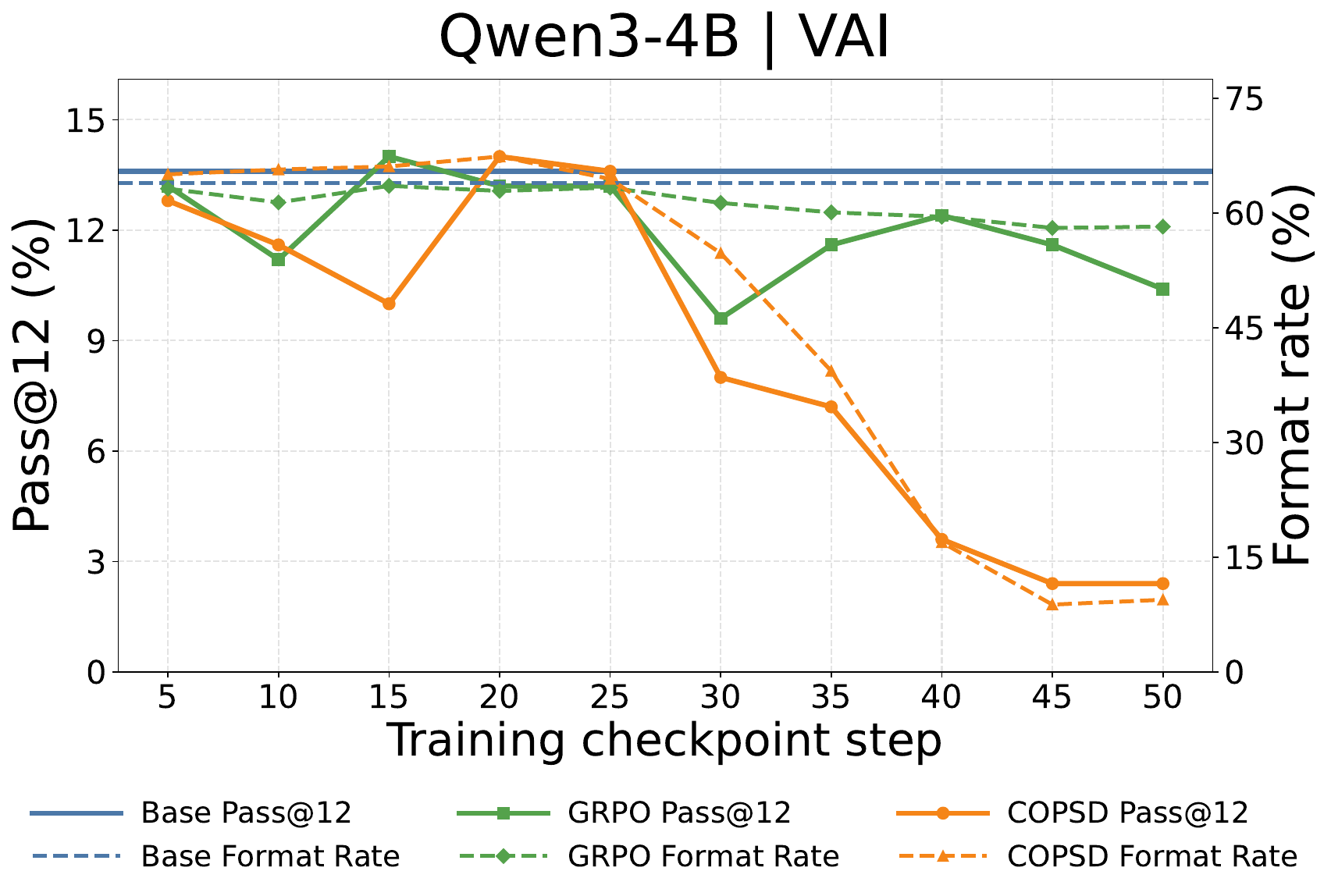}
    \includegraphics[width=0.32\linewidth]{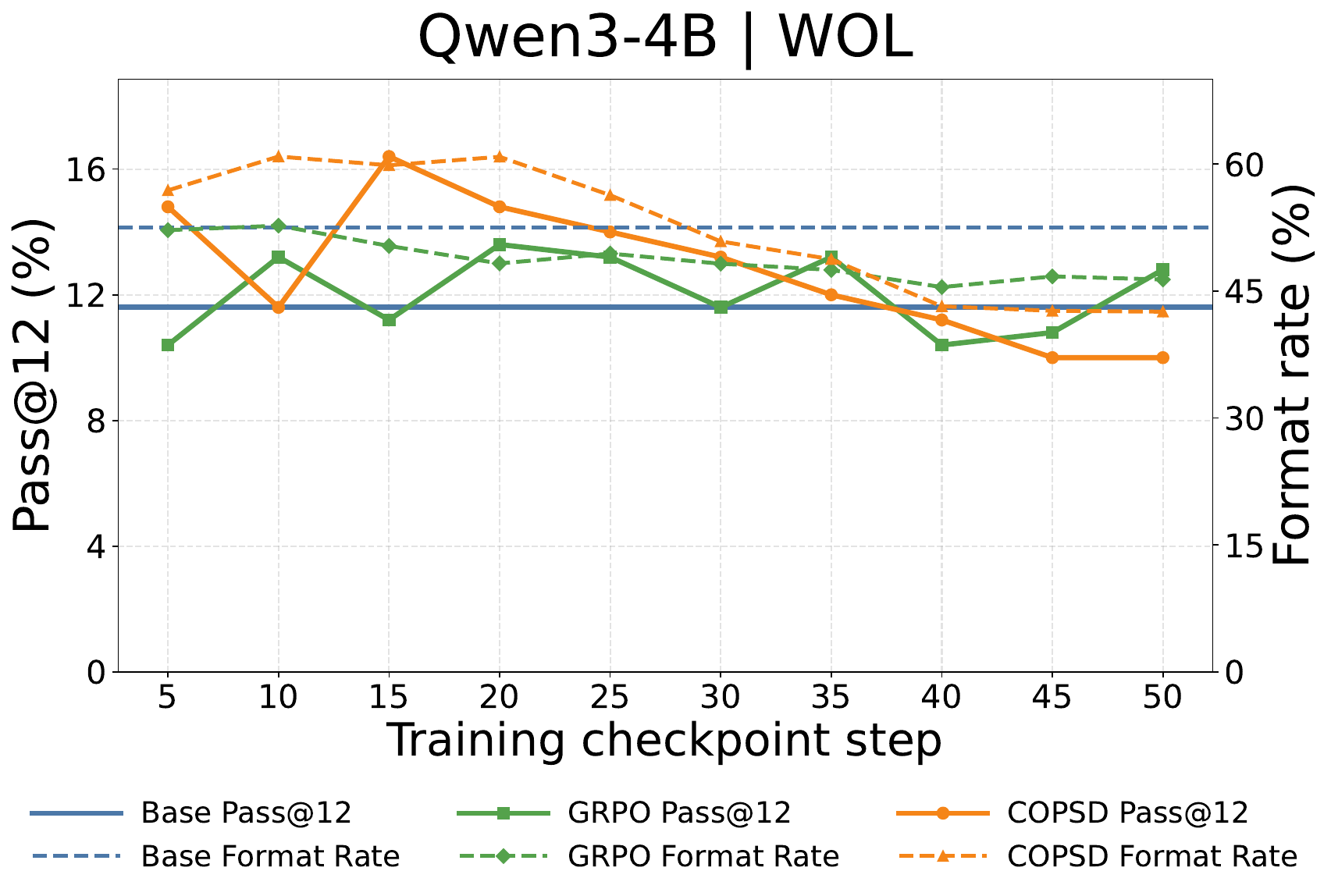}
    \includegraphics[width=0.32\linewidth]{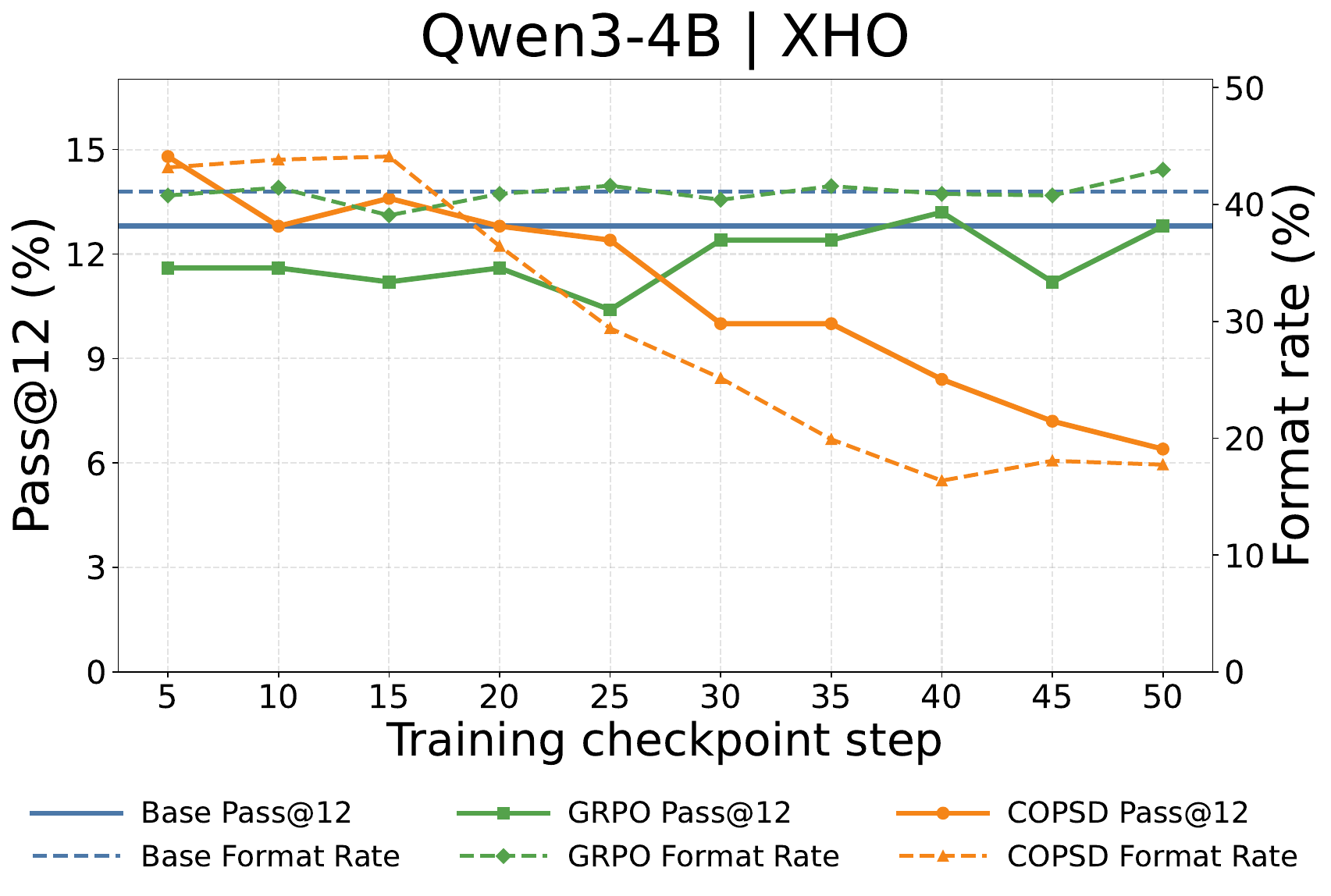}

    \includegraphics[width=0.32\linewidth]{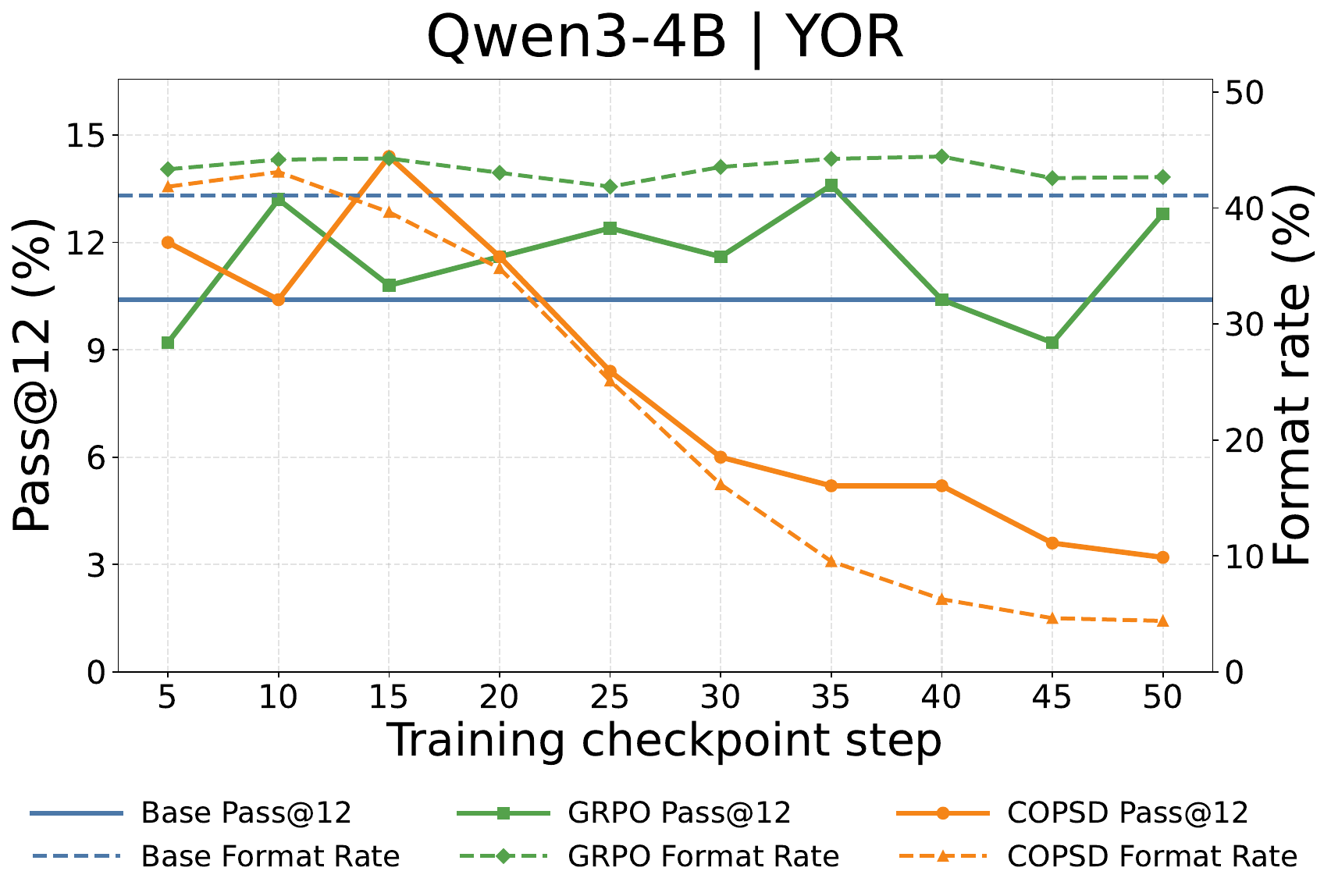}
    \includegraphics[width=0.32\linewidth]{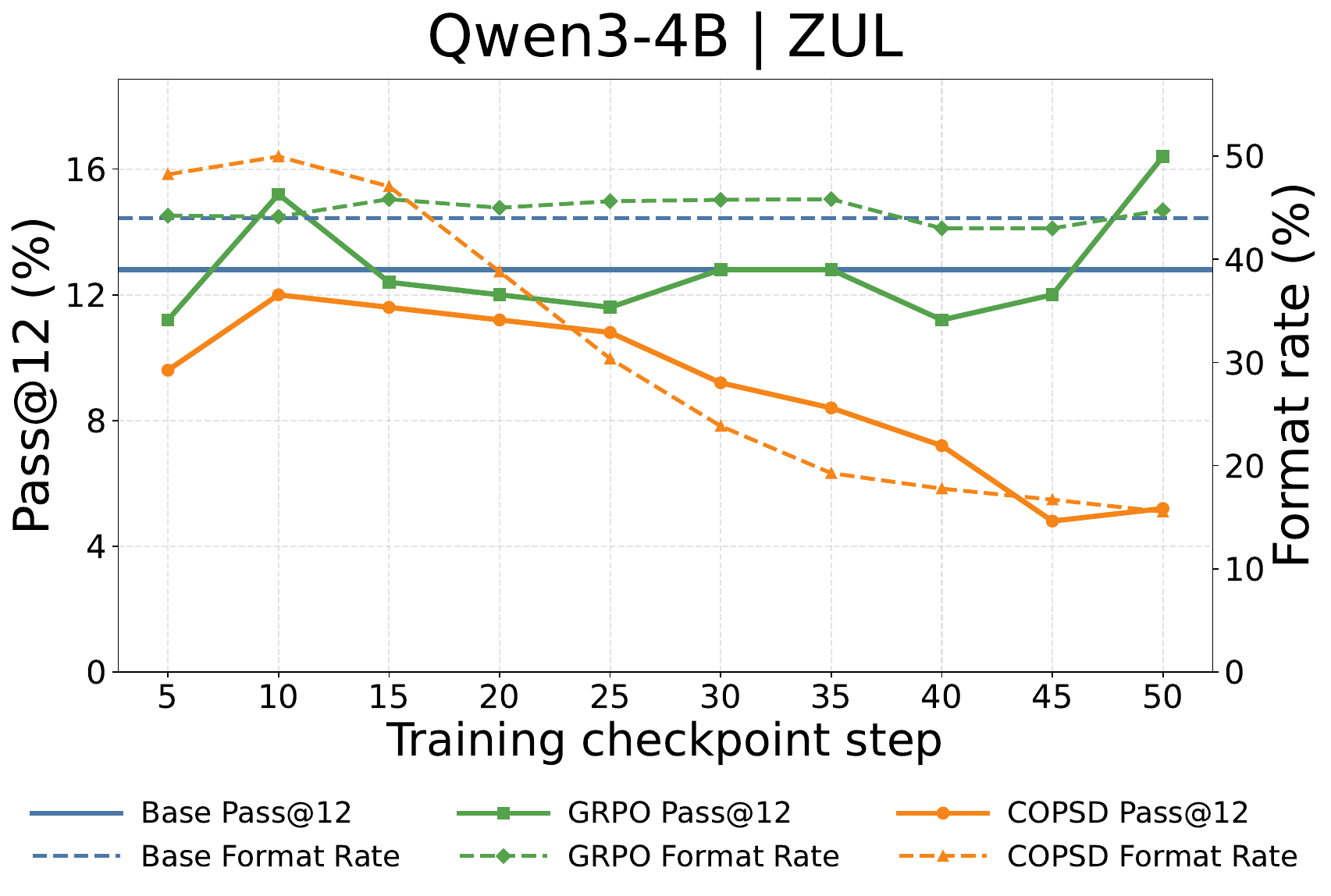}

    \caption{Per-language training dynamics for \texttt{Qwen3-4B} across all African languages under a 1024-token generation budget. Solid lines show Pass@12 and dashed lines show format rate.}
    \label{fig:training_dynamics_qwen3_4b_all_languages}
\end{figure*}

\begin{figure*}[t]
    \centering
    \includegraphics[width=0.32\linewidth]{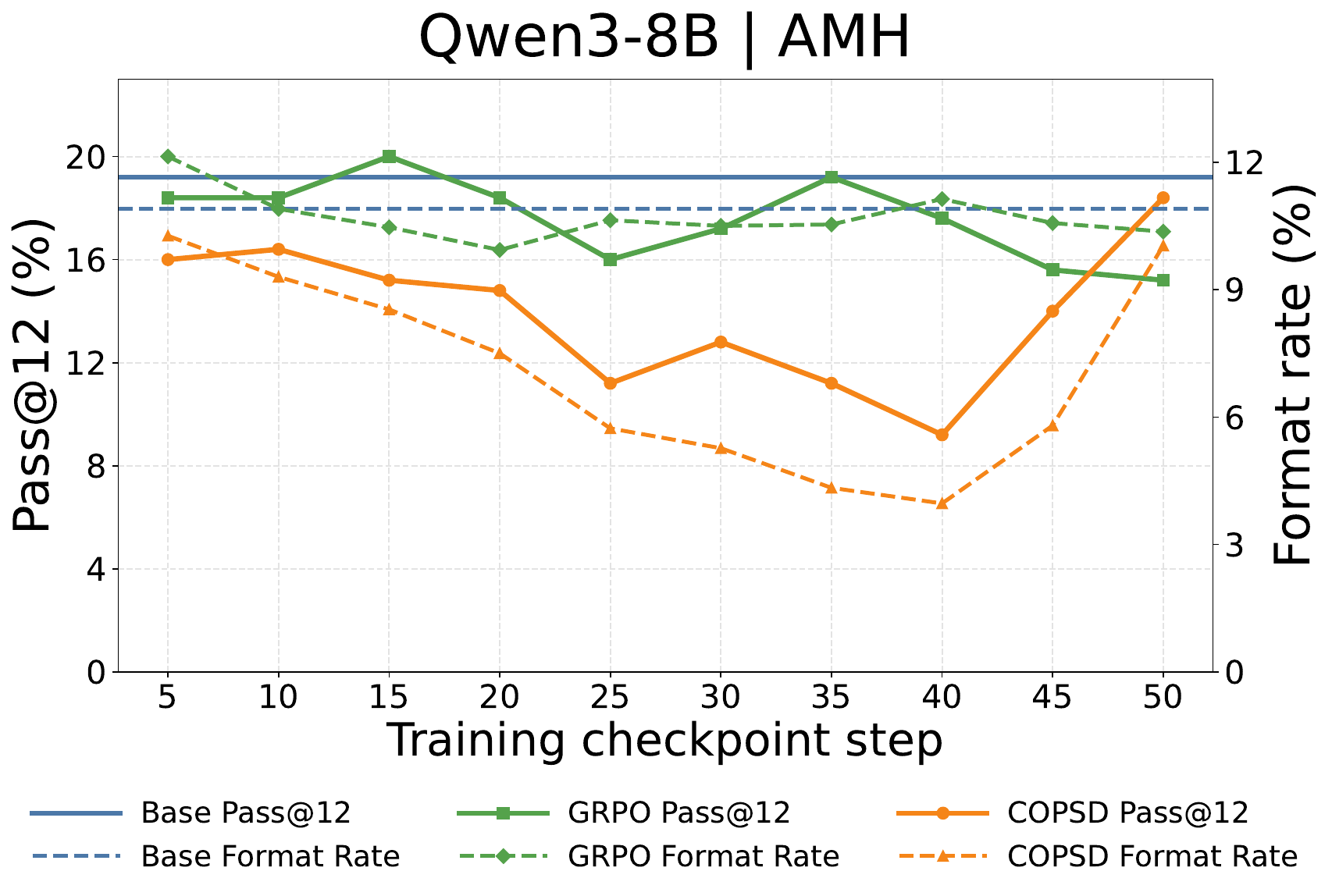}
    \includegraphics[width=0.32\linewidth]{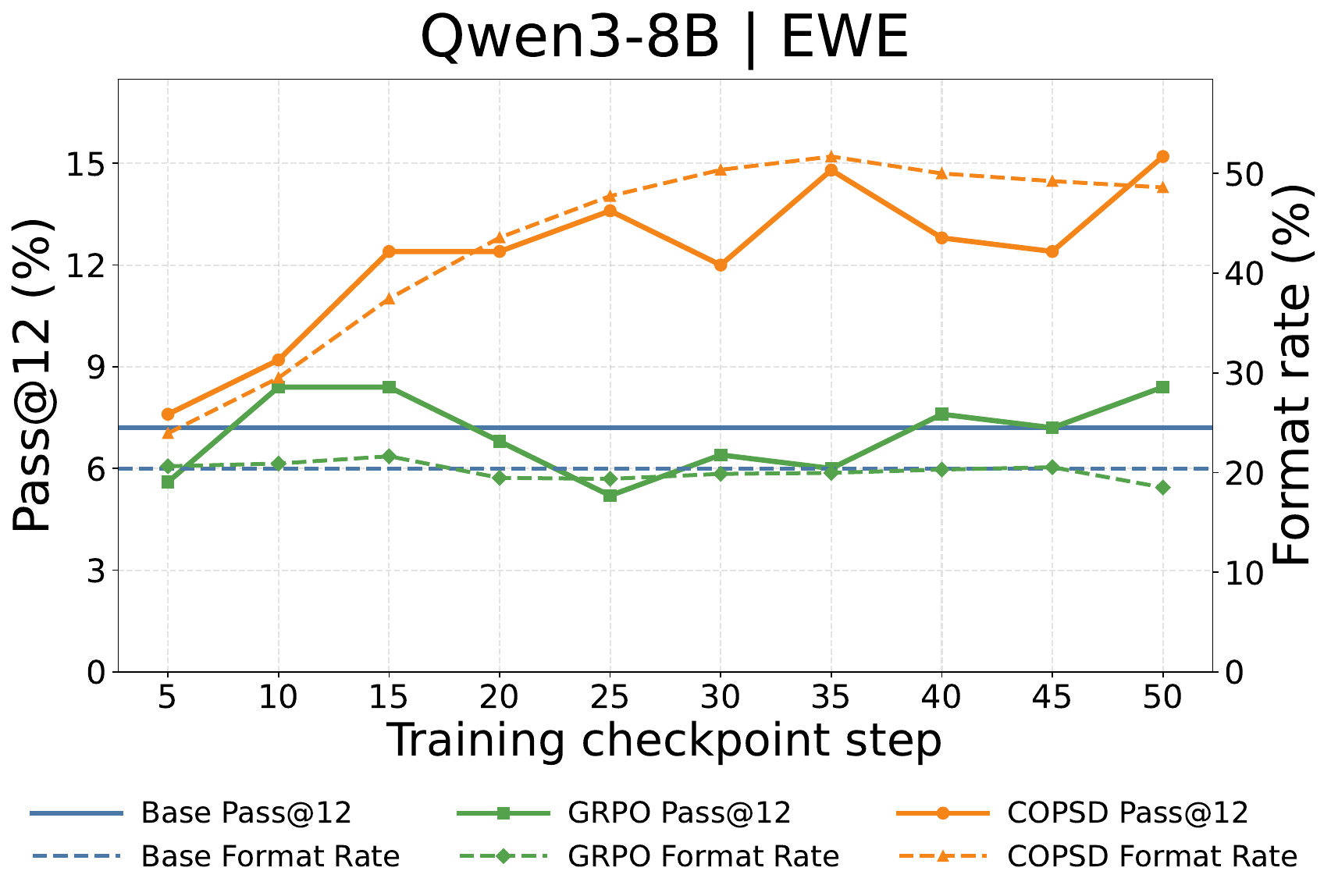}
    \includegraphics[width=0.32\linewidth]{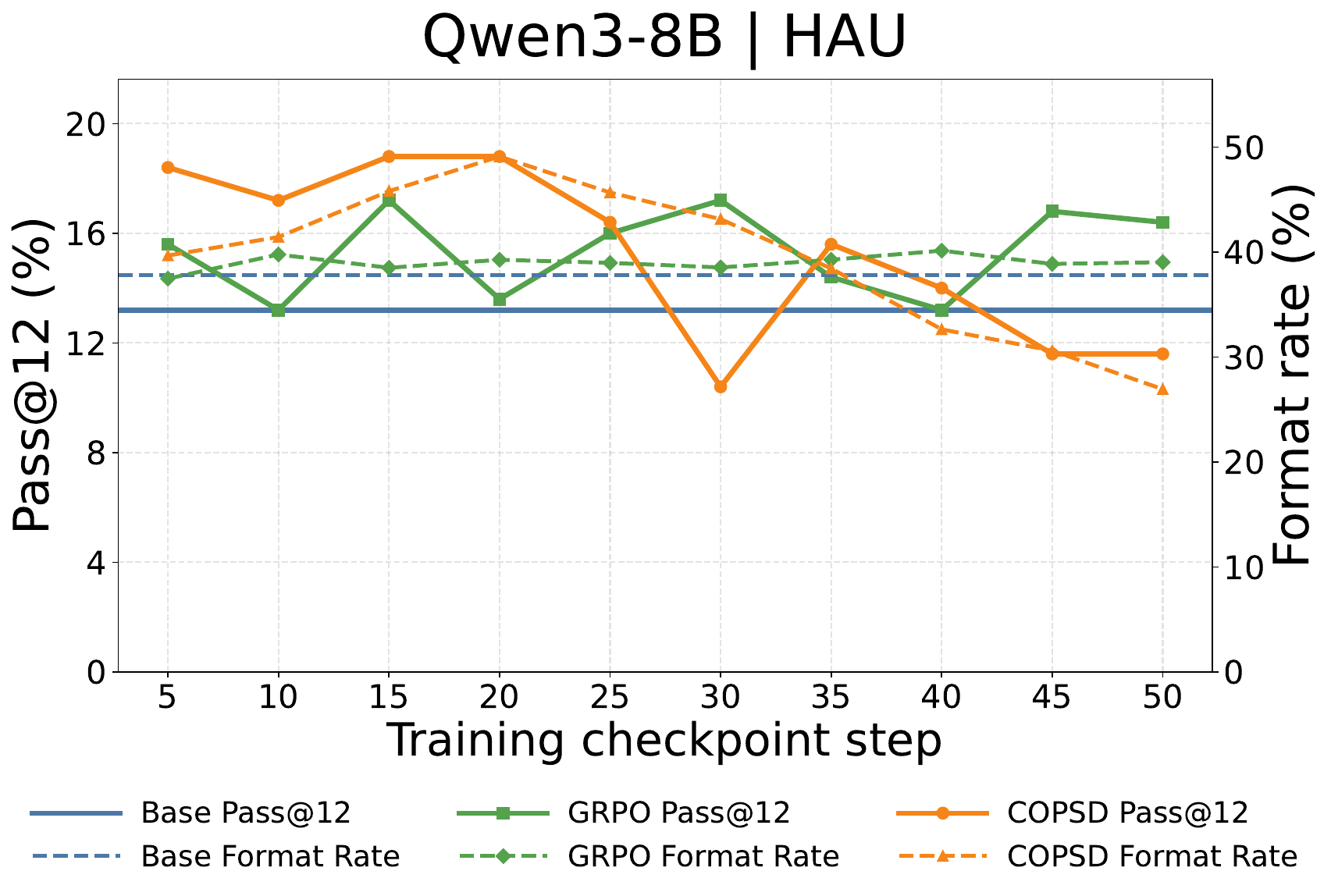}

    \includegraphics[width=0.32\linewidth]{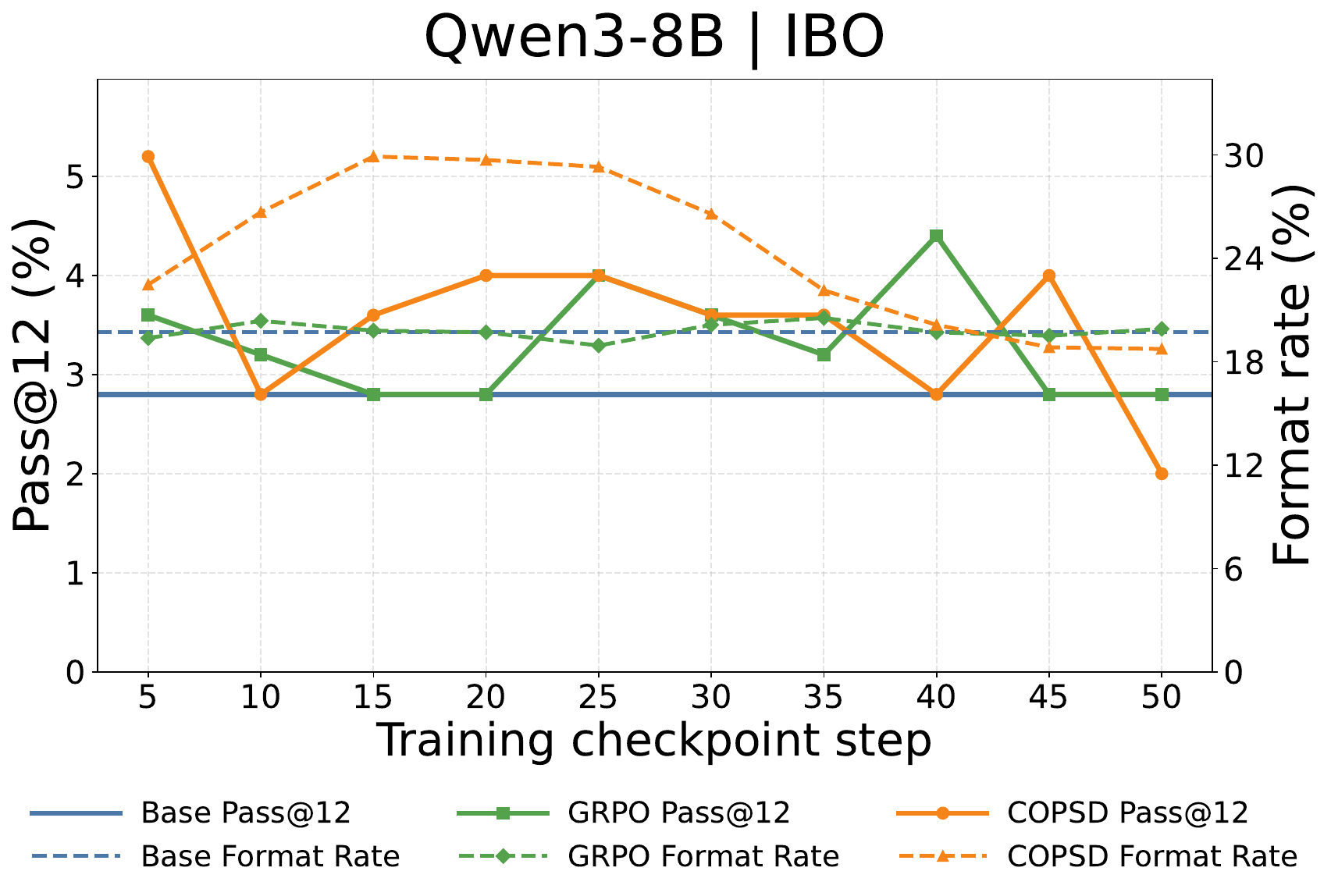}
    \includegraphics[width=0.32\linewidth]{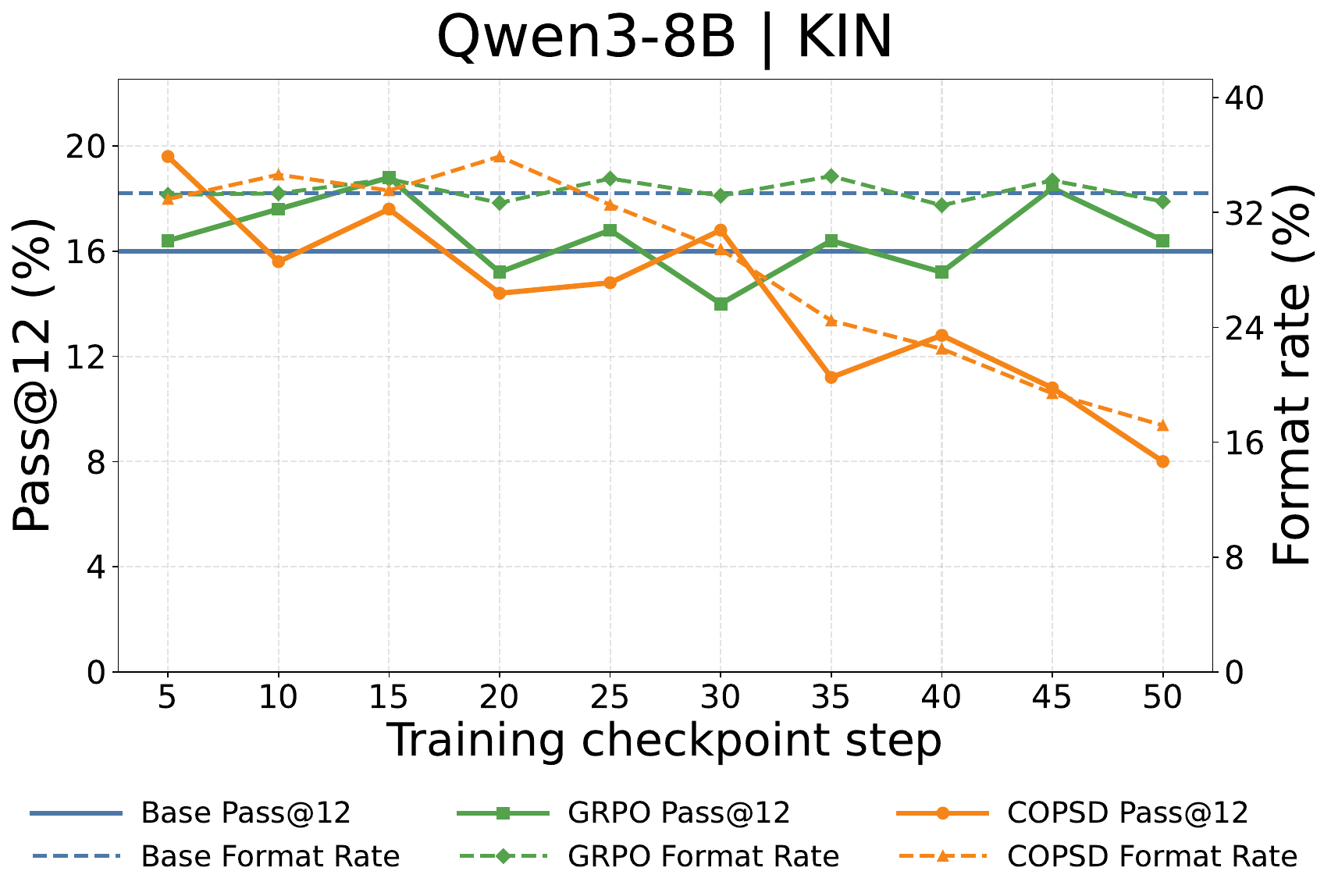}
    \includegraphics[width=0.32\linewidth]{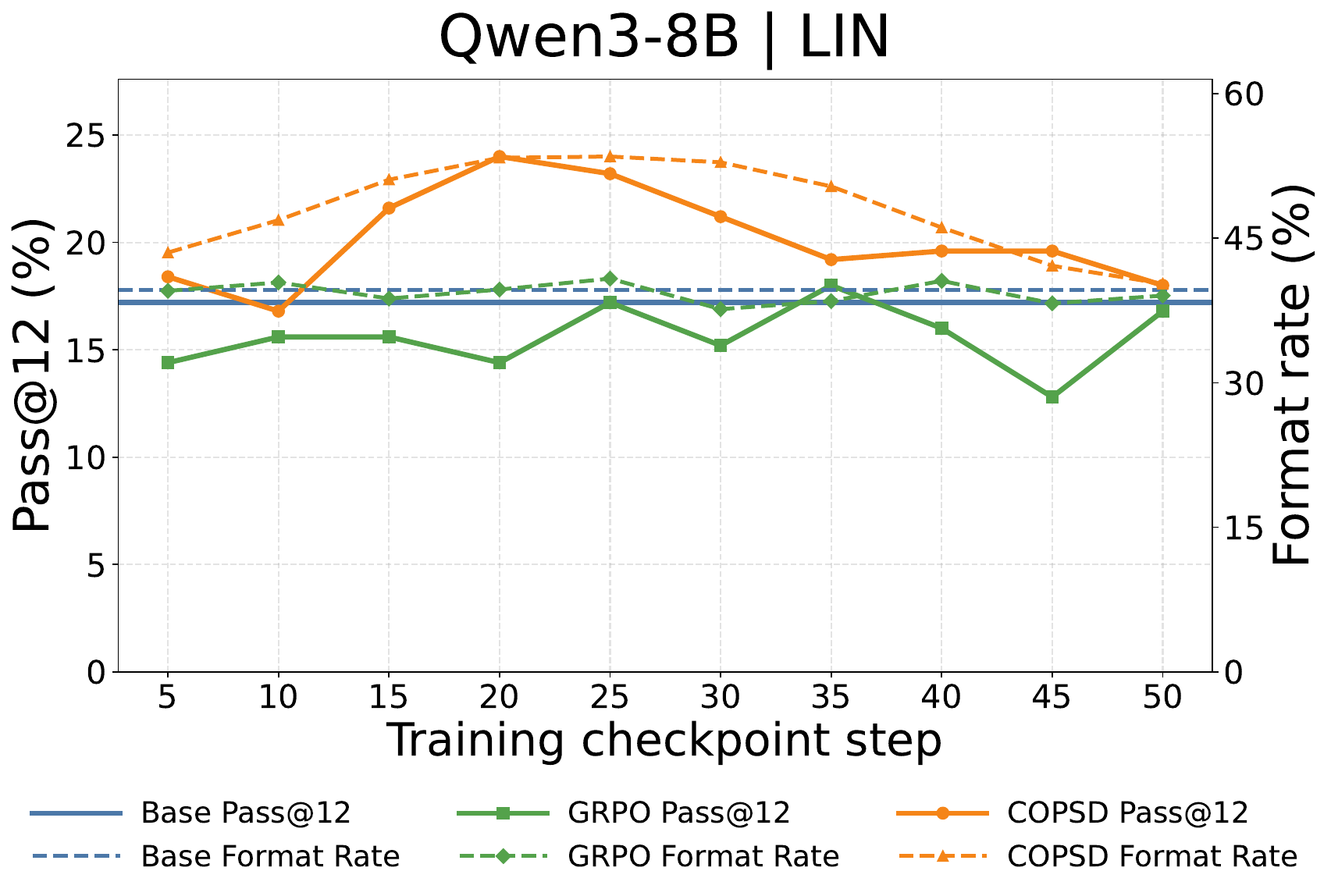}

    \includegraphics[width=0.32\linewidth]{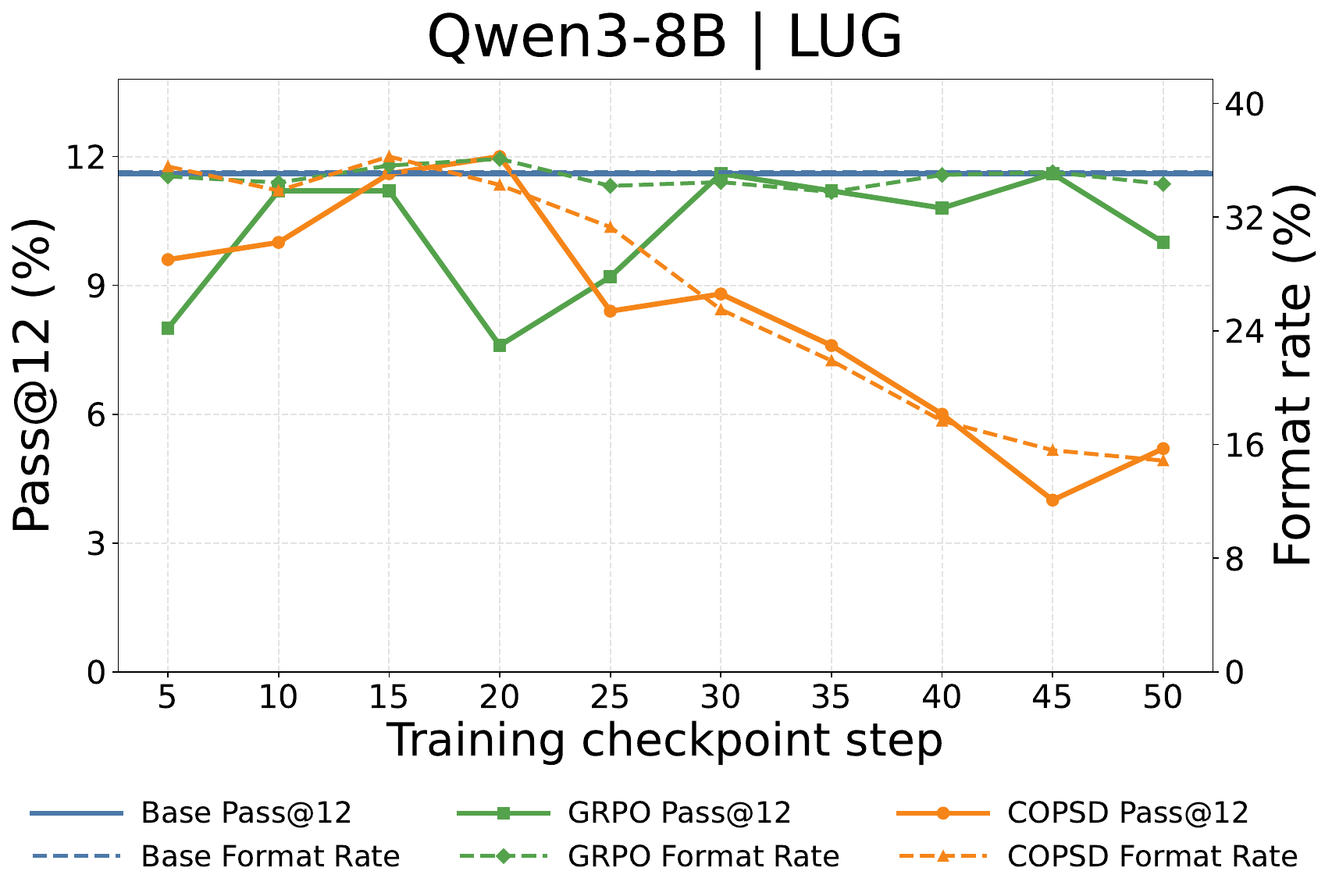}
    \includegraphics[width=0.32\linewidth]{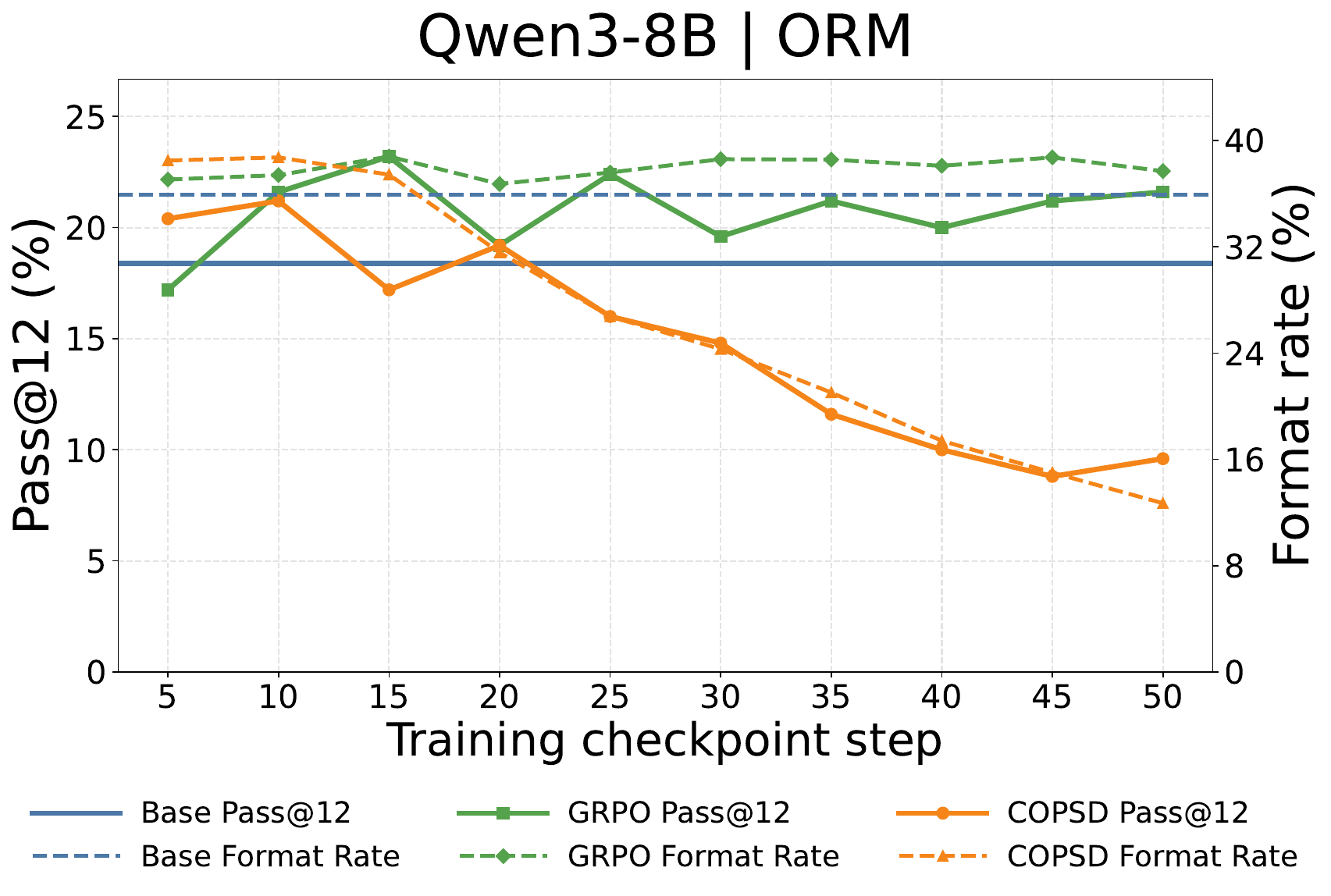}
    \includegraphics[width=0.32\linewidth]{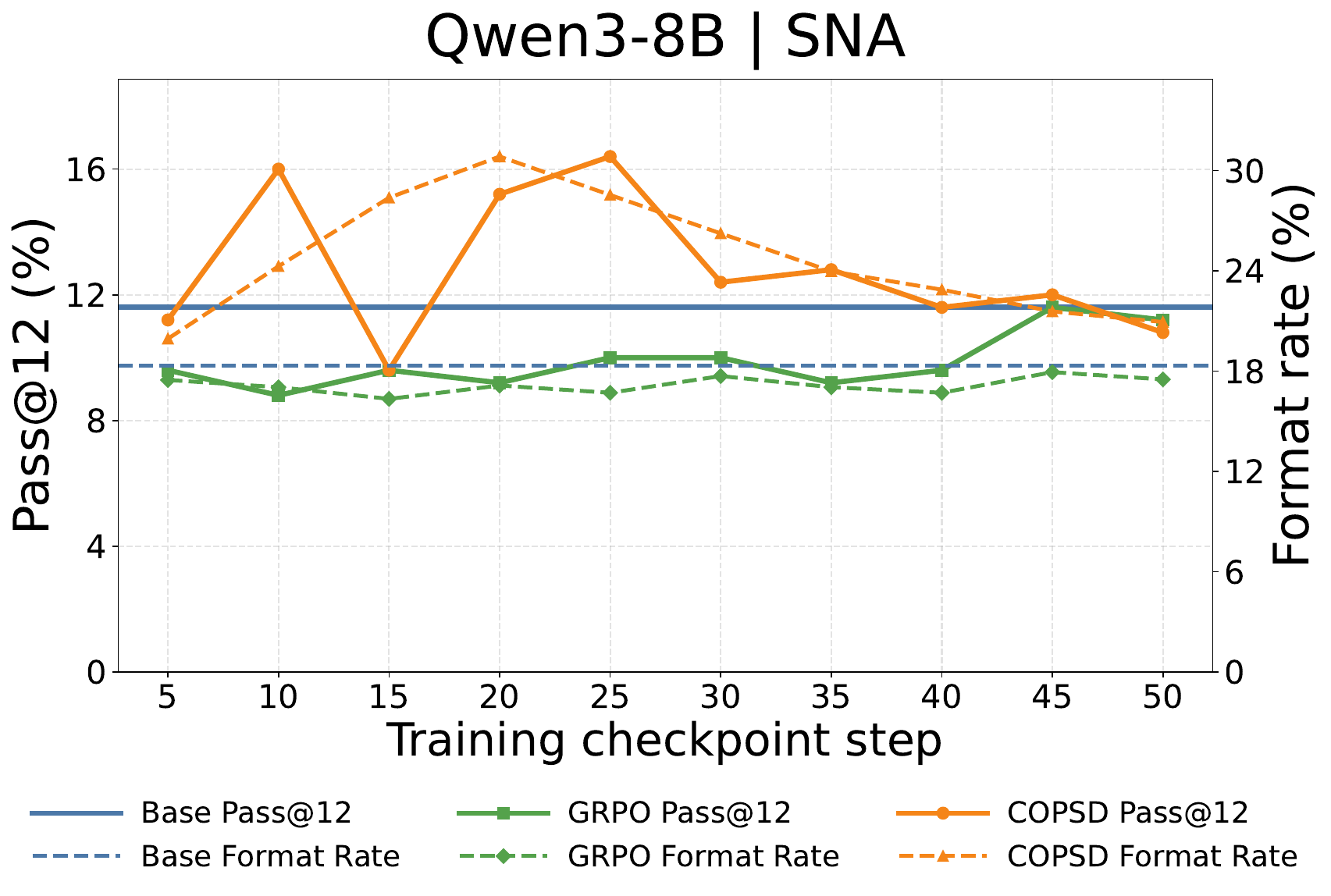}

    \includegraphics[width=0.32\linewidth]{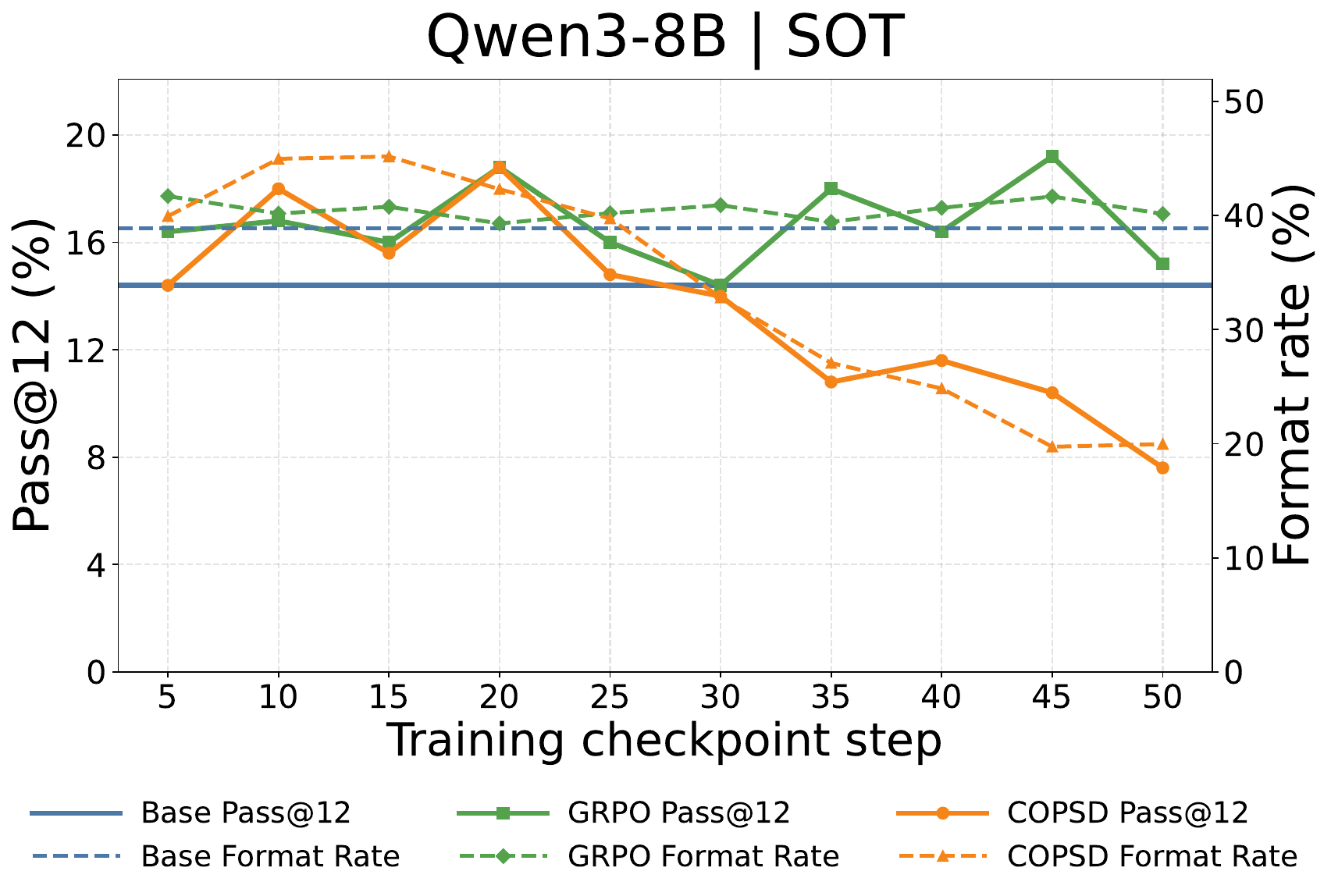}
    \includegraphics[width=0.32\linewidth]{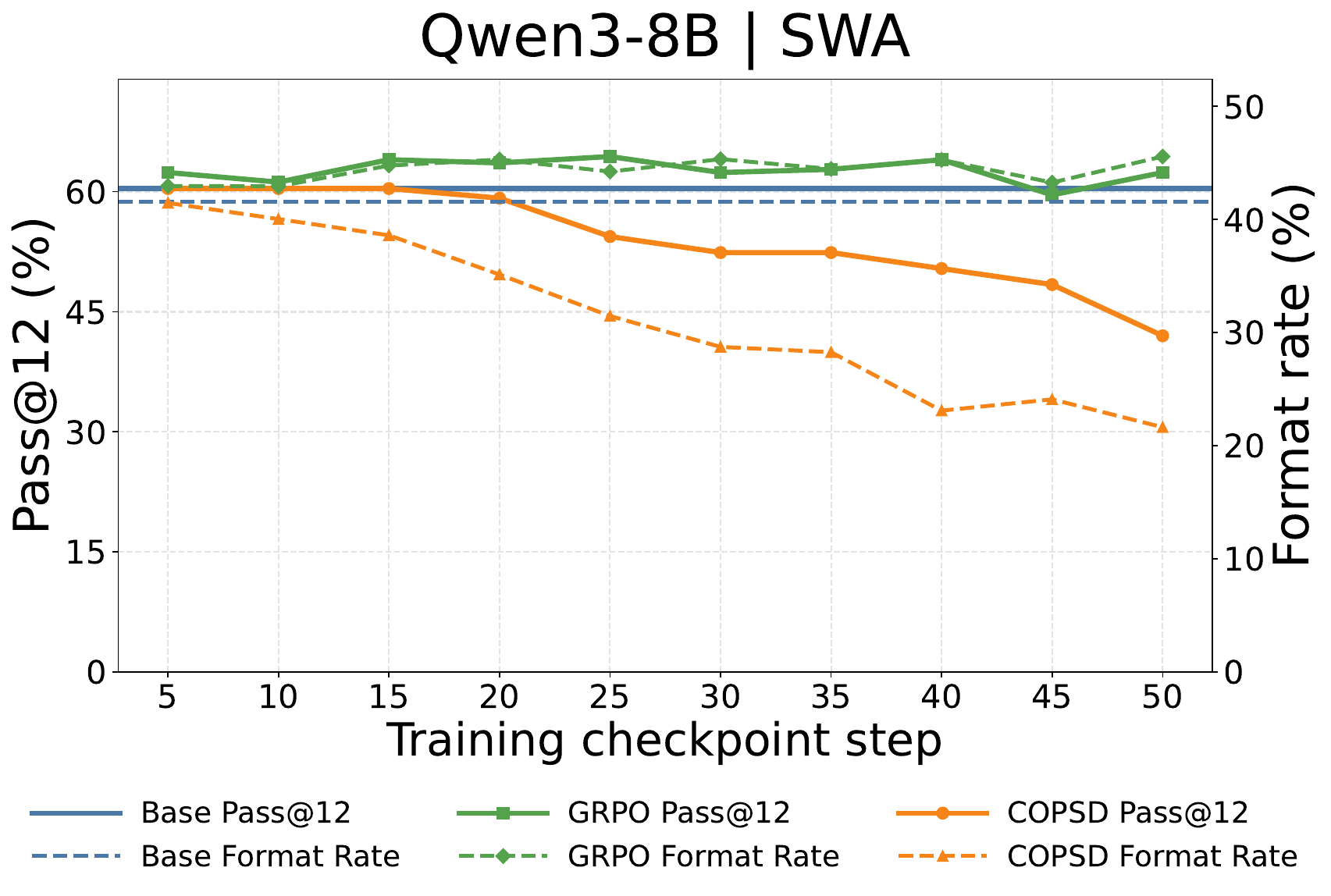}
    \includegraphics[width=0.32\linewidth]{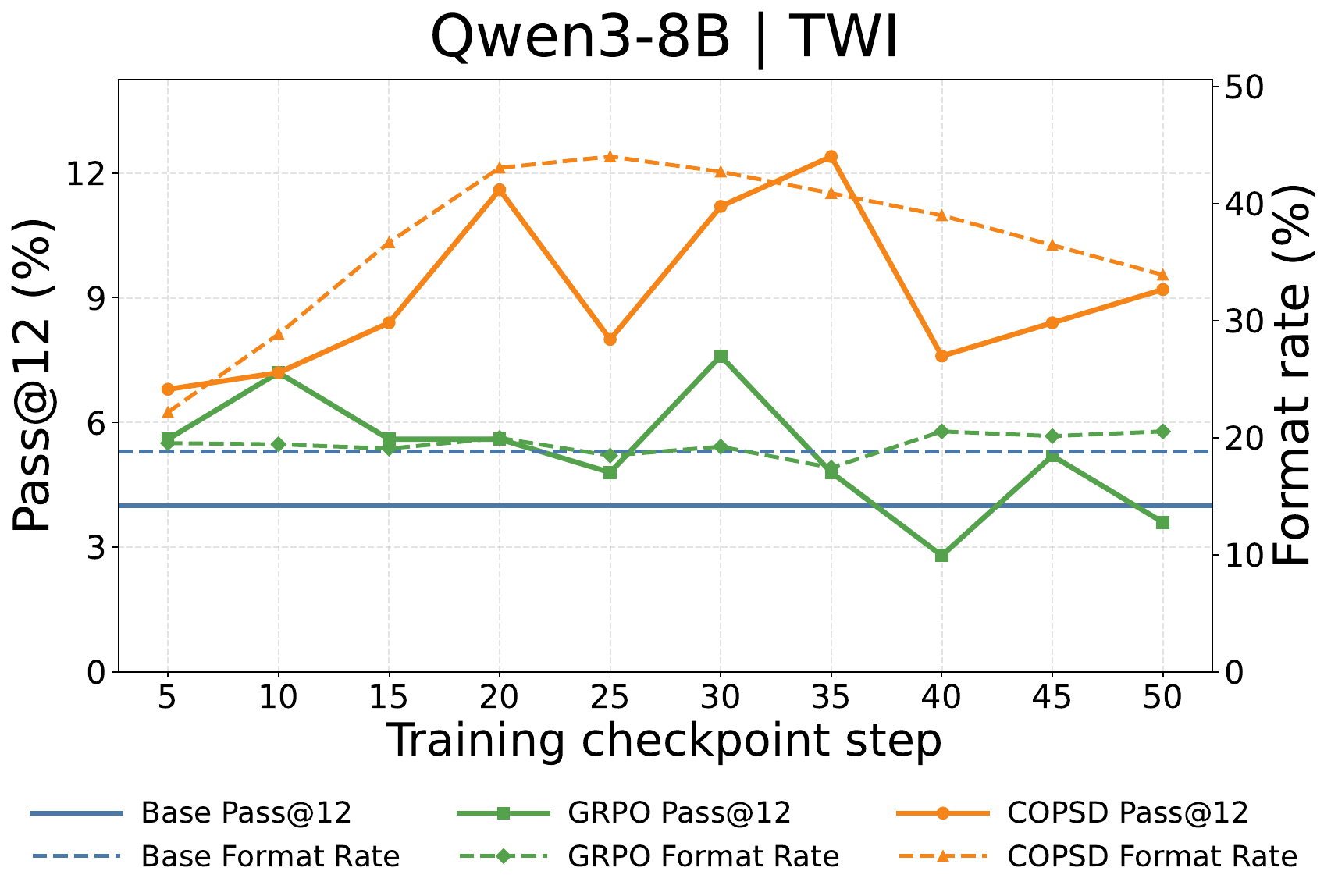}

    \includegraphics[width=0.32\linewidth]{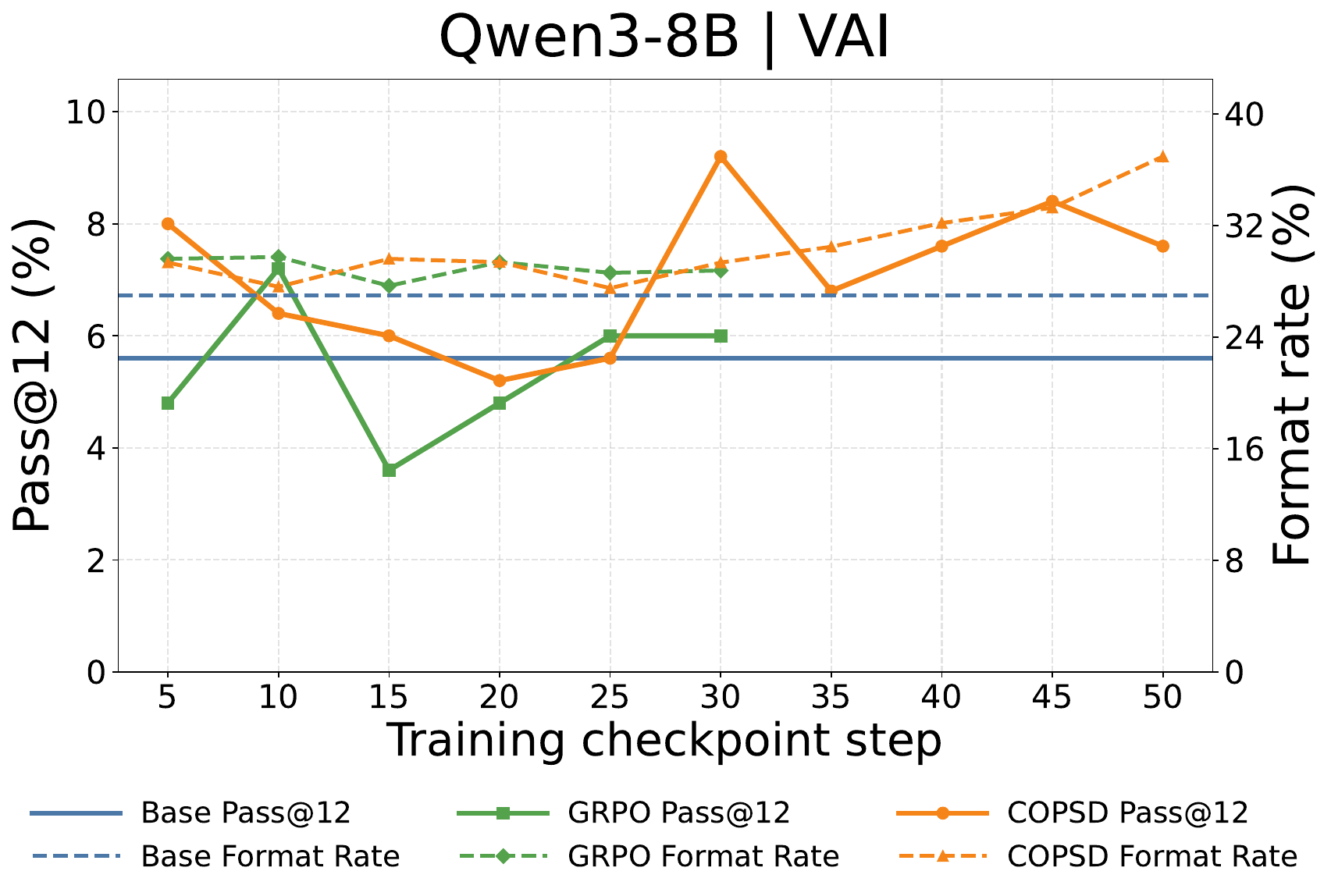}
    \includegraphics[width=0.32\linewidth]{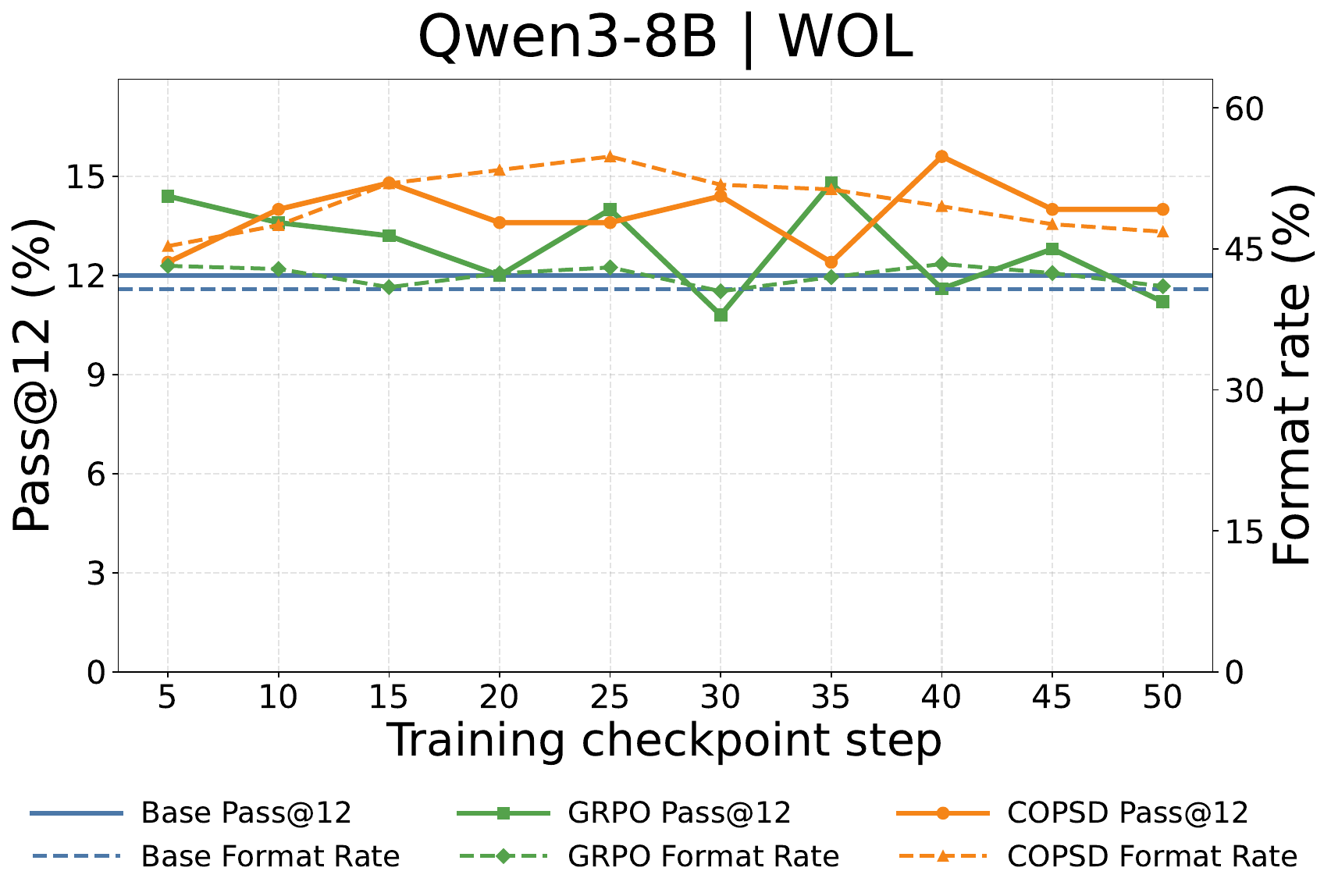}
    \includegraphics[width=0.32\linewidth]{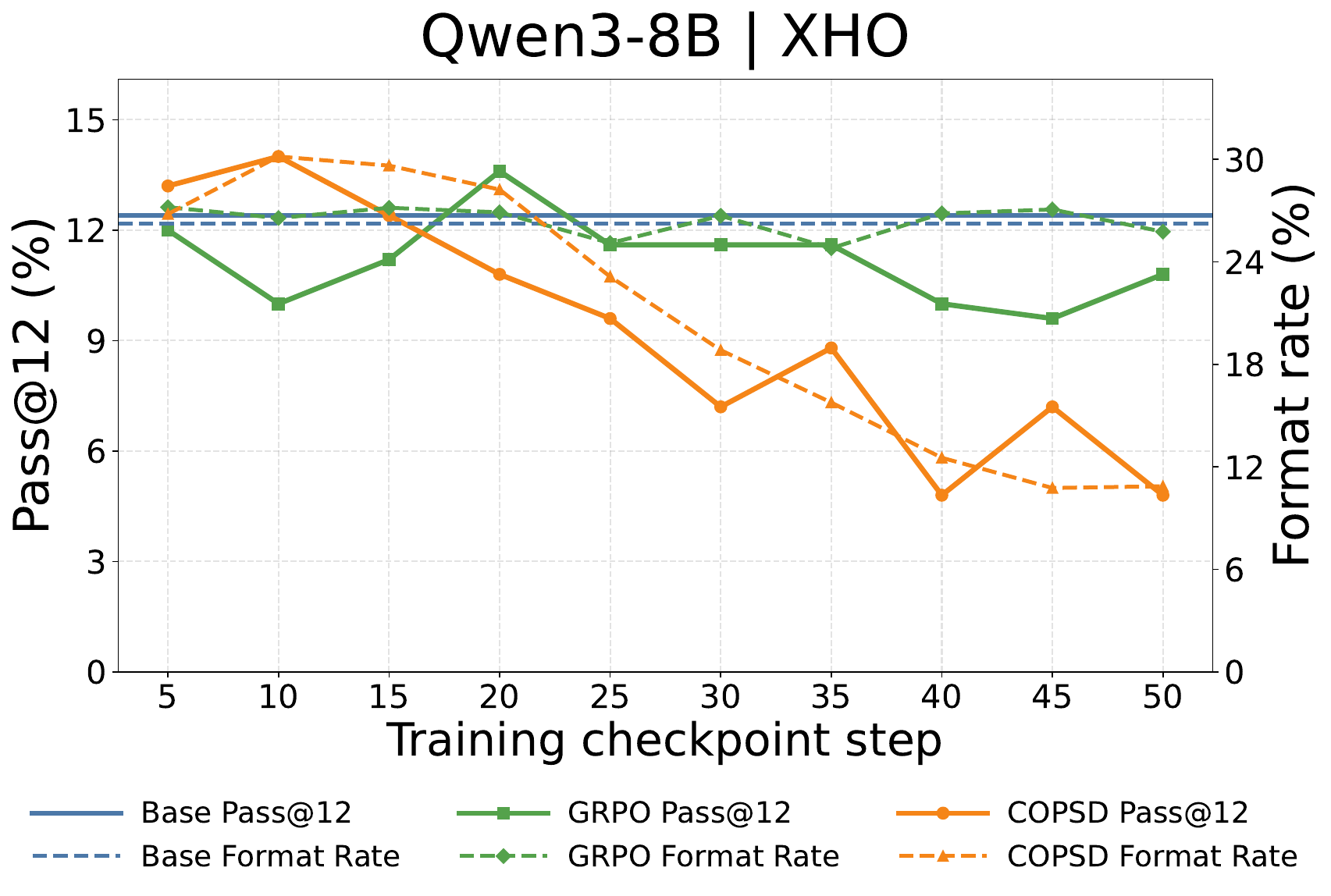}

    \includegraphics[width=0.32\linewidth]{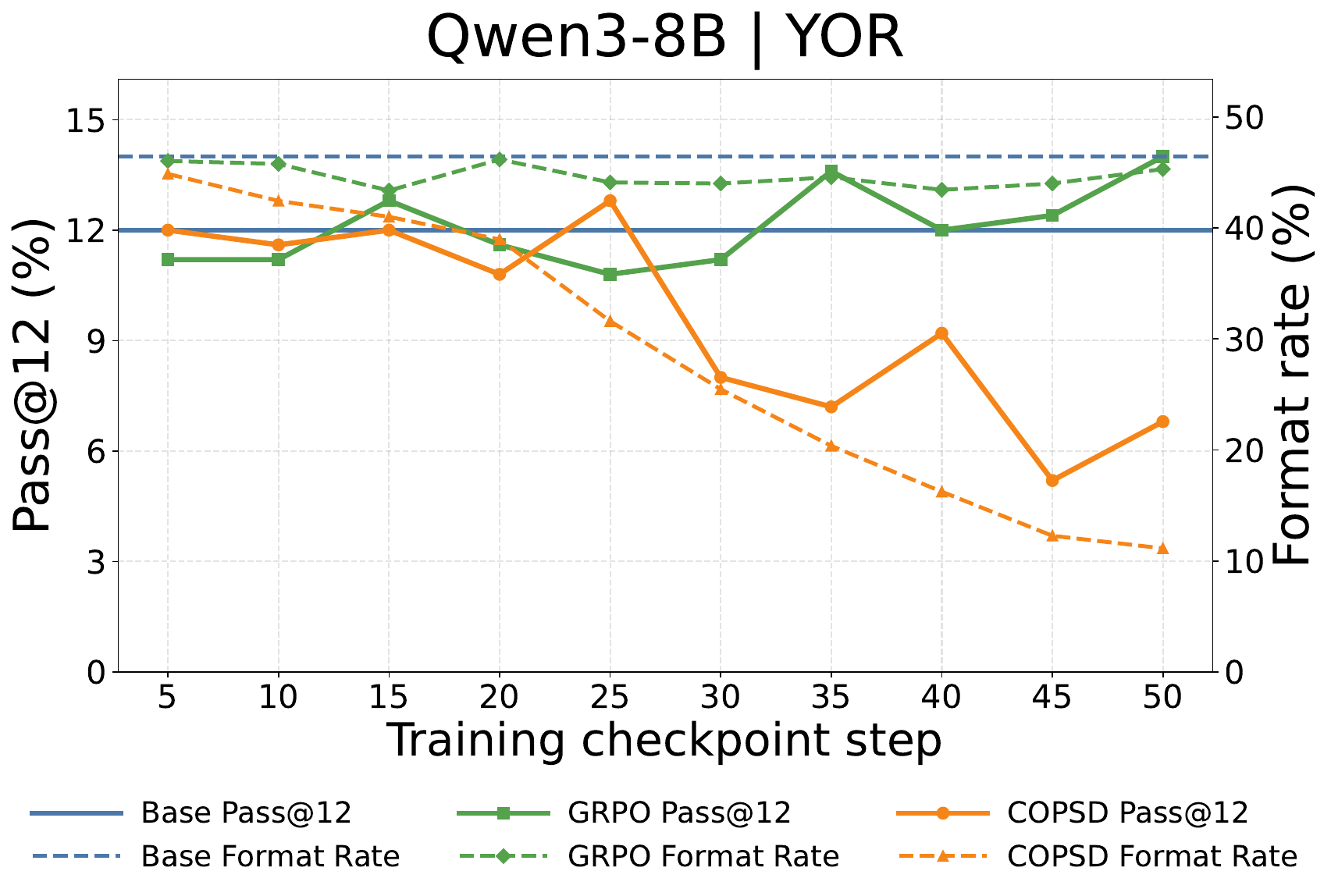}
    \includegraphics[width=0.32\linewidth]{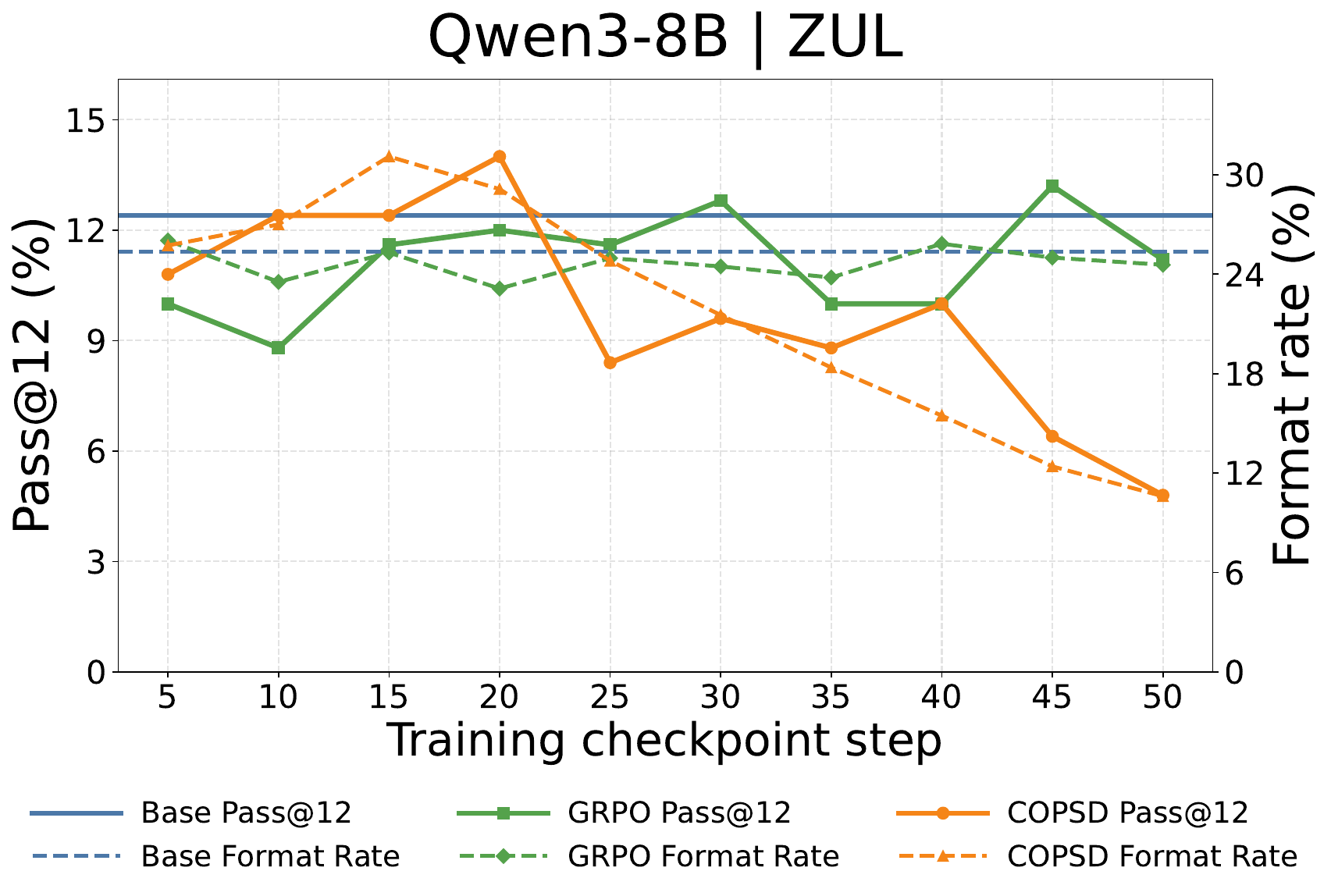}

    \caption{Per-language training dynamics for \texttt{Qwen3-8B} across all African languages under a 1024-token generation budget. Solid lines show Pass@12 and dashed lines show format rate.}
    \label{fig:training_dynamics_qwen3_8b_all_languages}
\end{figure*}

We provide the complete per-language training dynamics for all three model sizes in Figure~\ref{fig:training_dynamics_qwen3_1.7b_all_languages}, Figure~\ref{fig:training_dynamics_qwen3_4b_all_languages}, and Figure~\ref{fig:training_dynamics_qwen3_8b_all_languages}.
These figures complement the averaged results in Figure~\ref{fig:training_dynamics_average} and show that the main trends are broadly consistent across languages: \copsd typically improves Pass@12 and format rate within the early training steps, while GRPO often exhibits flatter or more unstable trajectories.
At the same time, the language-level plots reveal substantial variation across languages, suggesting that the effectiveness and saturation point of \copsd depend on both model scale and target-language generation quality.

\begin{figure*}[t]
    \centering
    \includegraphics[width=0.24\linewidth]{./figures/test_time_scaling/qwen3_1.7b/amh_test_time_scaling_pass_at_n_pct.pdf}
    \includegraphics[width=0.24\linewidth]{./figures/test_time_scaling/qwen3_1.7b/ewe_test_time_scaling_pass_at_n_pct.pdf}
    \includegraphics[width=0.24\linewidth]{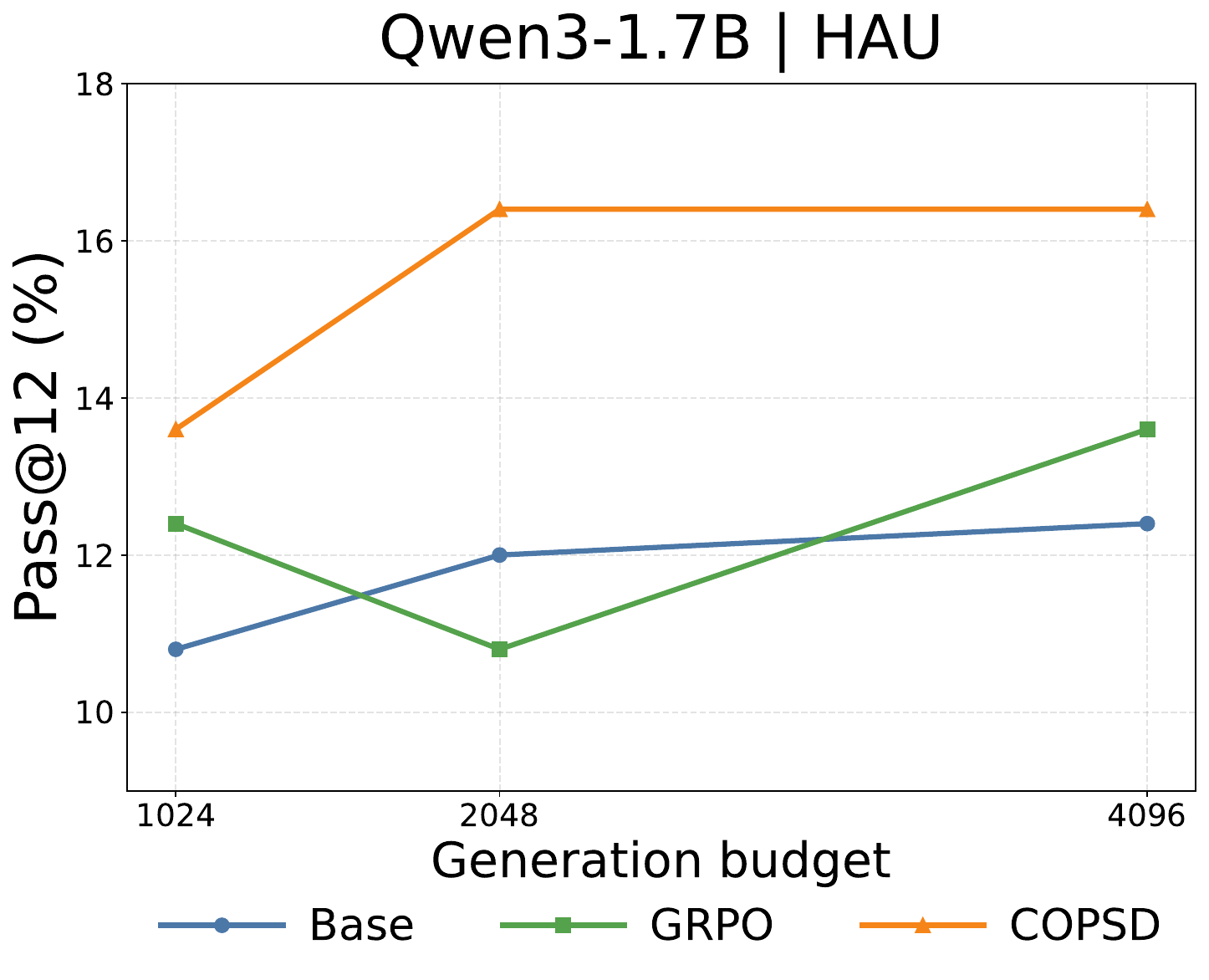}
    \includegraphics[width=0.24\linewidth]{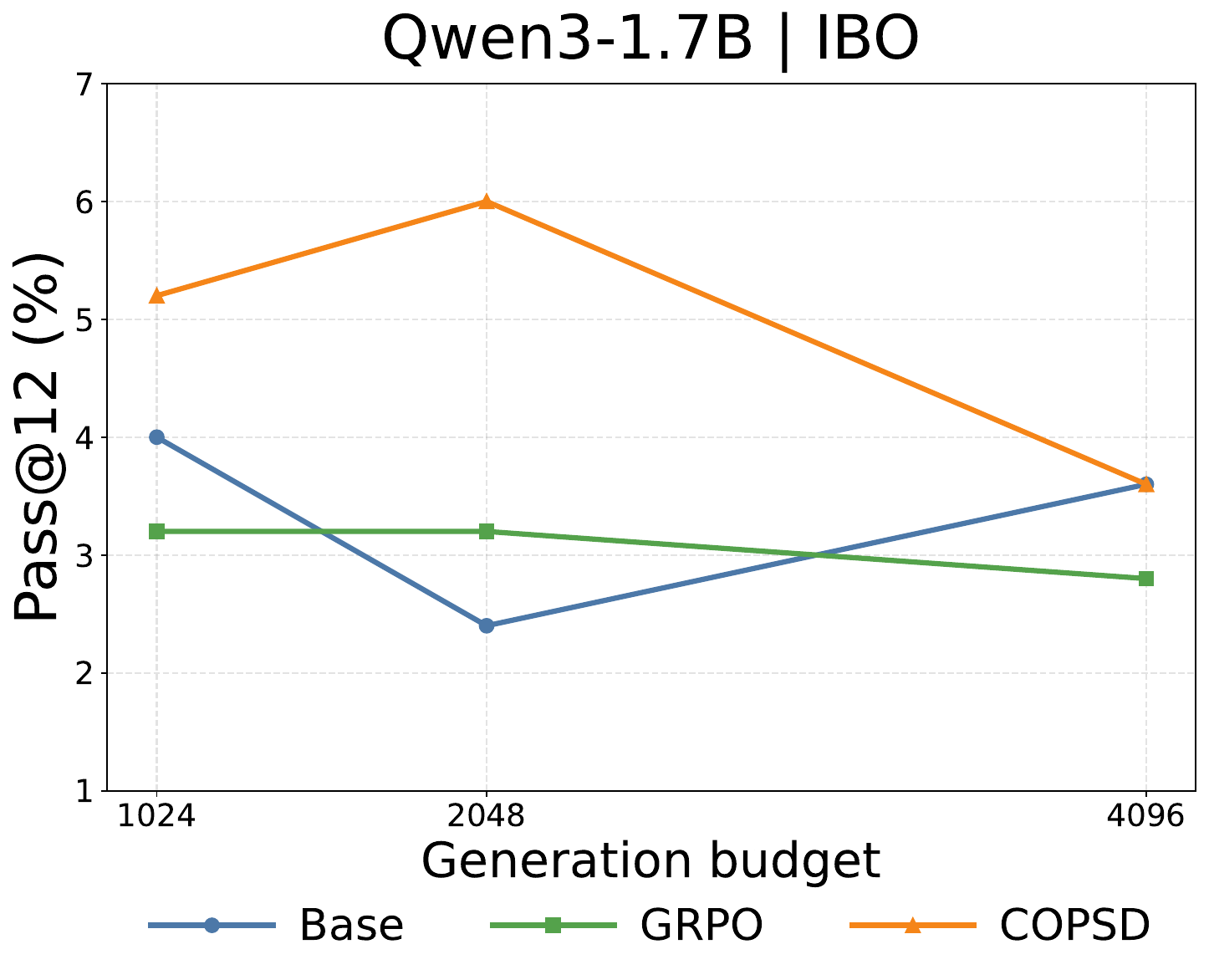}
    \includegraphics[width=0.24\linewidth]{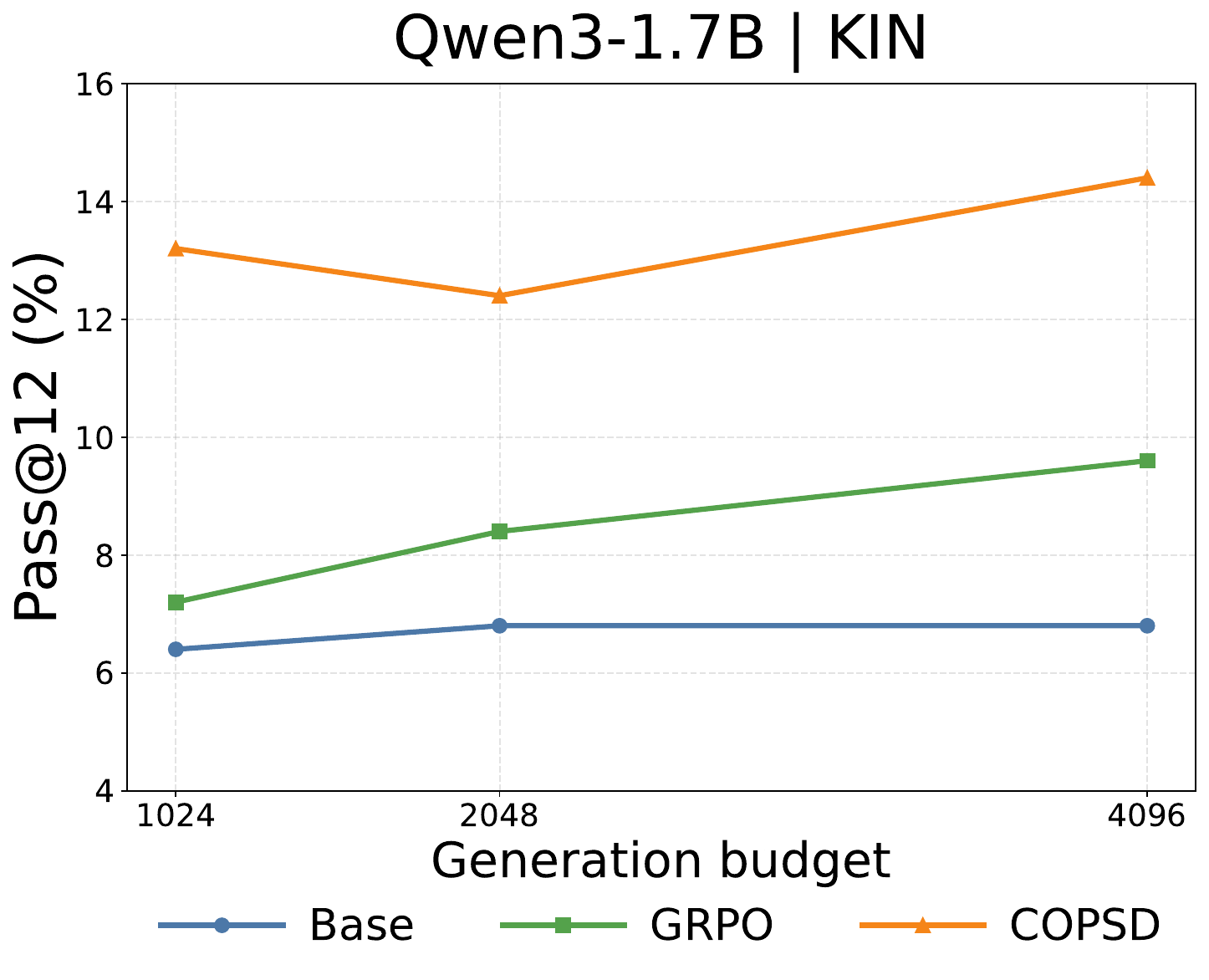}
    \includegraphics[width=0.24\linewidth]{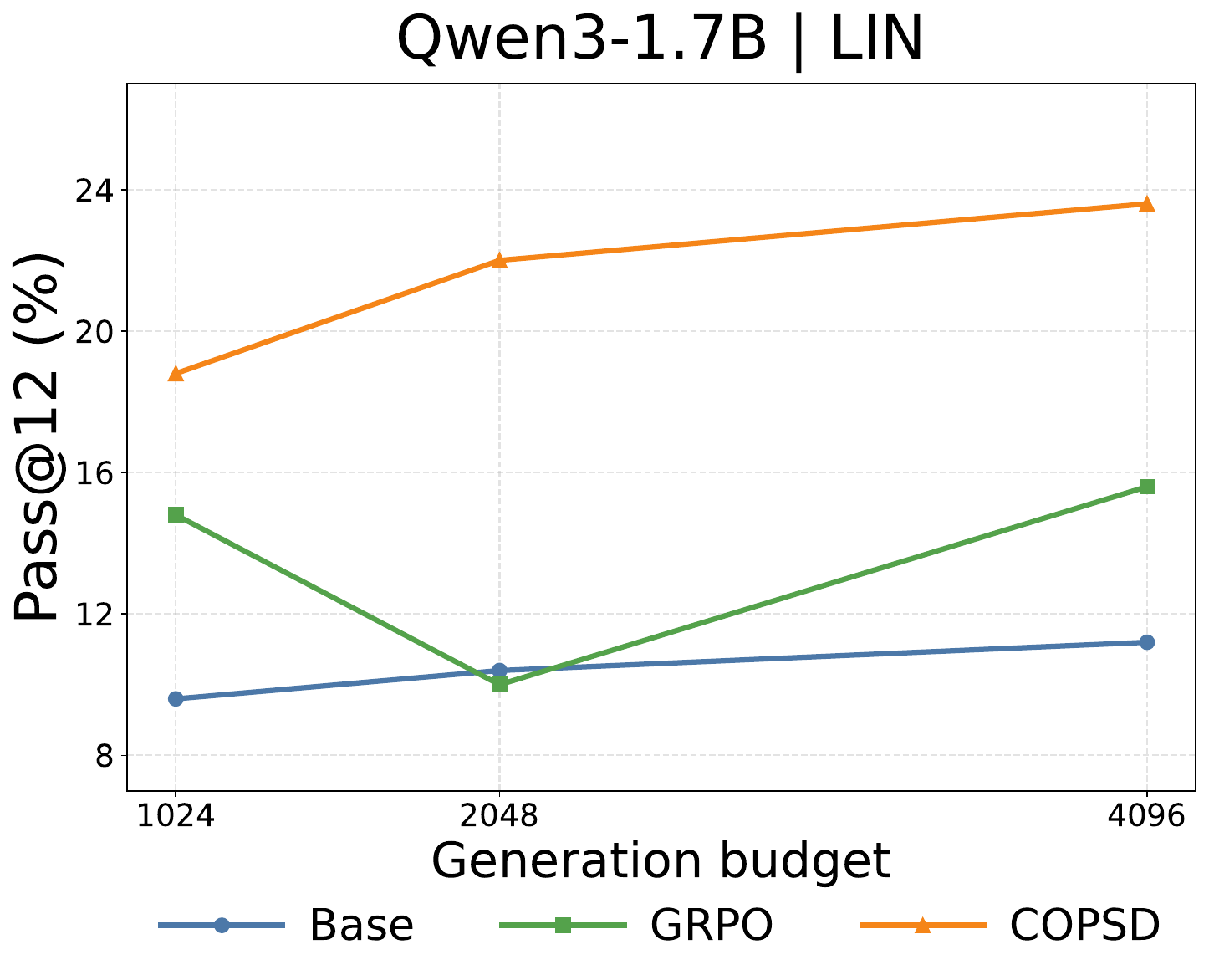}
    \includegraphics[width=0.24\linewidth]{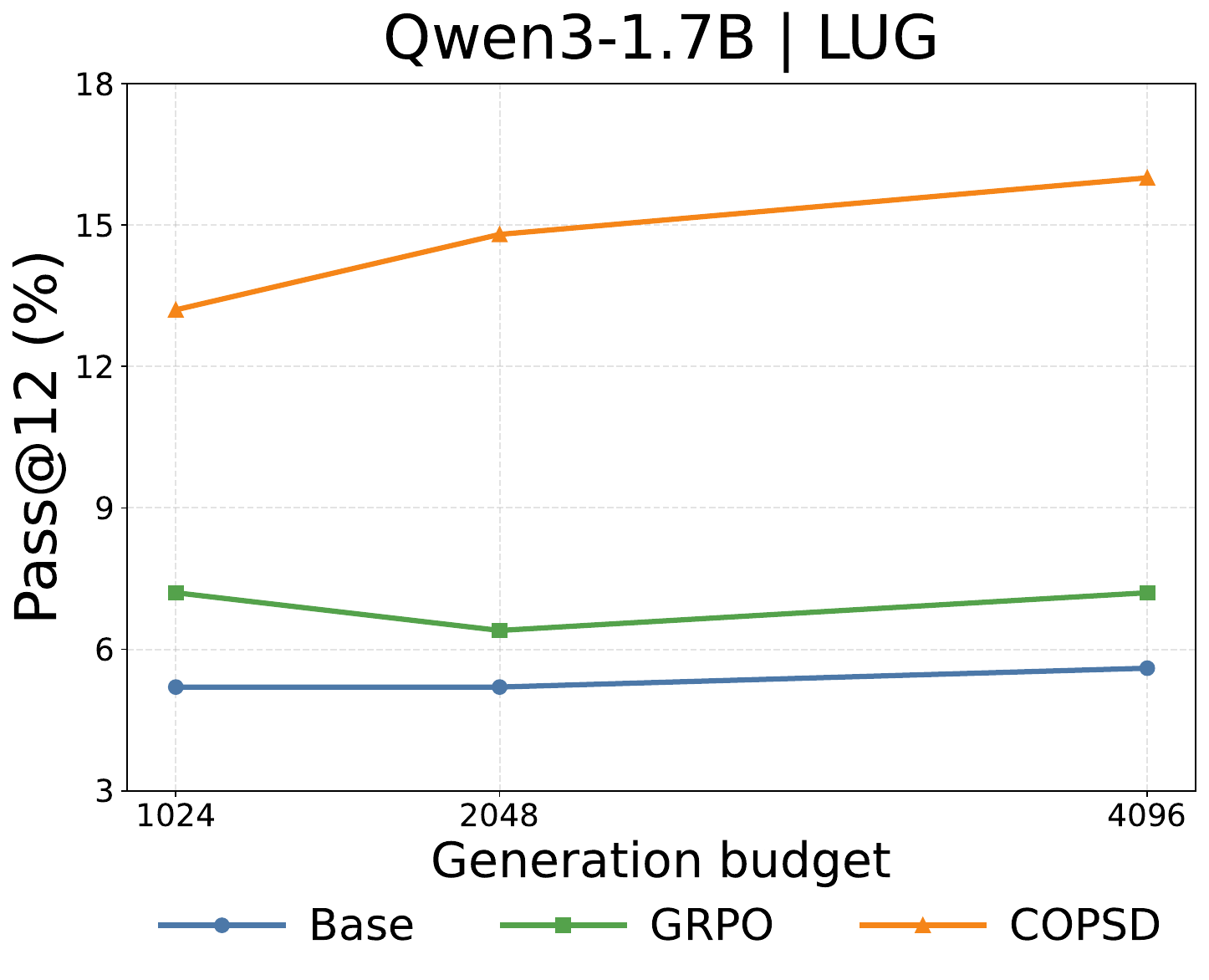}
    \includegraphics[width=0.24\linewidth]{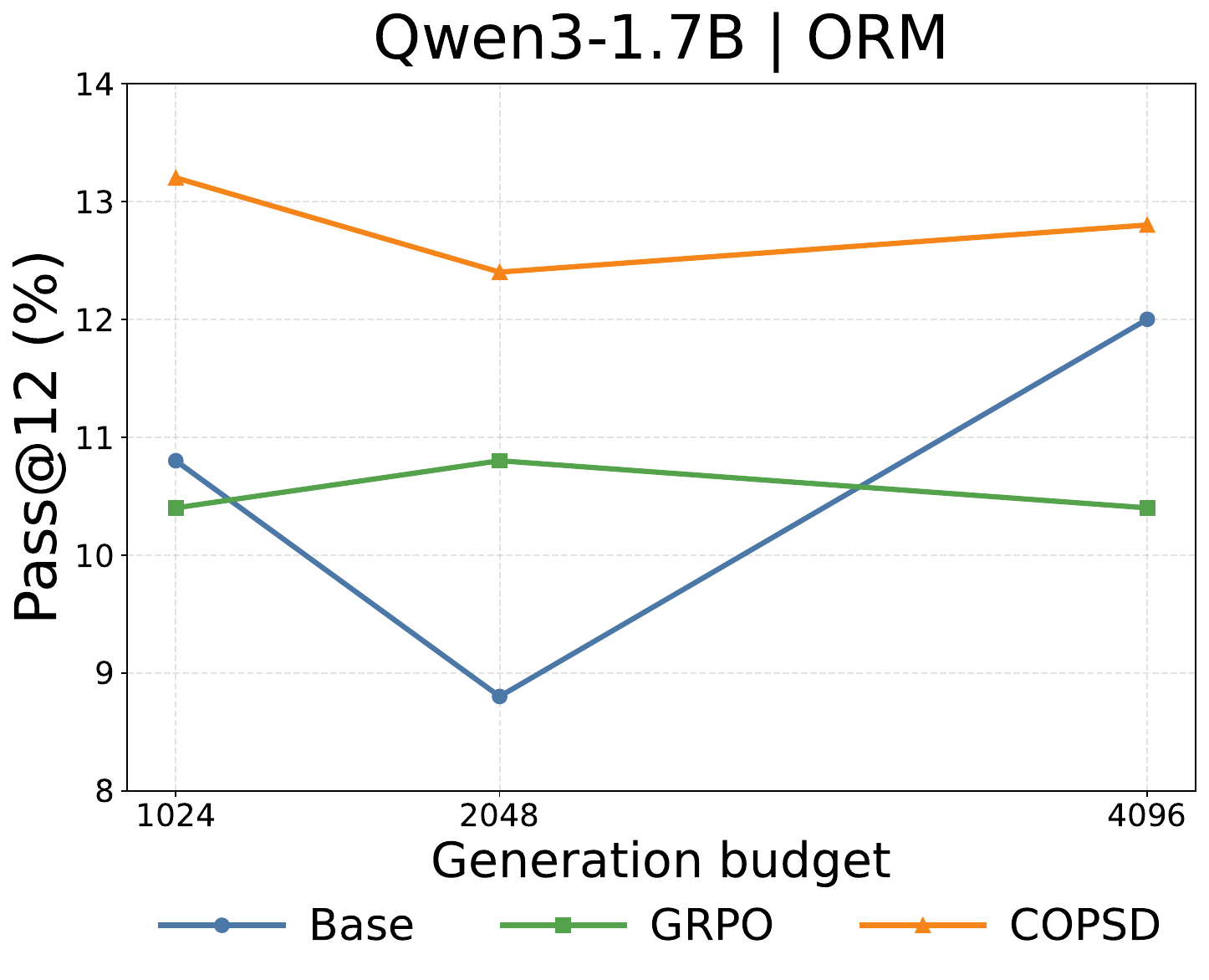}
    \includegraphics[width=0.24\linewidth]{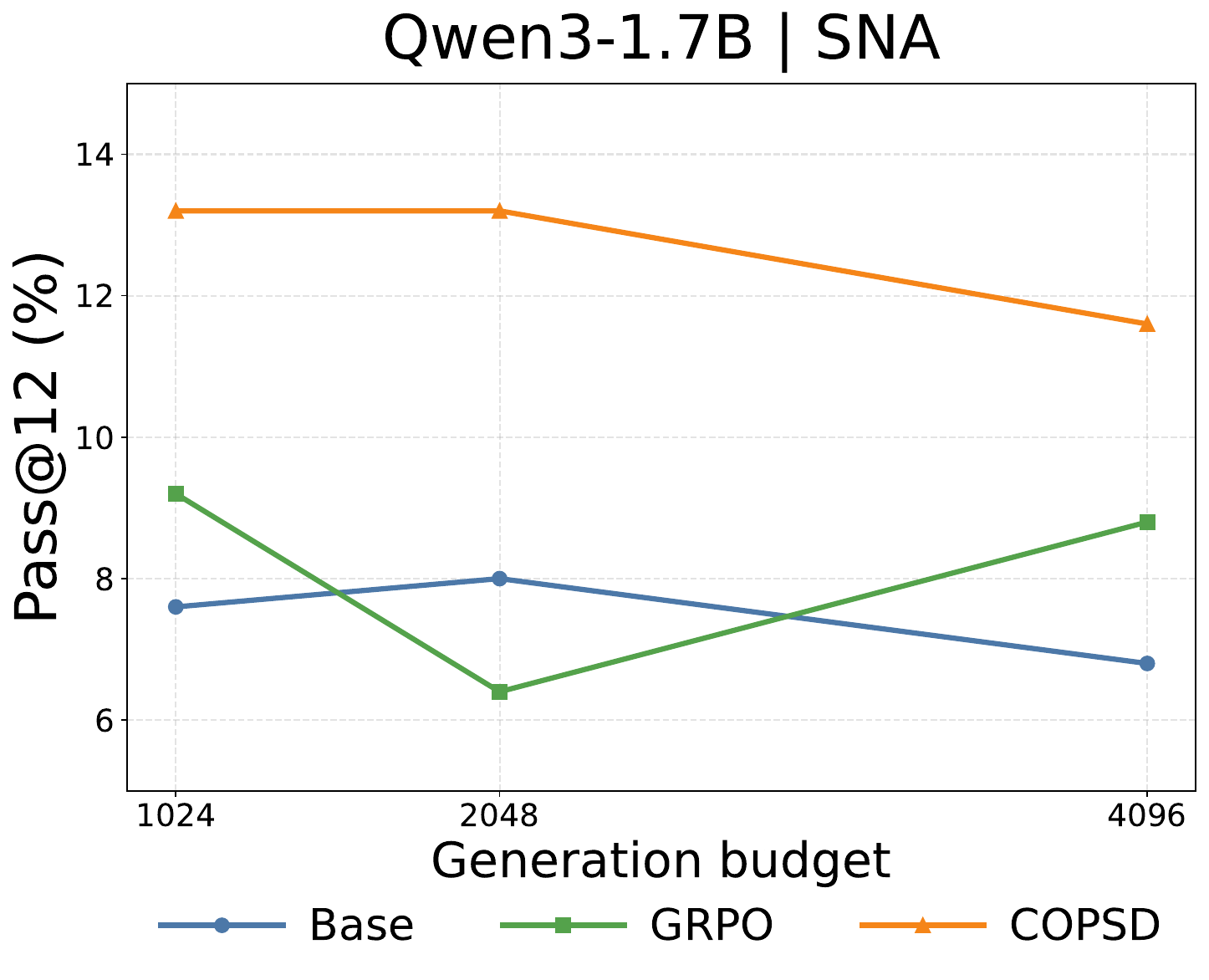}
    \includegraphics[width=0.24\linewidth]{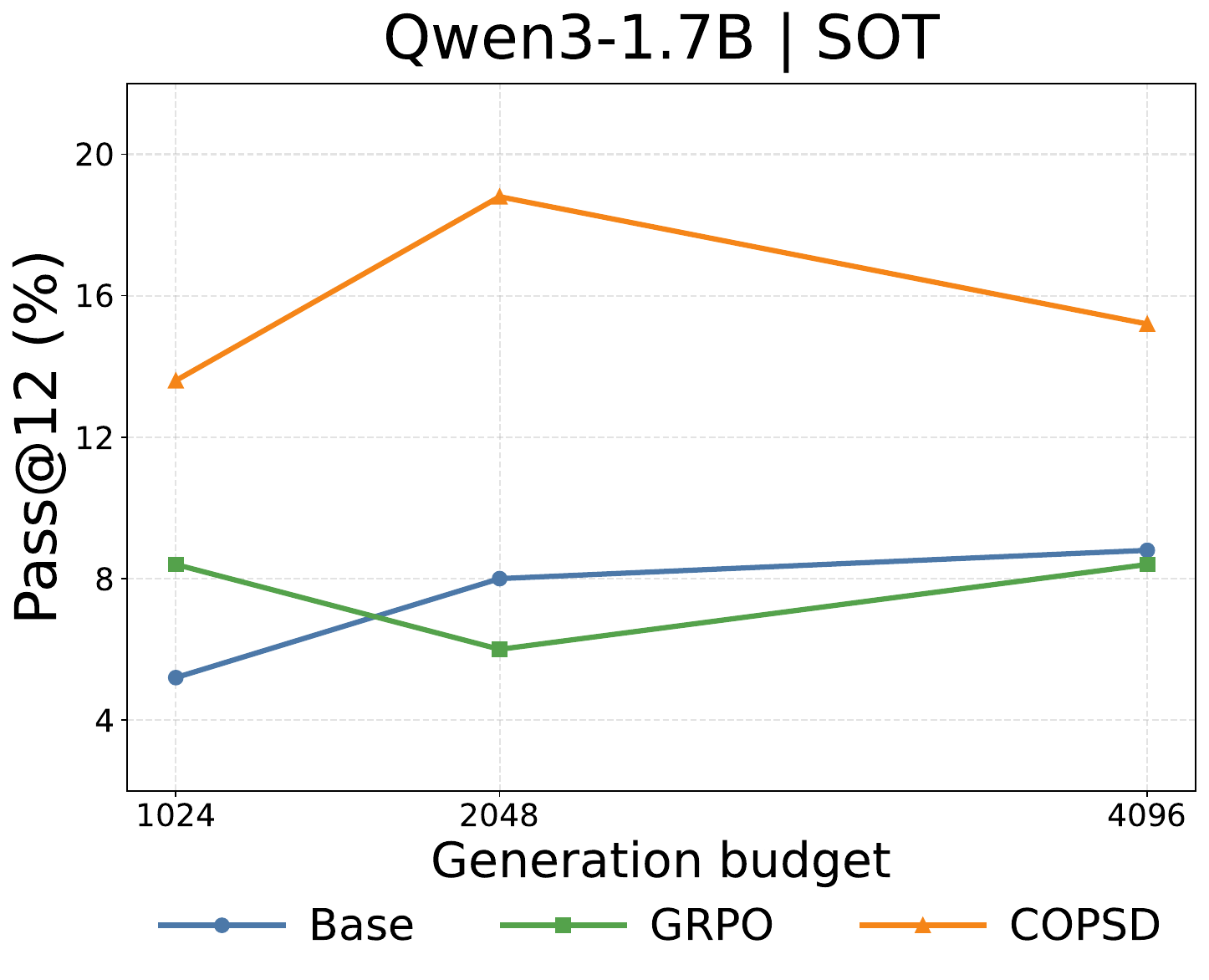}
    \includegraphics[width=0.24\linewidth]{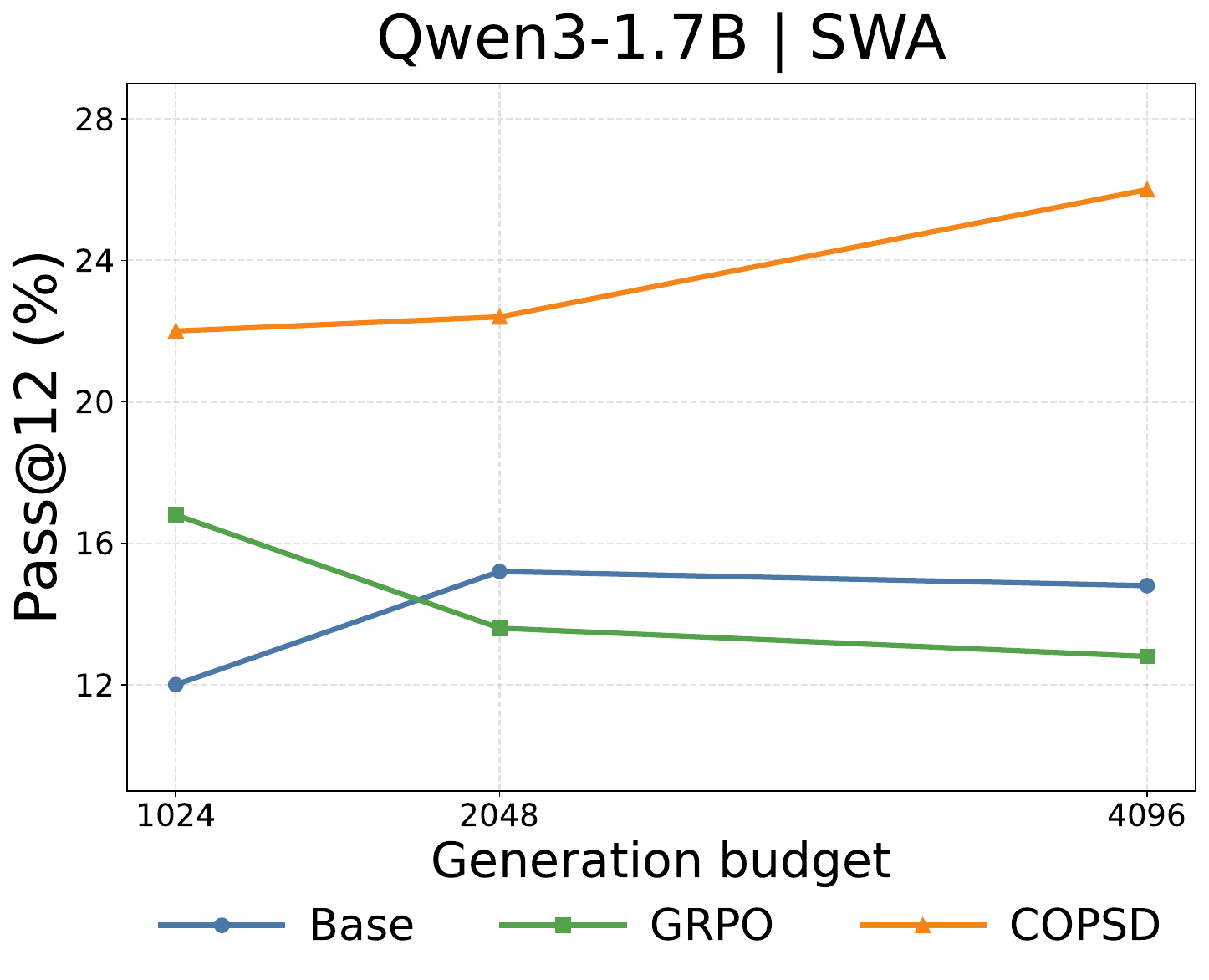}
    \includegraphics[width=0.24\linewidth]{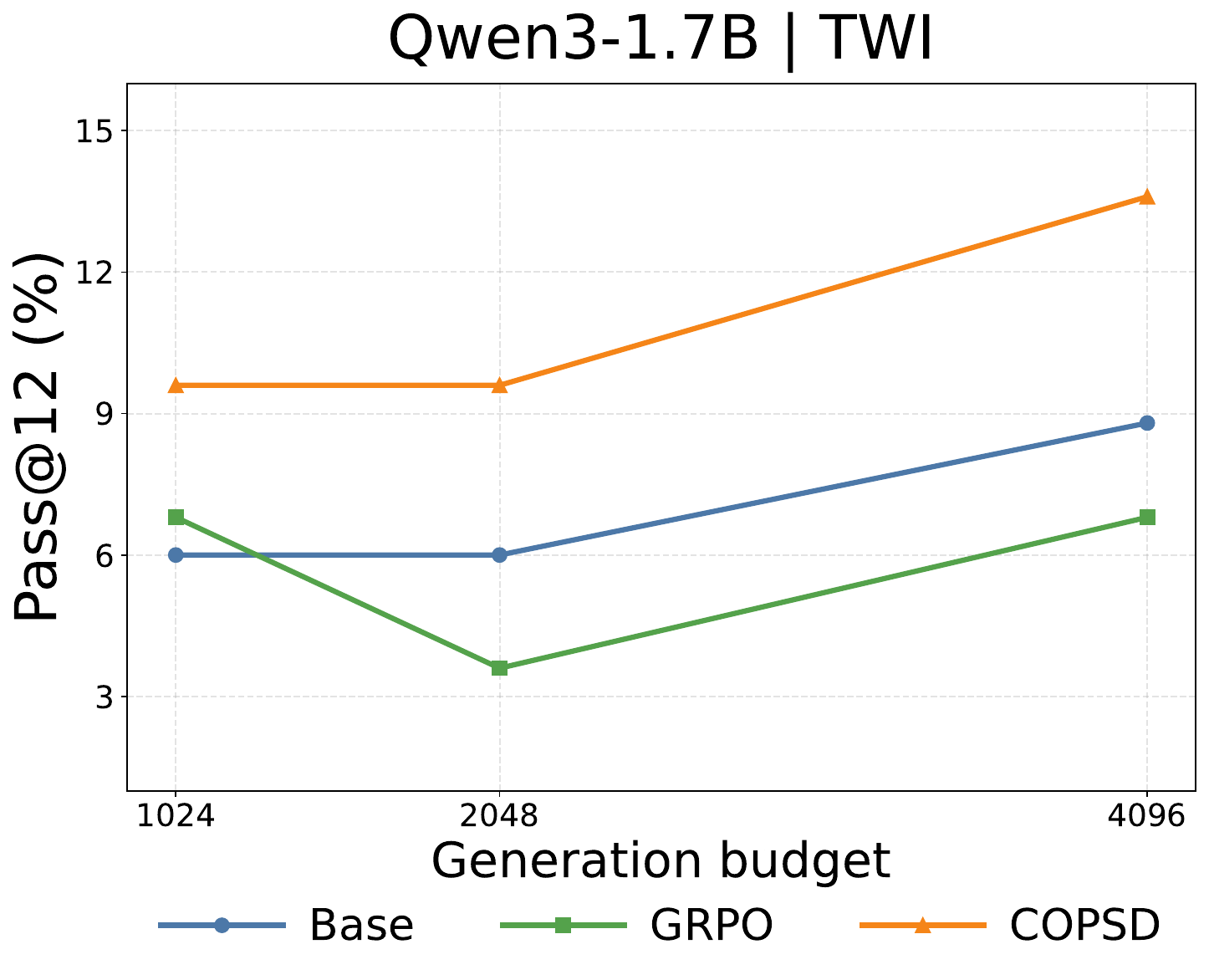}
    \includegraphics[width=0.24\linewidth]{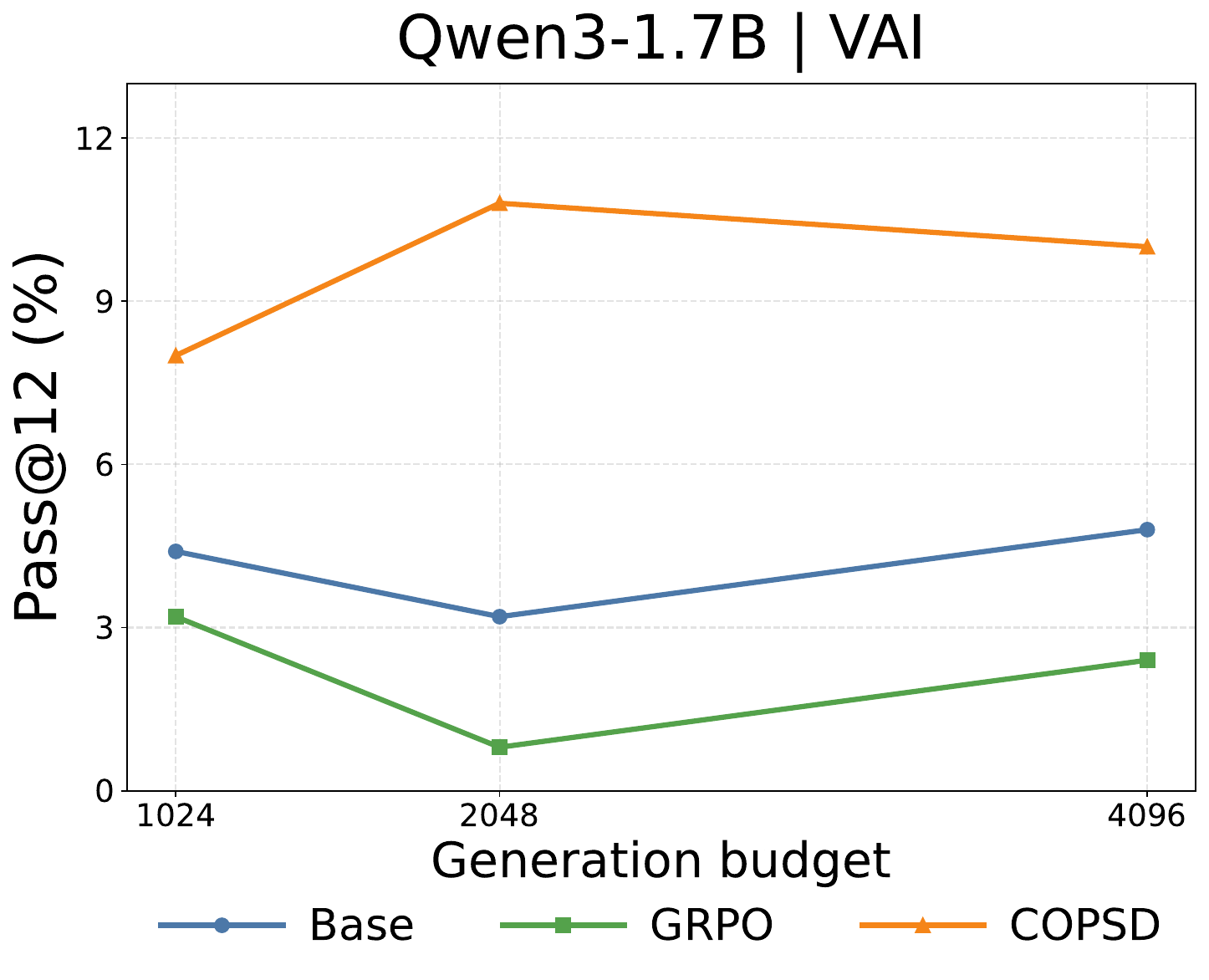}
    \includegraphics[width=0.24\linewidth]{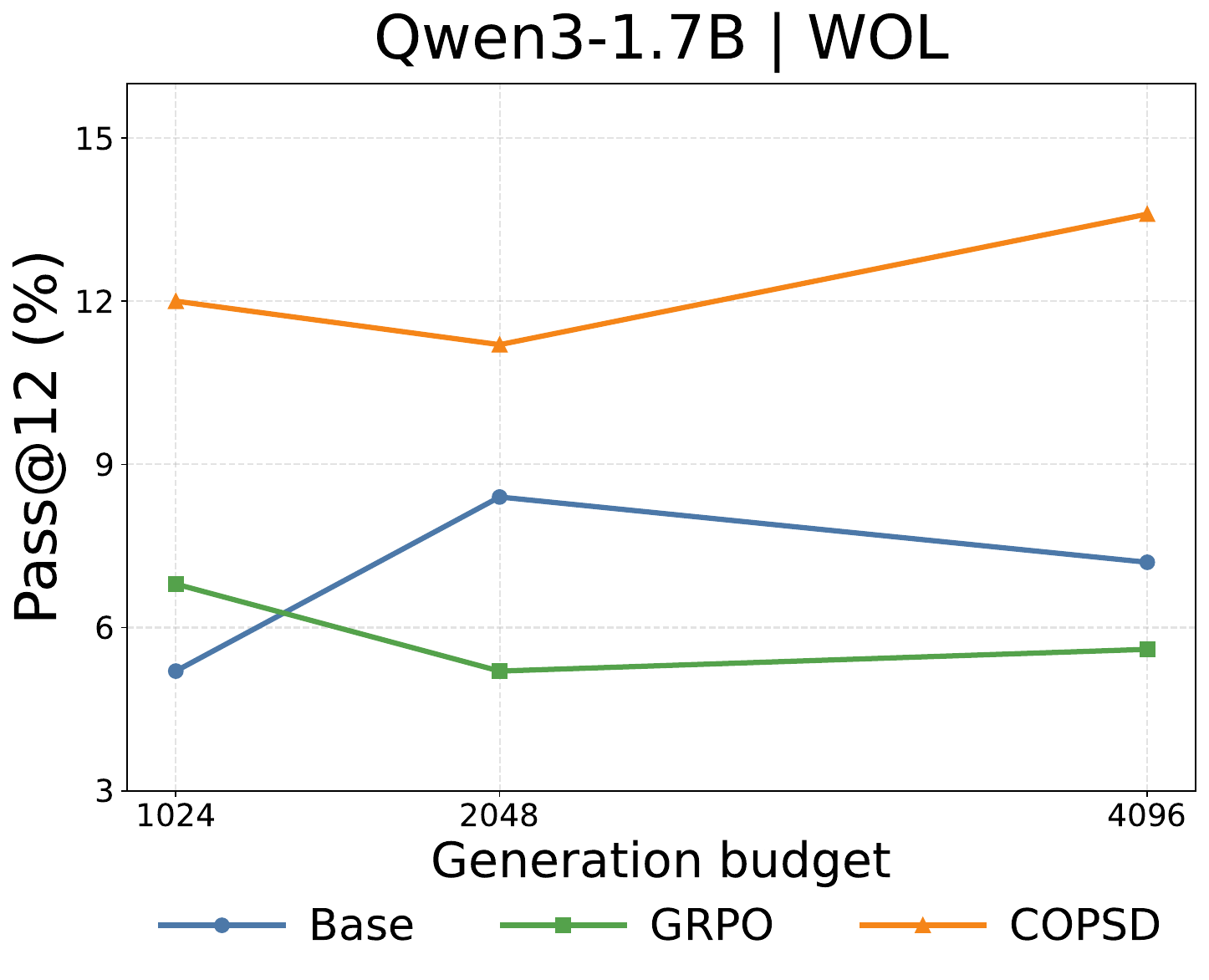}
    \includegraphics[width=0.24\linewidth]{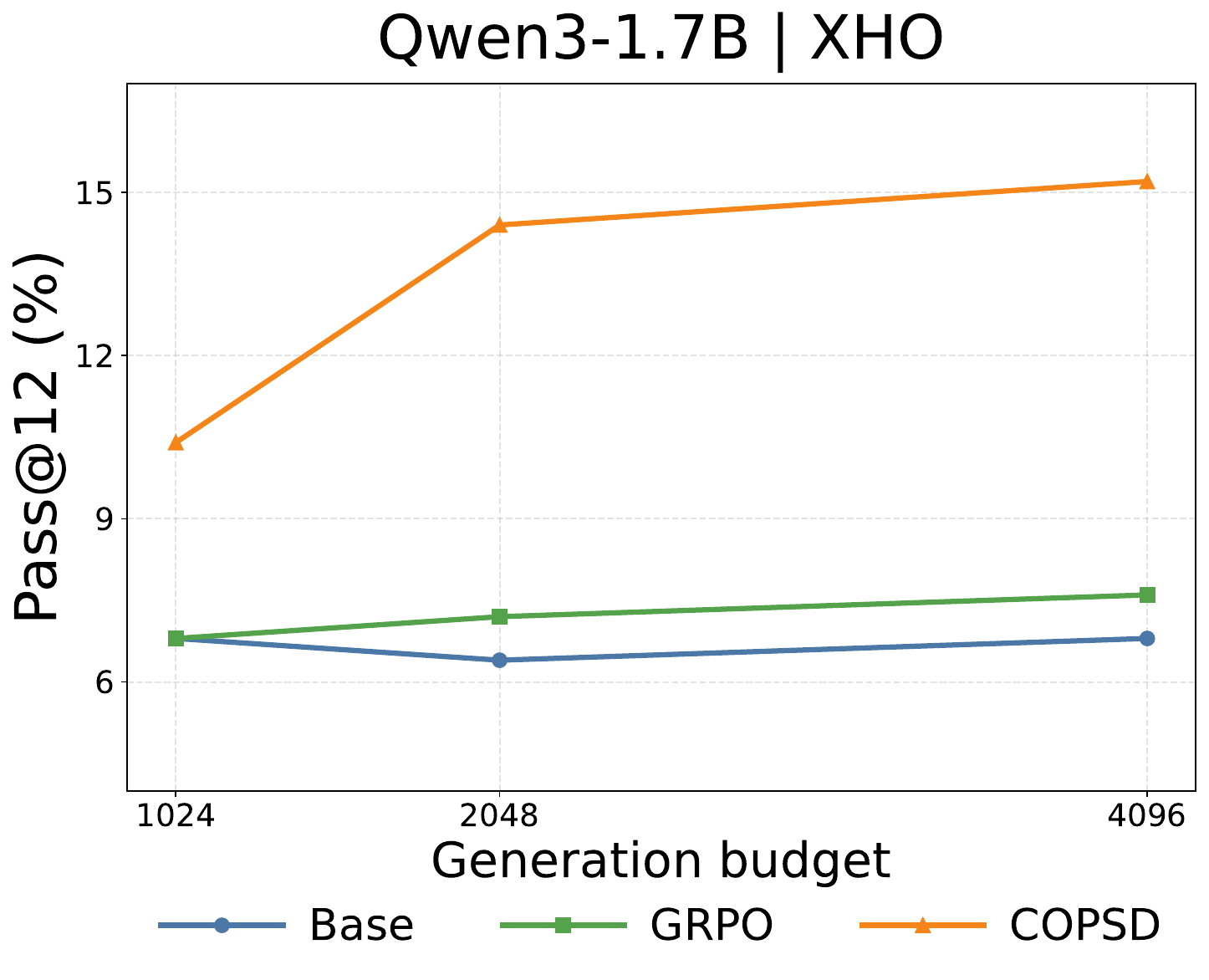}
    \includegraphics[width=0.24\linewidth]{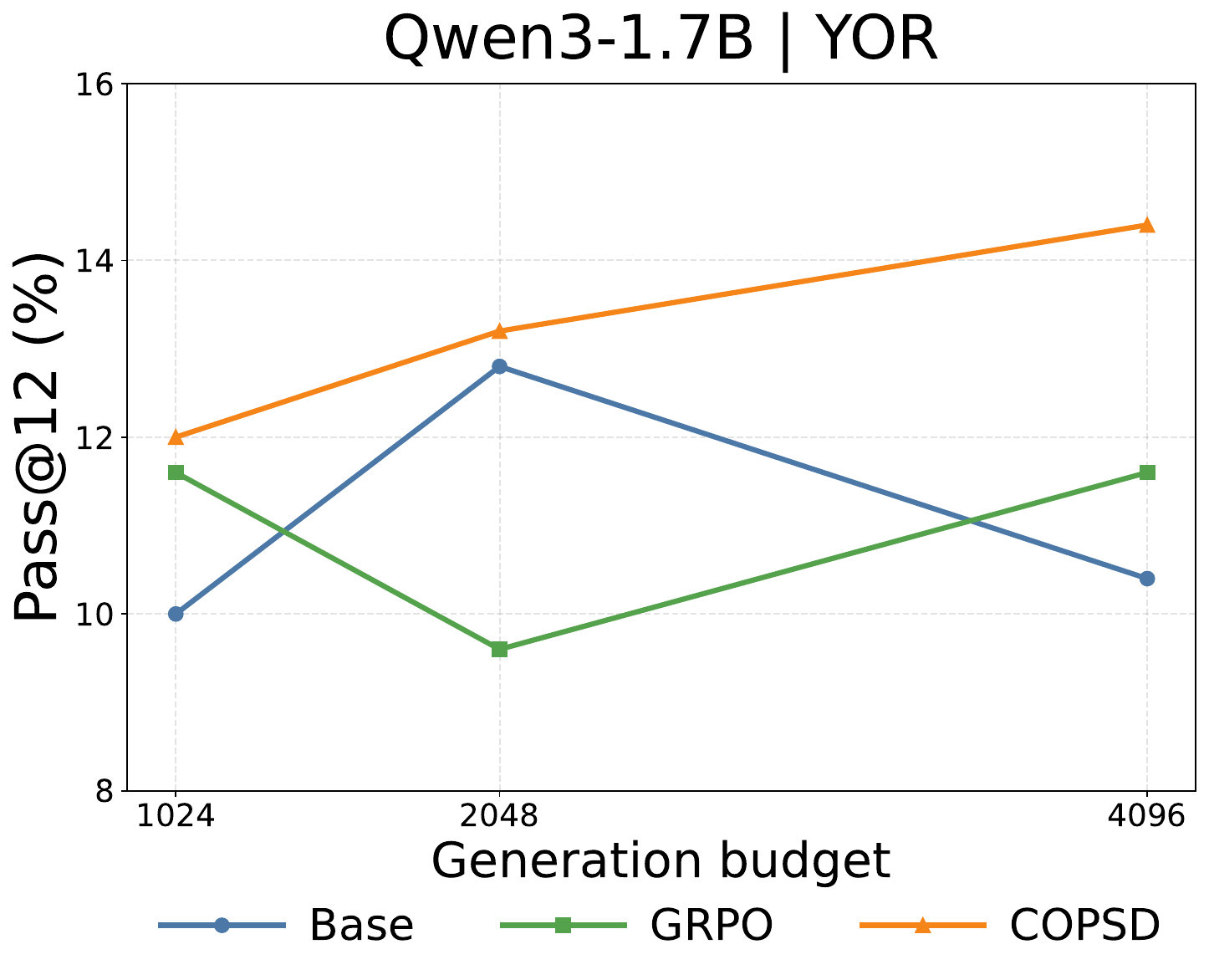}
    \includegraphics[width=0.24\linewidth]{./figures/test_time_scaling/qwen3_1.7b/zul_test_time_scaling_pass_at_n_pct.pdf}
    \caption{Per-language test-time scaling results on Pass@12 for \texttt{Qwen3-1.7B} across all African languages, under generation budgets of 1024, 2048, and 4096 tokens. Overall, the trends are mixed across languages, but \copsd generally achieves stronger performance than both the Base model and GRPO under different generation budgets.}
    \label{fig:test_time_scaling_qwen3_1.7b_all_languages}
\end{figure*}

\begin{figure*}[t]
    \centering
    \includegraphics[width=0.24\linewidth]{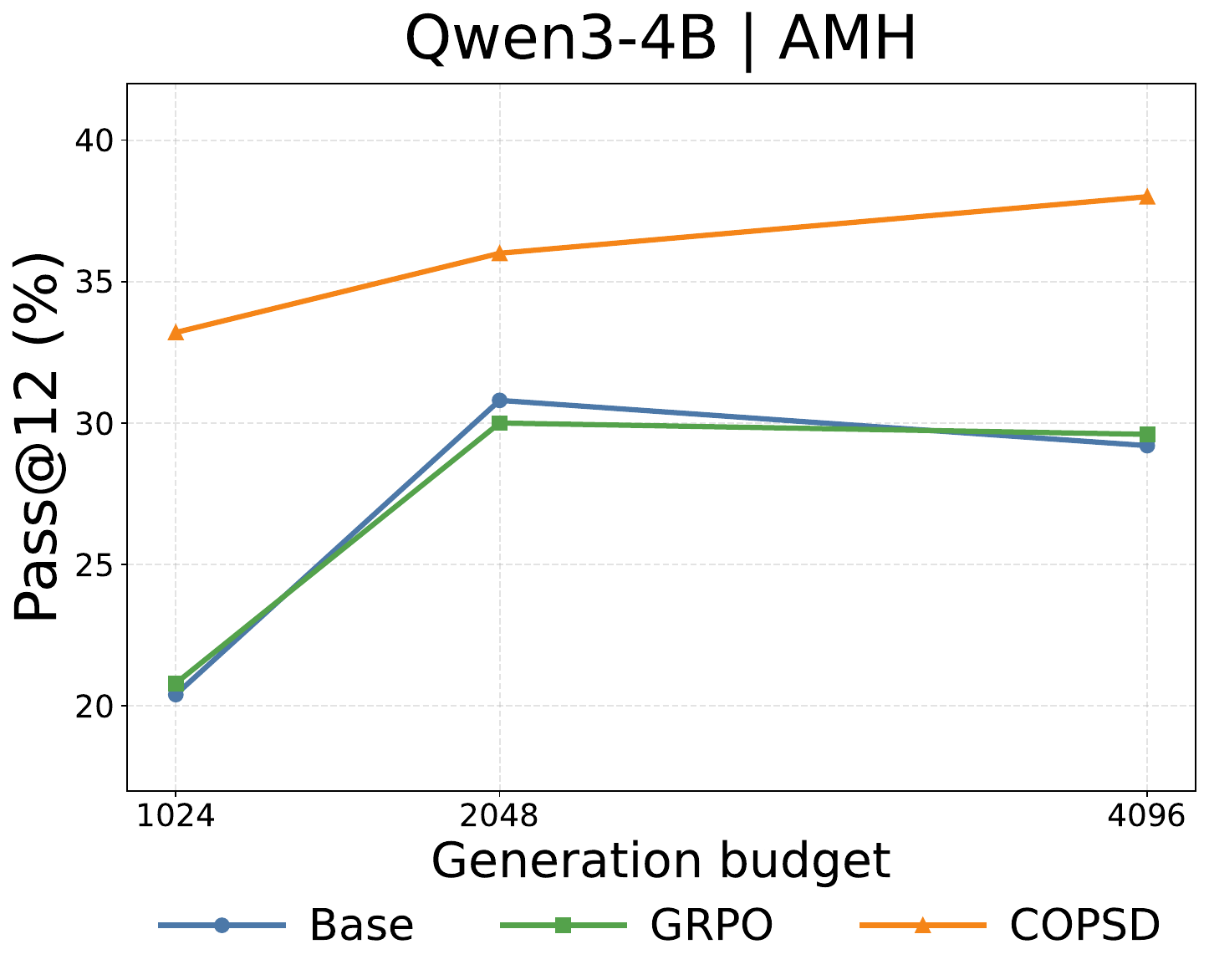}
    \includegraphics[width=0.24\linewidth]{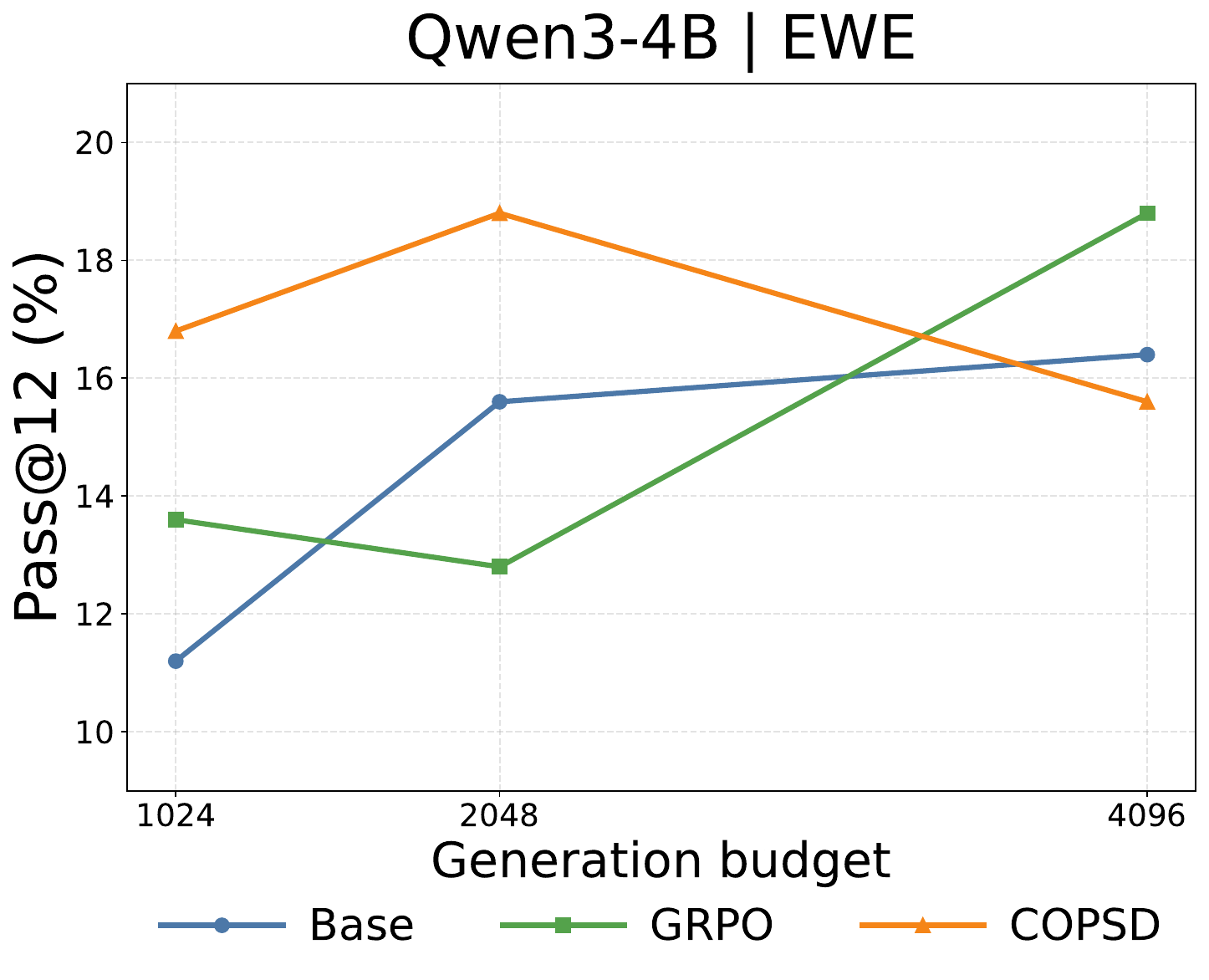}
    \includegraphics[width=0.24\linewidth]{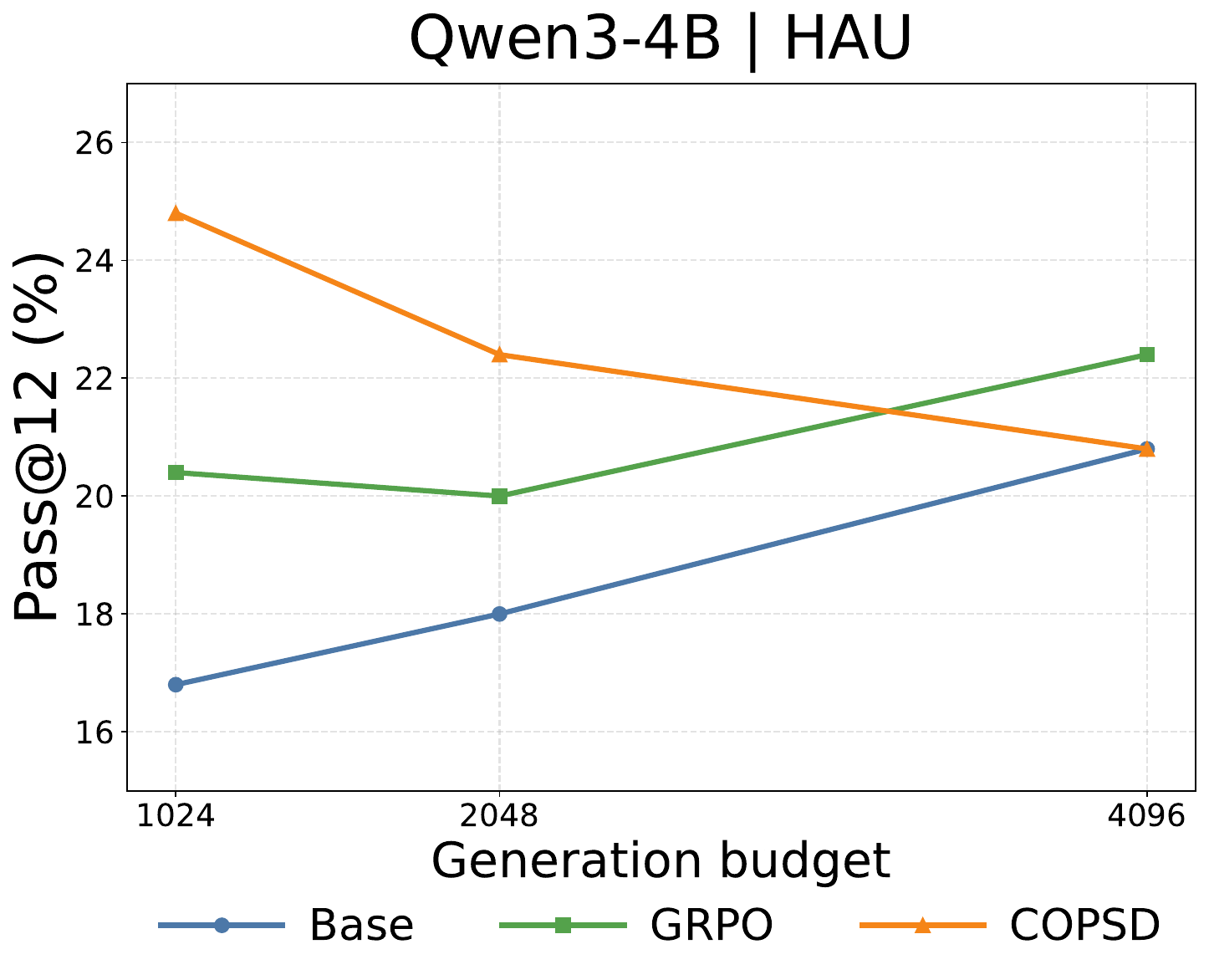}
    \includegraphics[width=0.24\linewidth]{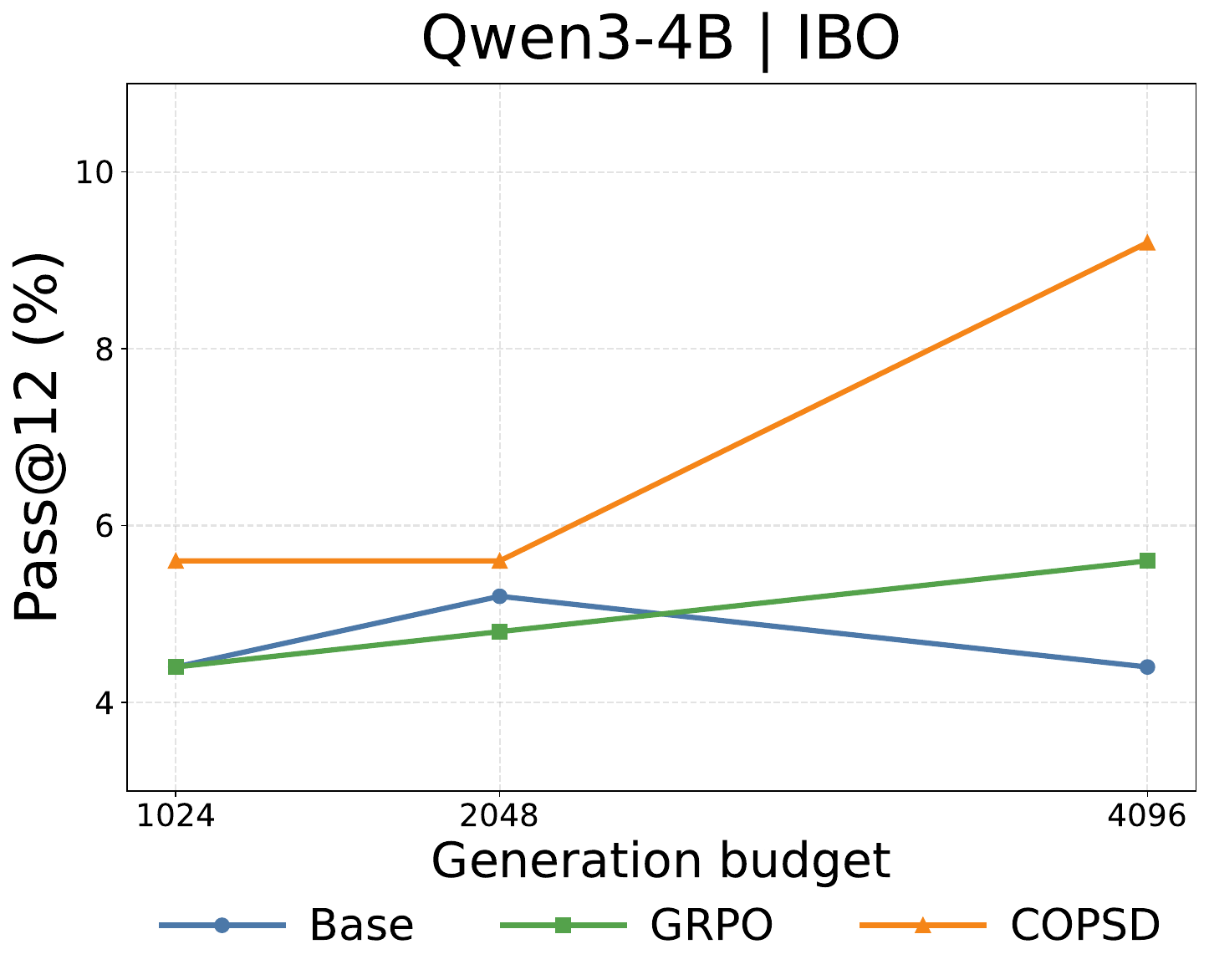}
    \includegraphics[width=0.24\linewidth]{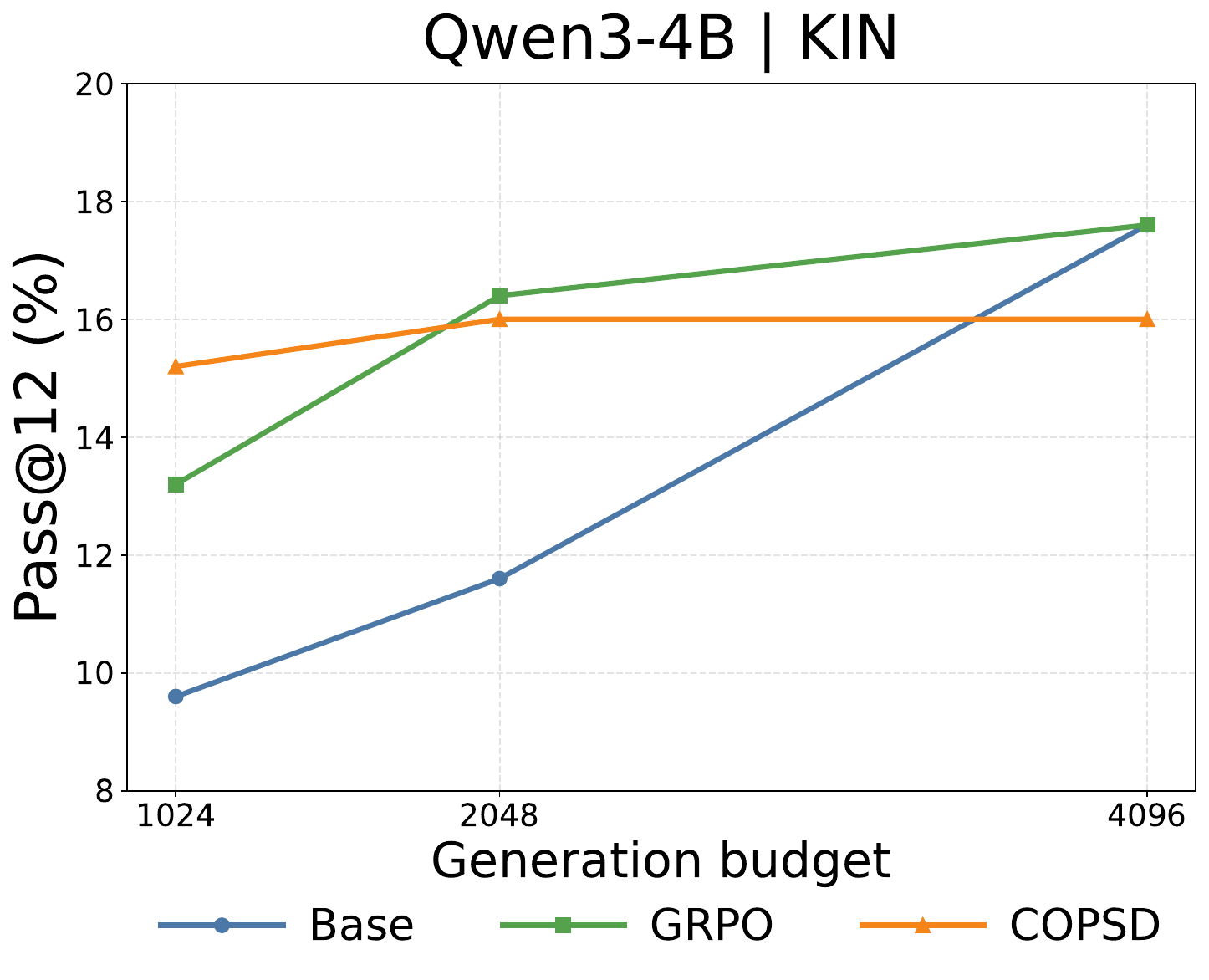}
    \includegraphics[width=0.24\linewidth]{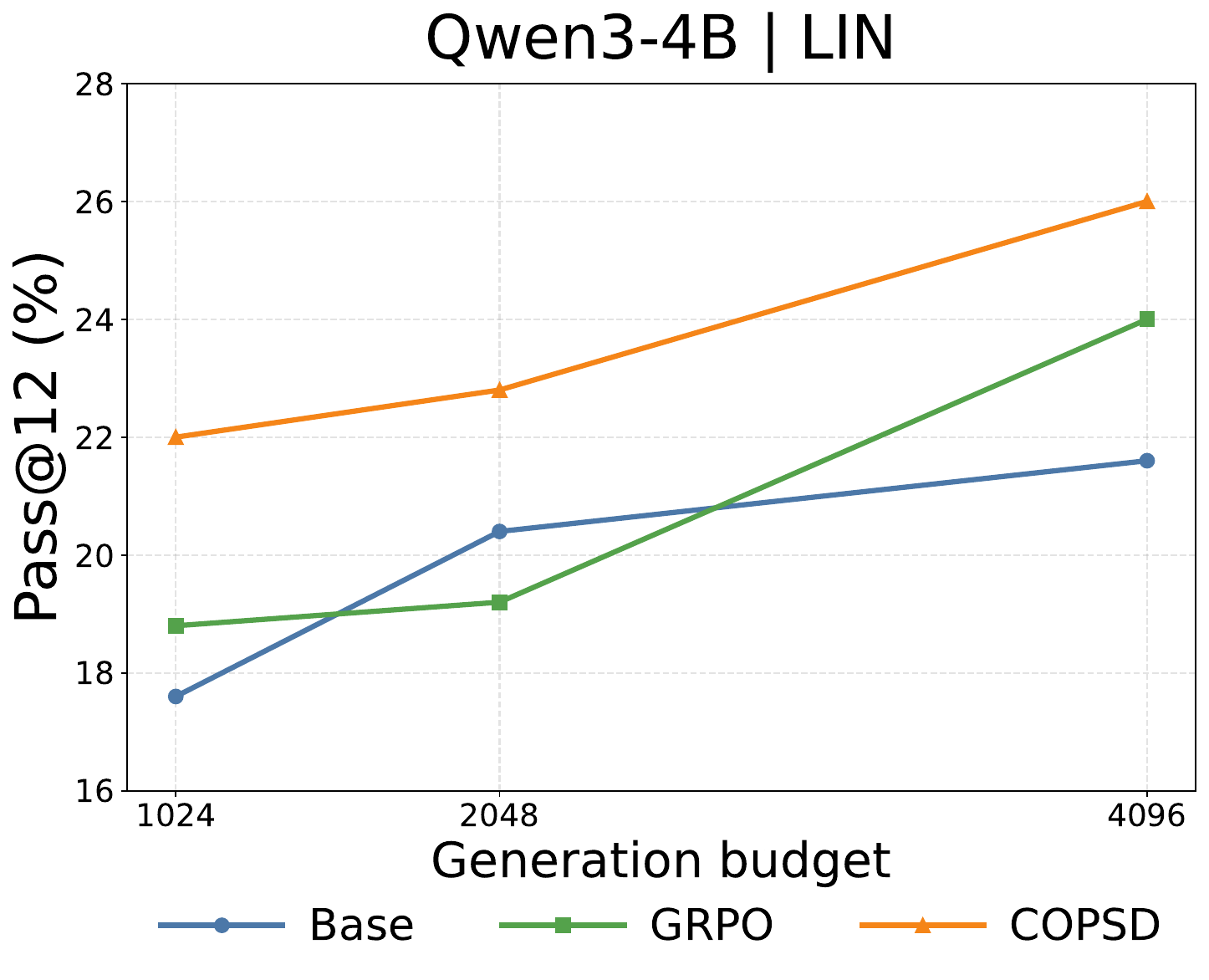}
    \includegraphics[width=0.24\linewidth]{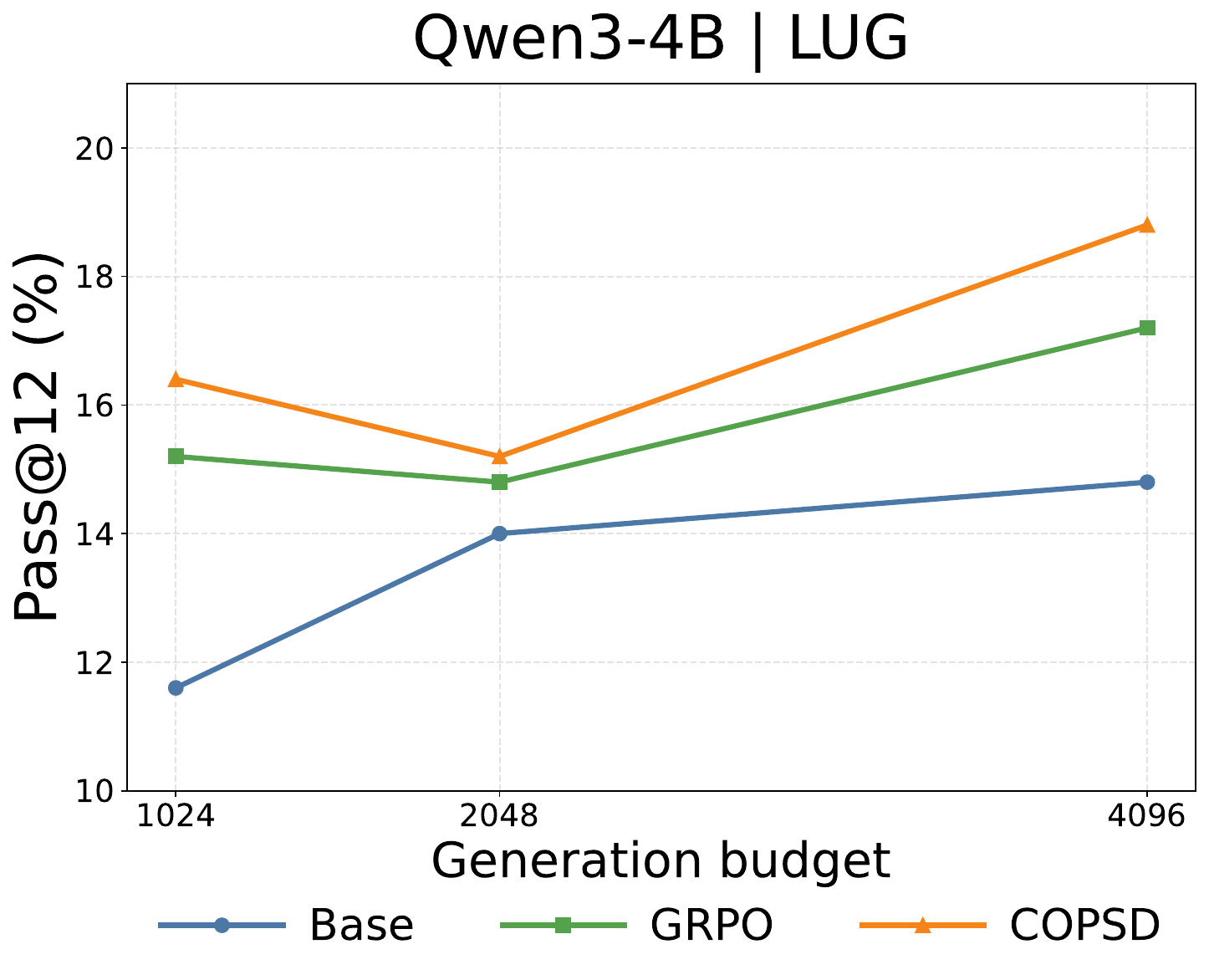}
    \includegraphics[width=0.24\linewidth]{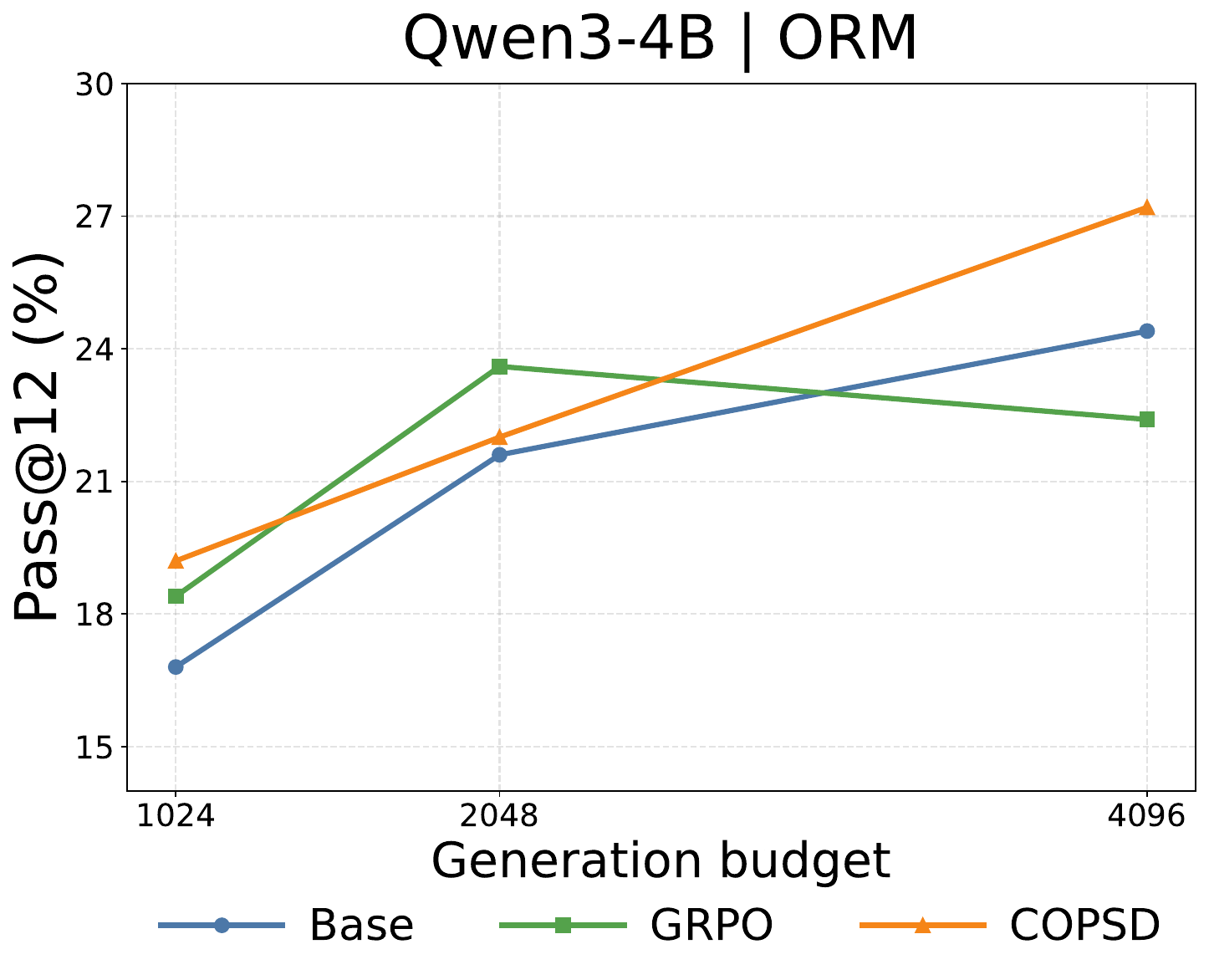}
    \includegraphics[width=0.24\linewidth]{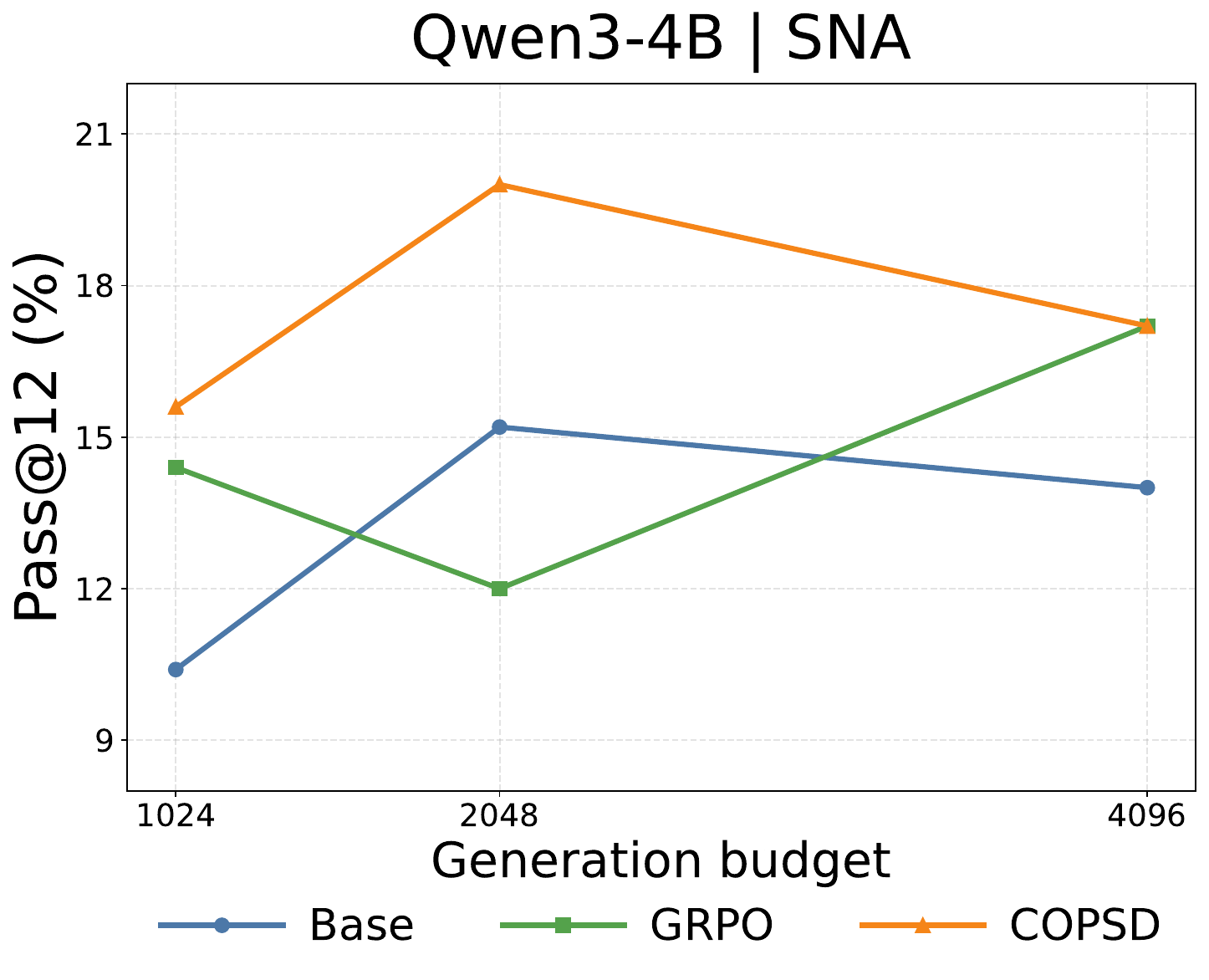}
    \includegraphics[width=0.24\linewidth]{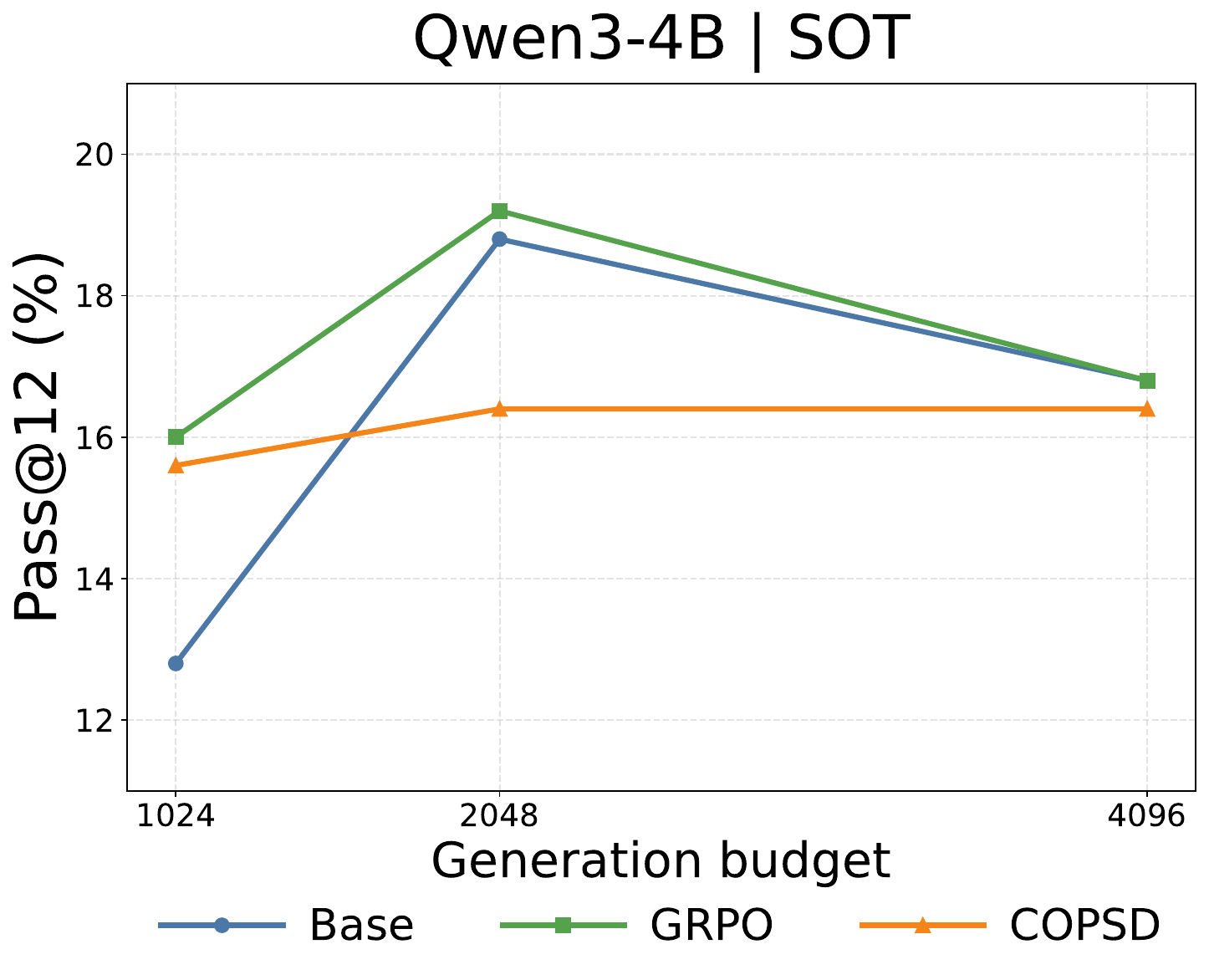}
    \includegraphics[width=0.24\linewidth]{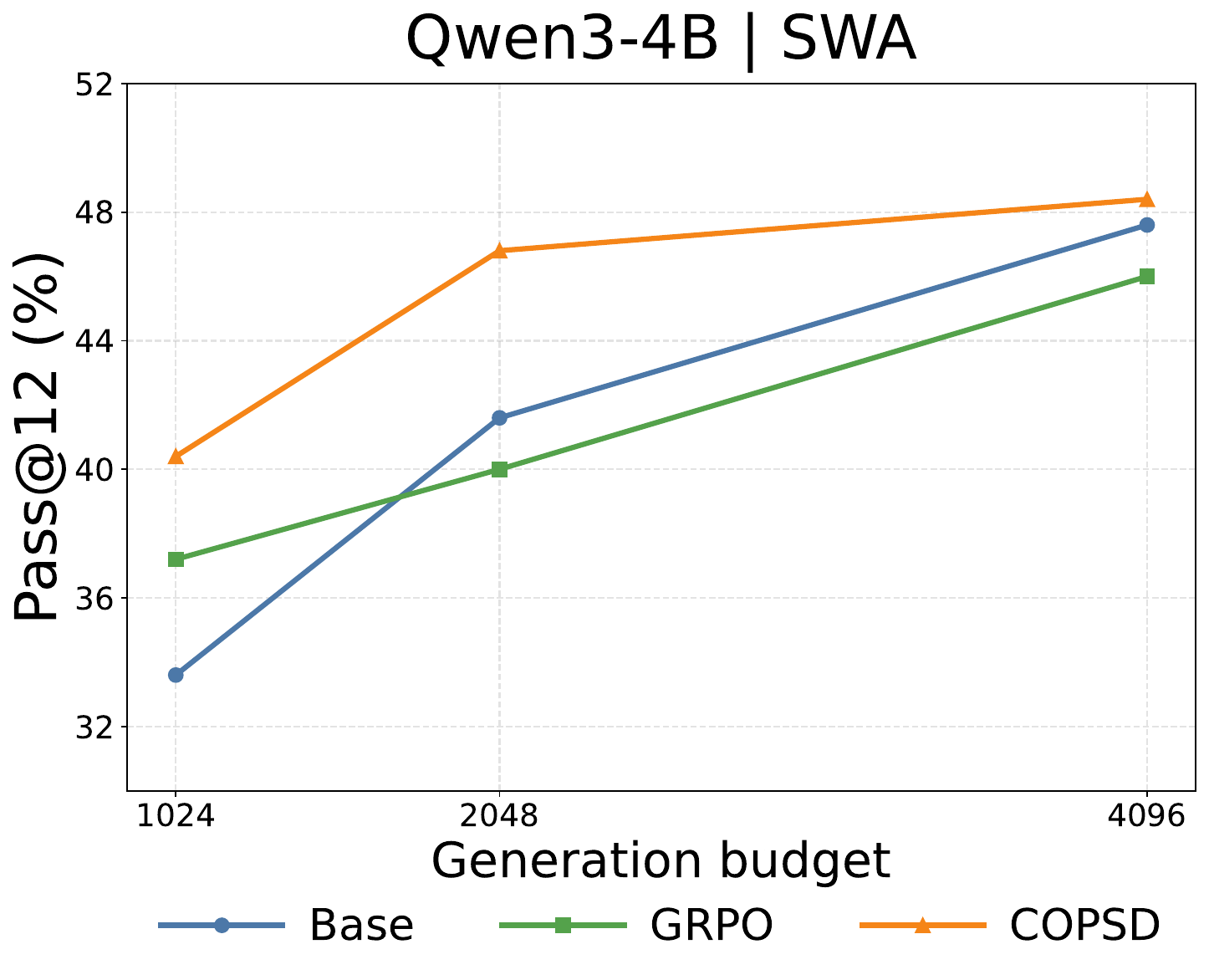}
    \includegraphics[width=0.24\linewidth]{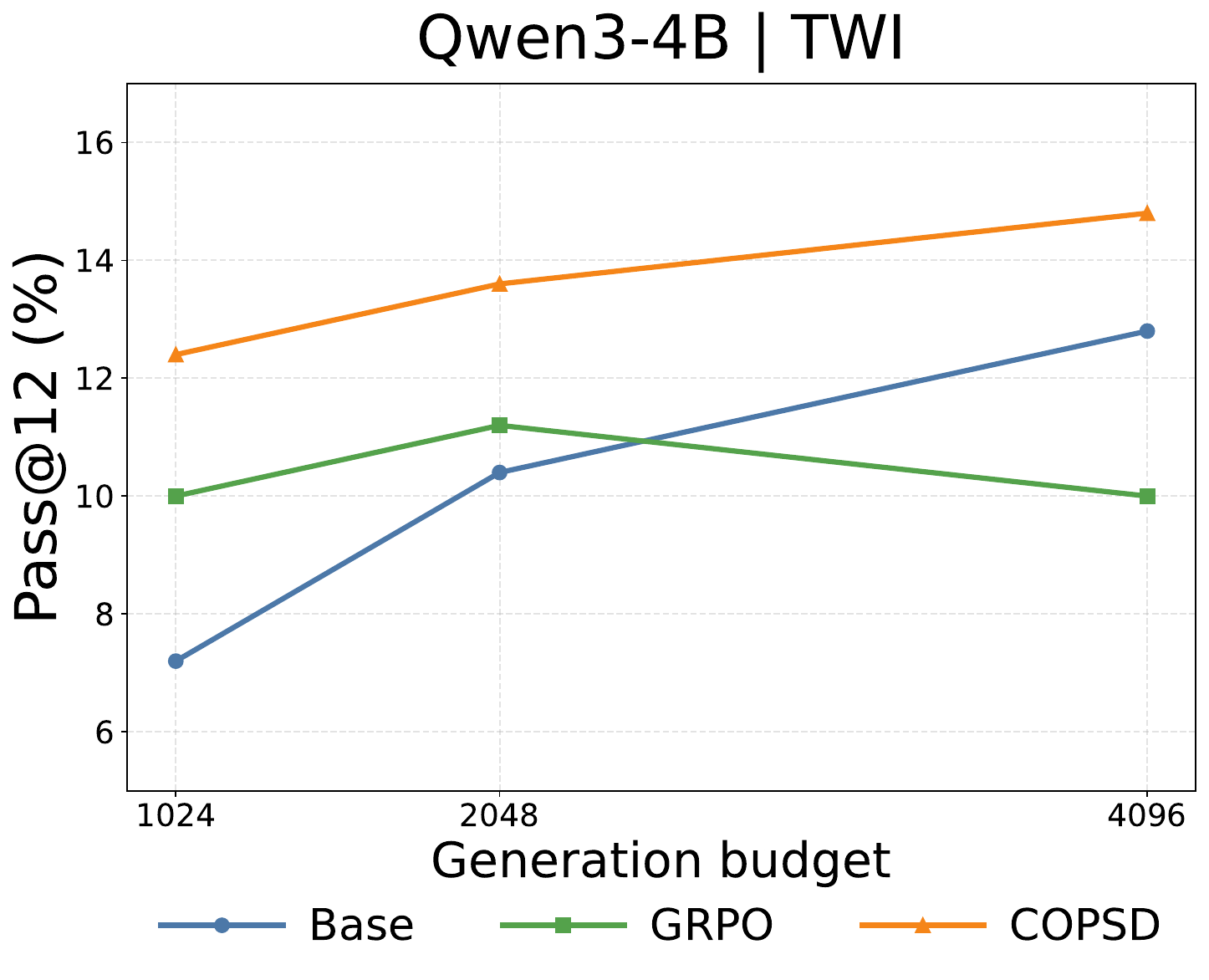}
    \includegraphics[width=0.24\linewidth]{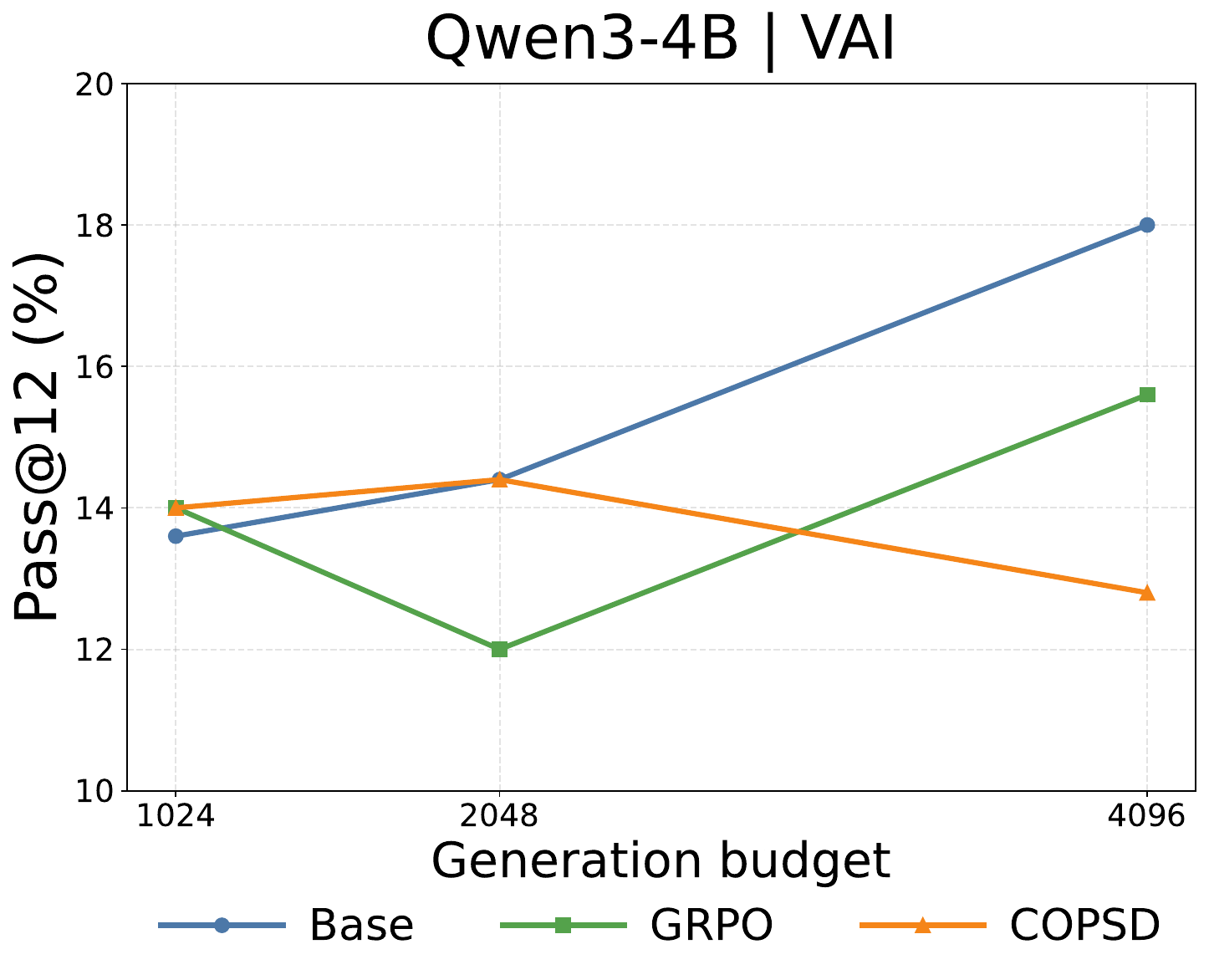}
    \includegraphics[width=0.24\linewidth]{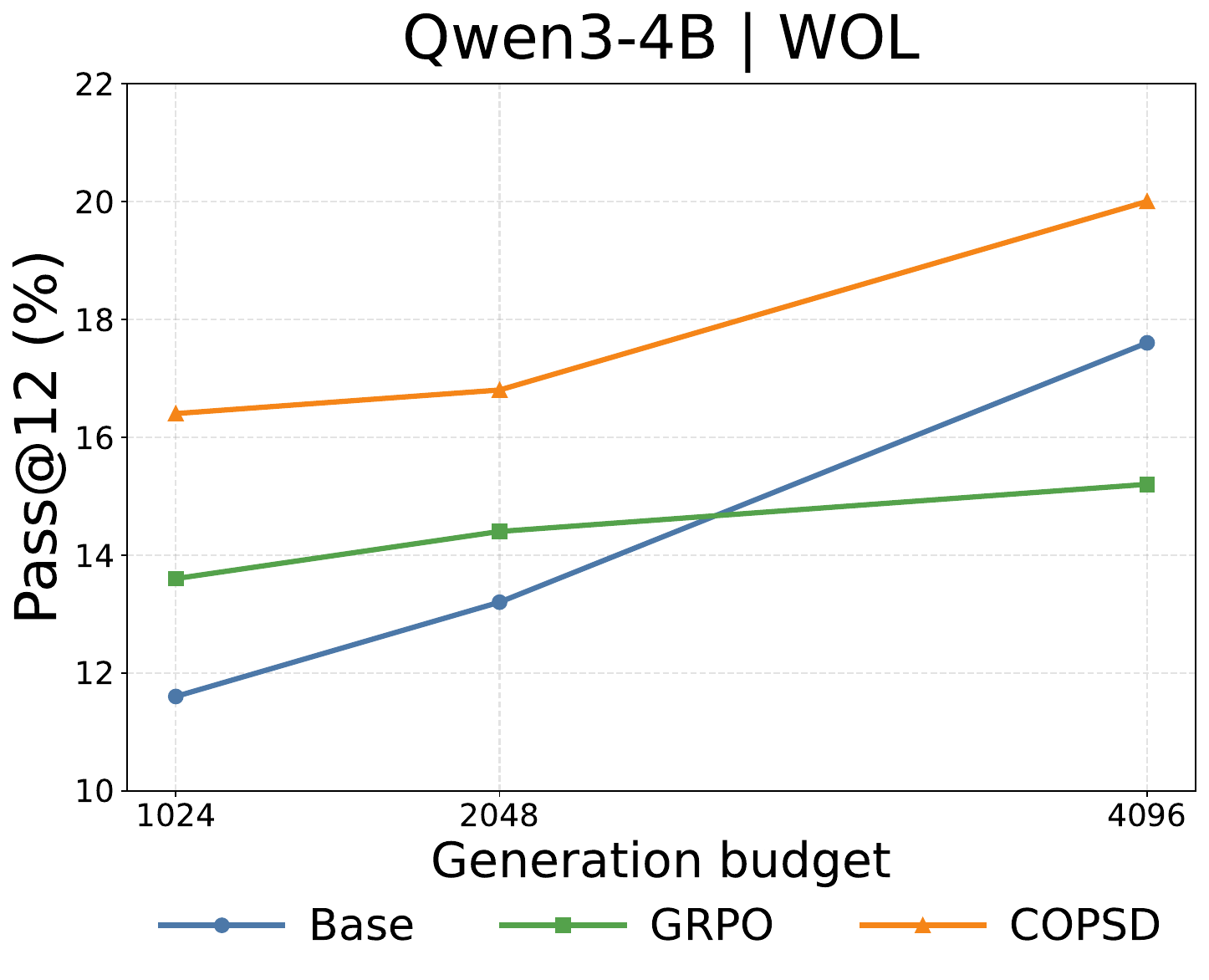}
    \includegraphics[width=0.24\linewidth]{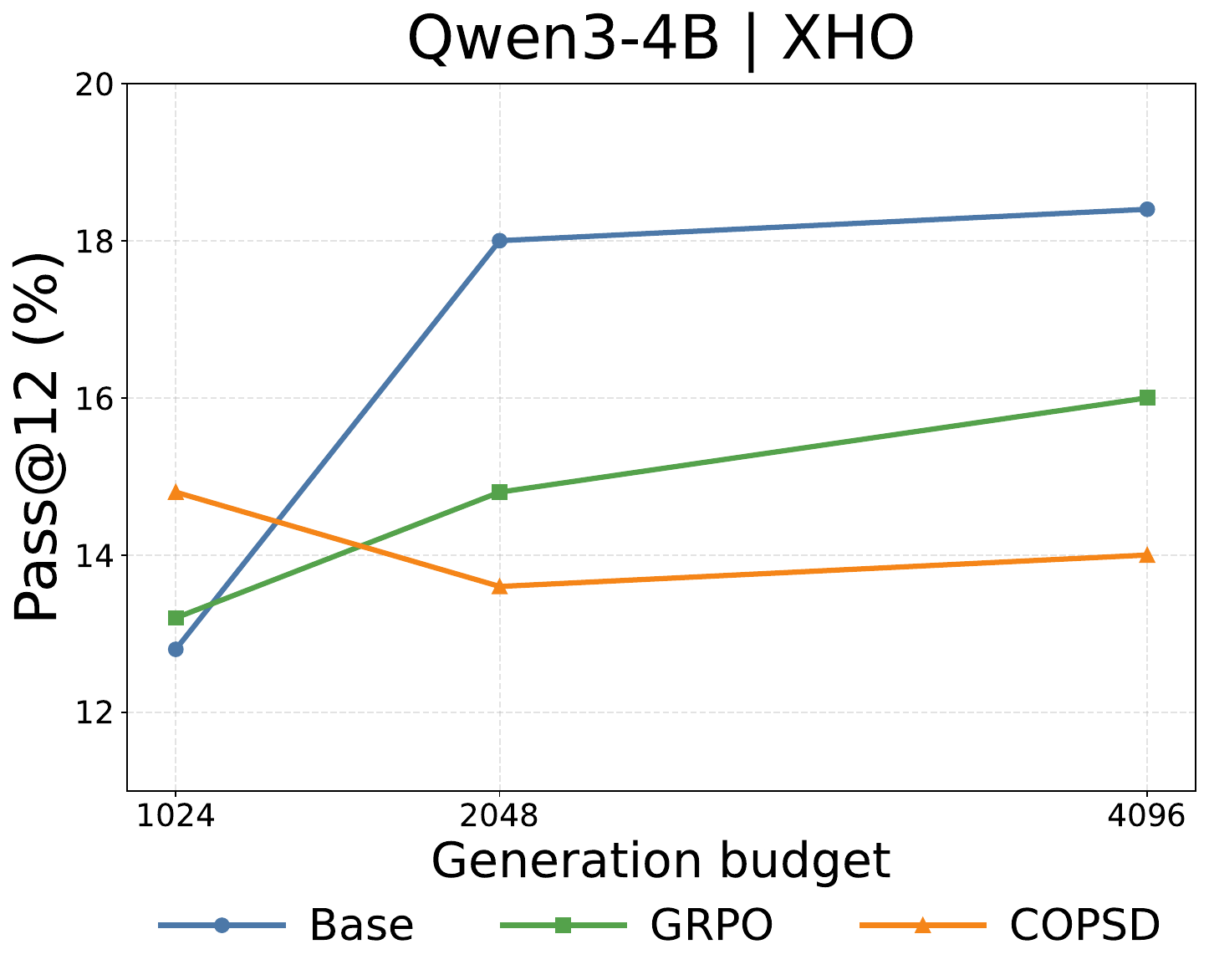}
    \includegraphics[width=0.24\linewidth]{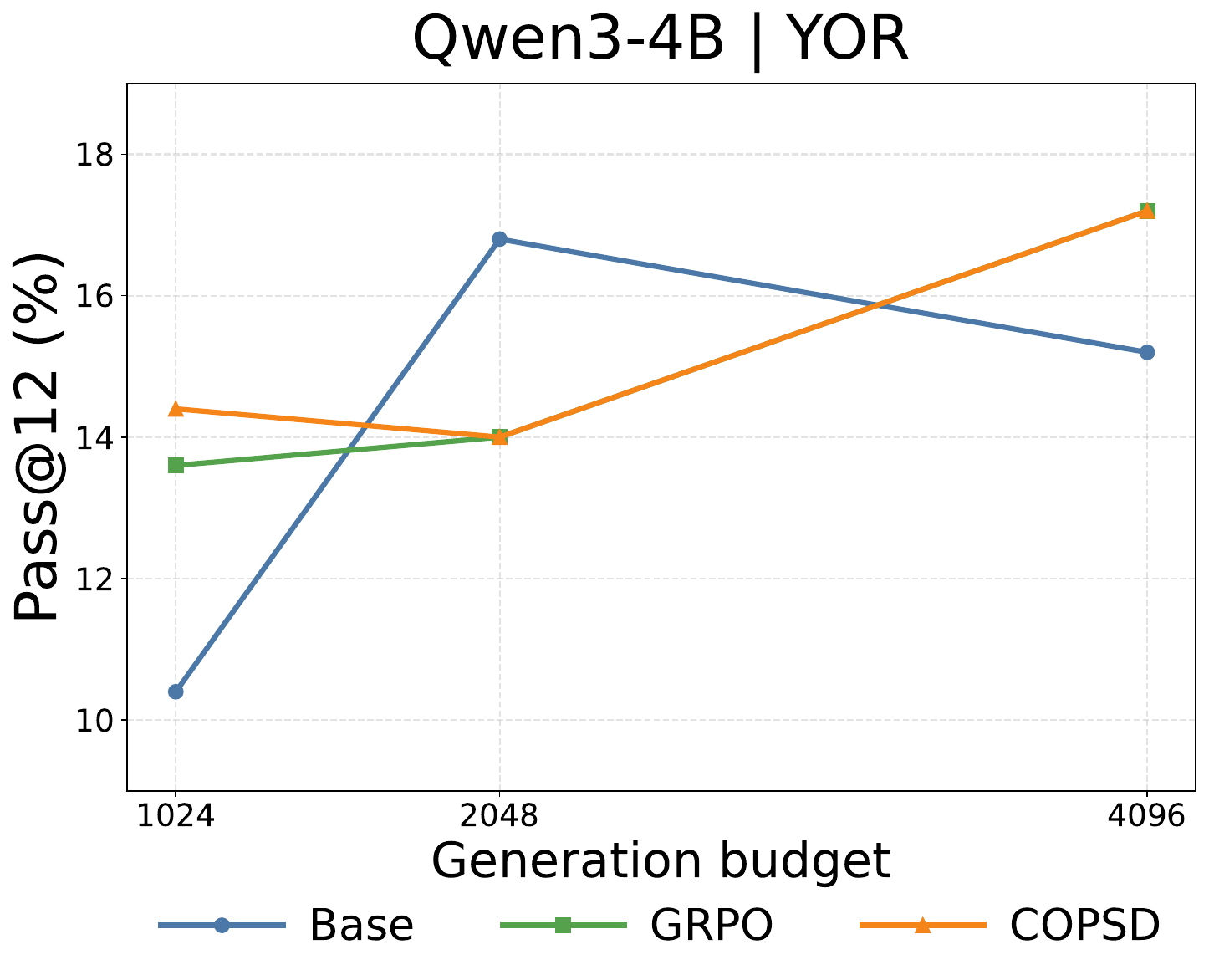}
    \includegraphics[width=0.24\linewidth]{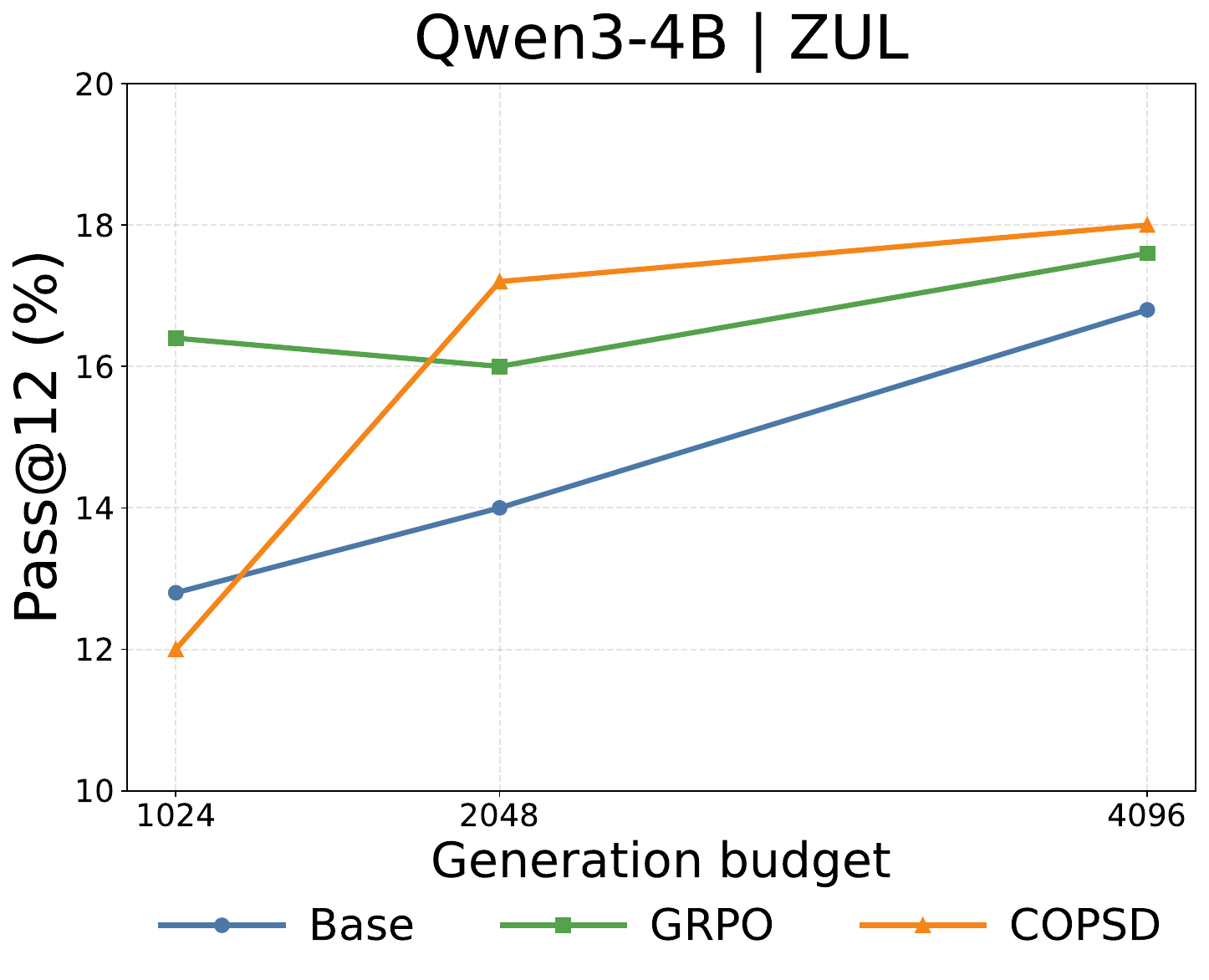}
    \caption{Per-language test-time scaling results on Pass@12 for \texttt{Qwen3-4B} across all African languages, under generation budgets of 1024, 2048, and 4096 tokens. Overall, the trends are mixed across languages, but \copsd generally achieves stronger performance than both the Base model and GRPO under different generation budgets.}
    \label{fig:test_time_scaling_qwen3_4b_all_languages}
\end{figure*}

\begin{figure*}[t]
    \centering
    \includegraphics[width=0.24\linewidth]{./figures/test_time_scaling/qwen3_8b/amh_test_time_scaling_pass_at_n_pct.pdf}
    \includegraphics[width=0.24\linewidth]{./figures/test_time_scaling/qwen3_8b/ewe_test_time_scaling_pass_at_n_pct.pdf}
    \includegraphics[width=0.24\linewidth]{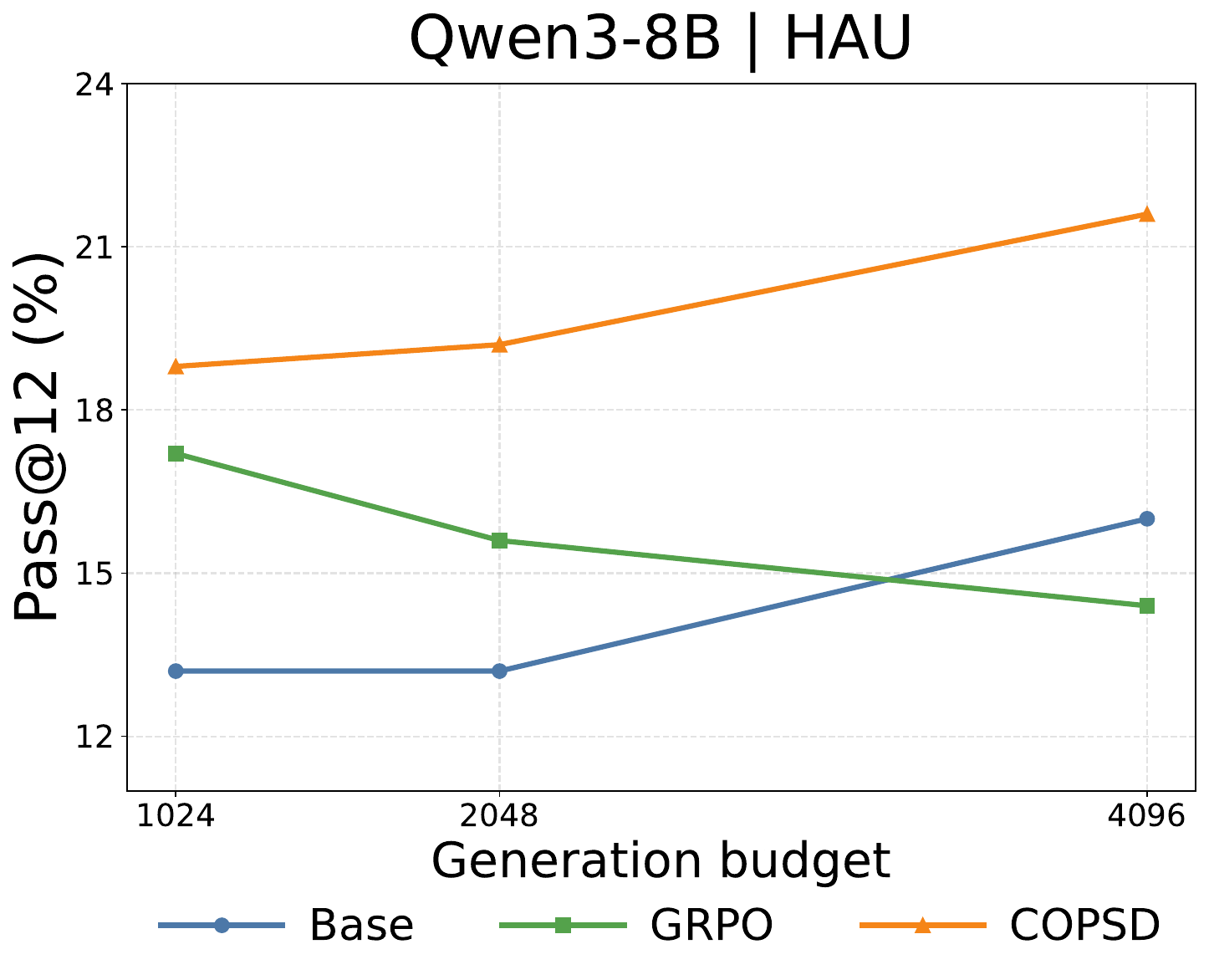}
    \includegraphics[width=0.24\linewidth]{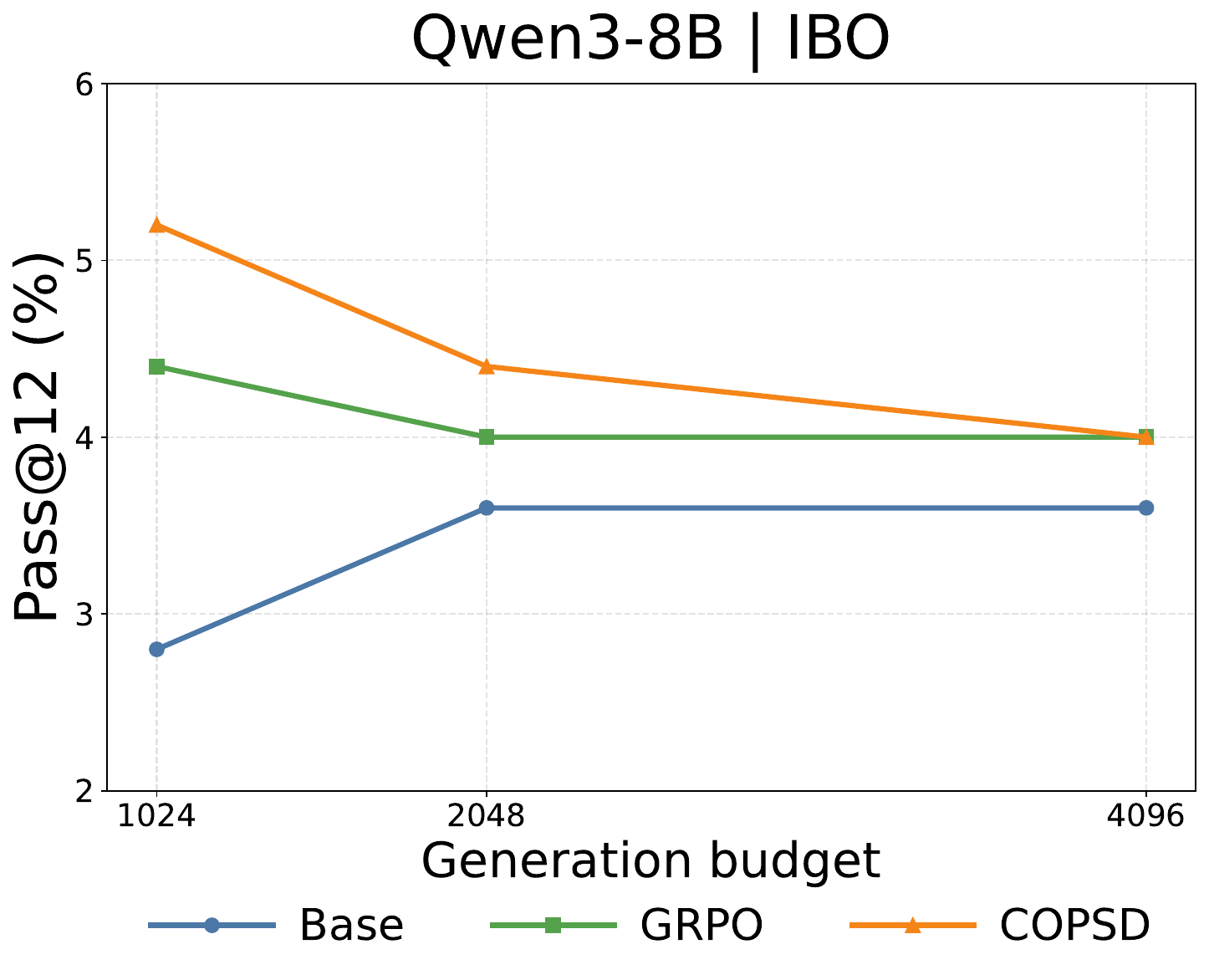}
    \includegraphics[width=0.24\linewidth]{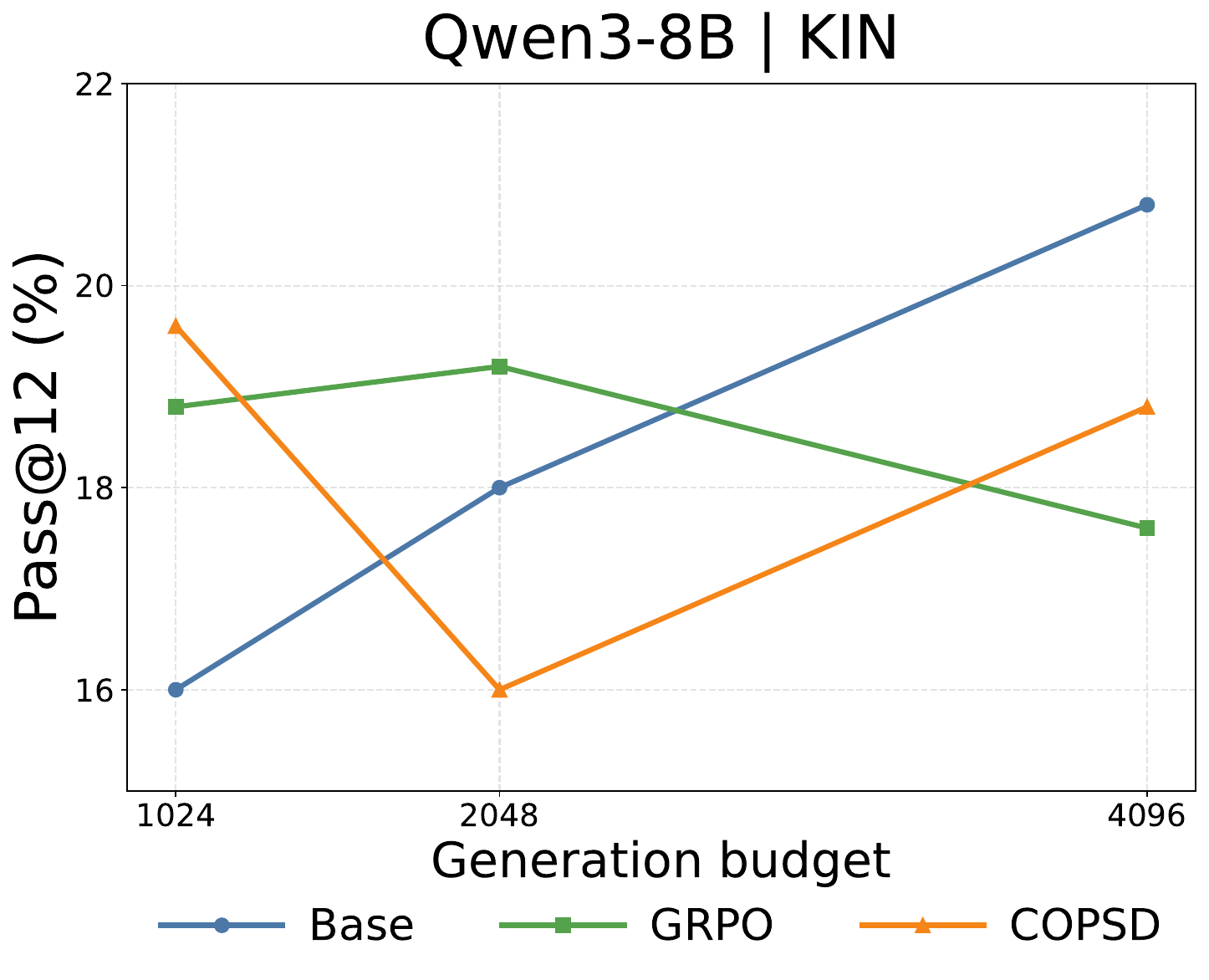}
    \includegraphics[width=0.24\linewidth]{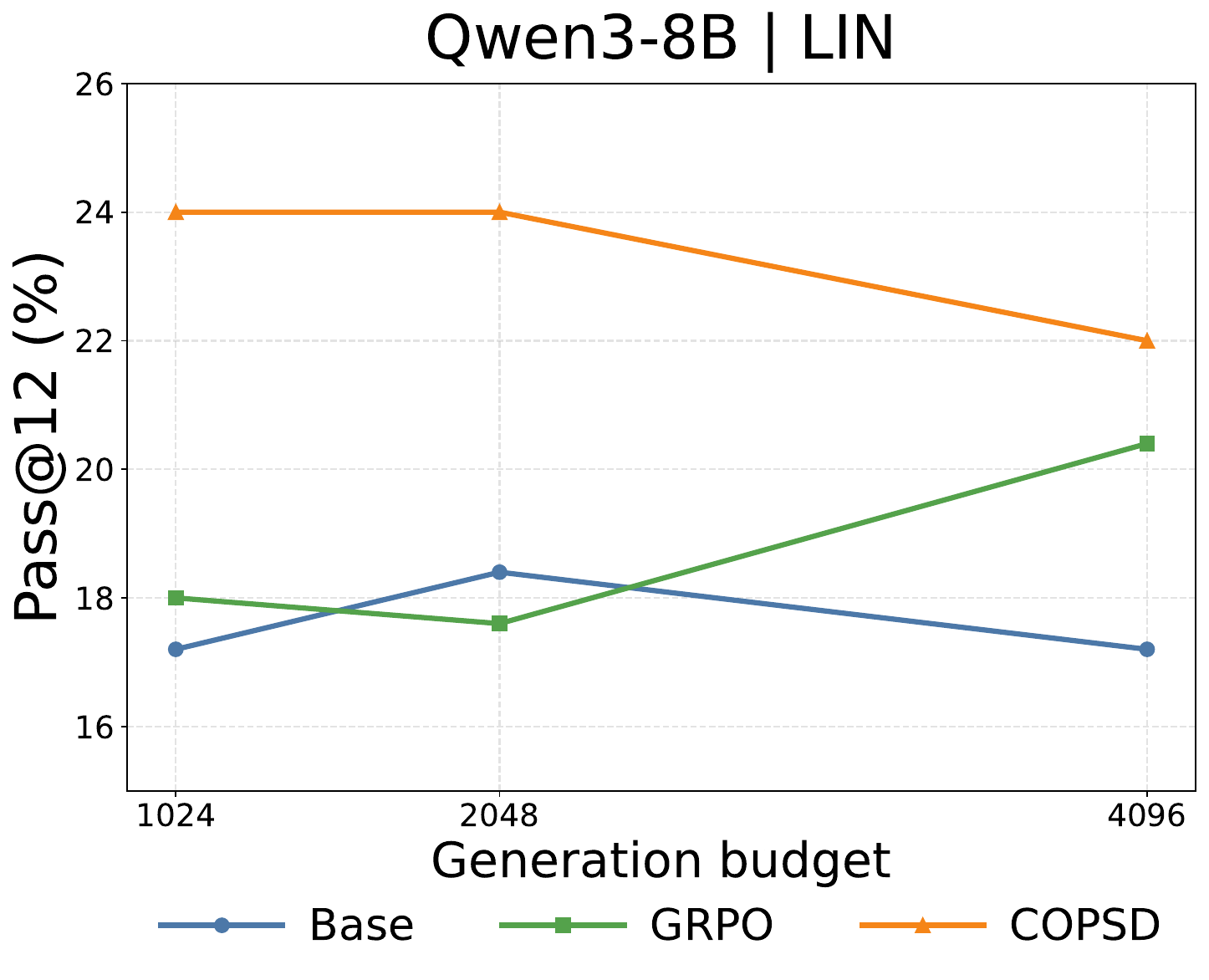}
    \includegraphics[width=0.24\linewidth]{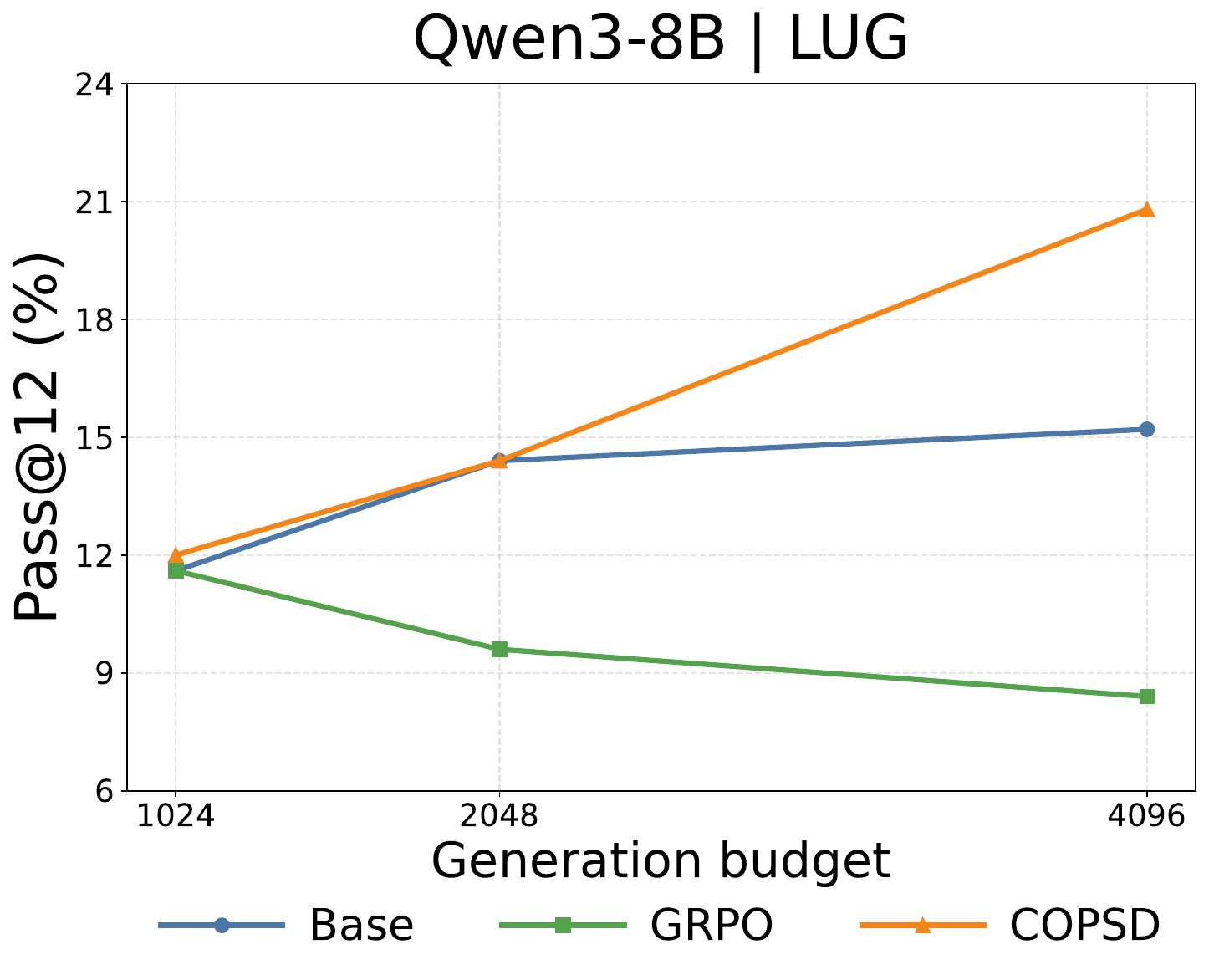}
    \includegraphics[width=0.24\linewidth]{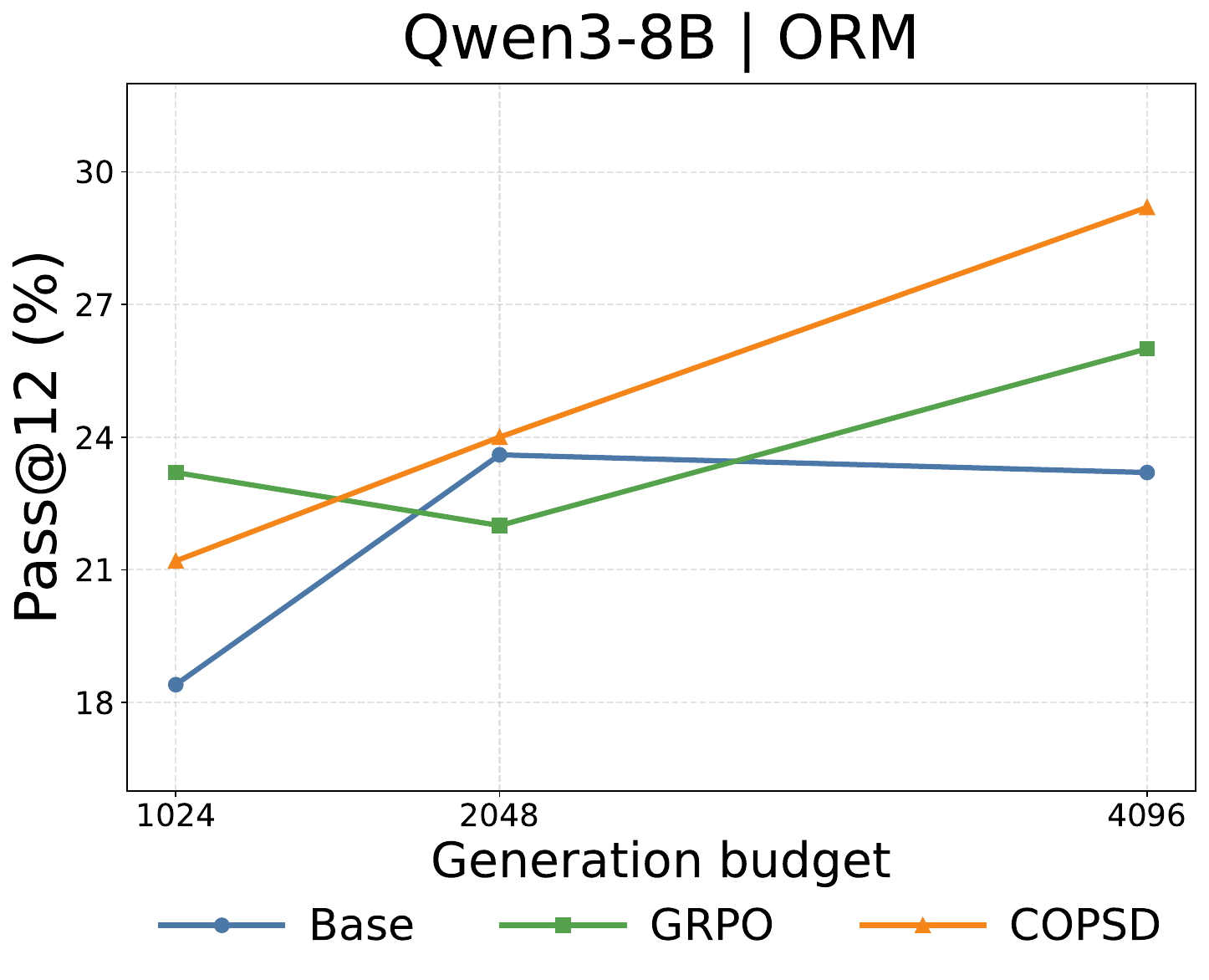}
    \includegraphics[width=0.24\linewidth]{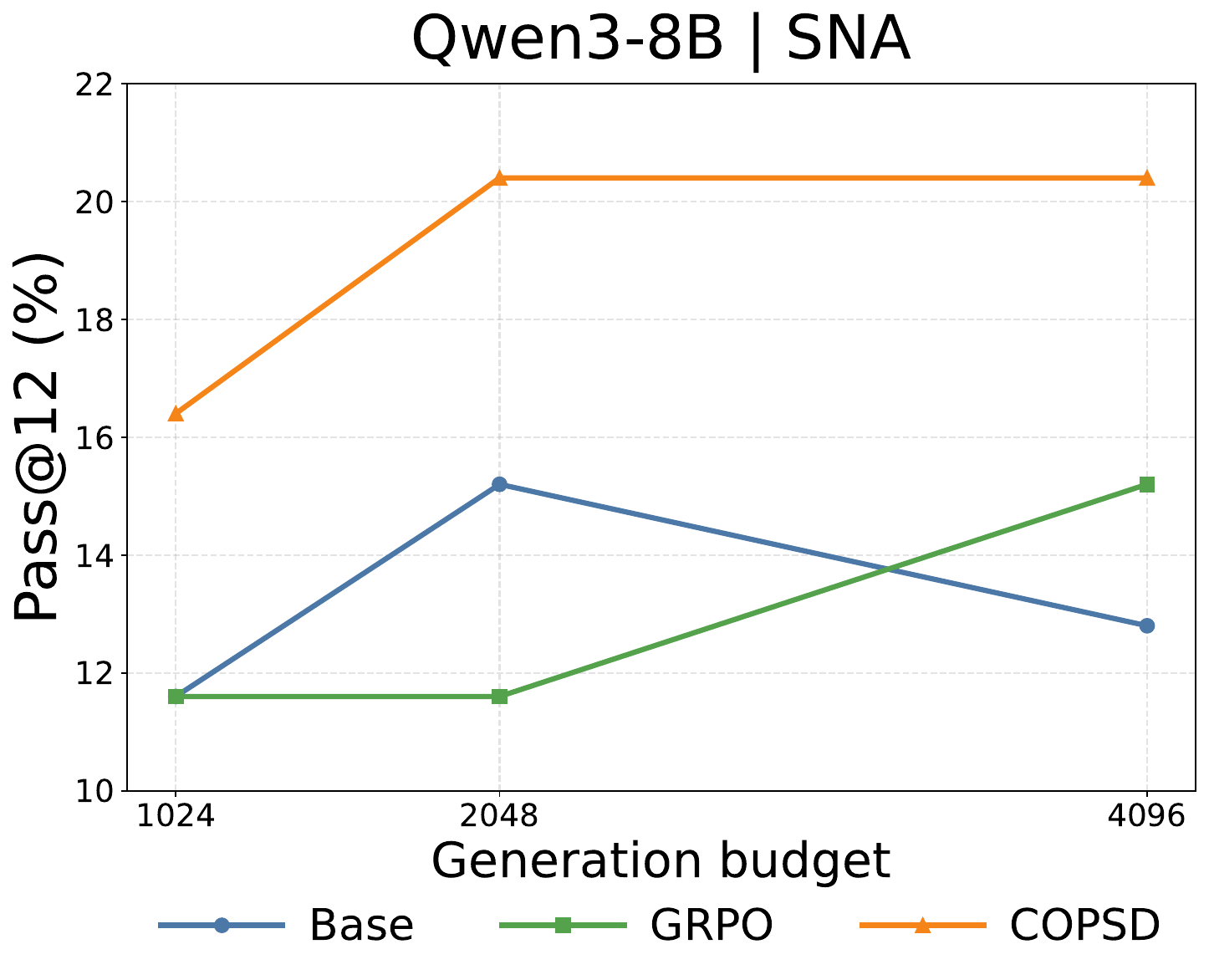}
    \includegraphics[width=0.24\linewidth]{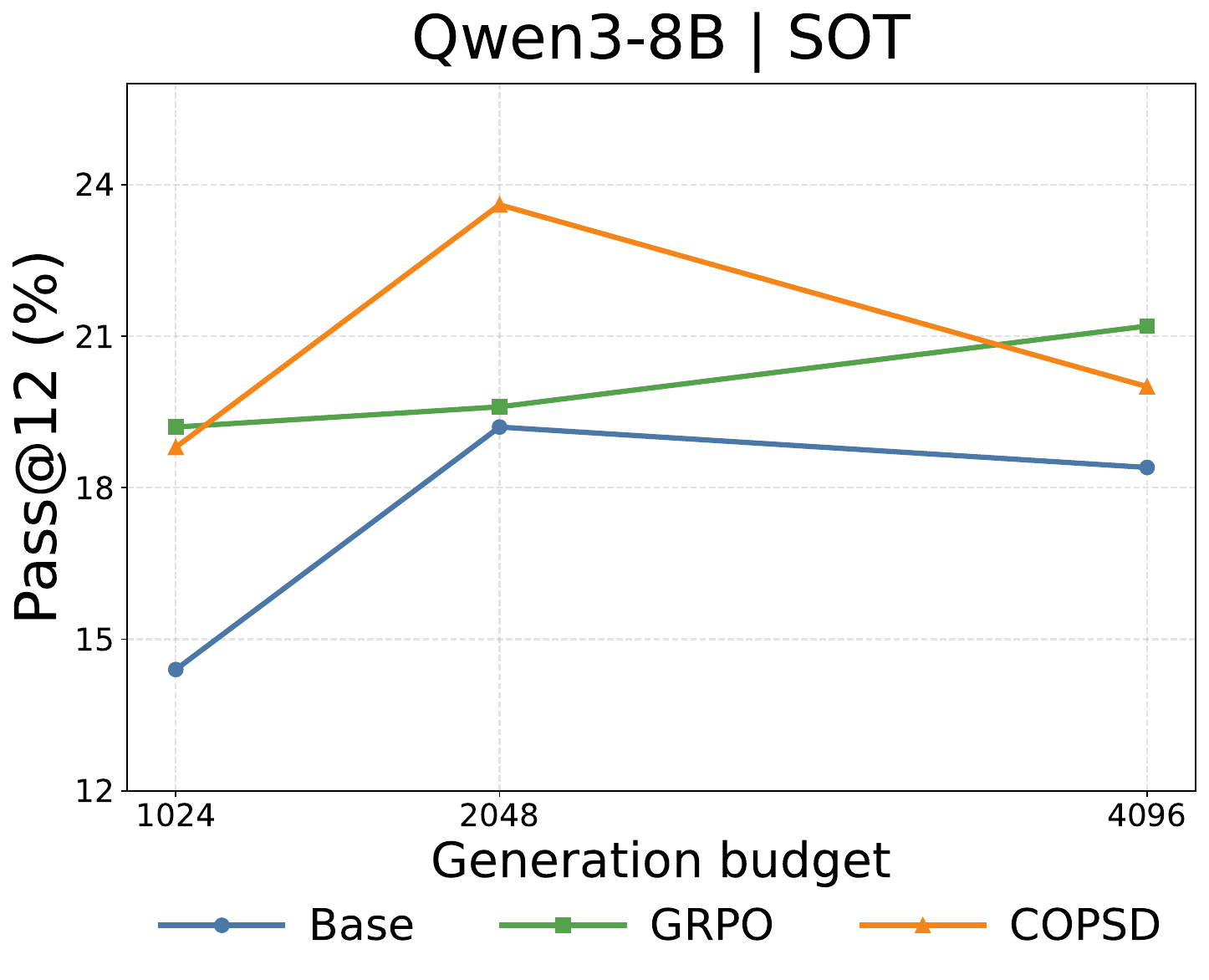}
    \includegraphics[width=0.24\linewidth]{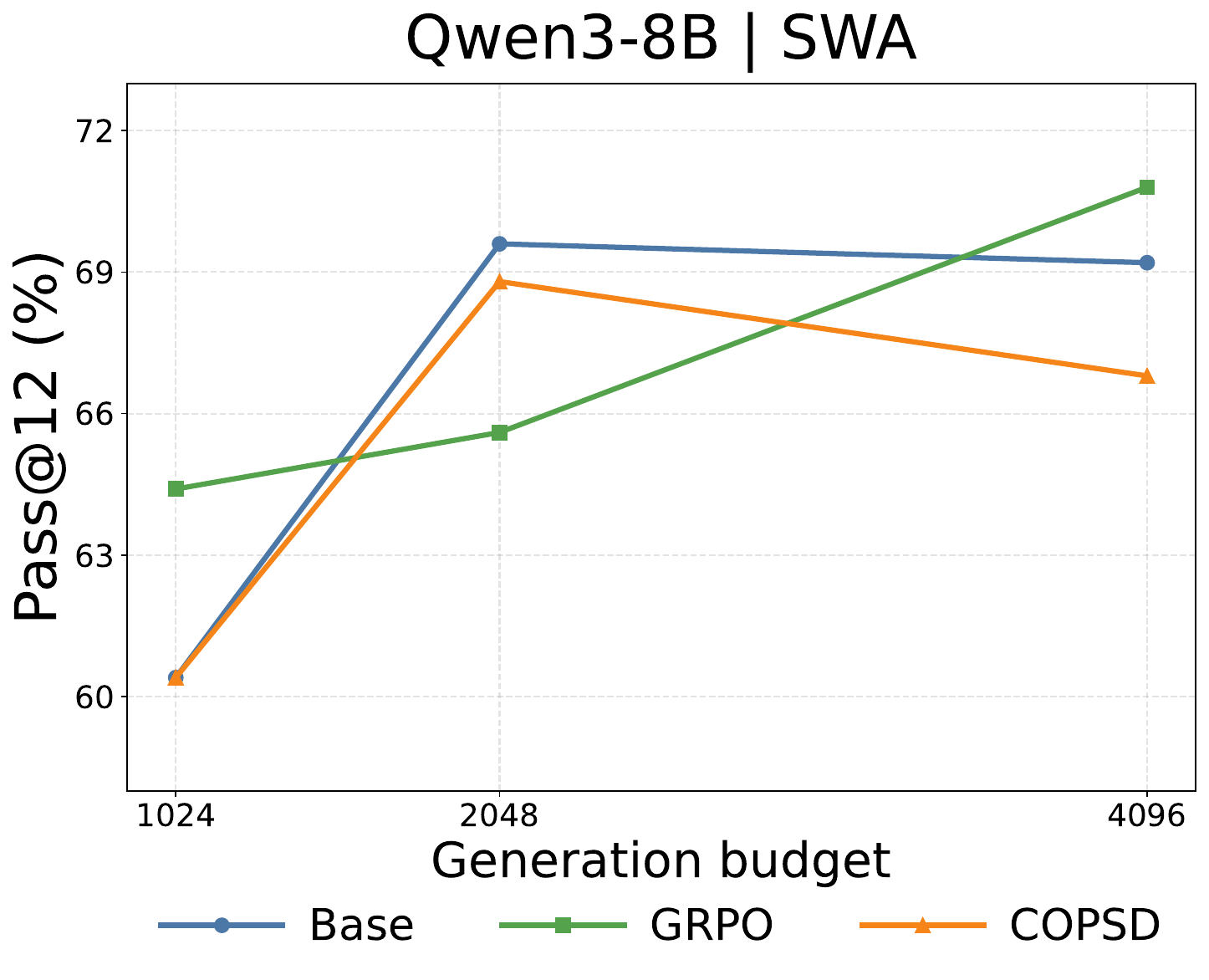}
    \includegraphics[width=0.24\linewidth]{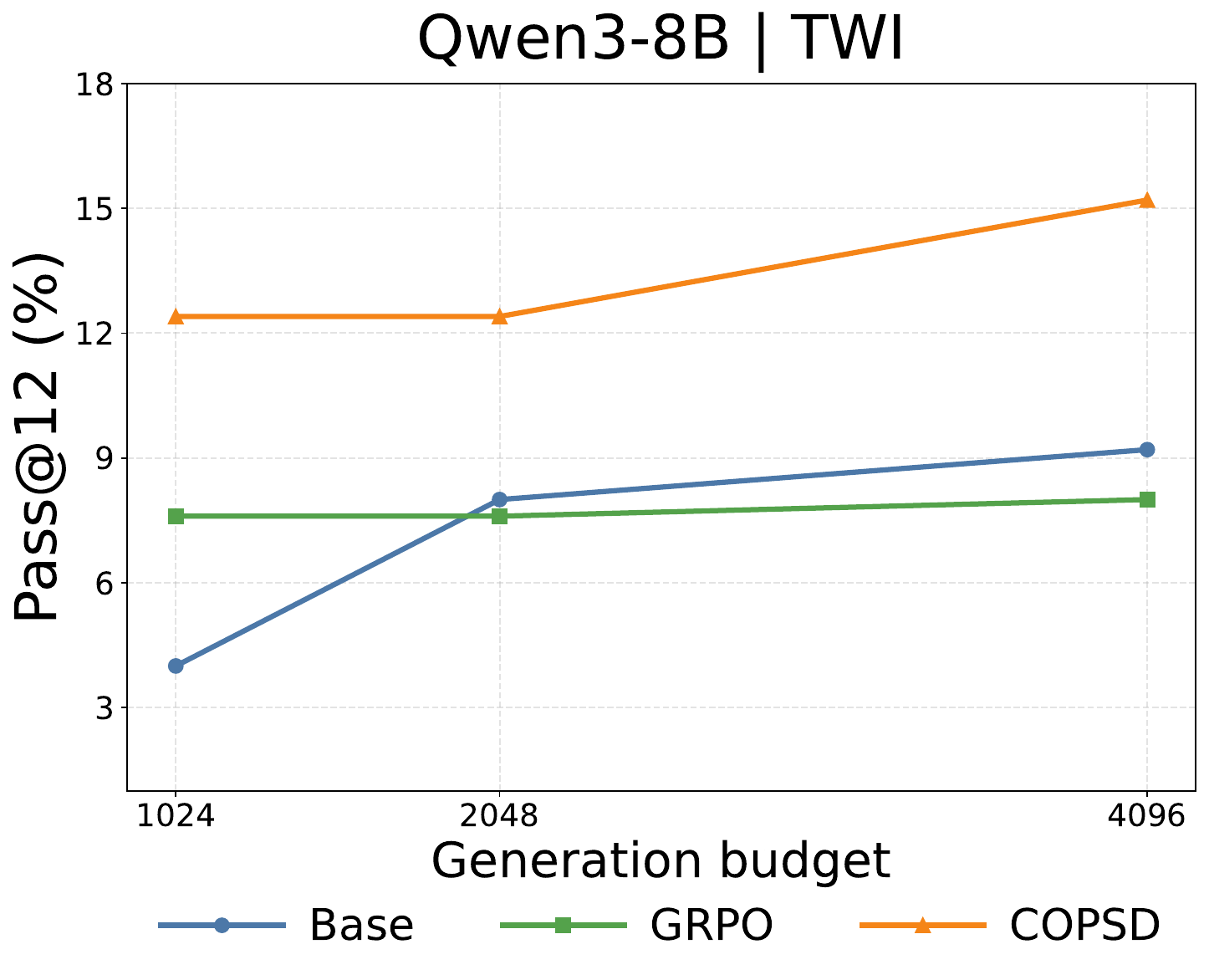}
    \includegraphics[width=0.24\linewidth]{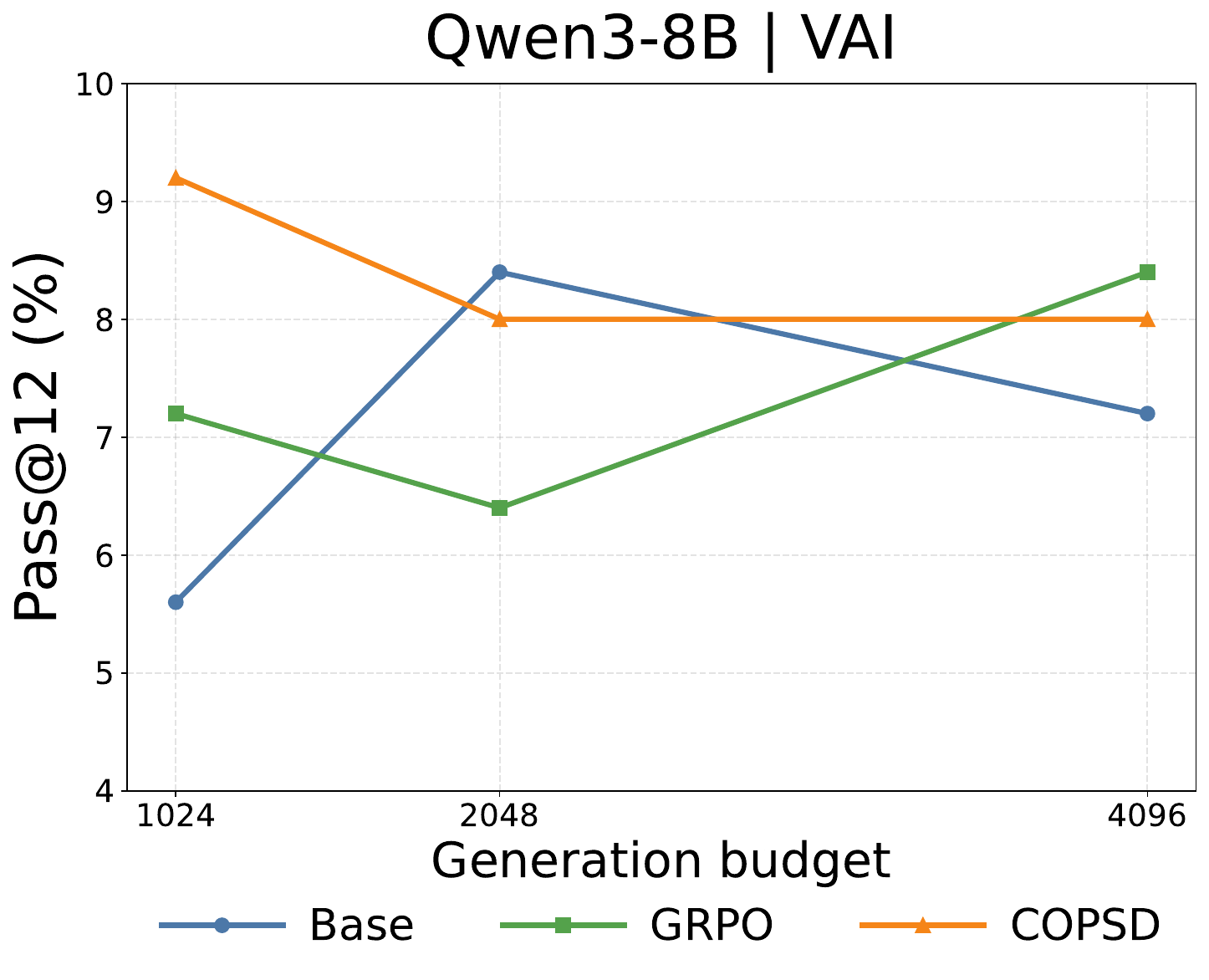}
    \includegraphics[width=0.24\linewidth]{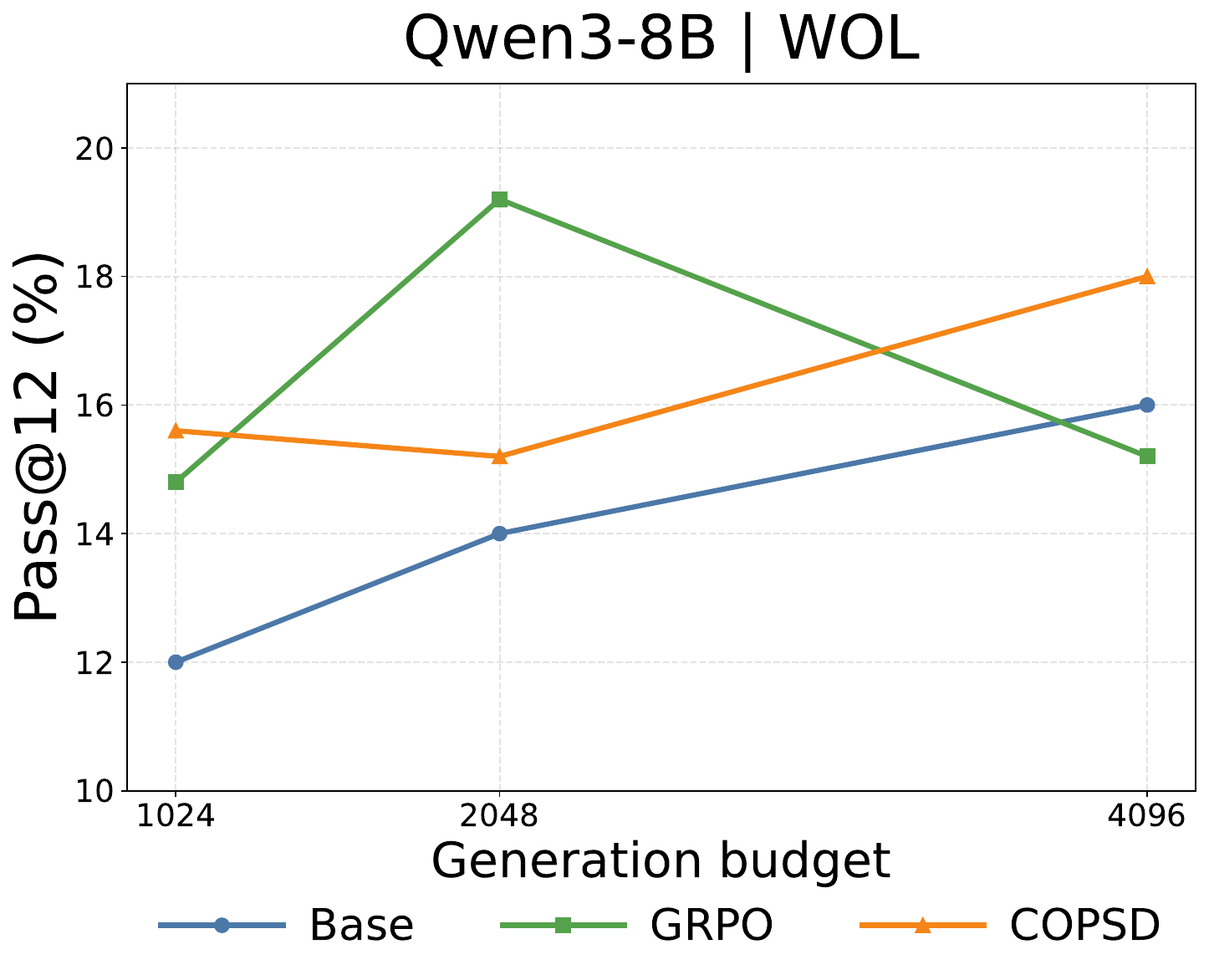}
    \includegraphics[width=0.24\linewidth]{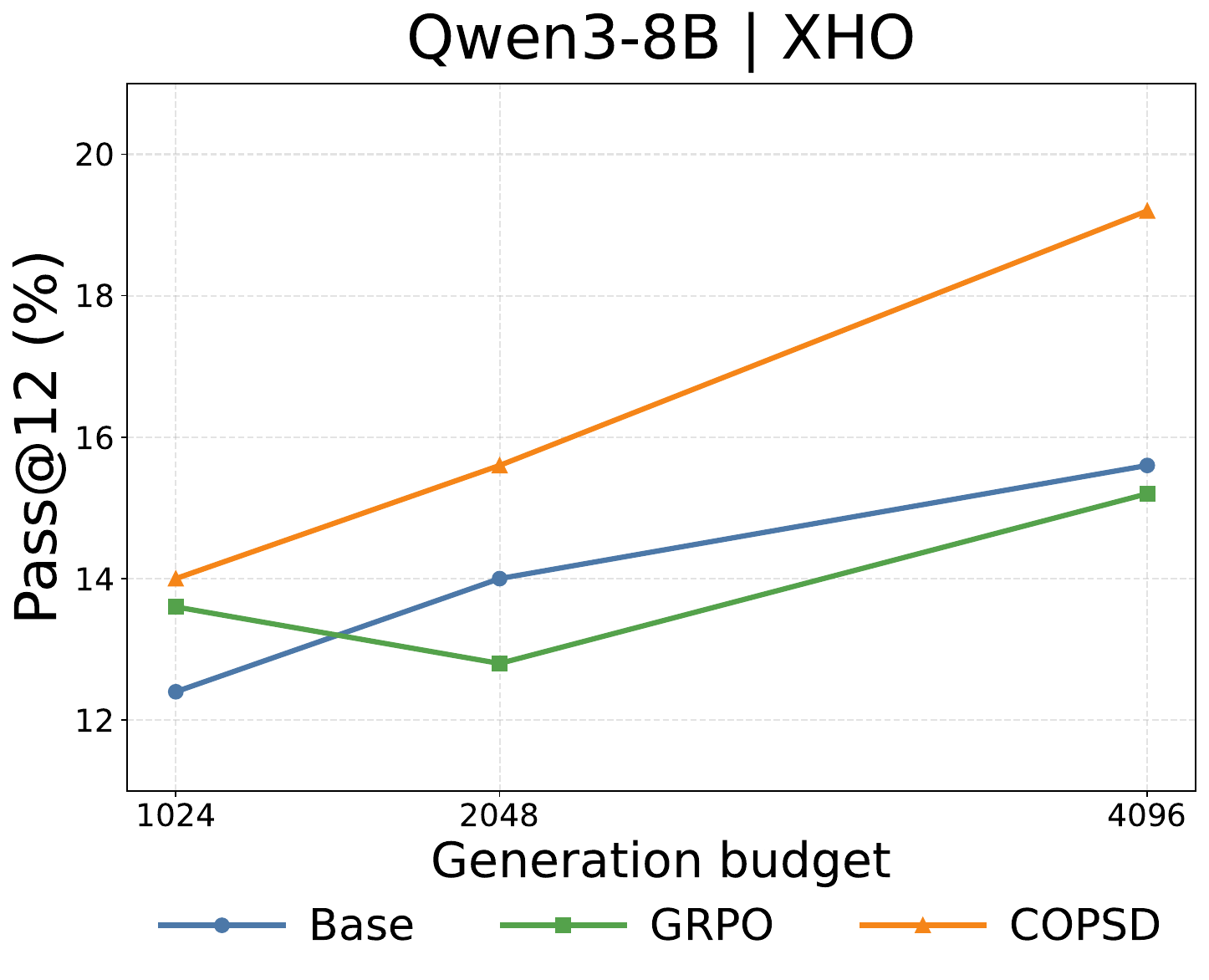}
    \includegraphics[width=0.24\linewidth]{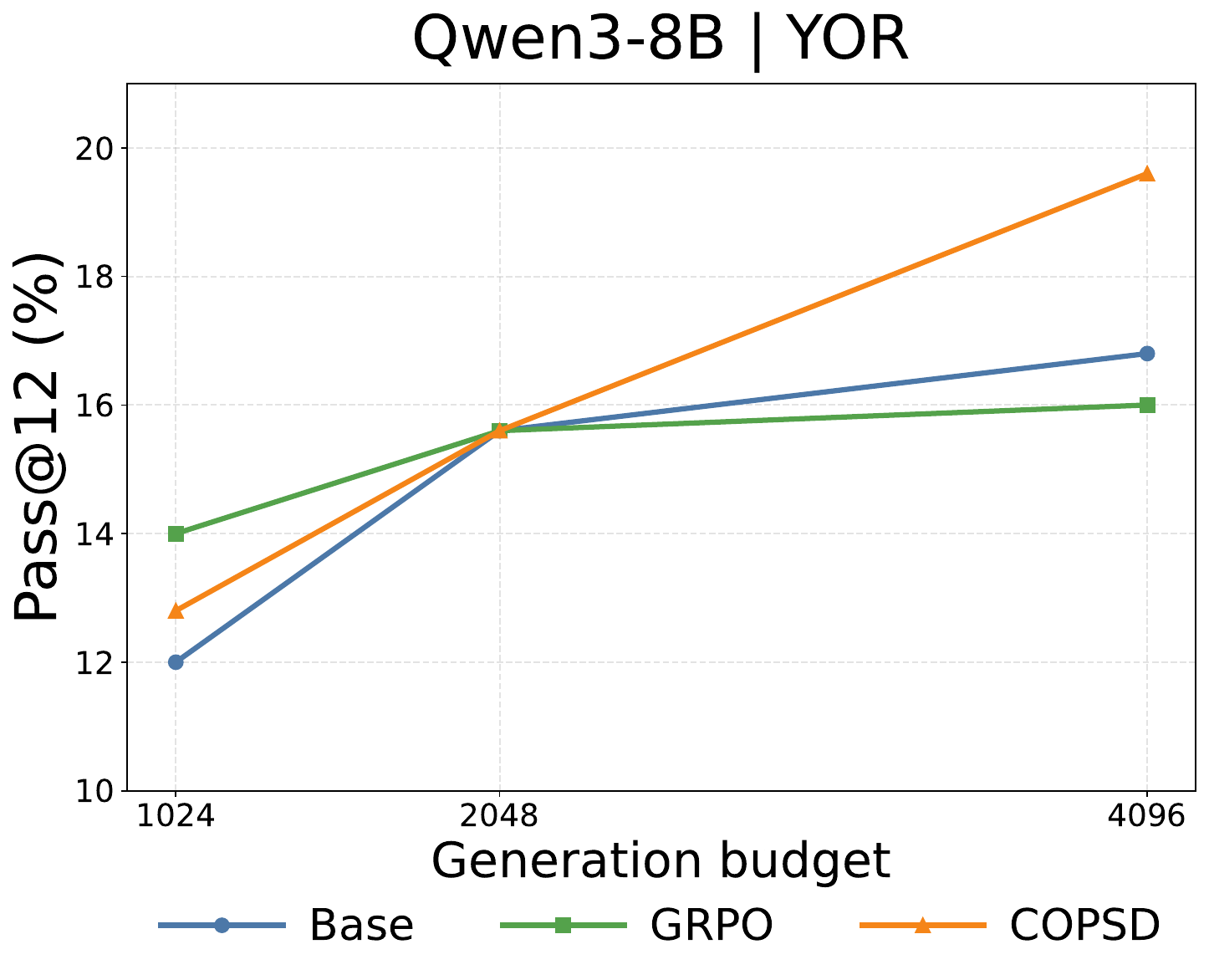}
    \includegraphics[width=0.24\linewidth]{./figures/test_time_scaling/qwen3_8b/zul_test_time_scaling_pass_at_n_pct.pdf}
    \caption{Per-language test-time scaling results on Pass@12 for \texttt{Qwen3-8B} across all African languages, under generation budgets of 1024, 2048, and 4096 tokens. Overall, the trends are mixed across languages, but \copsd generally achieves stronger performance than both the Base model and GRPO under different generation budgets.}
    \label{fig:test_time_scaling_qwen3_8b_all_languages}
\end{figure*}

Figure~\ref{fig:test_time_scaling_qwen3_1.7b_all_languages}, Figure~\ref{fig:test_time_scaling_qwen3_4b_all_languages}, and Figure~\ref{fig:test_time_scaling_qwen3_8b_all_languages} report the complete per-language test-time scaling results under generation budgets of 1024, 2048, and 4096 tokens for all three model sizes.
Overall, \copsd tends to outperform the base and GRPO-trained models across budgets, although the magnitude of improvement varies by language and model size.
The benefits of increased generation budget are more consistent for larger models, especially \texttt{Qwen3-8B}, supporting the observation in \secref{test_time} that effective crosslingual test-time scaling requires sufficient model capacity.

\begin{figure*}[t]
    \centering

    \includegraphics[width=0.32\linewidth]{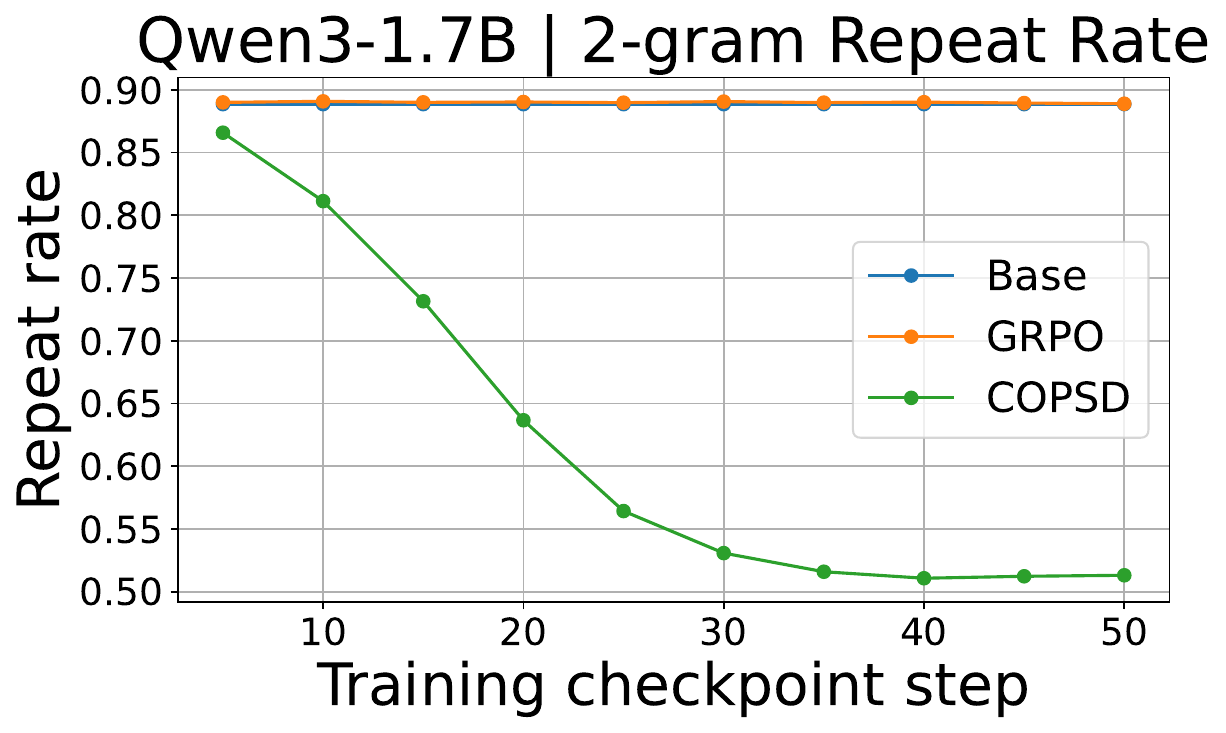}
    \includegraphics[width=0.32\linewidth]{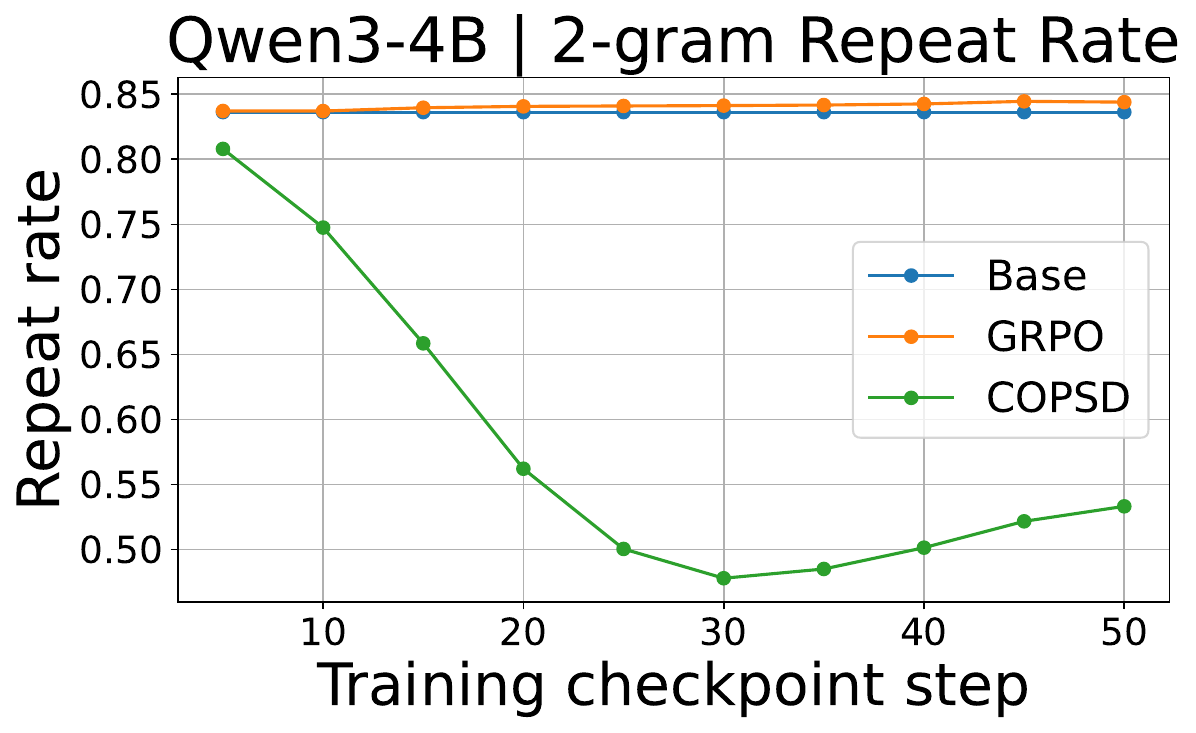}
    \includegraphics[width=0.32\linewidth]{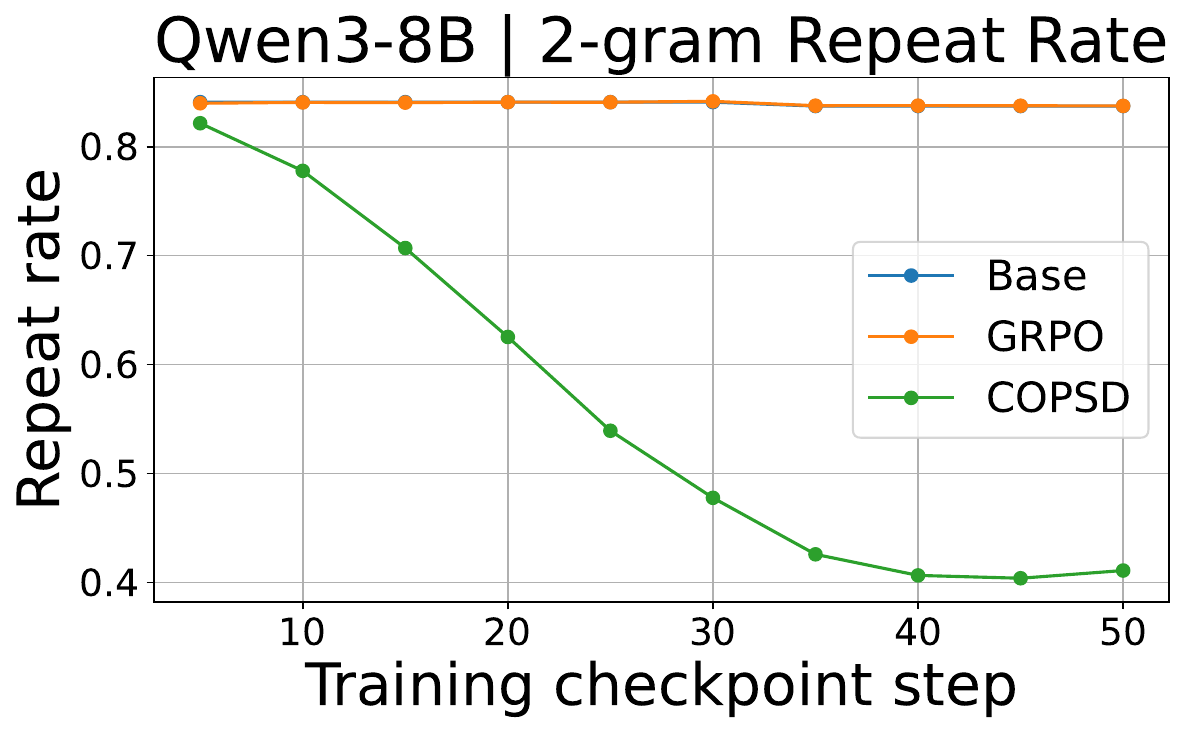}

    \vspace{0.3em}

    \includegraphics[width=0.32\linewidth]{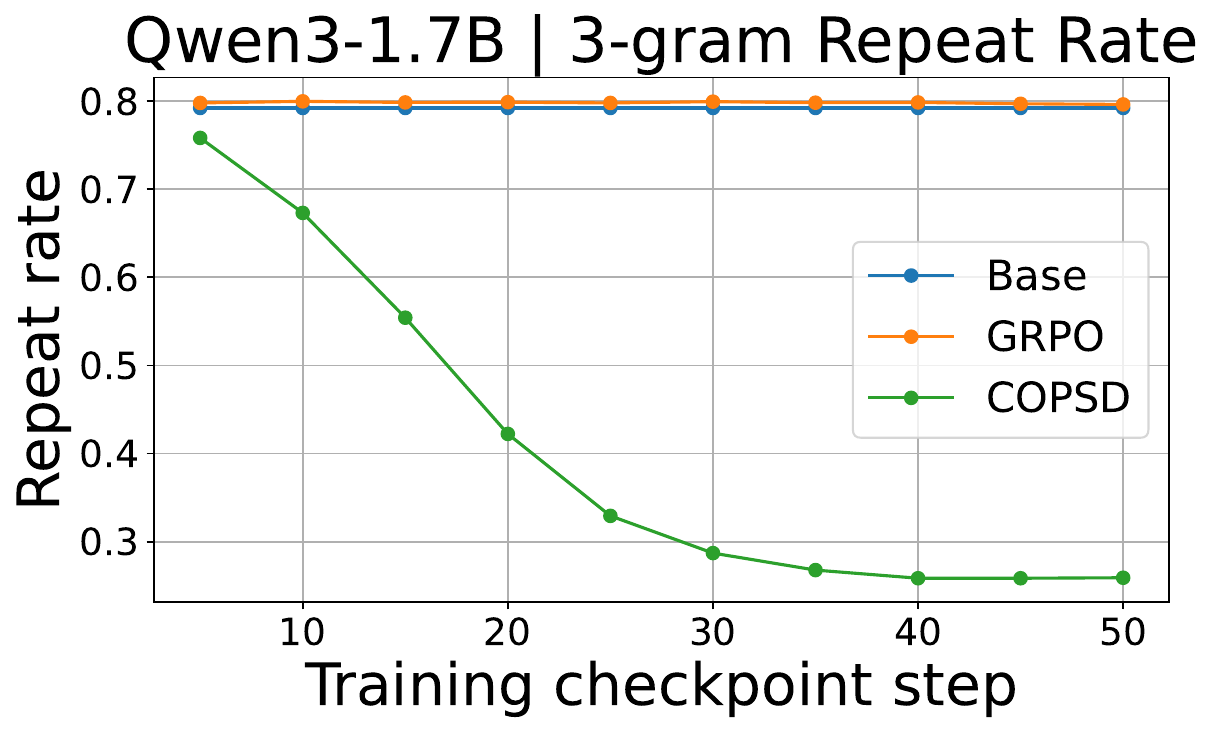}
    \includegraphics[width=0.32\linewidth]{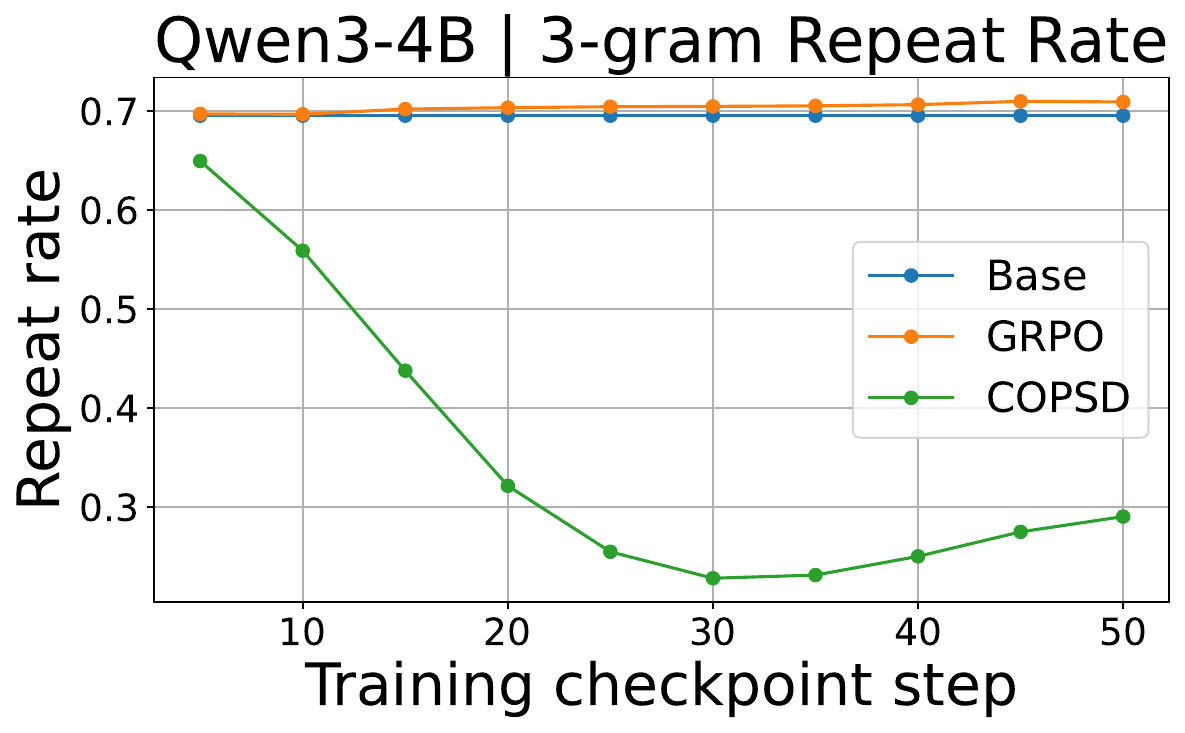}
    \includegraphics[width=0.32\linewidth]{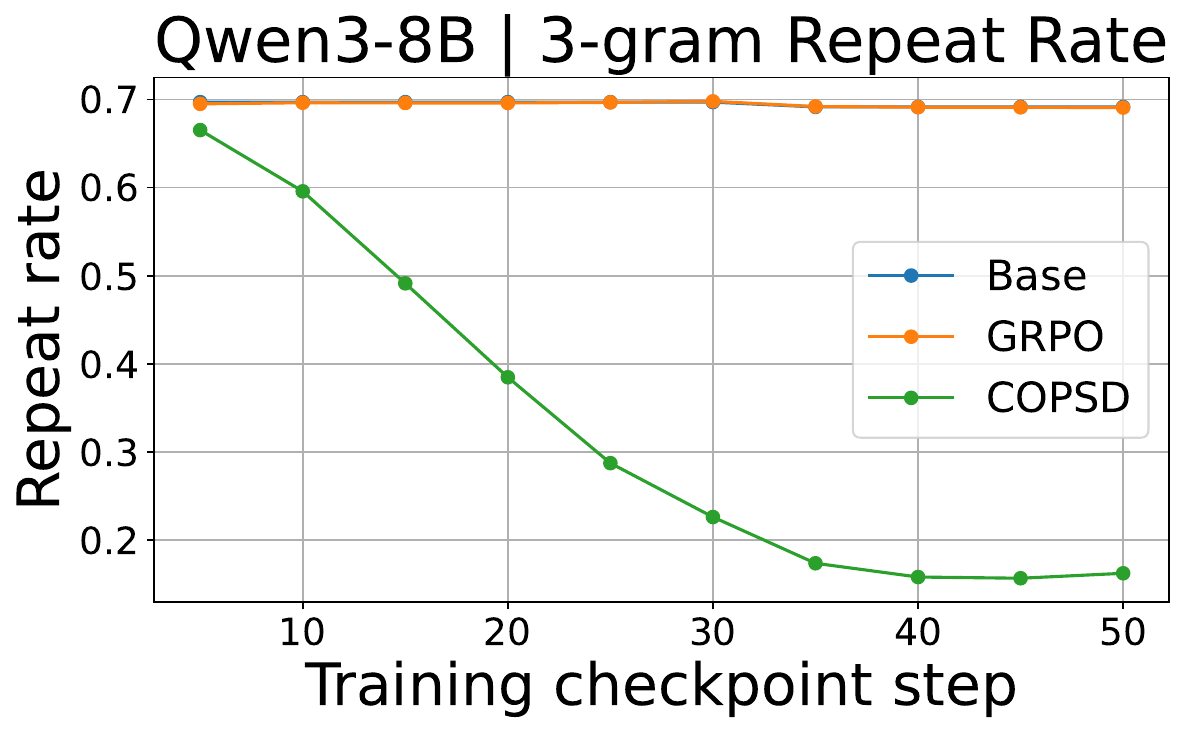}

    \vspace{0.3em}

    \includegraphics[width=0.32\linewidth]{./figures/ngram/Qwen3-1.7B_4gram_repeat_rate.pdf}
    \includegraphics[width=0.32\linewidth]{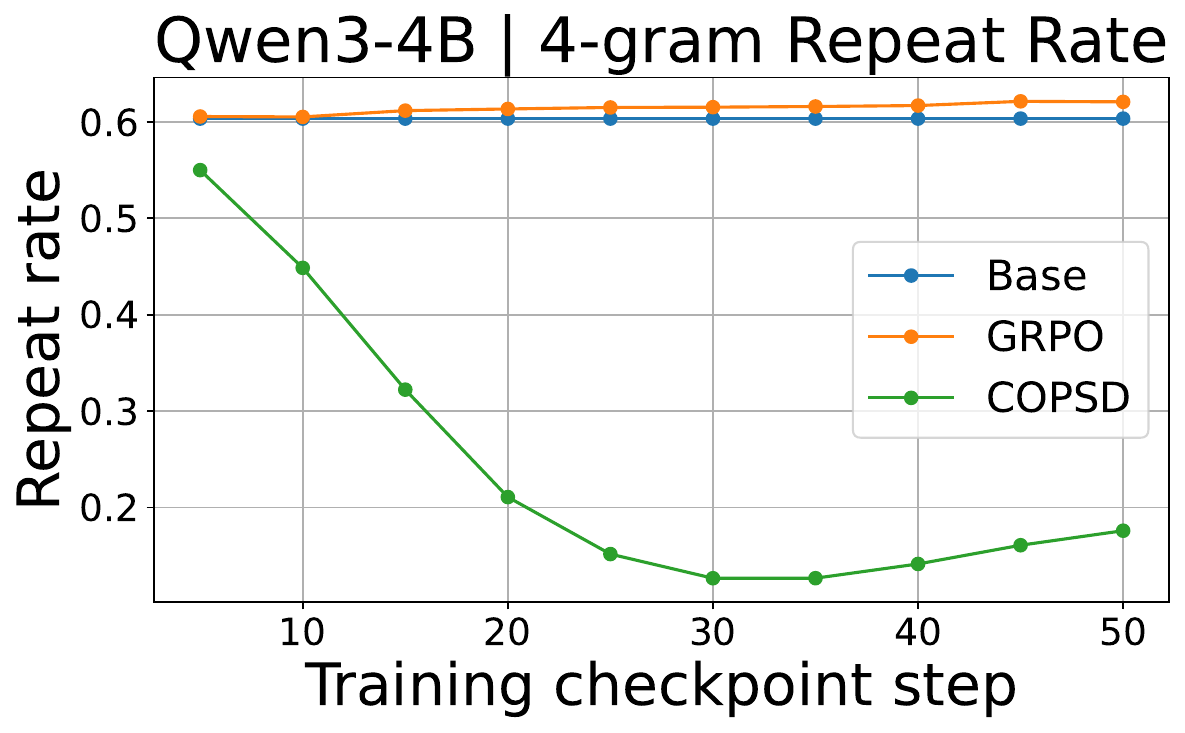}
    \includegraphics[width=0.32\linewidth]{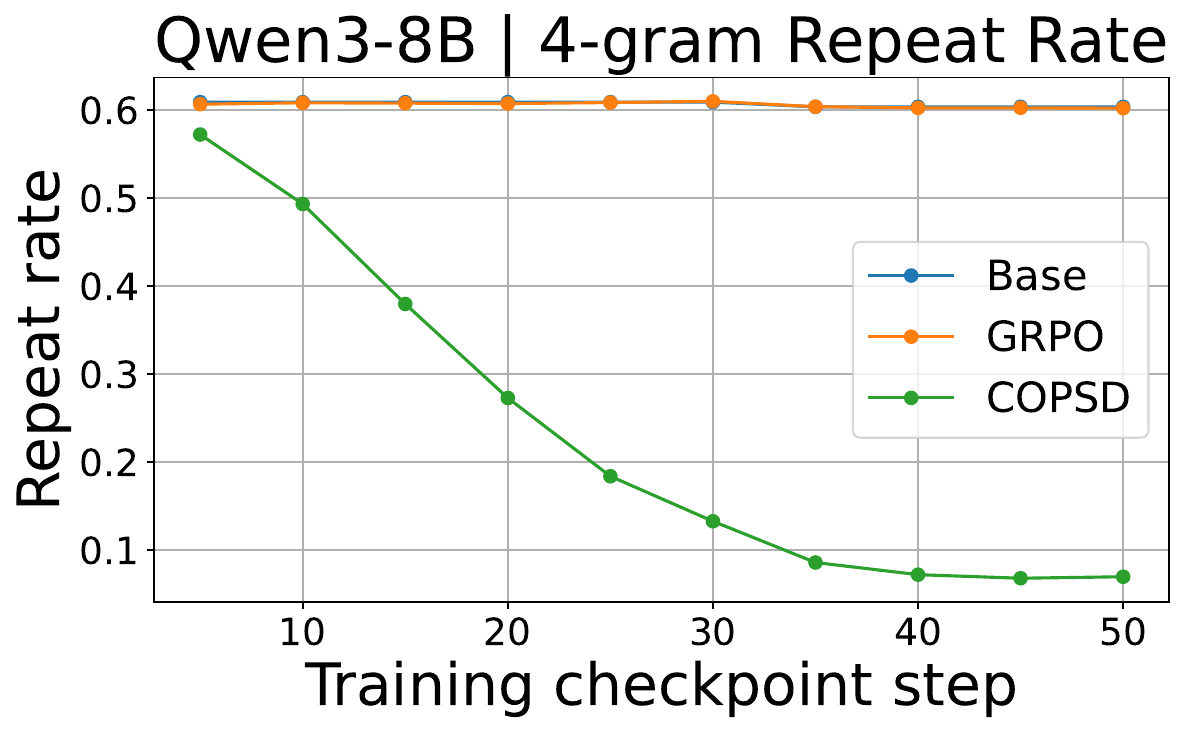}

    \vspace{0.3em}

    \includegraphics[width=0.32\linewidth]{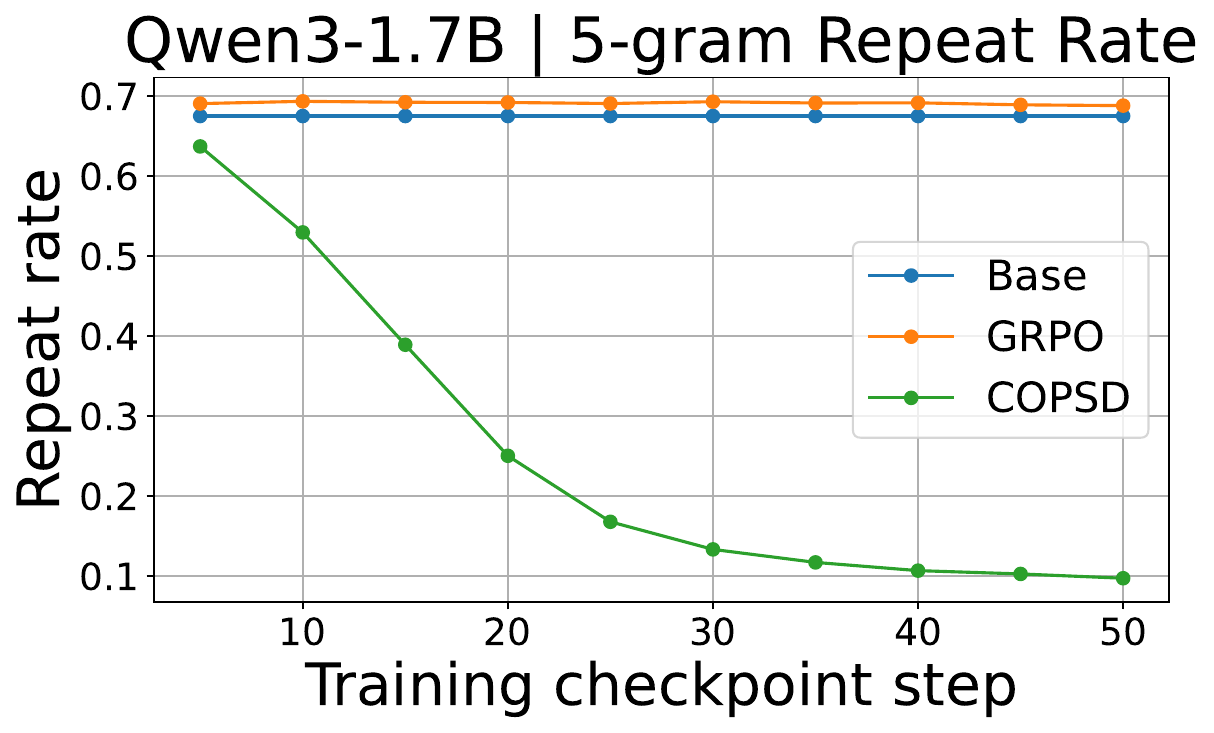}
    \includegraphics[width=0.32\linewidth]{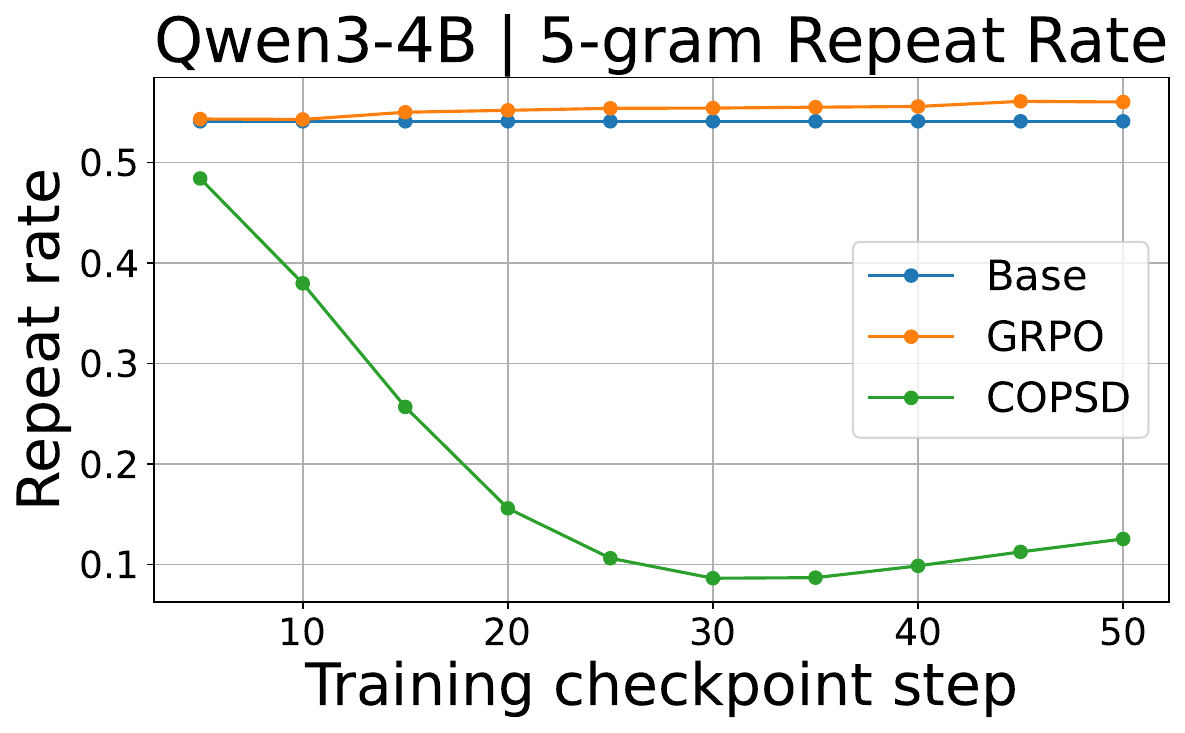}
    \includegraphics[width=0.32\linewidth]{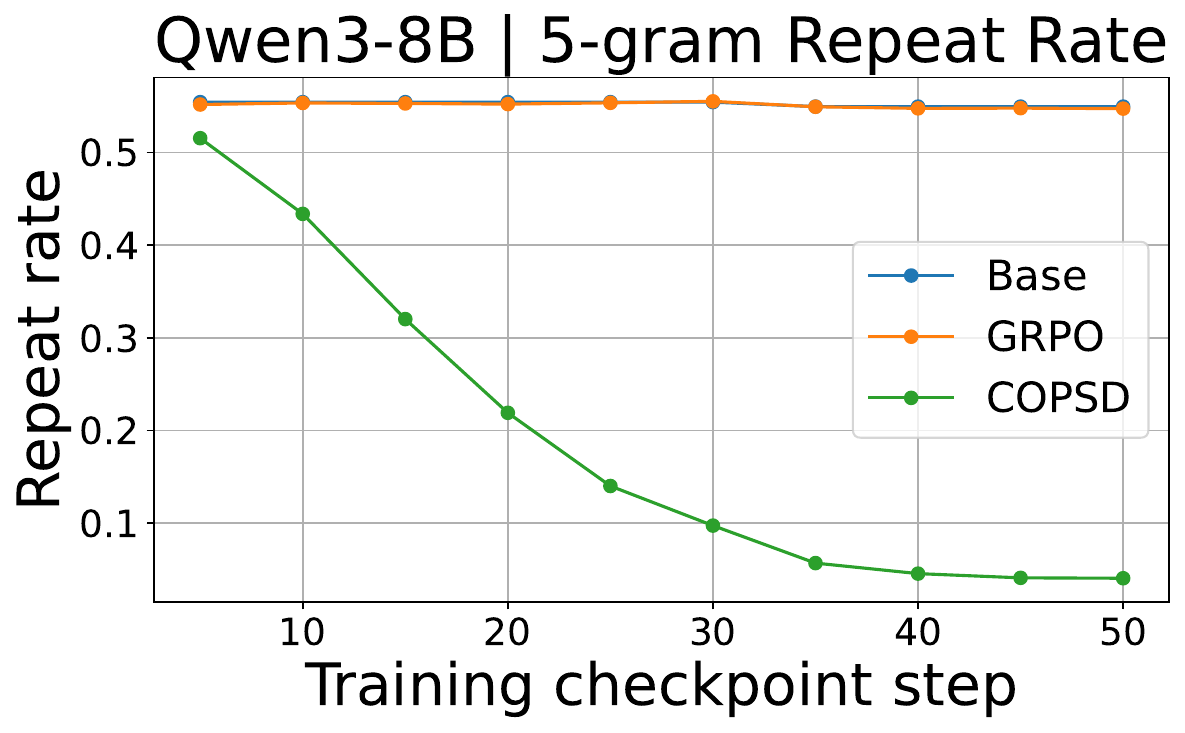}

    \vspace{0.3em}

    \includegraphics[width=0.32\linewidth]{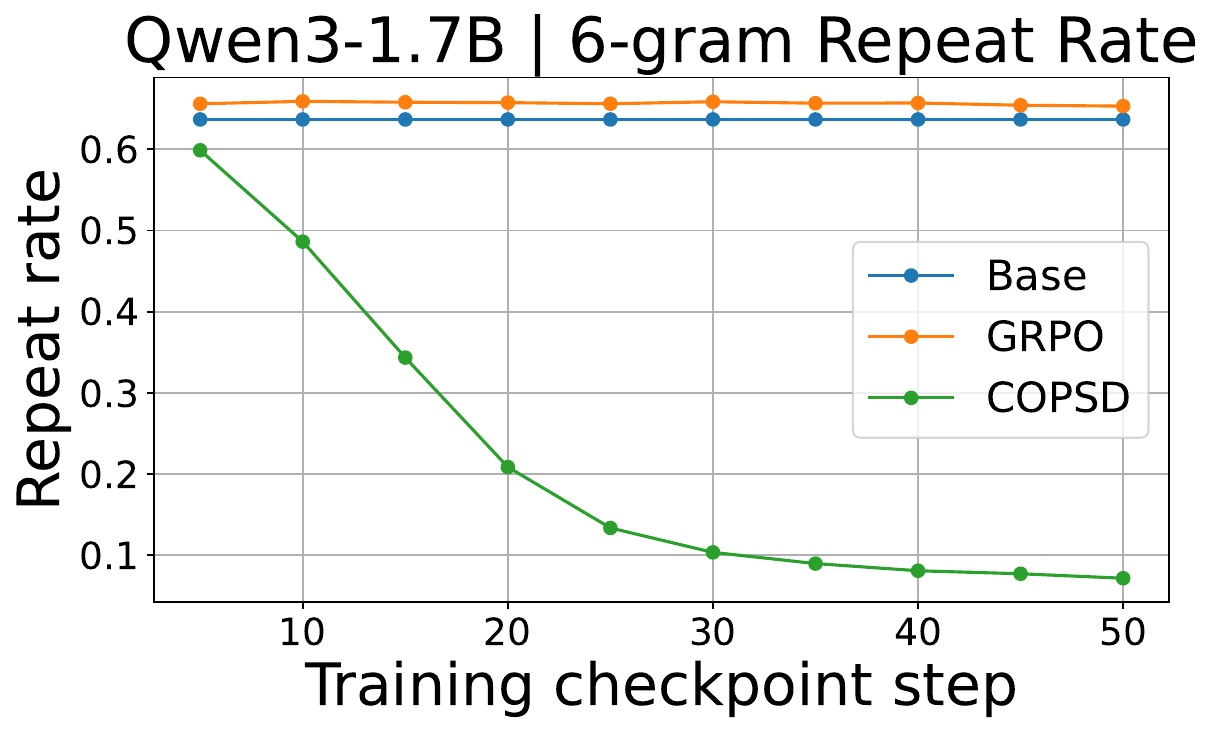}
    \includegraphics[width=0.32\linewidth]{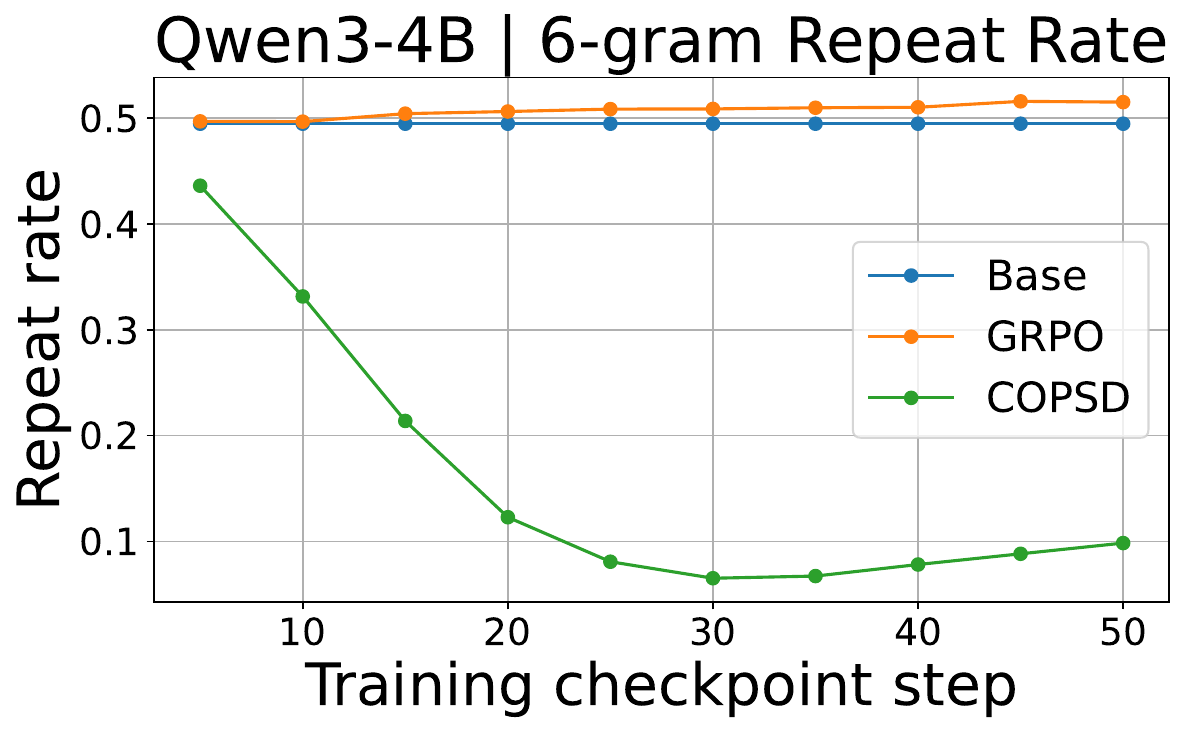}
    \includegraphics[width=0.32\linewidth]{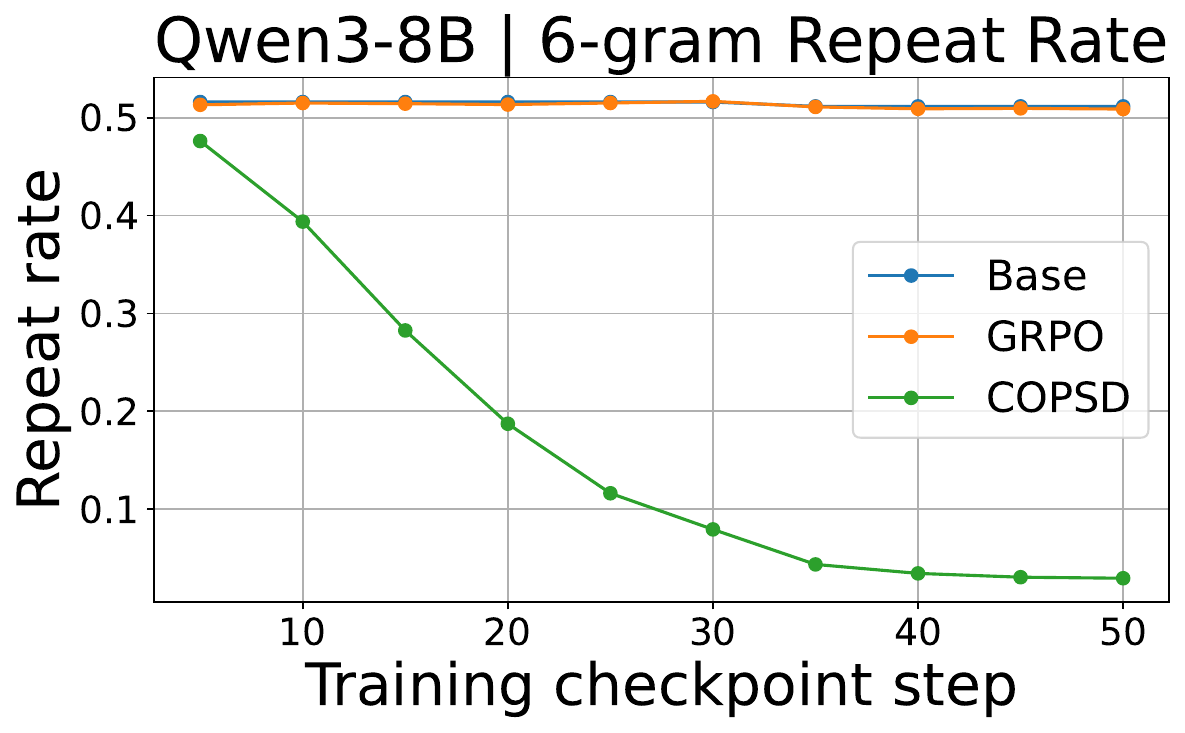}

    \caption{
    Repeat rate across different $n$-gram settings ($n=2$ to $6$) and model sizes. 
    \copsd consistently reduces repetition compared to both the base and GRPO.
    }
    \label{fig:repeat_grid}
\end{figure*}

Figure~\ref{fig:repeat_grid} presents repeat rate comparisons for $n = 1$ to $6$.
Across all settings, \copsd consistently achieves lower repeat rates than both the base model and GRPO. 
This pattern holds across different model scales and $n$-gram granularities, indicating that the reduction in repetition is robust and \copsd effectively improves the quality of low-resource reasoning.

\section{Prompt Template}\seclabel{prompt}

This section summarizes the three prompt templates used in our experiments: 
(i) the translation prompt for constructing low-resource training questions from OpenThoughts \citep{guha2025openthoughtsdatarecipesreasoning}, 
(ii) the student-policy prompt, and 
(iii) the teacher-policy prompt.
The student-policy and the teacher-policy are adapted from prompts used by \citet{zhao2026selfdistilledreasoneronpolicyselfdistillation}. 
For the student and teacher policies, we illustrate the instantiated prompts using Swahili as an example.
For readability, we additionally provide English reference translations of the Swahili prompts.
The complete prompt templates for all languages are available in our GitHub repository.\footnote{\url{https://github.com/cisnlp/COPSD}}

\subsection{Translation Prompt}

We use a translation prompt to convert English mathematical problems into target languages from 17 low-resource African languages.
This prompt is used with \texttt{Gemini-3-Flash} to translate only the problem text while preserving mathematical content, numbers, and \LaTeX{} expressions.\footnote{\url{https://aistudio.google.com/models/gemini-3}}

\begin{figure*}[t]
\centering
\fbox{
\begin{minipage}{0.95\textwidth}
\small
\textbf{Translation prompt template (used with Gemini-3-Flash)}\\[0.5em]
\texttt{
Translate the following competition math problem from English into \{language\_name\}.\\[0.25em]
Requirements:\\
- Return only the translated problem text.\\
- Do not add explanations, notes, quotation marks, or formatting wrappers.\\
- Preserve all numbers exactly.\\
- Preserve all LaTeX expressions exactly as they appear.\\
- Preserve the meaning and the final asked quantity exactly.\\
- Do not solve the problem.\\[0.25em]
English problem:\\
\{problem\}
}
\end{minipage}
}
\caption{
Translation prompt template used to translate English training questions into a target language.
The prompt explicitly constrains the model to preserve mathematical content and return only the translated problem text.
}
\label{fig:translation_prompt}
\end{figure*}

\subsection{Student-Policy Prompt}

The student policy receives only the low-resource problem and a language-specific instruction asking the model to reason step by step in the target language.
Figure~\ref{fig:student_prompt} shows the instantiated prompt for Swahili, together with an English reference translation.

\begin{figure*}[t]
\centering

\fbox{
\begin{minipage}{0.47\textwidth}
\small
\textbf{Student-policy prompt (Swahili)}\\[0.5em]
\textbf{Swali}: \textit{[problem\_target]}\\[0.5em]
\textit{Tafadhali fikiri hatua kwa hatua, na uweke jibu lako la mwisho ndani ya \texttt{\textbackslash boxed\{\}}.}
\end{minipage}
}
\hfill
\fbox{
\begin{minipage}{0.47\textwidth}
\small
\textbf{English reference translation}\\[0.5em]
\textbf{Question}: \textit{[problem\_target]}\\[0.5em]
\textit{Please think step by step, and place your final answer inside \texttt{\textbackslash boxed\{\}}.}
\end{minipage}
}

\caption{
Student-policy prompt.
The left panel shows the instantiated Swahili prompt used in training and inference; the right panel provides an English reference translation.
}
\label{fig:student_prompt}
\end{figure*}

\subsection{Teacher-Policy Prompt}

The teacher policy is given privileged crosslingual information, including the target-language problem, the English translation of the problem, and the English reference solution.
It is then asked to solve the original low-resource problem in the target language.
Figure~\ref{fig:teacher_prompt} shows the teacher prompt instantiated for Swahili, together with an English reference translation.

\begin{figure*}[t]
\centering

\fbox{
\begin{minipage}{0.47\textwidth}
\small
\textbf{Teacher-policy prompt (Swahili)}\\[0.5em]
\textbf{Swali}: \textit{[problem\_target]}\\[0.5em]
\textbf{Tafsiri ya Kiingereza ya swali}: \textit{[problem\_english]}\\[0.5em]
\textbf{Suluhisho sahihi la rejeleo kwa Kiingereza}:\\
\texttt{=== Mwanzo wa Suluhisho la Rejeleo ===}\\
\textit{[solution\_english]}\\
\texttt{=== Mwisho wa Suluhisho la Rejeleo ===}\\[0.5em]
\textit{Baada ya kusoma suluhisho la rejeleo la Kiingereza hapo juu, hakikisha umeelewa kweli mantiki ya kila hatua---usilinakili wala kulifafanua upya tu. Sasa, kwa kutumia maneno yako mwenyewe na hoja huru, tatua swali la asili kwa Kiswahili. Fikiri hatua kwa hatua, jaribu mbinu tofauti, na usiogope kurudi nyuma au kufikiria upya ikiwa jambo fulani halifanyi kazi:}\\[0.5em]
\textit{Tafadhali fikiri hatua kwa hatua kwa Kiswahili, na uweke jibu lako la mwisho ndani ya \texttt{\textbackslash boxed\{\}}.}
\end{minipage}
}
\hfill
\fbox{
\begin{minipage}{0.47\textwidth}
\small
\textbf{English reference translation}\\[0.5em]
\textbf{Question}: \textit{[problem\_target]}\\[0.5em]
\textbf{English translation of the question}: \textit{[problem\_english]}\\[0.5em]
\textbf{Correct reference solution in English}:\\
\texttt{=== Begin Reference Solution ===}\\
\textit{[solution\_english]}\\
\texttt{=== End Reference Solution ===}\\[0.5em]
\textit{After reading the English reference solution above, make sure you truly understand the logic of each step---do not simply copy or paraphrase it. Now, using your own words and independent reasoning, solve the original question in Swahili. Think step by step, try different approaches, and do not be afraid to backtrack or rethink if something does not work:}\\[0.5em]
\textit{Please think step by step in Swahili, and place your final answer inside \texttt{\textbackslash boxed\{\}}.}
\end{minipage}
}

\caption{
Teacher-policy prompt.
The teacher receives privileged information, including the English translation of the problem and the English reference solution, before solving the original low-resource problem.
The left panel shows the instantiated Swahili prompt; the right panel provides an English reference translation.
}
\label{fig:teacher_prompt}
\end{figure*}

For all other languages, the same prompt structure is used with language-specific instructions, labels, and reasoning prefixes.
The complete prompt templates for every language in our experiments are provided in our GitHub repository.
\section{Environment and Hyperparameters}\seclabel{environment}

We largely follow the training configuration of \citet{zhao2026selfdistilledreasoneronpolicyselfdistillation} for both GRPO and OPSD-style training.
The main difference is that we set the maximum completion length for \copsd to 2048 tokens, instead of the 1024-token budget used in the original OPSD setup.
Unlike \citet{zhao2026selfdistilledreasoneronpolicyselfdistillation}, we enable thinking mode for both the student and teacher policies.
This is necessary for eliciting language-specific reasoning traces, as our language-control strategy inserts a target-language prefix immediately after the \texttt{<think>} token, as described in \secref{language_control}.

We train a separate model for each language and model scale, resulting in $17 \times 3 = 51$ models in total.
All experiments are conducted on either 8 NVIDIA A100 GPUs or 4 NVIDIA H200 GPUs.
We use LoRA \citep{hu2022lora} for parameter-efficient fine-tuning, AdamW \citep{kingma2015adam,Loshchilov2019adamw} as the optimizer, and bfloat16 precision for all training runs.
By default, \copsd uses full-vocabulary logit distillation with a fixed teacher policy.
For both \copsd and GRPO, we save checkpoints every 5 training steps.
Table~\ref{tab:hyperparameters} summarizes the main training hyperparameters.

For evaluation, we use the same decoding configuration for all models and methods to ensure fair comparison, as shown in Table~\ref{tab:inference_hyperparameters}.
We enable thinking mode and sample 12 responses per problem with temperature $1.0$ and top-$p=0.95$.
For AfriMGSM, we evaluate under maximum new-token budgets of 1,024, 2,048, and 4,096 tokens.
These budgets are sufficient for AfriMGSM because the benchmark consists of relatively short mathematical reasoning problems.
Final answers are extracted from \texttt{\textbackslash boxed\{\}} and verified as described in \secref{experiments}.
For each language and method, we select the checkpoint that achieves the best performance under the 1,024-token budget, and then report that checkpoint's performance under the other generation budgets.

\begin{table}[t]
\centering
\small
\begin{tabular}{lcc}
\toprule
\textbf{Parameter} & \textbf{GRPO} & \textbf{\copsd} \\
\midrule
Learning Rate & $5 \times 10^{-6}$ & $5 \times 10^{-6}$ \\
Effective Batch Size & 32 & 32 \\
LoRA Rank ($r$) & 64 & 64 \\
LoRA Alpha ($\alpha$) & 128 & 128 \\
Max Completion Length & 16,000 & 2048 \\
Generations per Prompt & 8 & 1 \\
Sampling Temperature & 1.2 & 1.1 \\
KL Coefficient ($\beta$) & 0.0 & -- \\
Training Steps & 500 & 100 \\
\bottomrule
\end{tabular}
\caption{
Training hyperparameters for GRPO and \copsd.
We follow the original OPSD configuration \citep{zhao2026selfdistilledreasoneronpolicyselfdistillation} except that \copsd uses a maximum completion length of 2048 tokens to allow more supervision on low-resource generations.
}
\label{tab:hyperparameters}
\end{table}

\begin{table}[t]
\centering
\small
\begin{tabular}{lc}
\toprule
\textbf{Parameter} & \textbf{Value} \\
\midrule
Max New Tokens &  1,024, 2,048, or 4,096\\
Thinking Mode & Enabled \\
Top-$p$ & 0.95 \\
Top-$k$ & -1 \\
Min-$p$ & 0.0 \\
Presence Penalty & 0.0 \\
Samples per Prompt & 12 \\
Temperature & 1.0 \\
\bottomrule
\end{tabular}
\caption{
Inference hyperparameters used for evaluation.
We use the same decoding configuration for all models and methods.
}
\label{tab:inference_hyperparameters}
\end{table}

\end{document}